\documentclass[acmtog]{acmart}

\usepackage{booktabs} %
\usepackage{hhline}
\usepackage[ruled]{algorithm2e} %
\usepackage{subcaption} %
\usepackage{pifont}
\usepackage{arydshln}
\usepackage{multirow}
\usepackage{multicol}
\usepackage{xcolor}
\usepackage{cleveref}
\usepackage{xspace}
\usepackage{rotating}
\usepackage{xhfill}

\usepackage[toc,page,header]{appendix}

\def\etal{\emph{et al}\onedot}

\definecolor{darkgreen}{HTML}{008000}
\definecolor{jazzberryjam}{rgb}{0.65, 0.04, 0.37}
\definecolor{lightgreen}{HTML}{82cf8e}
\definecolor{lightblue}{HTML}{618fab}
\definecolor{lightpurple}{HTML}{ac85cc}

\def\Bezier{B\'{e}zier\xspace}

\makeatletter
\DeclareRobustCommand\onedot{\futurelet\@let@token\@onedot}
\def\@onedot{\ifx\@let@token.\else.\null\fi\xspace}

\def\etal{\emph{et al}\onedot}

\def\etal{\emph{et al}\onedot}

\setcopyright{none}
\makeatletter
\let\@authorsaddresses\@empty
\makeatother

\makeatletter
\def\@copyrightspace{\relax}
\makeatother

\setcopyright{none}
\renewcommand\footnotetextcopyrightpermission[1]{}
\makeatletter
\def\runningfoot{\def\@runningfoot{}}
\def\firstfoot{\def\@firstfoot{}}
\makeatother

\title{NeuralSVG: An Implicit Representation for Text-to-Vector Generation}

\author{Sagi Polaczek}
\affiliation{
\institution{Tel Aviv University}
\country{Israel}
}
\author{Yuval Alaluf}
\affiliation{
\institution{Tel Aviv University}
\country{Israel}
}
\author{Elad Richardson}
\affiliation{
\institution{Tel Aviv University}
\country{Israel}
}
\author{Yael Vinker}
\affiliation{
\institution{Tel Aviv University}
\country{Israel}
}
\affiliation{
\institution{MIT CSAIL}
\country{USA}
}
\author{Daniel Cohen-Or}
\affiliation{
\institution{Tel Aviv University}
\country{Israel}
}

\begin{document}

\begin{abstract}
Vector graphics are essential in design, providing artists with a versatile medium for creating resolution-independent and highly editable visual content. Recent advancements in vision-language and diffusion models have fueled interest in text-to-vector graphics generation. However, existing approaches often suffer from over-parameterized outputs or treat the layered structure --- a core feature of vector graphics --- as a secondary goal, diminishing their practical use. Recognizing the importance of layered SVG representations, we propose NeuralSVG, an implicit neural representation for generating vector graphics from text prompts. Inspired by Neural Radiance Fields (NeRFs), NeuralSVG encodes the entire scene into the weights of a small MLP network, optimized using Score Distillation Sampling (SDS). To encourage a layered structure in the generated SVG, we introduce a dropout-based regularization technique that strengthens the standalone meaning of each shape. We additionally demonstrate that utilizing a neural representation provides an added benefit of inference-time control, enabling users to dynamically adapt the generated SVG based on user-provided inputs, all with a single learned representation. Through extensive qualitative and quantitative evaluations, we demonstrate that NeuralSVG outperforms existing methods in generating structured and flexible SVG. 
Project page: \url{https://sagipolaczek.github.io/NeuralSVG/}.
    
\end{abstract}

\begin{teaserfigure}
\centering
\includegraphics[width=0.95\textwidth]{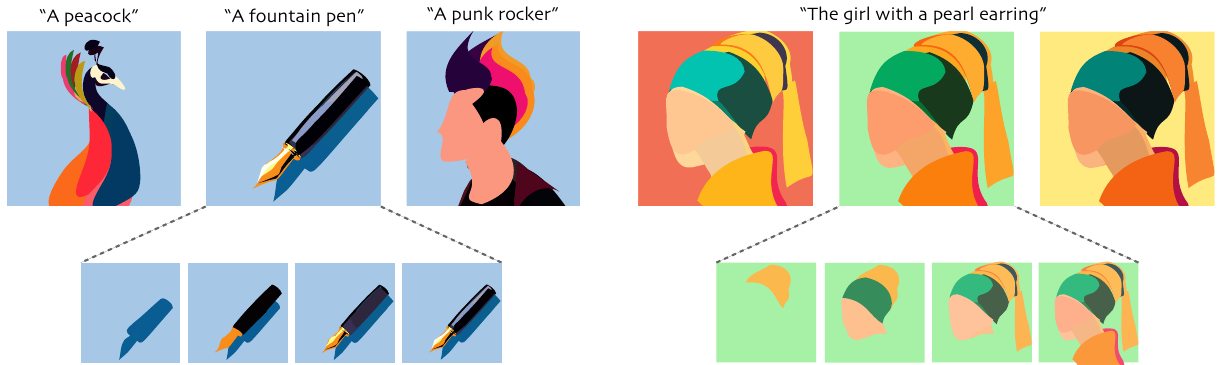} \\[-0.2cm]
\caption{
NeuralSVG generates vector graphics from text prompts with ordered and editable shapes. Our method supports dynamic background color conditioning, facilitating the generation of multiple color palettes for a single learned representation (right).
}
\label{fig:teaser}
\end{teaserfigure}

\maketitle

\section{Introduction}~\label{sec:intro}
Vector graphics represent images using parametric shapes, such as circles, polygons, lines, and curves, in contrast to rasterized images, which rely on pixel-level representations. Unlike raster images, vector graphics are resolution-independent, easily editable, and particularly effective for creating simplified visuals. These advantages make vector graphics a preferred choice in fields such as design, web development, and data visualization. Recent research has sought to automate the generation of vector graphics, aiming to create high-quality, scalable visual content accessible to both experts and non-experts alike.

\begin{figure}
    \centering
    \includegraphics[width=1\linewidth]{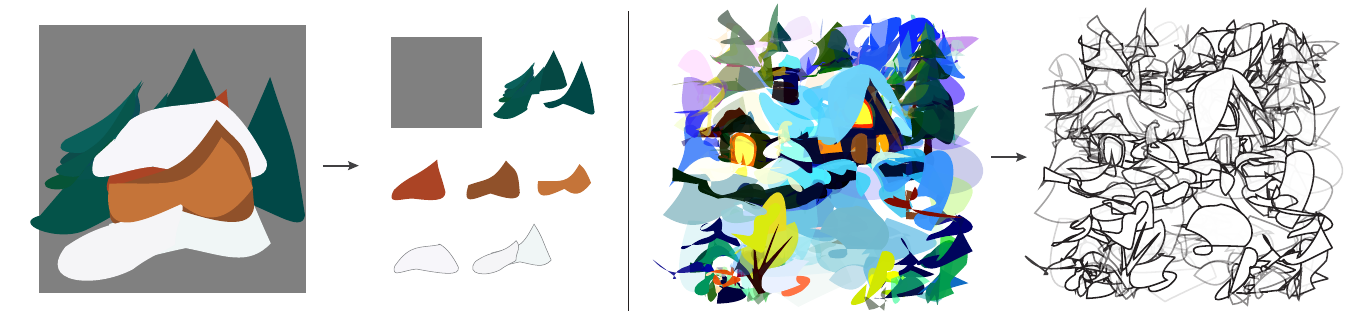}
    \vspace{-0.65cm}
    \caption{\textbf{The Importance of Layers and Compact Shapes in SVGs for Editability.} Left: SVGs are typically composed of ordered layers (e.g., the gray background and trees are placed behind the house) and individual shapes that represent complete components in an editable manner (e.g., snow can be removed or adjusted by modifying a few shapes). Right: An SVG that may appear visually appealing when rendered but lacks practical use for editing or control, as its individual components are difficult to modify.
    }
    \label{fig:intro_layered_svgs}
\end{figure}

With recent advancements in large-scale vision-language models~\cite{yin2024survey} and image diffusion models~\cite{po2023state}, there has been a growing interest in introducing these strong priors to directly generate vector graphics from text prompts~\cite{jain2023vectorfusion,xing2024svgdreamer,zhang2024text,thamizharasan2024nivel}. However, existing methods, while technically producing vector graphics, often result in over-parameterized outputs composed of almost pixel-like shapes, thus losing the original motivation and core advantages of editable vector graphics (see \Cref{fig:intro_layered_svgs}).

Notably, the editable nature of SVGs is inherently linked to their layered representation. These layers separate elements like backgrounds, text, and shapes for easier navigation, enable independent editing without affecting other components, and provide a hierarchical structure for stacking and visual clarity.
Motivated by this, several works have proposed methods for generating layer-based SVG representations~\cite{thamizharasan2024nivel,zhang2024text}. However, these approaches often depend on multiple post-processing stages to construct a meaningful layered structure.
Ideally, the SVG generation process itself should account for the hierarchical nature of SVGs, promoting the creation of shapes that possess standalone semantic meaning while contributing to the overall composition.

In this work, we introduce \textit{NeuralSVG}, an implicit neural representation for text-to-vector generation that takes into account the layered structure of vector graphics and offers greater flexibility in the generation process. 
Inspired by Neural Radiance Fields (NeRFs), which output individual points in space that are then aggregated into a scene, we propose a network that outputs individual shapes which are then aggregated to form the complete SVG.
Following prior work, the network weights are optimized using the standard Score Distillation Sampling (SDS) loss~\cite{poole2022dreamfusion}. In this formulation, the entire network encodes the complete SVG as an implicit neural representation, defined by its learned weights. To promote a semantic and ordered representation, we further introduce a dedicated dropout-based regularization method during the optimization process. This method encourages each learned shape to have a meaningful and ordered role in the overall composition.

Importantly, using a neural representation introduces greater flexibility in utilizing and extending SVGs. Specifically, we demonstrate that our implicit representation enables inference-time control over the generated asset. 
For instance, by conditioning the generation on a target background color, our network can learn to produce a color palette for the SVG that best complements this background. As illustrated in~\Cref{fig:teaser}, this enables the creation of dynamic SVGs that adapt to user-specific preferences.

We evaluate NeuralSVG through comprehensive qualitative and quantitative experiments, demonstrating improved performance across a diverse range of inputs compared to existing methods. Notably, we show that our single-stage framework generates meaningful individual shapes, providing users with a well-structured and layered representation. 
Additionally, we demonstrate that NeuralSVG can be adapted to produce vectorized sketches without any modifications. 
Finally, as a key distinguishing feature, we highlight how NeuralSVG supports additional user inputs, creating an adaptive SVG representation that can be dynamically adjusted at inference time beyond the capabilities of standard SVG representations.

\section{Related Work}
\label{sec:related_work}

\subsection{Vector Representation}
Scalable Vector Graphics (SVGs)~\cite{jackson2005scalable} offer a flexible and powerful medium for representing visual concepts, leveraging primitives such as \Bezier curves~\cite{bezier1986courbes}. 
Extensive research has focused on learning neural-based representations of SVGs. 
SketchRNN~\cite{ha2017neural} uses a recurrent neural network (RNN) to generate vector paths for sketches, while DeepSVG~\cite{carlier2020deepsvg} adopts a hierarchical Transformer model to create vector icons with multiple paths. More recently, IconShop~\cite{wu2023iconshop} represents SVGs as token sequences.

\subsection{Text-to-Image Generation}
Recent advancements in large-scale generative models~\cite{po2023state,yin2024survey} have rapidly transformed content creation, especially in visual content generation. 
Among these, large-scale diffusion models~\cite{ramesh2022hierarchical,nichol2021glide,balaji2023ediffi,kandinsky2,ding2022cogview2,saharia2022photorealistic,rombach2021highresolution} have achieved unprecedented levels of quality, diversity, and fidelity in their outputs.
These models have also spurred the development of various text-guided tasks. 
Central to this progress is the Score Distillation Sampling (SDS) loss, introduced by Poole~\etal~\shortcite{poole2022dreamfusion}, which has proven highly effective for extracting meaningful signals from pretrained text-to-image diffusion models.
SDS has enabled a wide range of applications, including text-to-3D generation~\cite{poole2022dreamfusion,richardson2023texture,metzer2023latent,lin2023magic3d,wang2023score,wang2024prolificdreamer}, image editing~\cite{hertz2023delta,koo2024posterior,kim2025dreamsampler}, sketch generation~\cite{iluz2023word,Gal_2024_CVPR,mo2024text,xing2023diffsketcher,kim2023collaborative}, and text-to-SVG generation.

\subsection{Vector Graphics Generation}
Early vector graphics generation approaches relied on sequence-based learning applied to vector representations~\cite{carlier2020deepsvg,ha2017neural,lopes2019learned,ganin2018synthesizing,wang2023deepvecfont,wu2023iconshop}, but their dependence on vector datasets limited their generalization to more complex generations.
Advances in differentiable rendering~\cite{zheng2018strokenet,mihai2021differentiable,li2020differentiable,reddy2020discovering} have enabled vector synthesis using raster-based losses~\cite{shen2021clipgen,reddy2021im2vec,ma2022towards,xing2023diffsketcher}.
Additionally, the emergence of large-scale vision-language models, such as CLIP~\cite{radford2021learning}, had led to innovative methods for sketch and vector generation~\cite{jain21vector,frans2022clipdraw,vinker2022clipasso,vinker2023clipascene,mirowski2022clip,song2023clipvg,tian2022modern,rodriguez2023starvector,vinker2024sketchagent}.

Recent research has focused on integrating diffusion models into vector graphics generation. A key approach optimizes geometric and color parameters of primitives using diffusion model priors with SDS-based losses~\cite{jain2023vectorfusion,thamizharasan2024nivel,xing2024svgdreamer,zhang2024text}. However, these methods often suffer from redundant and degraded geometry due to the absence of ordering constraints. For instance, SVGDreamer~\cite{xing2024svgdreamer} requires numerous shapes (e.g., $\{\sim\}512$) and supports only basic scene decomposition into background and foreground.
Text-to-Vector~\cite{zhang2024text} trains a Variational Autoencoder (VAE) to encode valid geometric properties into a path latent space. Their method employs a two-stage path optimization process for text-to-vector generation, utilizing the learned latent space with an SDS-based loss. As a post-processing step, they simplify the obtained paths to produce a layer-wise representation. 
NIVeL~\cite{thamizharasan2024nivel} trains an MLP to learn decomposable SVG layers, generating layered outputs in pixel space that are vectorized into \Bezier curves via marching squares in post-processing.

Several methods combine text-to-image generation with image vectorization techniques~\cite{ma2022towards,wang2024layered,hirschorn2024optimize} to produce vector graphics~\cite{kopf2011depixelizing,chen2023editable,du2023image}. 
For instance, LIVE~\cite{ma2022towards} employs a differentiable rasterizer to iteratively optimize closed \Bezier paths. Wang~\etal~\shortcite{wang2024layered} combined Score Distillation Sampling and semantic segmentation to iteratively simplify the input image into vectorized layers.

In this work, we propose a novel implicit neural representation for SVGs, encoding the SVG as the weights of a small MLP neural network. This neural representation provides a more interpretable generation process with enhanced user control, allowing customization of parameters such as the number of shapes, background color, and aspect ratio, all within a single network.

\vspace{-0.3cm}
\subsection{Ordered Representations}
Ordered representations, such as those obtained through Principal Component Analysis (PCA), where dimensions are ranked by their relative importance, are extensively used in machine learning and statistics. Rippel~\etal~\shortcite{rippel2014learning} demonstrated that neural networks could be encouraged to learn ordered representations by applying a specialized form of dropout on hidden units.

In the context of generative models, the exploration of ordered representations is still a developing area. Alaluf~\etal~\shortcite{alaluf2023neural} utilize ordered representations to personalize text-to-image models, enabling inference-time control over the reconstruction and editability of learned concepts. Zhang~\etal~\shortcite{zhang2024text} introduce a post-processing method for SVGs, where layer-wise structures are extracted from complete SVGs through a path simplification process.
In this work, we adopt an ordered-centric approach to SVG generation, integrating the layered structure directly into the generation process.

\setlength{\abovedisplayskip}{4pt}
\setlength{\belowdisplayskip}{4pt}

\vspace{-0.2cm}
\section{Preliminaries}
\paragraph{\textbf{Score-Distillation Sampling}}
Score-Distillation Sampling (SDS), introduced by Poole~\etal~\shortcite{poole2022dreamfusion}, has emerged as a prominent technique for extracting meaningful signals from pretrained text-to-image diffusion models. 
The authors demonstrated how the standard diffusion loss can be leveraged to optimize the parameters of a NeRF~\cite{mildenhall2021nerf} model for text-to-3D generation. 

Given an image $x$ (e.g., a radiance field rendered from a specific viewpoint) synthesized by a model with parameters $\phi$, the image is noised to an intermediate diffusion timestep $t$ as follows:
\begin{equation}
x_t = \alpha_t x + \sigma_t \epsilon
\end{equation}
where $\epsilon \sim \mathcal{N}(0,1)$ represents a noise sample, and $\alpha_t$ and $\sigma_t$ are parameters defined by the denoising scheduler.

The noised image is then passed through a pretrained, frozen denoising model conditioned on a prompt $p$, which aims to predict the added noise $\epsilon$. The deviation between the predicted noise and the true added noise serves as a measure of the difference between the input image $x$ and one that better matches the given prompt. The corresponding gradients can then be used to update the parameters $\phi$ of the original synthesis model, guiding it to generate outputs more aligned with the prompt. The loss function is given by:
\begin{equation}~\label{eq:sds}
\nabla_{\theta} \mathcal{L}{\text{SDS}} =
\mathbb{E}_{t, \epsilon}
\left[
w(t) \left( \hat{\epsilon}_{\phi}(\mathbf{S}t; p, t) - \epsilon \right)
\frac{\partial \mathbf{S}}{\partial \theta}
\right],
\end{equation}
where $\hat{\epsilon}_{\phi}$ is the noise predicted by the denoising model, and $w(t)$ is a weighting function that depends on the diffusion timestep $t$. 
Intuitively, this iterative process progressively aligns the synthesis model with the conditioning prompt $p$. 
Here, we adopt this approach to update the weights of our network representing the SVG scene.

\begin{figure}
    \centering
    \includegraphics[width=0.5\textwidth]{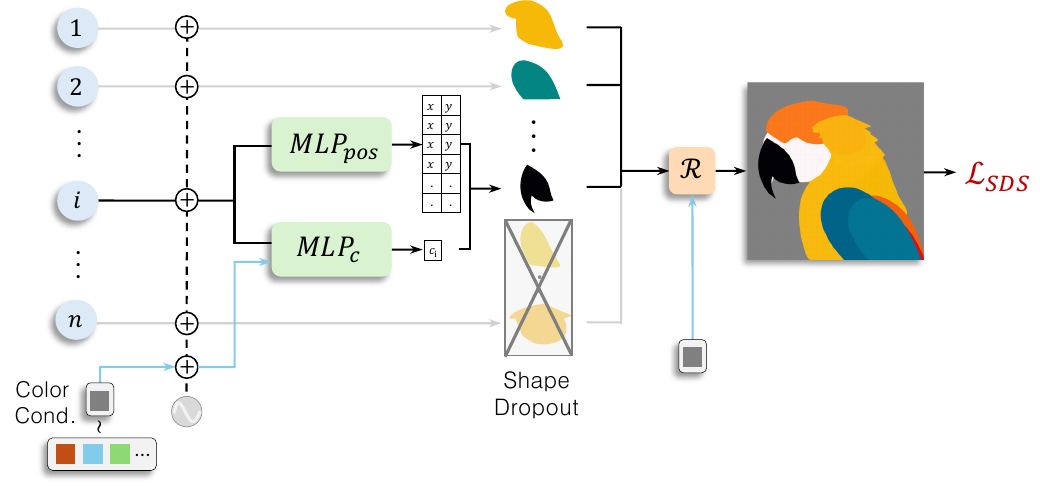} \\[-0.15cm]
    \caption{
    \textbf{NeuralSVG Overview.} Input indices $\{1,\dots,n\}$, each corresponding to a single shape, are processed through two parallel branches: $MLP_{\text{pos}}$, which predicts the control points of the shape, and $MLP_{\text{c}}$, which predicts its RGB color. The predicted shapes and colors are then aggregated and rendered using a differentiable rasterizer $\mathcal{R}$.
    To encourage a meaningful ordering of the shape primitives, a truncation index is randomly sampled during training, and all shapes above this index are dropped. 
    The final rendered vector graphic is optimized to align with the user-provided text prompt using an SDS loss~\cite{poole2022dreamfusion}, guided by a trained diffusion model.
    Additionally, random background colors are sampled during training, with their RGB values passed to $MLP_{\text{c}}$ and $\mathcal{R}$. 
    \\[-0.2cm]
    }
    \label{fig:method}
\end{figure}

\section{Method}\label{sec:method}
Given a user-provided text prompt, NeuralSVG learns an implicit neural representation of the corresponding vector graphics scene. We begin by describing our network architecture and training scheme. We then introduce a dropout-based regularization technique applied during optimization, which is designed to establish a meaningful ordering of the learned shape primitives. Finally, we demonstrate how our neural representation enables greater user flexibility, allowing users to better customize the generated SVGs using a single learned representation.
A high-level overview of NeuralSVG is illustrated in~\Cref{fig:method}.

\subsection{Neural SVG Representation}
Our neural SVG representation is inspired by the implicit representation of Neural Radiance Fields (NeRFs)~\cite{mildenhall2021nerf}, where 3D pixel coordinates are mapped to spatial points through a compact mapping network. Similarly, we represent an SVG implicitly as a set of indices, $\{1,2,\dots,n\}$, where each index $i$ corresponds to a single shape $z_i$ in the SVG. 
Each shape is defined by four concatenated cubic \Bezier curves, with their first and last control points being identical to form a closed shape. This results in $12$ control points $p_i=\{x_j,y_j\}_{j=1}^{12}$ per shape. Each shape is defined by its control points and fill color: $z_i=(p_i,c_i)$.
Specifically, we learn a function using a small MLP network, $f_\theta$ with learnable weights $\theta$:
\begin{equation}
    f_\theta : i \to (p_i, c_i),
\end{equation}
In essence, the MLP takes a shape index $i \in \{1,2,\dots,n\}$ as input and outputs the parameters defining the corresponding shape.
These individual shapes are aggregated to form the full set of shapes and are then rendered using a differentiable rasterizer~\cite{li2020differentiable} to produce the output in pixel space. 
  
In this formulation, the entire vector scene is encoded within the weights of the network. During inference, the network can then be queried to generate the SVG by feeding it with each of the $n$ indices. Additionally, this neural representation can be extended to accept additional input parameters, such as background color.

\subsection{Architecture.}
Our model consists of three primary components: a positional encoding layer and two Multi-Layer Perceptron (MLP) networks.

\paragraph{\textbf{Positional Encoding}}
Given the index of the $i^\text{th}$ shape, we first map the scalar value $i$ to a higher-dimensional space, following prior works on implicit representations~\cite{alaluf2023neural,Gal_2024_CVPR,mildenhall2021nerf,thamizharasan2024nivel}. 
Specifically, each input scalar is encoded using Random Fourier Features~\cite{rahimi2007random,tancik2020fourierfeaturesletnetworks} into a $128$-dimensional vector, $\gamma(i) \in \mathbb{R}^{128}$, modulated by $64$ random frequencies. The encoding function is defined as:
\begin{equation}
\gamma(i) = \left [ \cos(2\pi \mathbf{B}i) \sin(2\pi \mathbf{B}i) \right ]
\end{equation}
where $\mathbf{B}$ is a matrix of random frequencies.

\paragraph{\textbf{Network Architecture}}
Given the high-dimensional encoding of the shape index, we predict the shape's control parameters and color using an MLP network. To better disentangle the color and shape information, each is predicted using parallel branches.
In each branch, the input vector $\mathbf{v_i} = \gamma(i)$ is passed through two fully connected layers, each followed by a LayerNorm~\cite{ba2016layernormalization} normalization layer and a LeakyReLU activation. 
The resulting vector is then passed through a final fully connected layer to produce the outputs $\hat{p}_i$ or $\hat{c_i}_i$, where $\hat{p}_i$ is a $(12{\times}2)$-dimensional output representing the $(x,y)$ coordinates of the $12$ control points, and $\hat{c}_i$ is a $3$-dimensional output representing the RGB color values:
\begin{equation}
\hat{p}_i = \text{MLP}_{\text{pos}}(\mathbf{v}_i) \in \mathbb{R}^{12\times2} \qquad
\hat{c}_i = \text{MLP}_c(\mathbf{v}_i) \in [0, 1]^3.
\end{equation}
Finally, the output color values are additionally passed through a Sigmoid function to ensure the values are between $0$ and $1$.

\subsection{Training Scheme}
\paragraph{\textbf{Initialization}}
To calibrate the outputs of the mapping networks for generating points within the rendered canvas, we perform an initialization stage, as is common in text-to-vector approaches~\cite{jain2023vectorfusion,thamizharasan2024nivel,xing2024svgdreamer}.
Specifically, given the user-provided text prompt $p$, we first generate an image using an off-the-shelf text-to-image diffusion model~\cite{rombach2021highresolution}. We then adopt the saliency-based initialization technique proposed by Vinker \etal~\shortcite{vinker2023clipascene}, identifying salient regions in the image via an attention-based relevancy map. From this map, we sample $n$ points and convert them into a set of convex shapes with simple geometry. To initialize the corresponding RGB color values, we extract the colors from the relevant pixels in the generated image. This process provides an initial set of $n$ shape control points $p^{\text{init}}_i$, and their corresponding color values, $c^{\text{init}}_i$.

Next, we train our network to predict these extracted positions and colors. The network is trained using a simple $\mathcal{L}_2$ loss to encourage accurate reconstruction of the initialization values:
\begin{equation}
\begin{split}
    \mathcal{L}_{\text{pos}}(i) & = \| MLP_{\text{pos}}(i) - p^{\text{init}}_i \|_2^2, \\
    \mathcal{L}_{\text{c}}(i) & = \| MLP_{\text{c}}(i) - c^{\text{init}}_i \|_2^2.
\end{split}
\end{equation}
Having initialized the outputs of the network, we now turn to describe how the network can be trained to represent the desired vector graphics scene based on the user-provided prompt.

\vspace{-0.25cm}
\paragraph{\textbf{Training.}}
To guide the training process, we leverage a pretrained text-to-image diffusion model, specifically Stable Diffusion \cite{rombach2021highresolution}. 
To better capture the visual look of SVGs, we fine-tune a LoRA adapter using a small dataset of high-quality vector art images. We provide additional details on this fine-tuning in the supplementary.

Following prior works on text-to-vector generation, the training process is driven by an SDS loss~\cite{poole2022dreamfusion}.
At each training iteration, the full set of $n$ indices is passed through the network to predict all the control points and colors. These primitives are rendered using the differentiable renderer $\mathcal{R}$ to produce the current representation of the scene. Finally, we use the SDS loss defined in~\Cref{eq:sds} to update the parameters of our network. 
Intuitively, the SDS loss guides the network to learn a vector graphics scene that faithfully reflects the desired content specified by the text prompt.

\paragraph{\textbf{Encouraging an Ordered Representation}}
The above optimization process results in a generated SVG that aligns with the provided prompt. However, it does not inherently promote a layered representation of the scene. Specifically, there is no objective that explicitly encourages a meaningful ordering of shapes, where later shapes build upon earlier ones to enhance the overall composition. 
Prior works either (1) fail to explicitly address this~\cite{jain2023vectorfusion,xing2024svgdreamer}, resulting in unordered shapes being learned, or (2) decompose the SVG in a separate post-processing~\cite{zhang2024text}.

To address this, we explicitly encourage an ordered representation to be learned directly \textit{during} the optimization process. Specifically, as illustrated on the right side of~\Cref{fig:method}, we adopt a variant of the Nested Dropout technique~\cite{rippel2014learning}. 
At each iteration, before rendering the current scene, we sample a truncation value \(tr\) and drop all shapes above this value, yielding a simplified scene \(S_{tr}\):
\begin{equation}~\label{eq:dropout}
\begin{split}
    P_{tr} = \{p_i\}_{i < tr} & \qquad C_{tr} = \{c_i\}_{i < tr} \\
    S_{tr} = \mathcal{R} & \left ( P_{tr}, C_{tr} \right ),
\end{split}
\end{equation}
where \(P_{tr}\) and \(C_{tr}\) are the truncated sets of positions and colors.

By randomly dropping shapes during training, the model is encouraged to encode more semantic information into the earlier shapes, which are less likely to be dropped. 
This technique also provides an additional benefit: enhanced user flexibility at inference. By adjusting the truncation, users can control the number of shapes rendered, tailoring the scene's complexity to their preferences.

\begin{figure}
    \centering\addtolength{\belowcaptionskip}{-7.5pt}
    \includegraphics[width=0.915\linewidth]{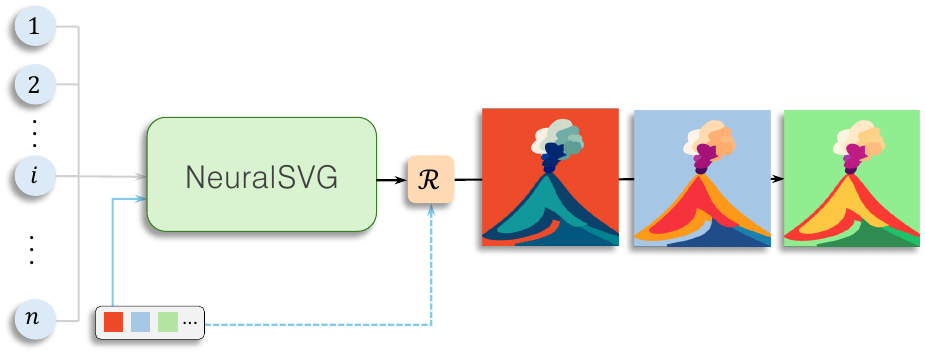} \\[-0.25cm]
    \caption{
    \textbf{Dynamic color palette control enabled by the NeuralSVG representation.}
    Given a learned representation of an SVG, users can dynamically adjust the color palette of the SVG by specifying new background colors. 
    \\[-0.35cm]
    }
    \label{fig:additional_controls}
\end{figure}

\subsection{Introducing Additional Controls}
Finally, leveraging a neural network to represent SVGs offers the additional benefit of introducing user inputs that can directly control the generated scene, all within a single learned representation.

As a motivating example, users can adjust the color palette of the generated SVG by specifying a desired background color, as illustrated in~\Cref{fig:additional_controls}. During training, we extend the previously described scheme as follows. At each training step, we sample a background color represented as RGB values. 
This sampled background color is passed through a positional encoding function and provided as an additional input to the MLP networks, alongside the encoding of the shape index. 
When rendering, the sampled background color is additionally passed to the renderer to generate the SVG with that background.
The sampled colors are chosen either from a set of predefined colors or taken as random RGB values.

At inference time, users can specify any background color to dynamically adjust the color palette of the SVG scene and better match their needs. 
We illustrate additional controls in~\Cref{sec:additional_apps}.

\section{Experiments}

\begin{figure}
    \centering
    \setlength{\tabcolsep}{1pt}
    \addtolength{\belowcaptionskip}{-5pt}
    {\small
    \begin{tabular}{c c c c c}

        \multicolumn{5}{c}{``an astronaut walking across a desert...''} \\
        \includegraphics[width=0.185\linewidth,height=0.185\linewidth]{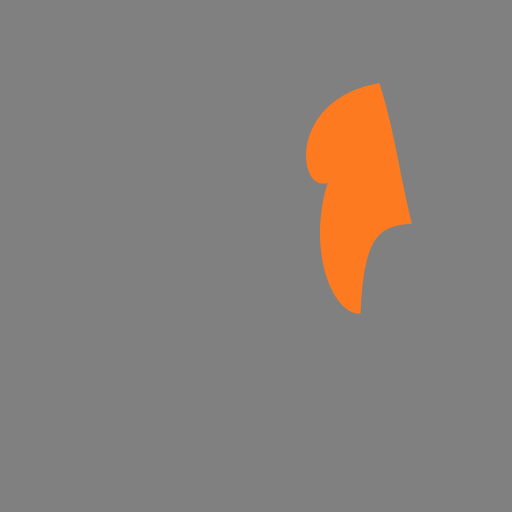} &
        \includegraphics[width=0.185\linewidth,height=0.185\linewidth]{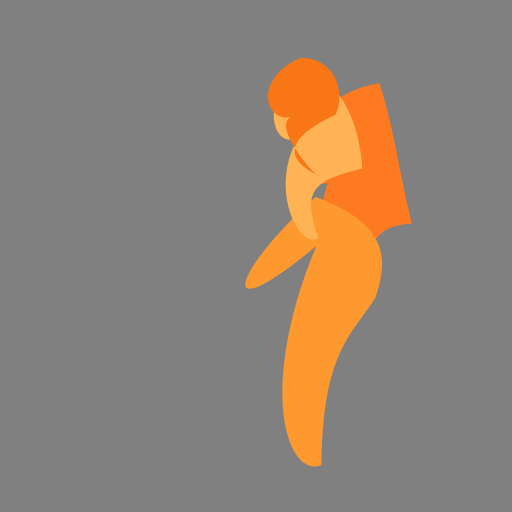} &
        \includegraphics[width=0.185\linewidth,height=0.185\linewidth]{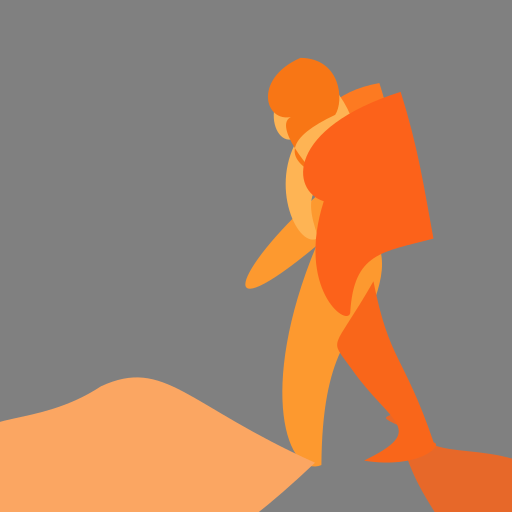} &
        \includegraphics[width=0.185\linewidth,height=0.185\linewidth]{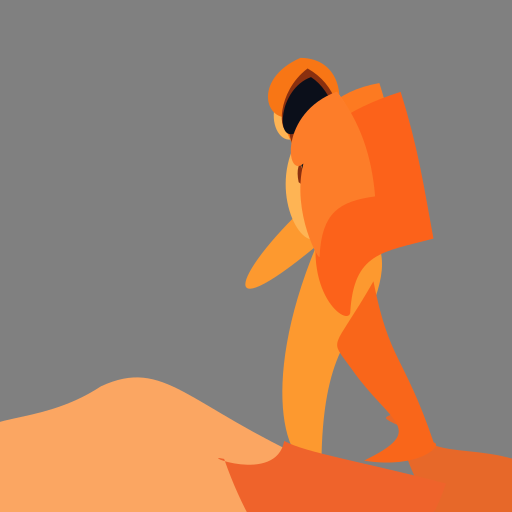} &
        \includegraphics[width=0.185\linewidth,height=0.185\linewidth]{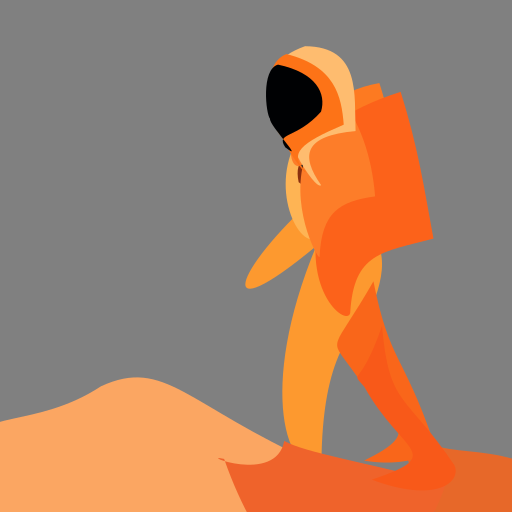} \\

        \multicolumn{5}{c}{``a family vacation to Walt Disney World''} \\
        \includegraphics[width=0.185\linewidth,height=0.185\linewidth]{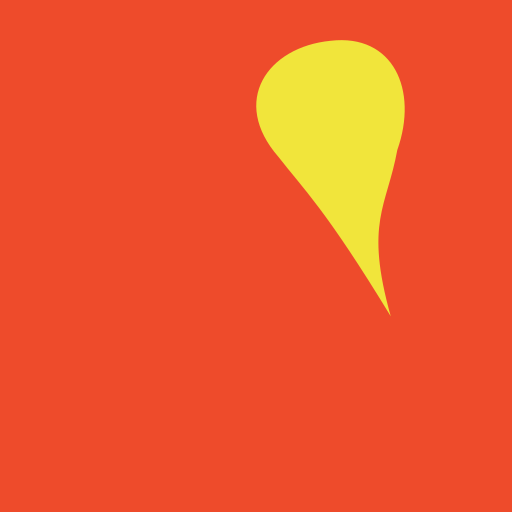} &
        \includegraphics[width=0.185\linewidth,height=0.185\linewidth]{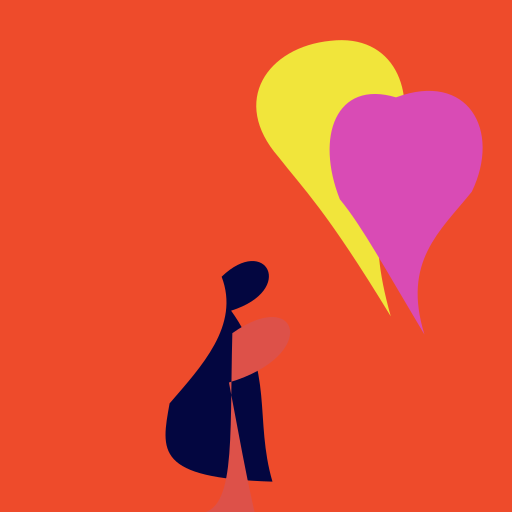} &
        \includegraphics[width=0.185\linewidth,height=0.185\linewidth]{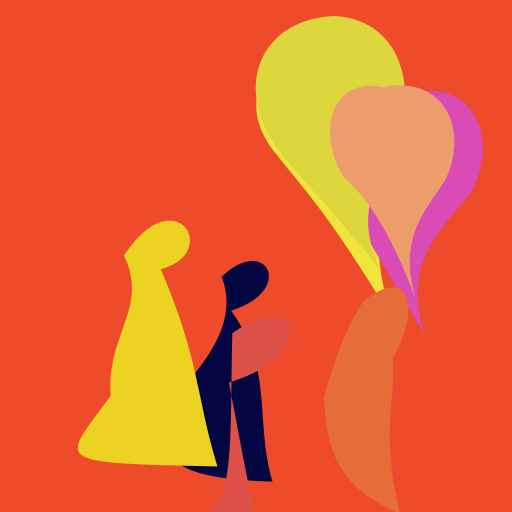} &
        \includegraphics[width=0.185\linewidth,height=0.185\linewidth]{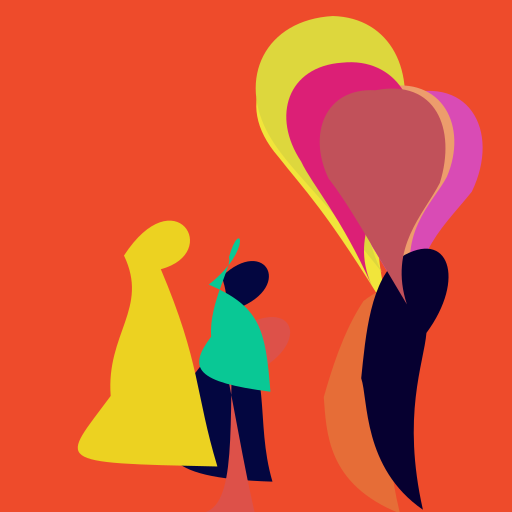} &
        \includegraphics[width=0.185\linewidth,height=0.185\linewidth]{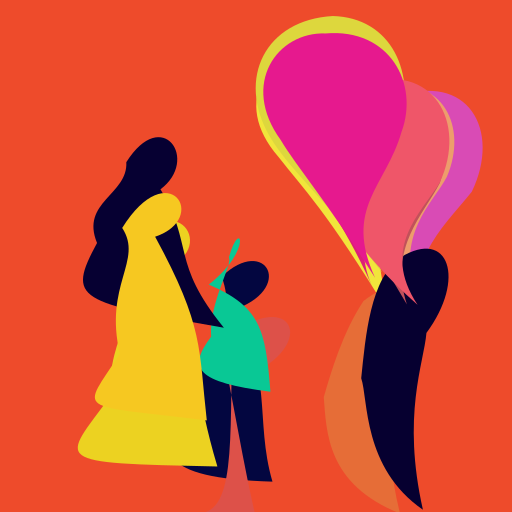} \\

        \multicolumn{5}{c}{``a colorful rooster''} \\
        \includegraphics[width=0.19\linewidth,height=0.19\linewidth]{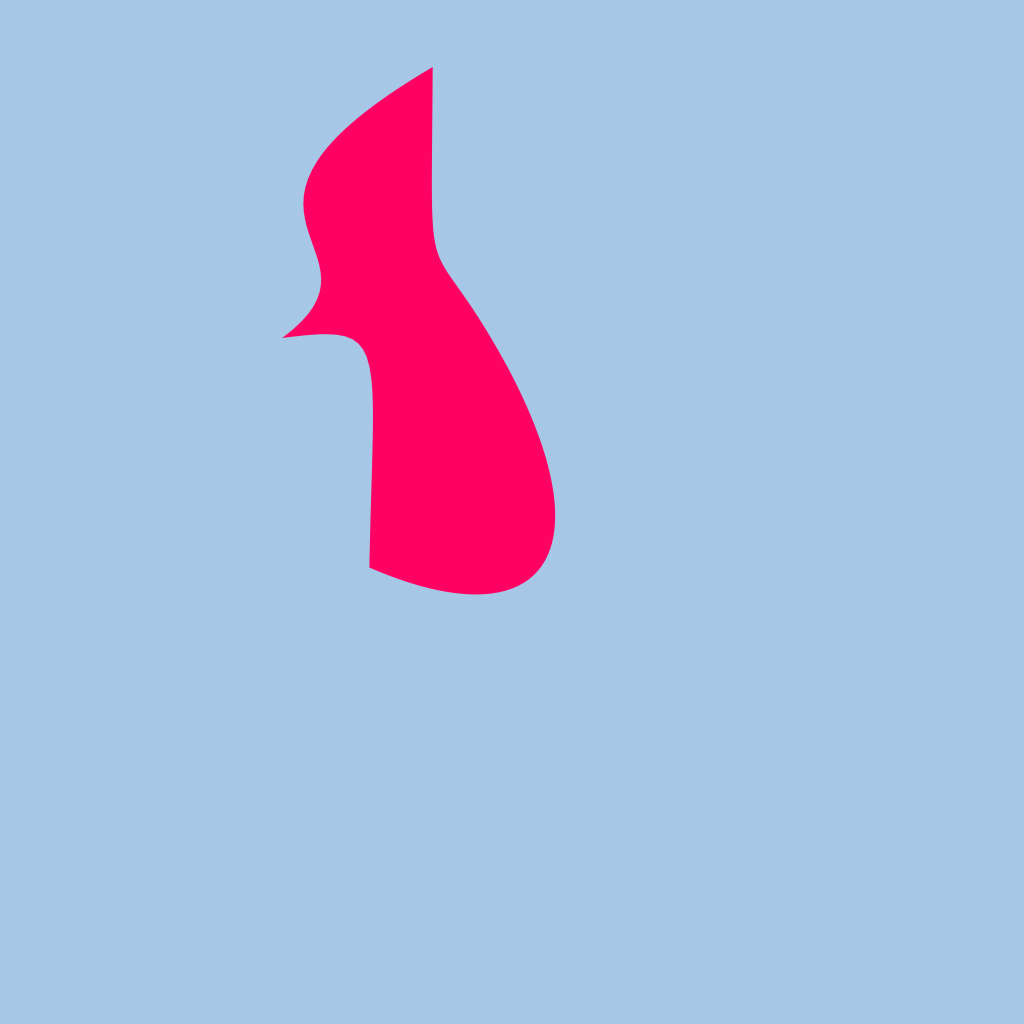} &
        \includegraphics[width=0.19\linewidth,height=0.19\linewidth]{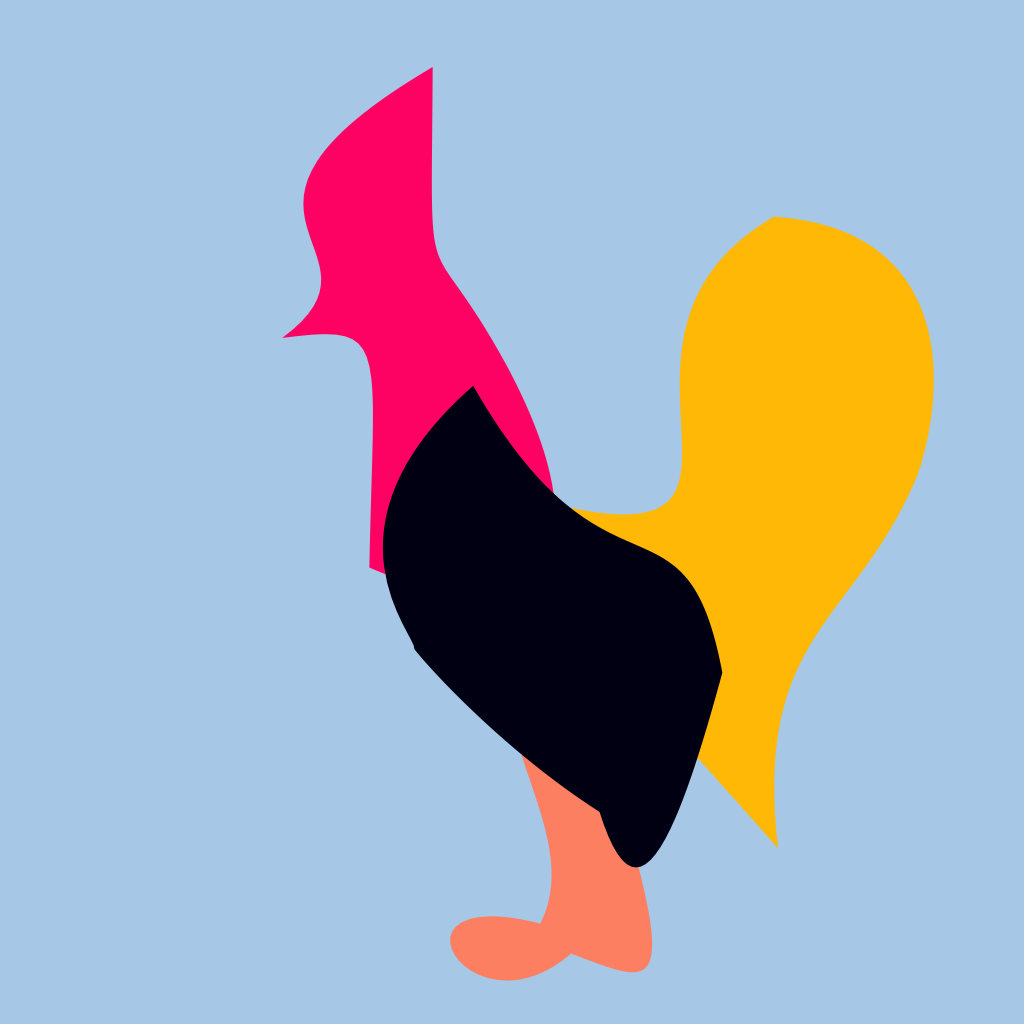} &
        \includegraphics[width=0.19\linewidth,height=0.19\linewidth]{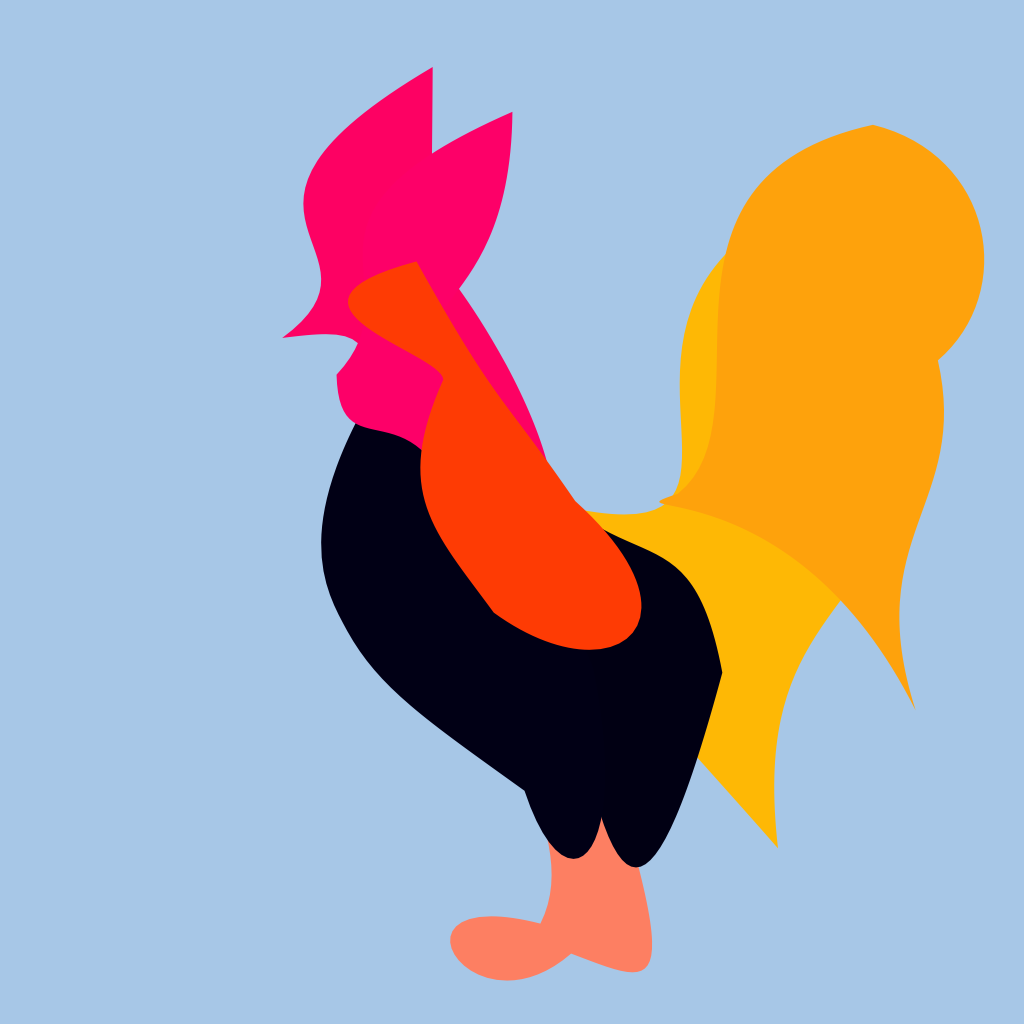} &
        \includegraphics[width=0.19\linewidth,height=0.19\linewidth]{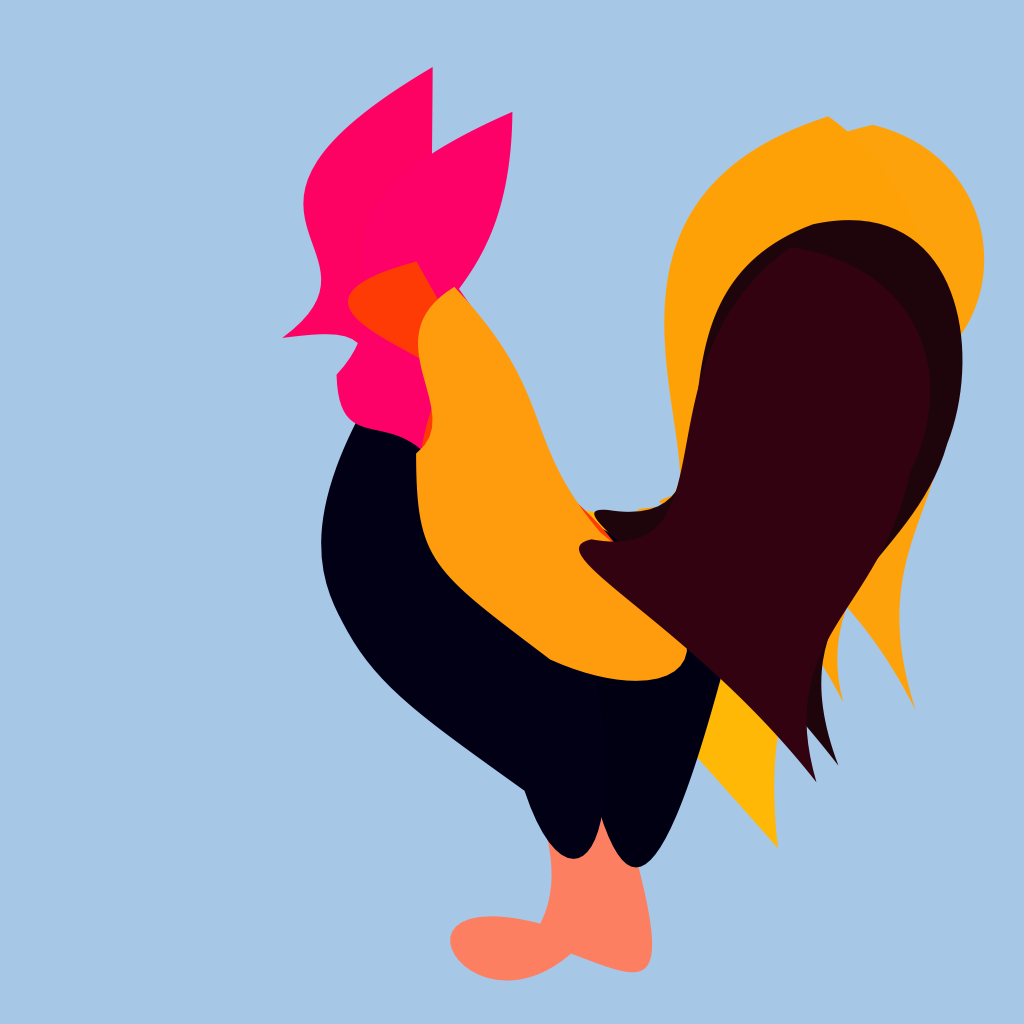} &
        \includegraphics[width=0.19\linewidth,height=0.19\linewidth]{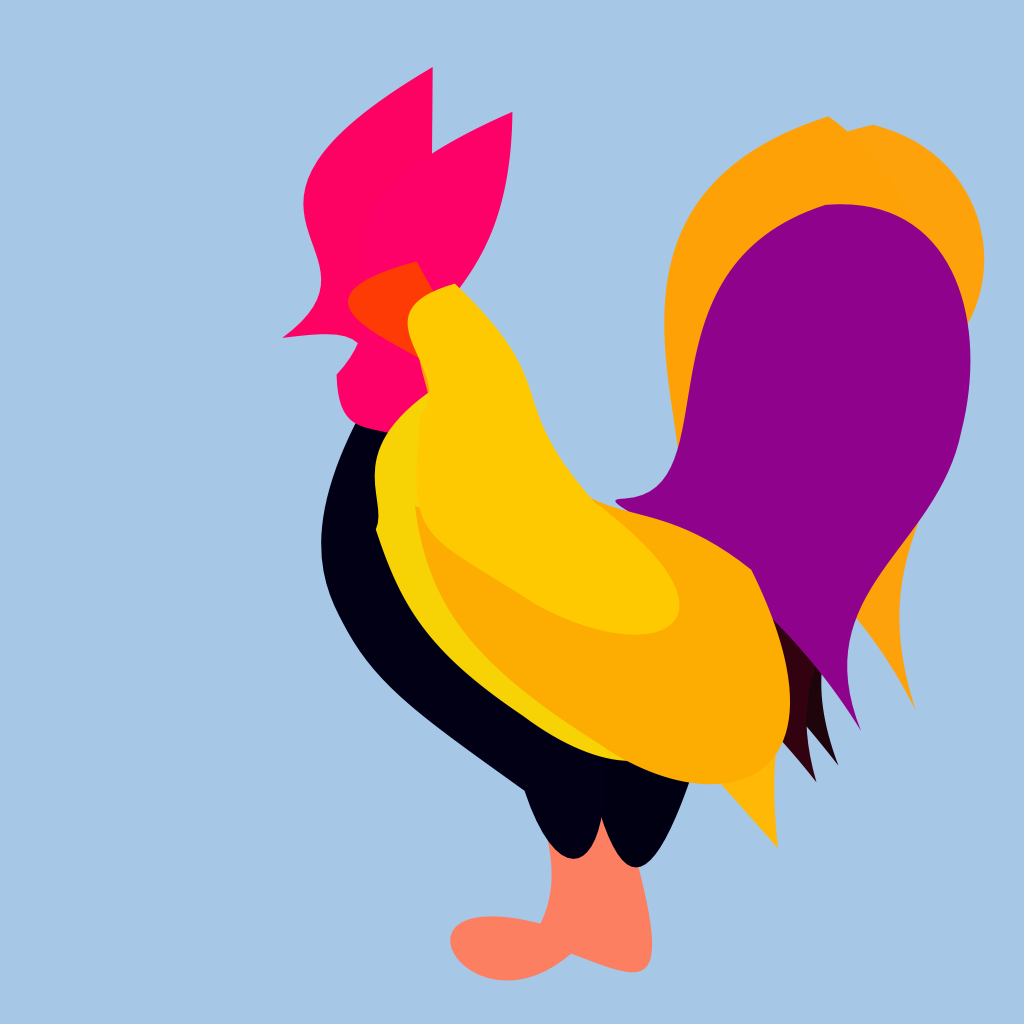} \\

        \multicolumn{5}{c}{``an erupting volcano''} \\
        \includegraphics[width=0.19\linewidth,height=0.19\linewidth]{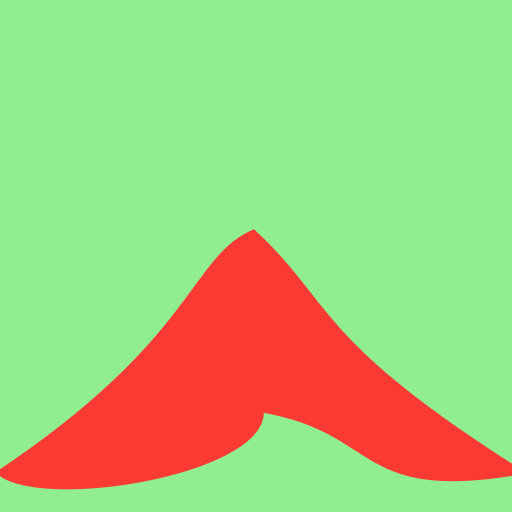} &
        \includegraphics[width=0.19\linewidth,height=0.19\linewidth]{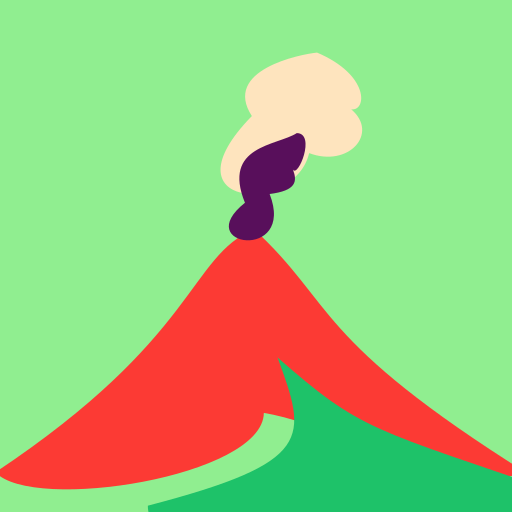} &
        \includegraphics[width=0.19\linewidth,height=0.19\linewidth]{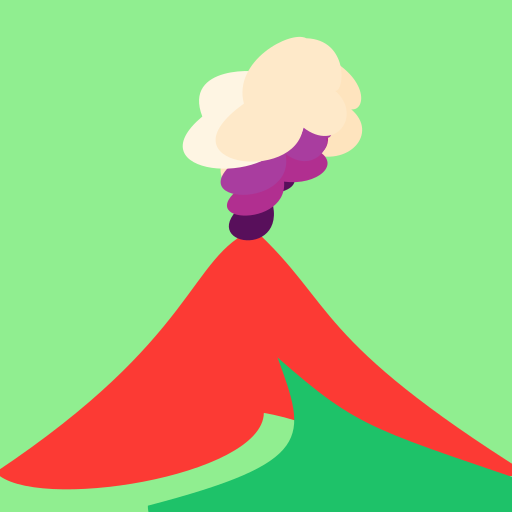} &
        \includegraphics[width=0.19\linewidth,height=0.19\linewidth]{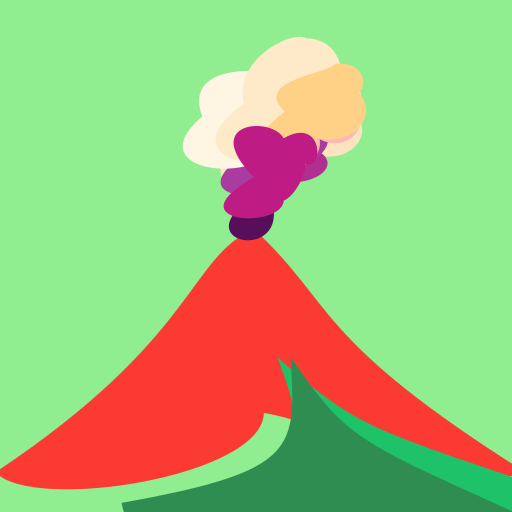} &
        \includegraphics[width=0.19\linewidth,height=0.19\linewidth]{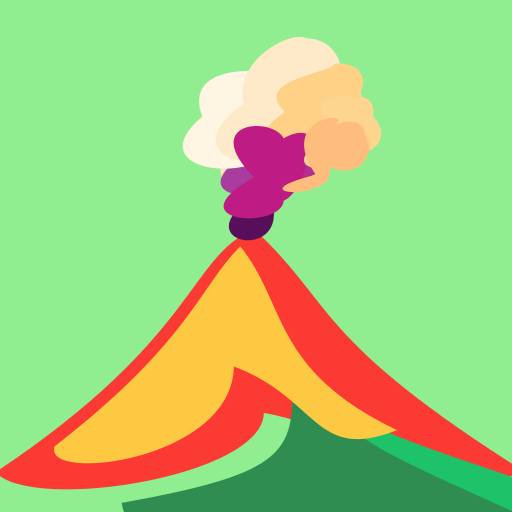} \\

        \multicolumn{5}{c}{``a peacock''} \\
        \includegraphics[width=0.19\linewidth,height=0.19\linewidth]{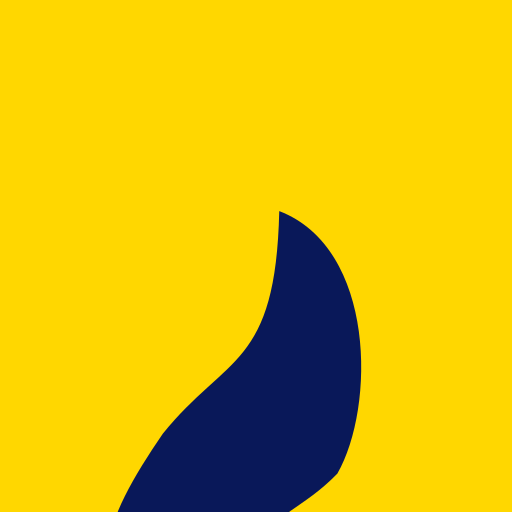} &
        \includegraphics[width=0.19\linewidth,height=0.19\linewidth]{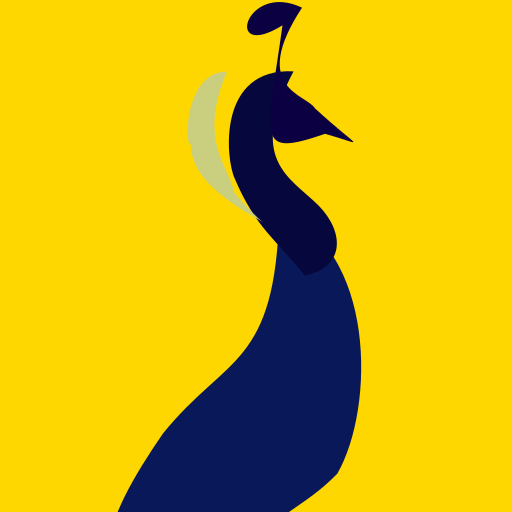} &
        \includegraphics[width=0.19\linewidth,height=0.19\linewidth]{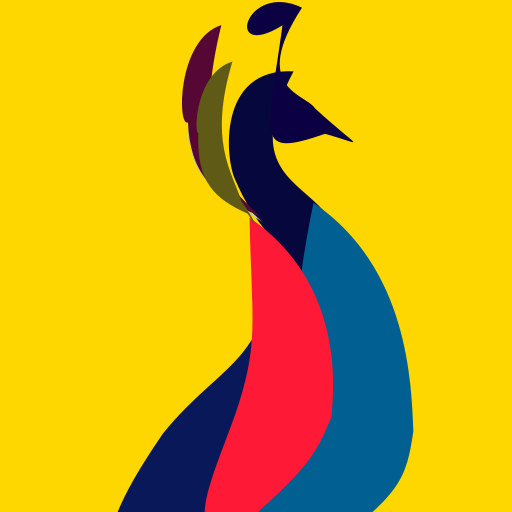} &
        \includegraphics[width=0.19\linewidth,height=0.19\linewidth]{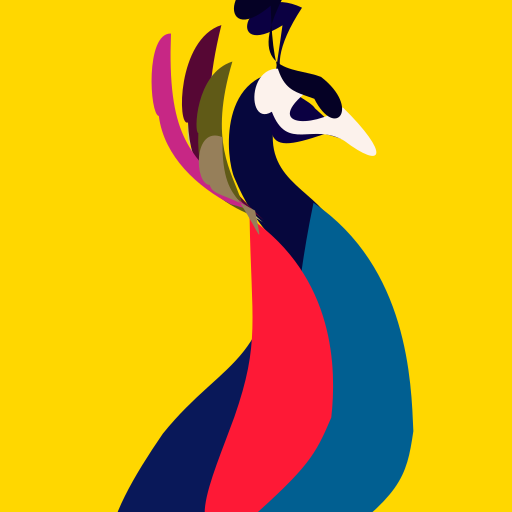} &
        \includegraphics[width=0.19\linewidth,height=0.19\linewidth]{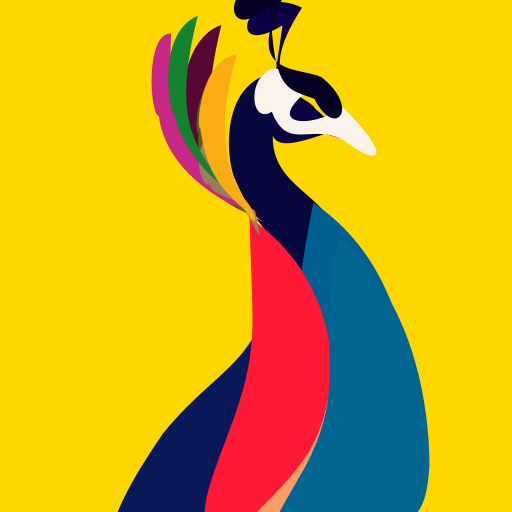} \\

        $1$ & $4$ & $8$ & $12$ & $16$ \\

    \\[-0.7cm]        
    \end{tabular}
    }
    \caption{\textbf{Qualitative Results Obtained with NeuralSVG.} We show results generated by our method when keeping a varying number of learned shapes in the final rendering. Even with a small number of shapes ($<4$), our approach effectively captures the coarse structure of the scene. Moreover, additional shapes progressively introduce finer details in an ordered manner. \\[-0.7cm]
    }
    \label{fig:our_results_dropout}
\end{figure}

In the following section, we demonstrate the effectiveness of NeuralSVG and the appealing properties of our implicit representation.

\paragraph{\textbf{Evaluation Setup}}
We evaluate NeuralSVG with respect to state-of-the-art text-to-vector methods including VectorFusion~\cite{jain2023vectorfusion} initialized using LIVE~\cite{ma2022towards}, SVGDreamer~\cite{xing2024svgdreamer}, NiVEL~\cite{thamizharasan2024nivel}, and Text-to-Vector~\cite{zhang2024text}. 
For our evaluations, we use the set of $128$ prompts from VectorFusion, as this is the only publicly available text-to-vector prompt evaluation set. For each prompt, we generate five SVGs using five different random seeds.
For VectorFusion and SVGDreamer, we evaluate two variants: one using 16 shapes (matching the number of shapes in our method) and another with additional shapes (64 shapes for VectorFusion and 256 for SVGDreamer). We note that SVGDreamer with 512 shapes was not evaluated due to the substantial computational overhead required (over 40GB of VRAM and several hours of runtime on a single A100 GPU).

Furthermore, we note that official implementations for NIVeL and Text-to-Vector are unavailable. Our comparison with these methods is based solely on the visual results reported in their respective papers. 
Finally, unless otherwise specified, we do not apply dropout during inference and output all $16$ shapes learned by NeuralSVG.

\subsection{Qualitative Evaluations and Comparisons}
\paragraph{\textbf{Qualitative Evaluation}}
In~\Cref{fig:our_results_dropout}, we demonstrate text-to-vector results obtained using NeuralSVG. We present outputs generated while retaining a different number of learned shapes in the final rendering: $1$, $4$, $8$, $12$, and all $16$ shapes. The results show that NeuralSVG effectively matches the given prompt even when using only a subset of shapes. Specifically, with just four shapes, the model captures the coarse structure of the scene, such as the outline of the volcano in the fourth row or the body of the peacock in the fifth row. As more shapes are gradually added, the model incorporates finer details in a hierarchical fashion, building upon previously learned shapes. This is most noticeable in the second row where additional people and balloons are gradually added to the complex scene.

\begin{figure}
    \centering
    \setlength{\tabcolsep}{1pt}
    \addtolength{\belowcaptionskip}{-5pt}
    {\small
    \begin{tabular}{c c c c c}

        \multicolumn{5}{c}{``a fox playing the cello''} \\
        \includegraphics[height=0.09\textwidth,width=0.09\textwidth]{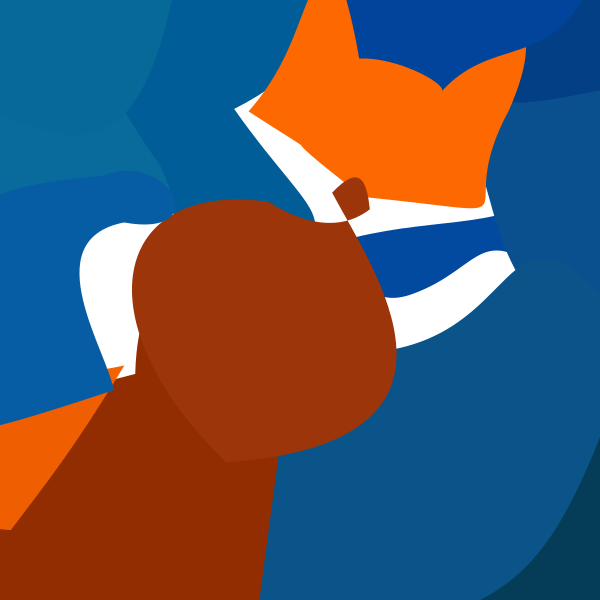} &
        \includegraphics[height=0.09\textwidth,width=0.09\textwidth]{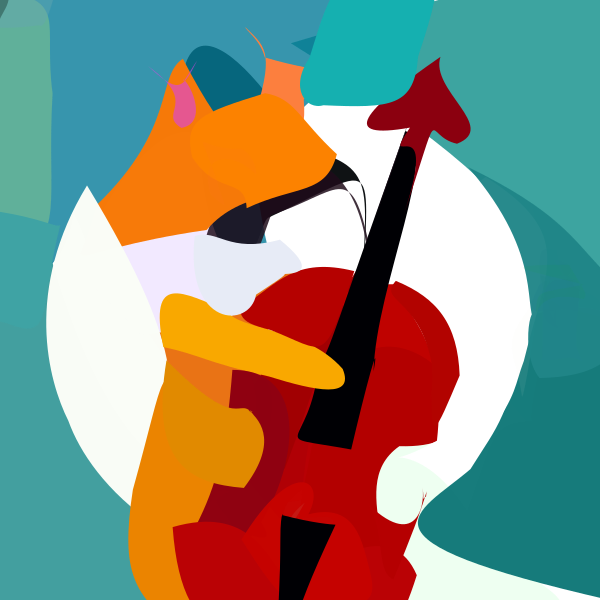} &
        \includegraphics[height=0.09\textwidth,width=0.09\textwidth]{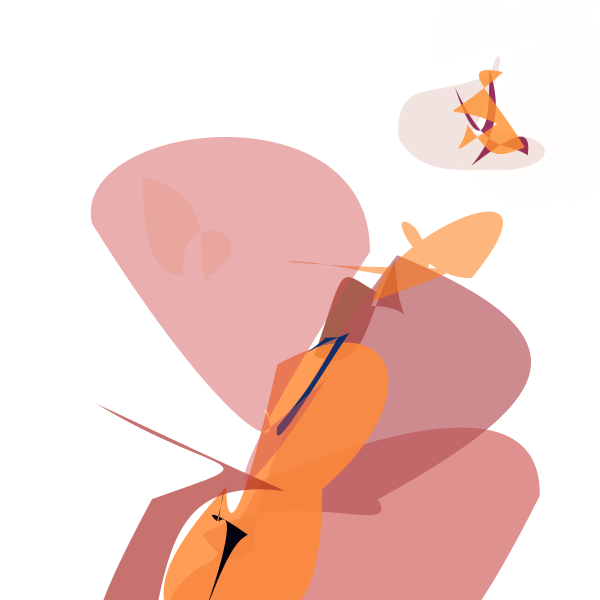} &
        \includegraphics[height=0.09\textwidth,width=0.09\textwidth]{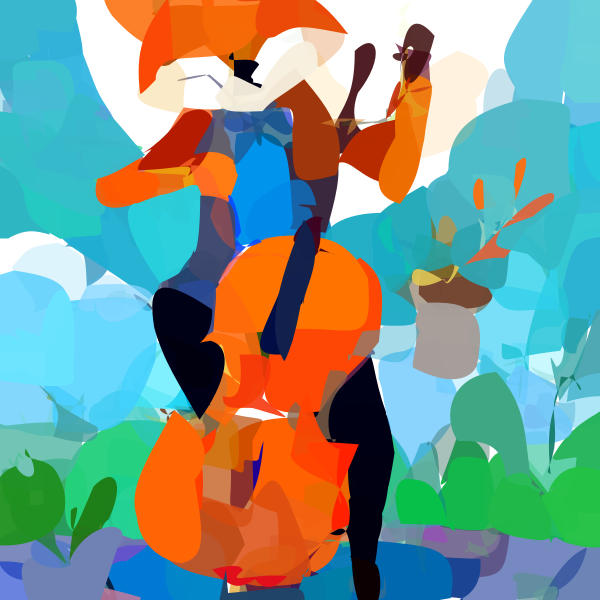} &
        \includegraphics[height=0.09\textwidth,width=0.09\textwidth]{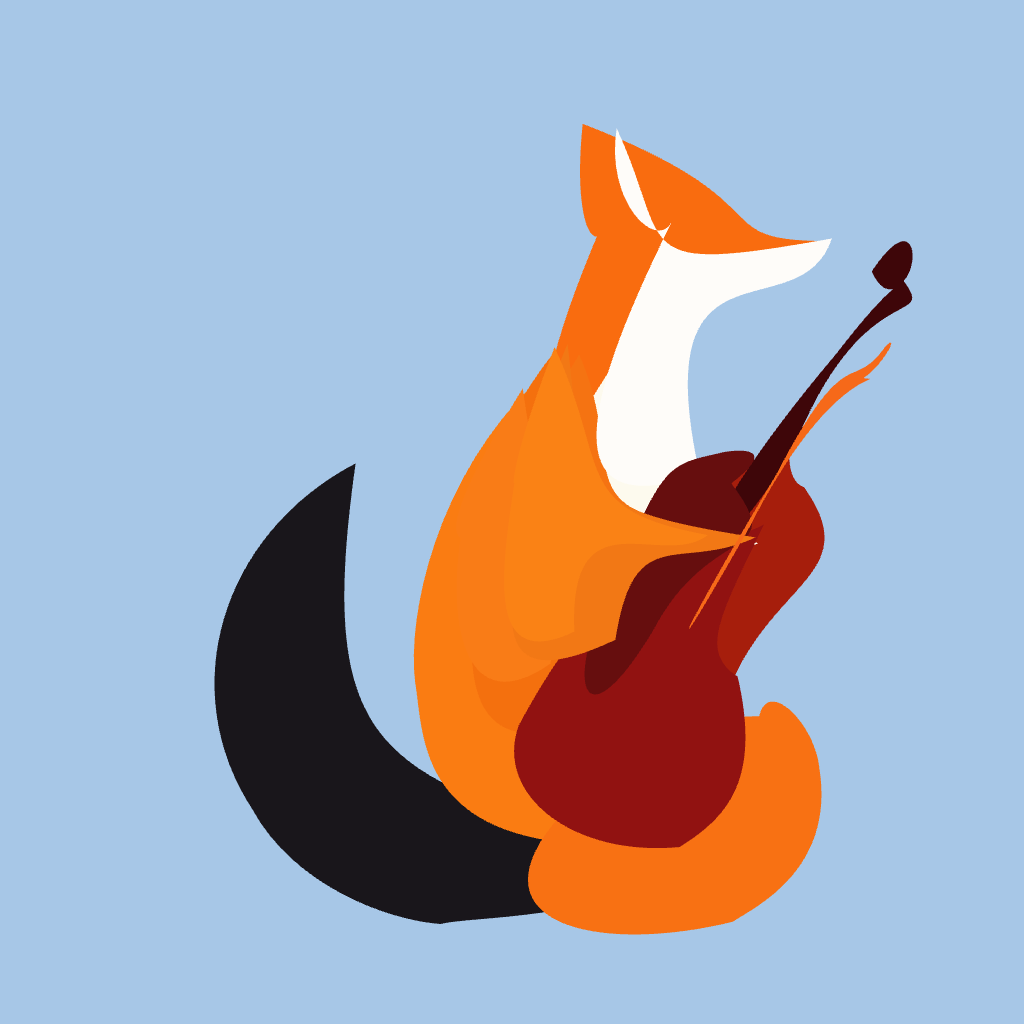} \\

        \multicolumn{5}{c}{``a child unraveling a roll of toilet paper''} \\
        \includegraphics[height=0.09\textwidth,width=0.09\textwidth]{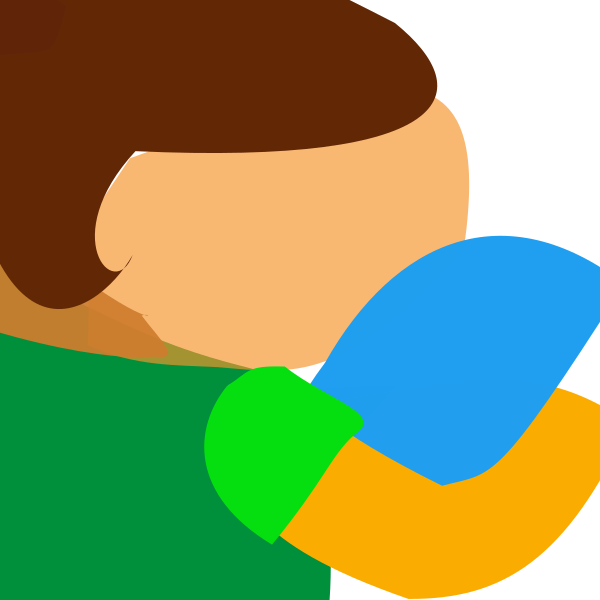} &
        \includegraphics[height=0.09\textwidth,width=0.09\textwidth]{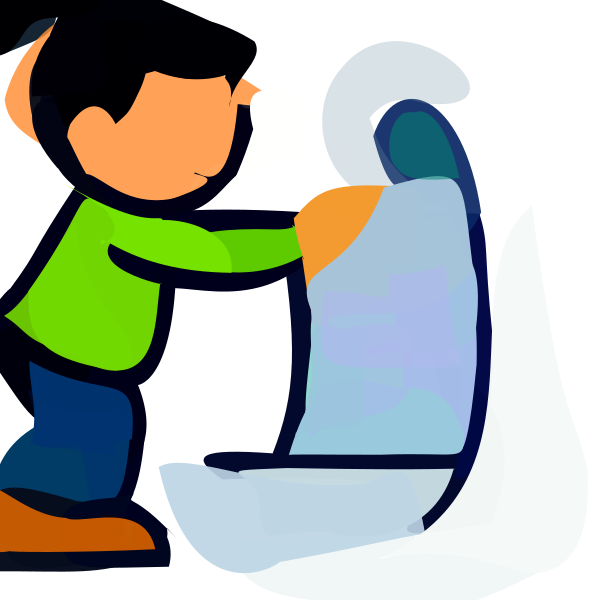} &
        \includegraphics[height=0.09\textwidth,width=0.09\textwidth]{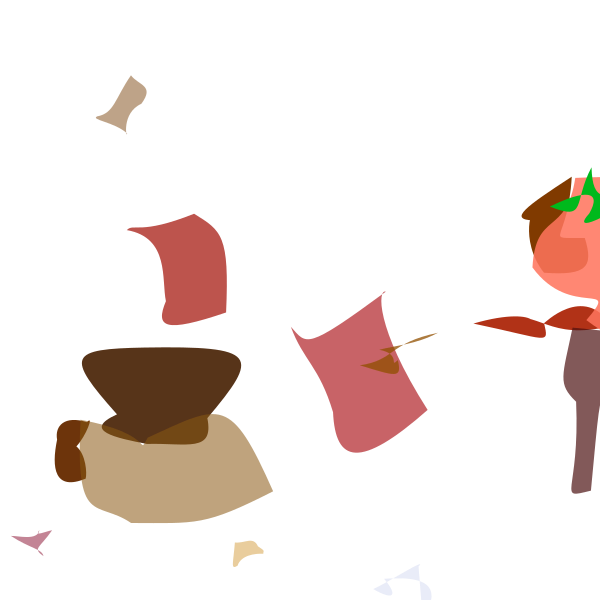} &
        \includegraphics[height=0.09\textwidth,width=0.09\textwidth]{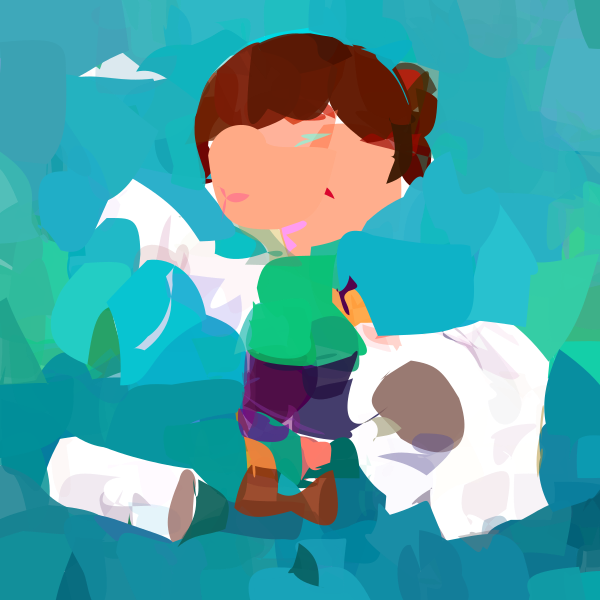} &
        \includegraphics[height=0.09\textwidth,width=0.09\textwidth]{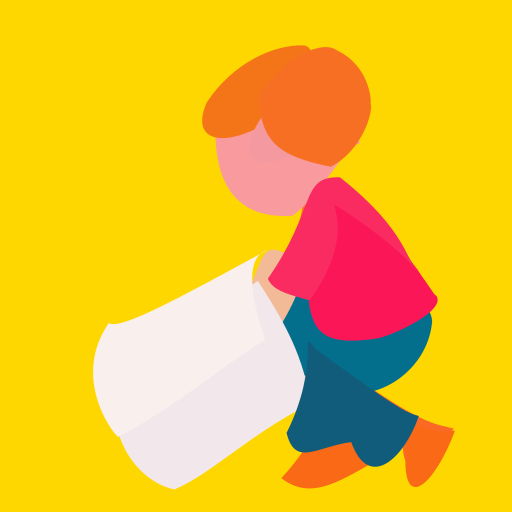} \\

        \multicolumn{5}{c}{``a wolf howling on top of the hill, with a full moon in the sky''} \\
        \includegraphics[height=0.09\textwidth,width=0.09\textwidth]{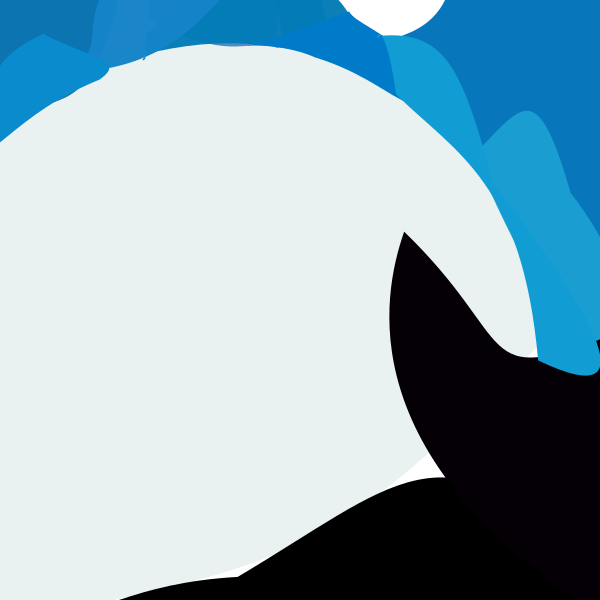} &
        \includegraphics[height=0.09\textwidth,width=0.09\textwidth]{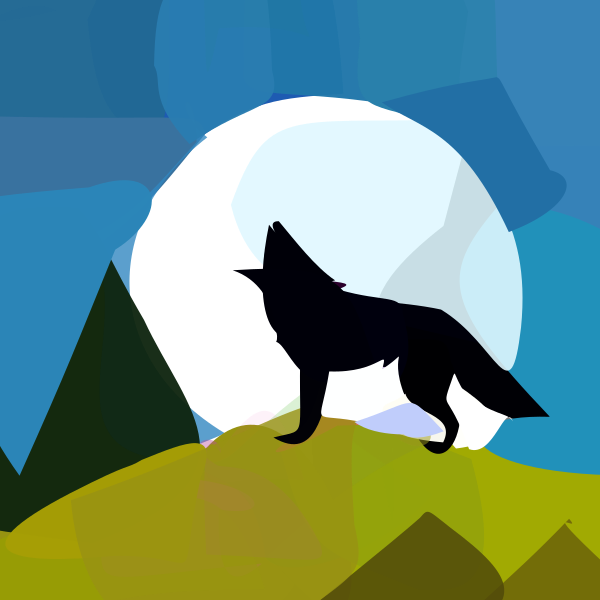} &
        \includegraphics[height=0.09\textwidth,width=0.09\textwidth]{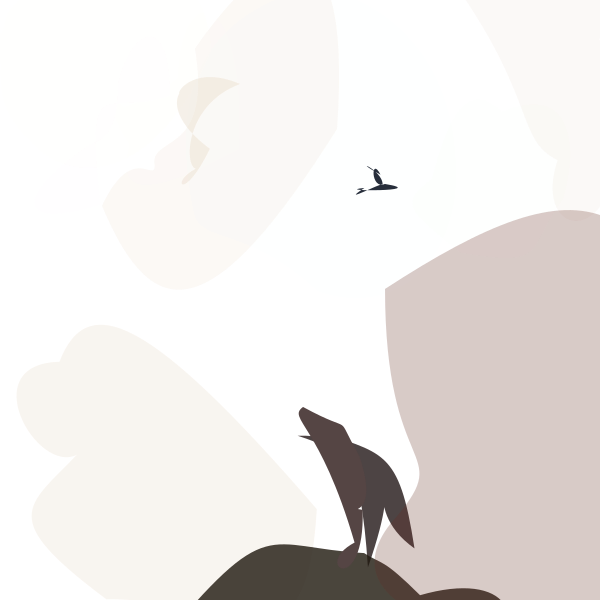} &
        \includegraphics[height=0.09\textwidth,width=0.09\textwidth]{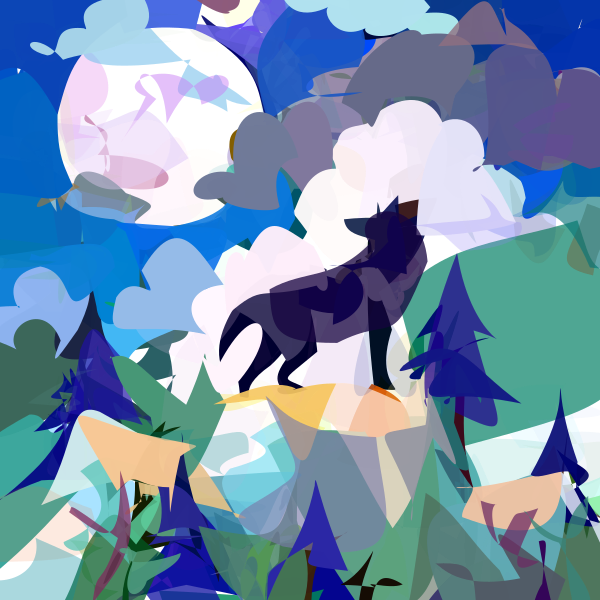} &
        \includegraphics[height=0.09\textwidth,width=0.09\textwidth]{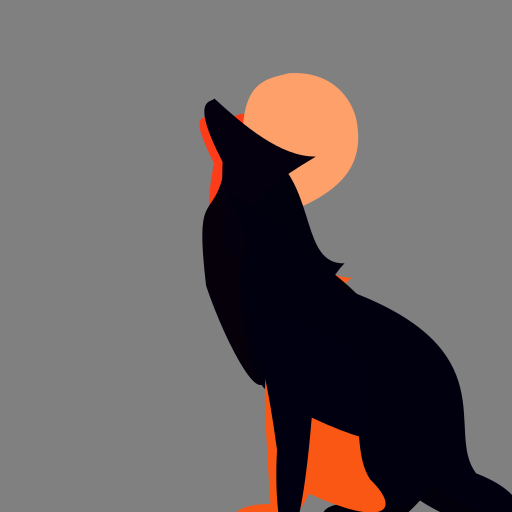} \\

        \multicolumn{5}{c}{``a rabbit cutting grass with a lawnmower''} \\
        \includegraphics[height=0.09\textwidth,width=0.09\textwidth]{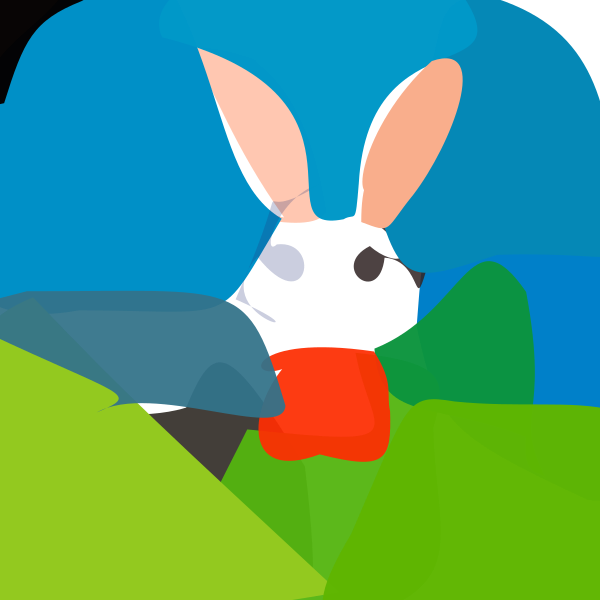} &
        \includegraphics[height=0.09\textwidth,width=0.09\textwidth]{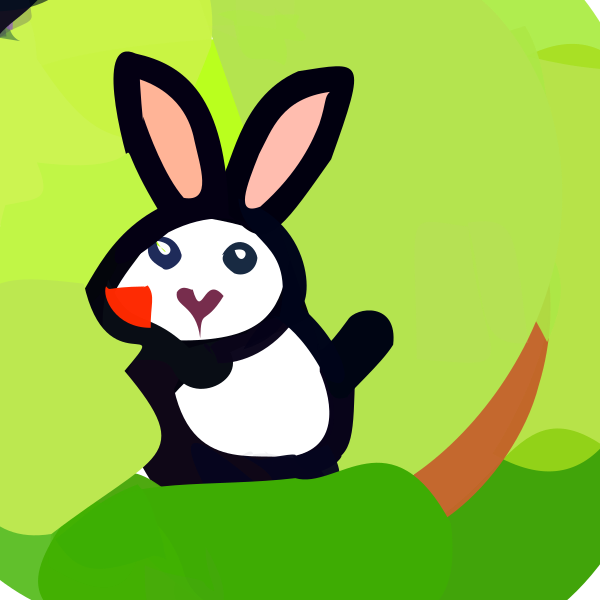} &
        \includegraphics[height=0.09\textwidth,width=0.09\textwidth]{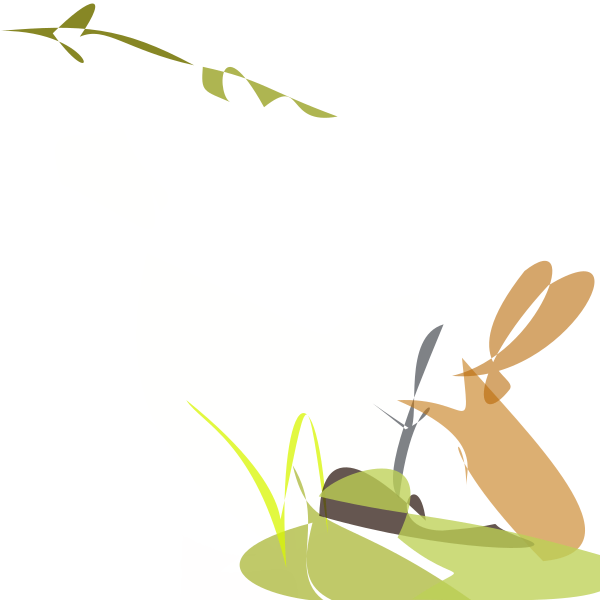} &
        \includegraphics[height=0.09\textwidth,width=0.09\textwidth]{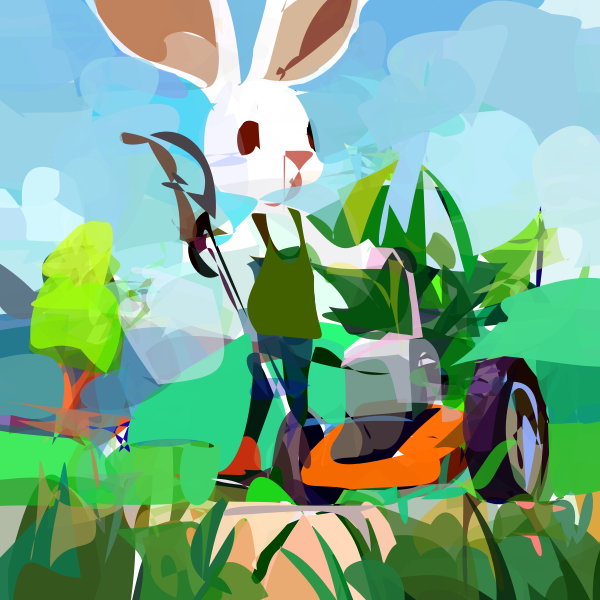} &
        \includegraphics[height=0.09\textwidth,width=0.09\textwidth]{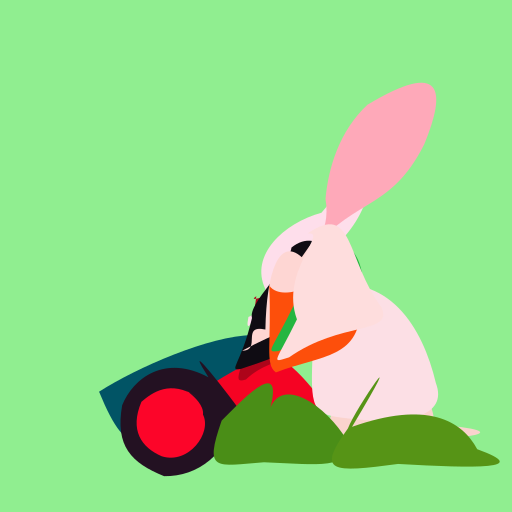} \\
        
        \multicolumn{5}{c}{``a yeti taking a selfie''} \\
        \includegraphics[height=0.09\textwidth,width=0.09\textwidth]{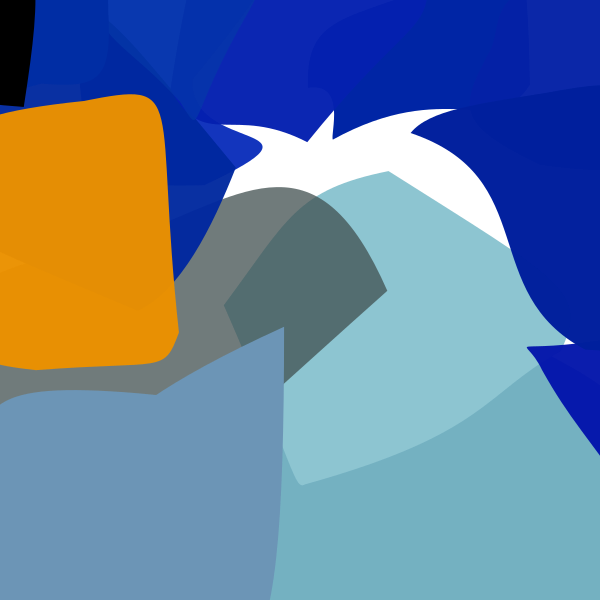} &
        \includegraphics[height=0.09\textwidth,width=0.09\textwidth]{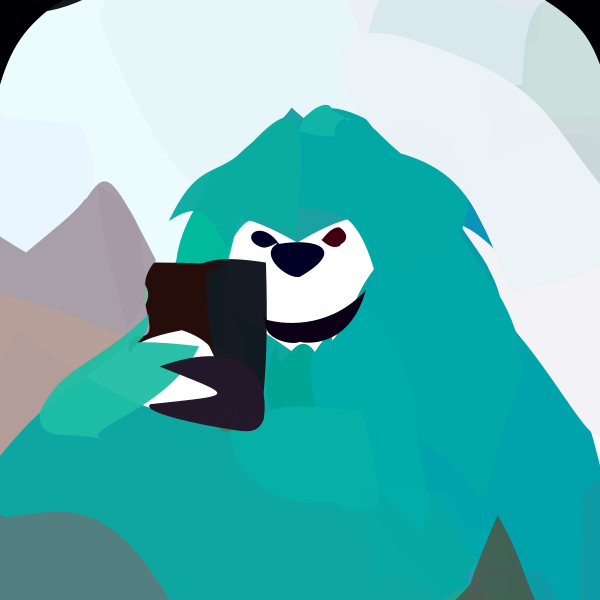} &
        \includegraphics[height=0.09\textwidth,width=0.09\textwidth]{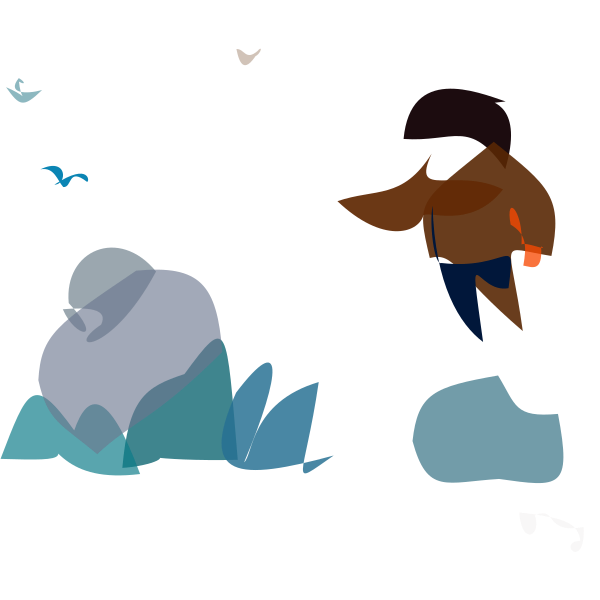} &
        \includegraphics[height=0.09\textwidth,width=0.09\textwidth]{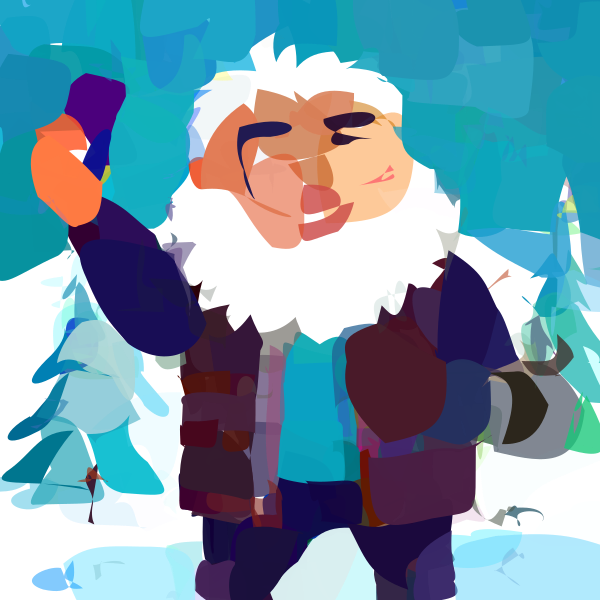} &
        \includegraphics[height=0.09\textwidth,width=0.09\textwidth]{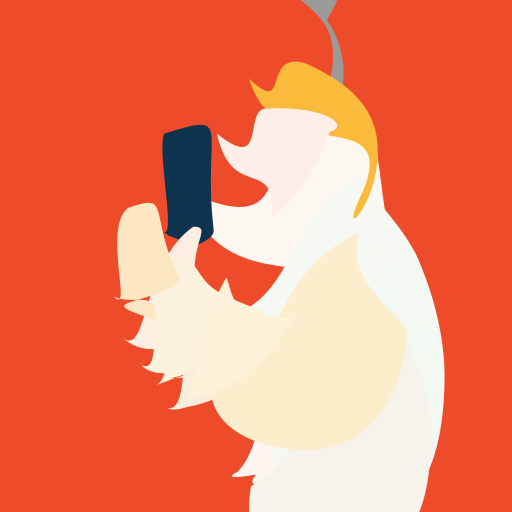} \\

        \begin{tabular}{c} \textbf{VectorFusion} \\ \textbf{(16 Shapes)} \end{tabular} &
        \begin{tabular}{c} VectorFusion \\ (64 Shapes) \end{tabular} &
        \begin{tabular}{c} \textbf{SVGDreamer}   \\ \textbf{(16 Shapes)} \end{tabular} &
        \begin{tabular}{c} SVGDreamer   \\ (256 Shapes) \end{tabular} &
        \begin{tabular}{c} \textbf{Ours}  \\ \textbf{(16 Shapes)} \end{tabular} \\
        
    \\[-0.5cm]        
    \end{tabular}
    }
    \caption{\textbf{Qualitative Comparisons.} Visual comparisons to VectorFusion~\cite{jain2023vectorfusion} and SVGDreamer~\cite{xing2024svgdreamer} using a varying number of shapes.}
    \label{fig:comparisons_vectorfusion_svgdream}
\end{figure}
\begin{figure}
    \centering
    \setlength{\tabcolsep}{1pt}
    \addtolength{\belowcaptionskip}{-5pt}
    {\small
    \begin{tabular}{c c c c c}

        \multicolumn{5}{c}{``a fox playing the cello''} \\
        \includegraphics[height=0.09\textwidth,width=0.09\textwidth]{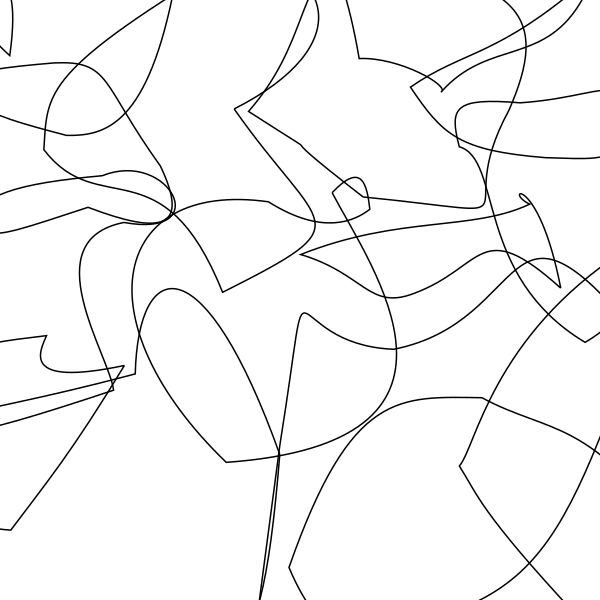} &
        \includegraphics[height=0.09\textwidth,width=0.09\textwidth]{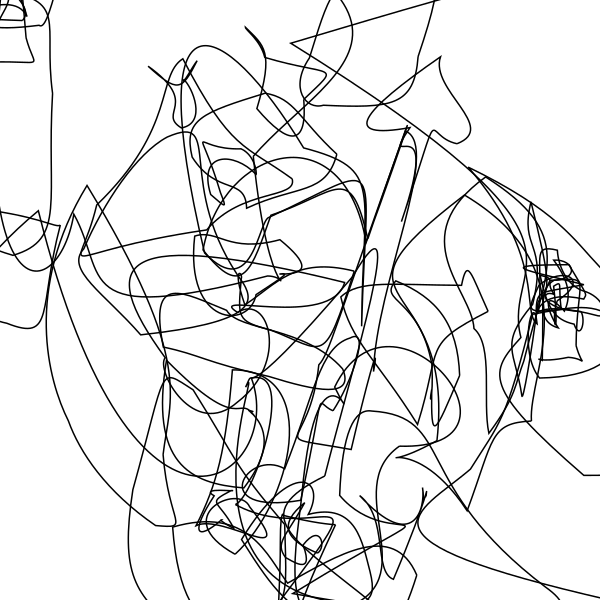} &
        \includegraphics[height=0.09\textwidth,width=0.09\textwidth]{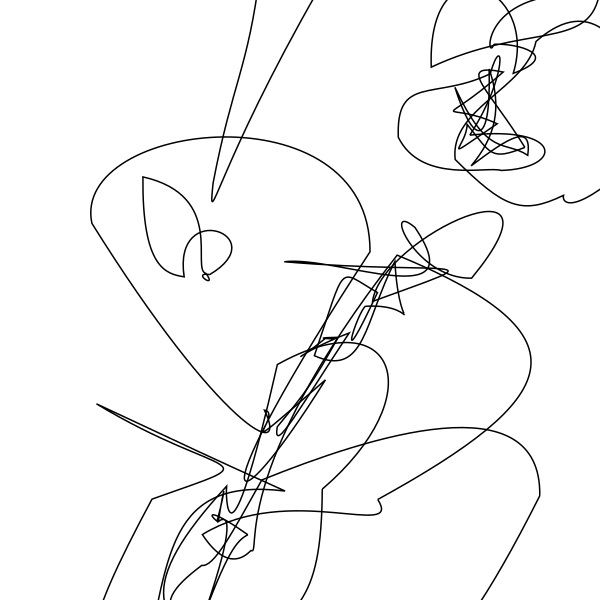} &
        \includegraphics[height=0.09\textwidth,width=0.09\textwidth]{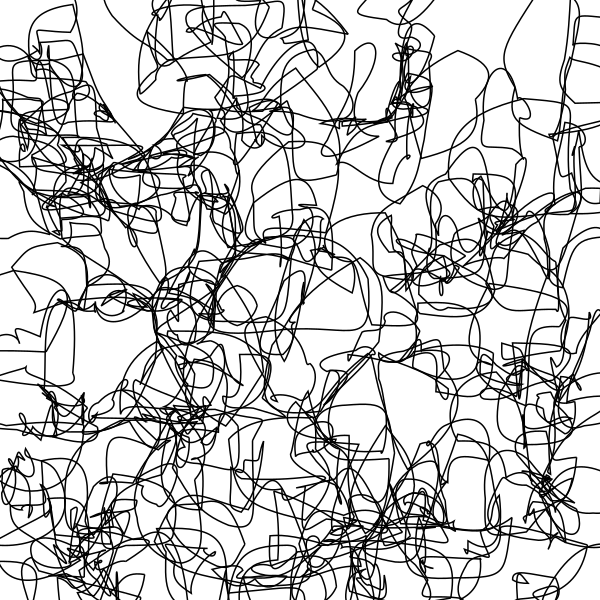} &
        \includegraphics[height=0.09\textwidth,width=0.09\textwidth]{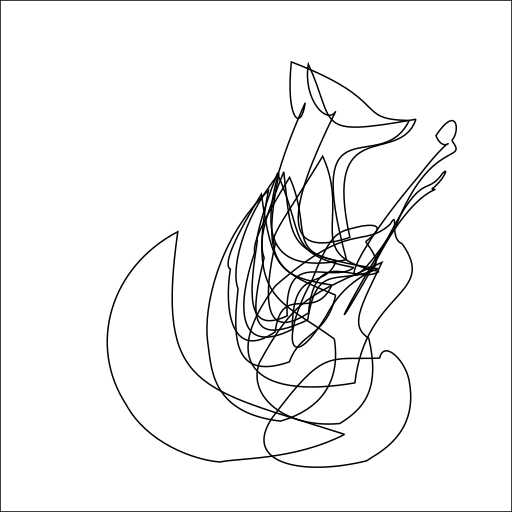} \\

        \multicolumn{5}{c}{``a child unraveling a roll of toilet paper''} \\
        \includegraphics[height=0.09\textwidth,width=0.09\textwidth]{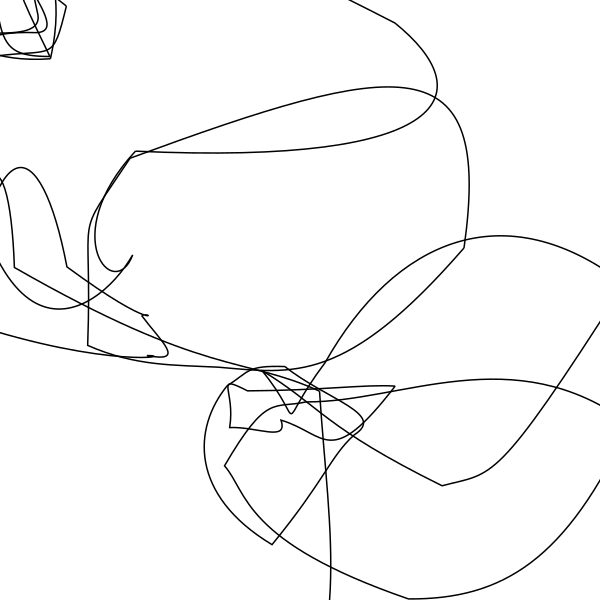} &
        \includegraphics[height=0.09\textwidth,width=0.09\textwidth]{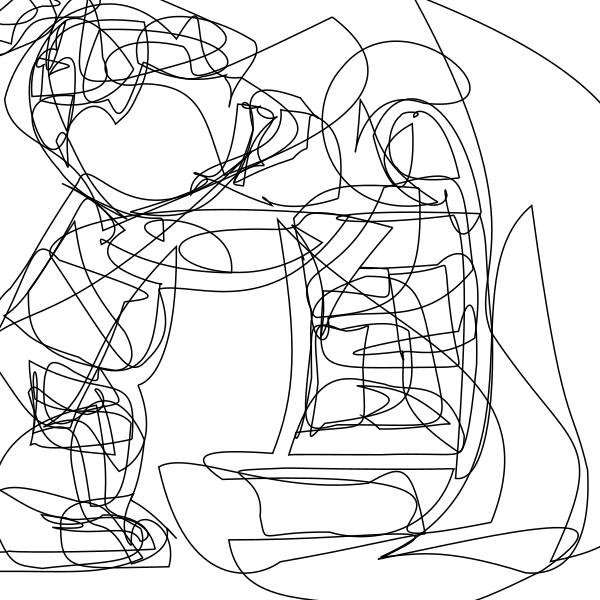} &
        \includegraphics[height=0.09\textwidth,width=0.09\textwidth]{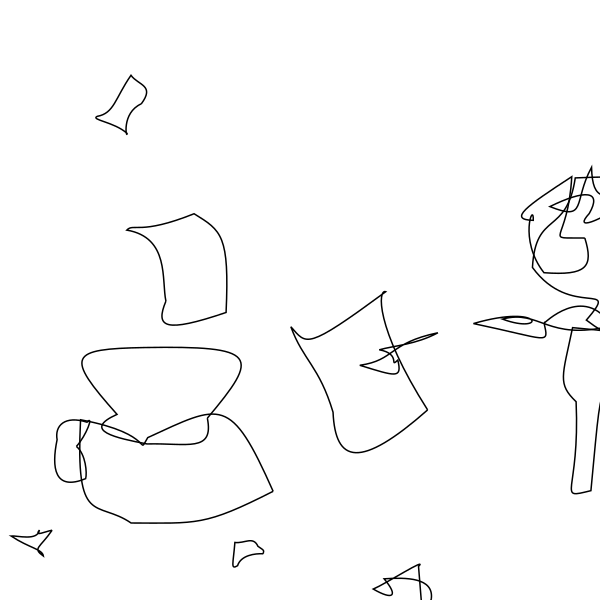} &
        \includegraphics[height=0.09\textwidth,width=0.09\textwidth]{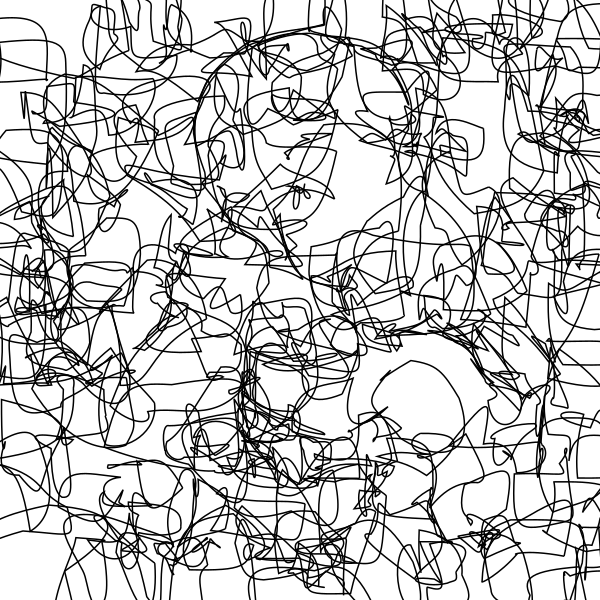} &
        \includegraphics[height=0.09\textwidth,width=0.09\textwidth]{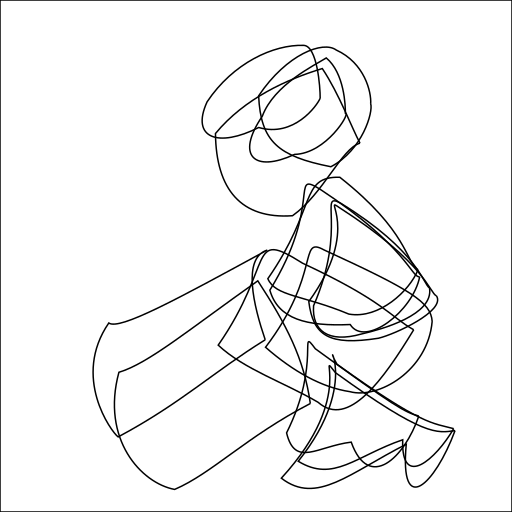} \\

        \multicolumn{5}{c}{``a wolf howling on top of the hill, with a full moon in the sky''} \\               
        \includegraphics[height=0.09\textwidth,width=0.09\textwidth]{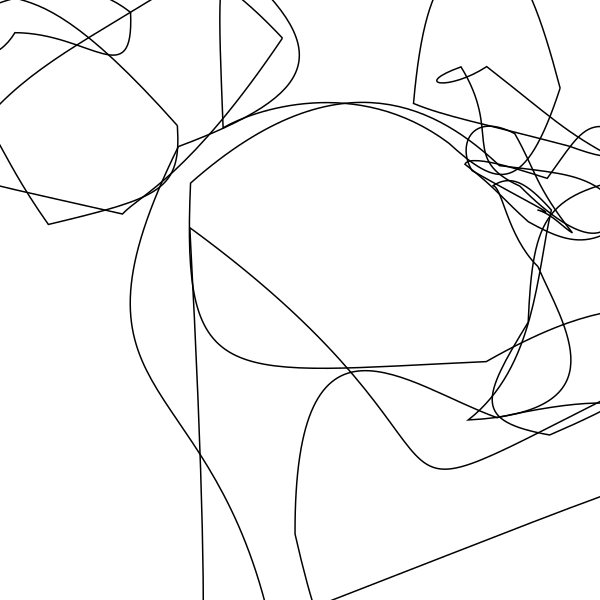} &
        \includegraphics[height=0.09\textwidth,width=0.09\textwidth]{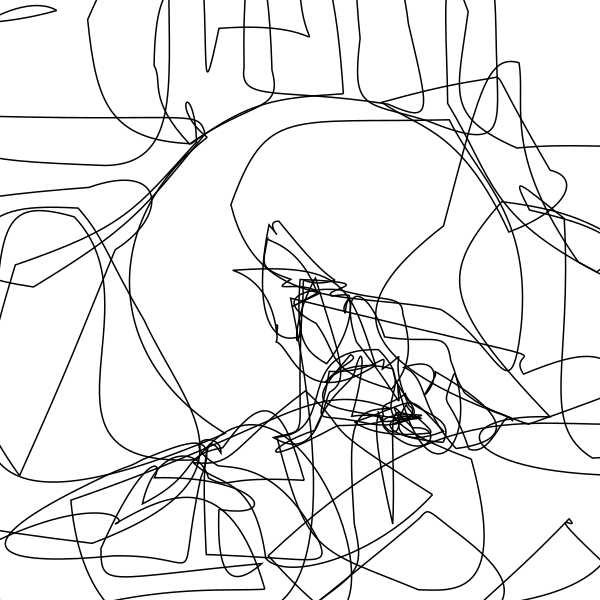} &
        \includegraphics[height=0.09\textwidth,width=0.09\textwidth]{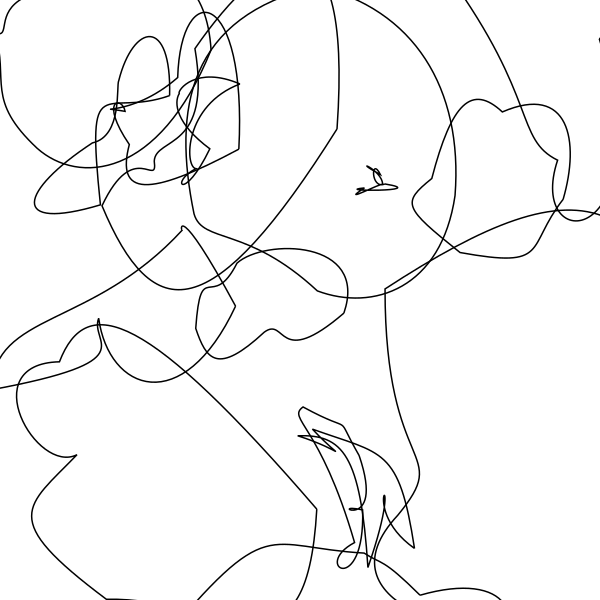} &
        \includegraphics[height=0.09\textwidth,width=0.09\textwidth]{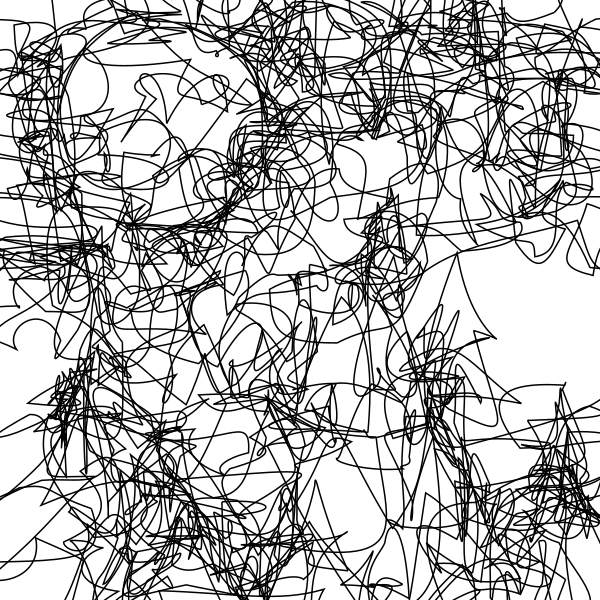} &
        \includegraphics[height=0.09\textwidth,width=0.09\textwidth]{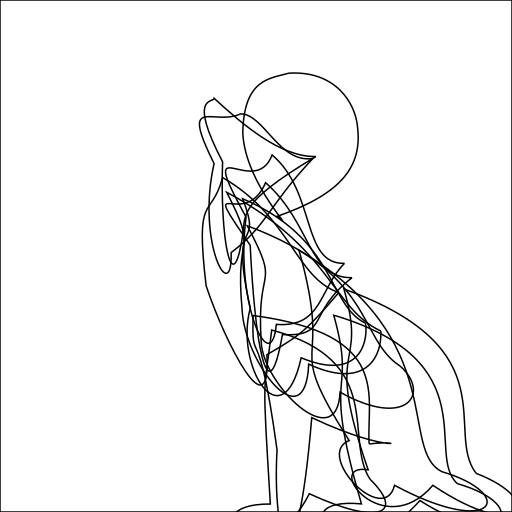} \\

        \begin{tabular}{c} VectorFusion \\ (16 Shapes) \end{tabular} &
        \begin{tabular}{c} VectorFusion \\ (64 Shapes) \end{tabular} &
        \begin{tabular}{c} SVGDreamer   \\ (16 Shapes) \end{tabular} &
        \begin{tabular}{c} SVGDreamer   \\ (256 Shapes) \end{tabular} &
        \begin{tabular}{c} \textbf{NeuralSVG}  \\ \textbf{(16 Shapes)} \end{tabular} \\
        
    \\[-0.5cm]        
    \end{tabular}
    }
    \caption{\textbf{Shape Outlines of the Generated SVGs.} We present the outlines of SVGs generated by NeuralSVG, VectorFusion, and SVGDreamer. The alternative methods often produce nearly pixel-like shapes that are difficult to modify manually. In contrast, NeuralSVG generates cleaner SVGs, making them more editable and practical.}
    \label{fig:outlines_comparisons}
\end{figure}

\vspace{-0.1cm}
\paragraph{\textbf{Qualitative Comparisons}}
In~\Cref{fig:comparisons_vectorfusion_svgdream}, we compare NeuralSVG to state-of-the-art open-source methods, VectorFusion and SVGDreamer. When constrained to the same number of shapes (16), both VectorFusion and SVGDreamer struggle to faithfully represent the desired scene, often missing critical details from the prompt. 
With increased shape counts --- $64$ for VectorFusion and $256$ for SVGDreamer --- the methods generate more detailed SVGs that better align with the prompt but exhibit noticeable artifacts. More importantly, both baselines produce uninterpretable and uneditable shapes, limiting their practical usability. We further highlight this redundancy in~\Cref{fig:outlines_comparisons}, where we show the outlines of the learned shapes for the results shown here. 
In contrast, using only $16$ shapes, NeuralSVG achieves high-quality results that adhere closely to the prompt, maintain smooth contours, and minimize artifacts, providing users with a more practical result. 

\begin{figure}
    \centering
    \setlength{\tabcolsep}{1pt}
    \addtolength{\belowcaptionskip}{-10pt}
    {\small
    \begin{tabular}{c c @{\hspace{0.2cm}} c c}

        \multicolumn{2}{c}{``a walrus smoking a pipe''} &
        \multicolumn{2}{c}{``a crown''} \\
        \includegraphics[height=0.0975\textwidth,width=0.0975\textwidth]{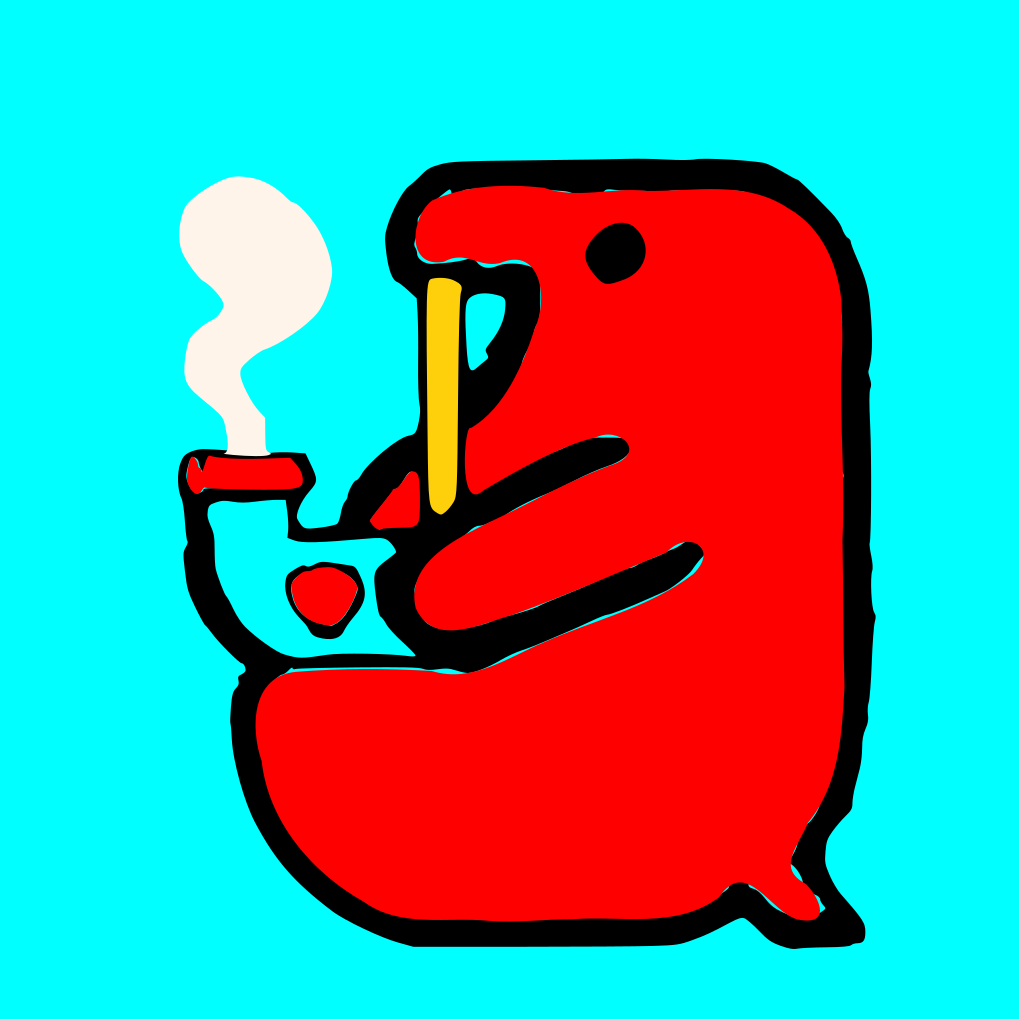} &
        \includegraphics[height=0.0975\textwidth,width=0.0975\textwidth]{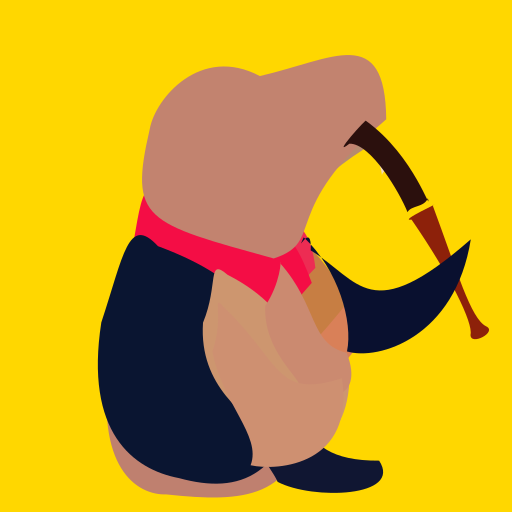} &
        \includegraphics[height=0.0975\textwidth,width=0.0975\textwidth]{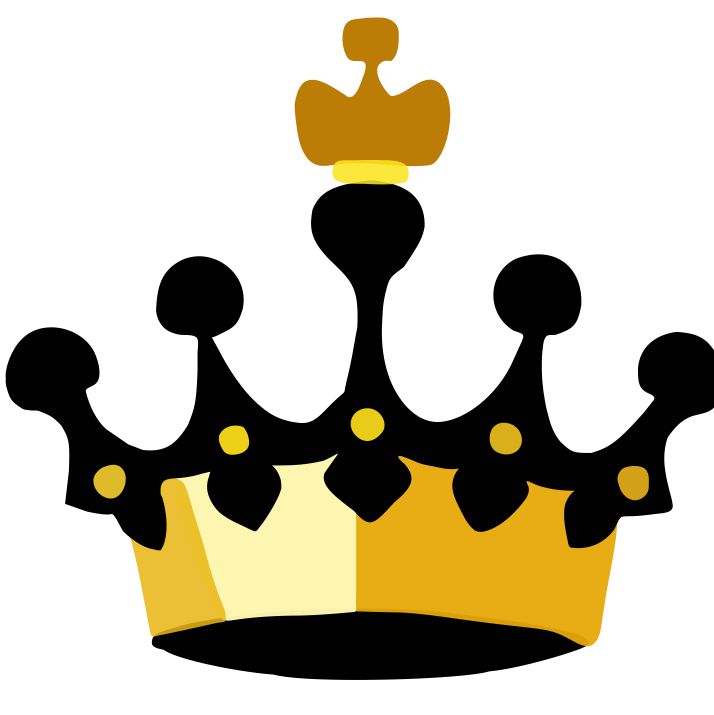} &
        \includegraphics[height=0.0975\textwidth,width=0.0975\textwidth]{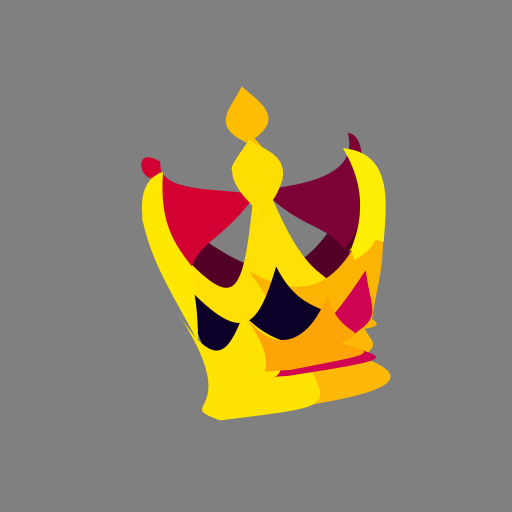} \\

        \multicolumn{2}{c}{``a spaceship''} &
        \multicolumn{2}{c}{``a Japanese sakura tree on a hill''} \\
        \includegraphics[height=0.0975\textwidth,width=0.0975\textwidth]{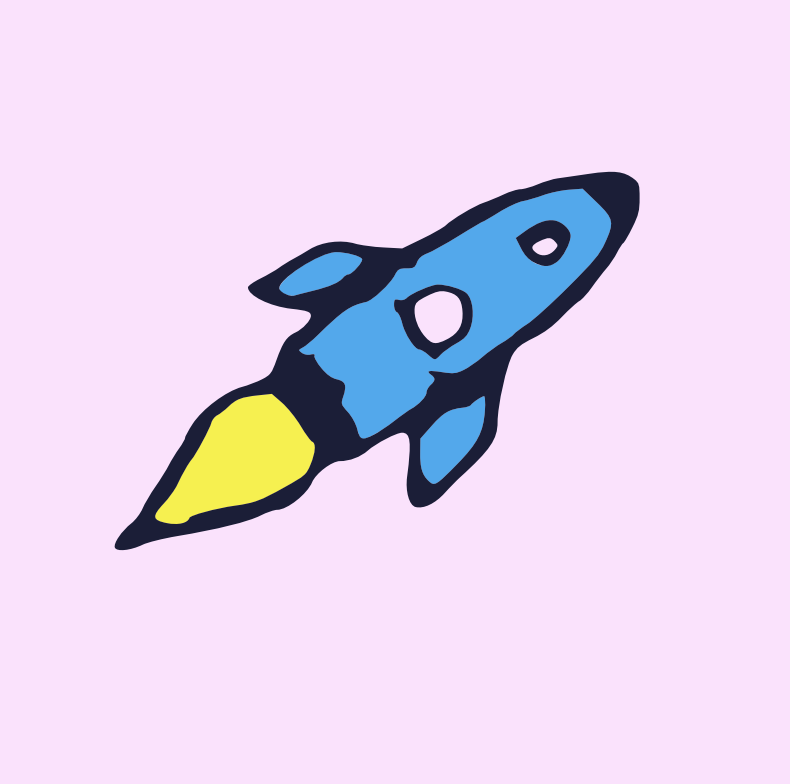} &
        \includegraphics[height=0.0975\textwidth,width=0.0975\textwidth]{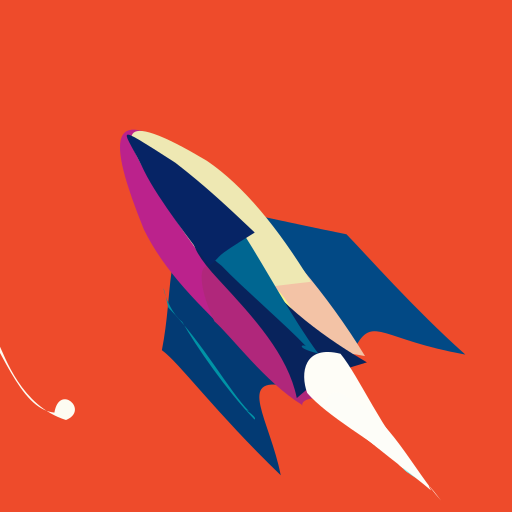} &
        \includegraphics[height=0.0975\textwidth,width=0.0975\textwidth]{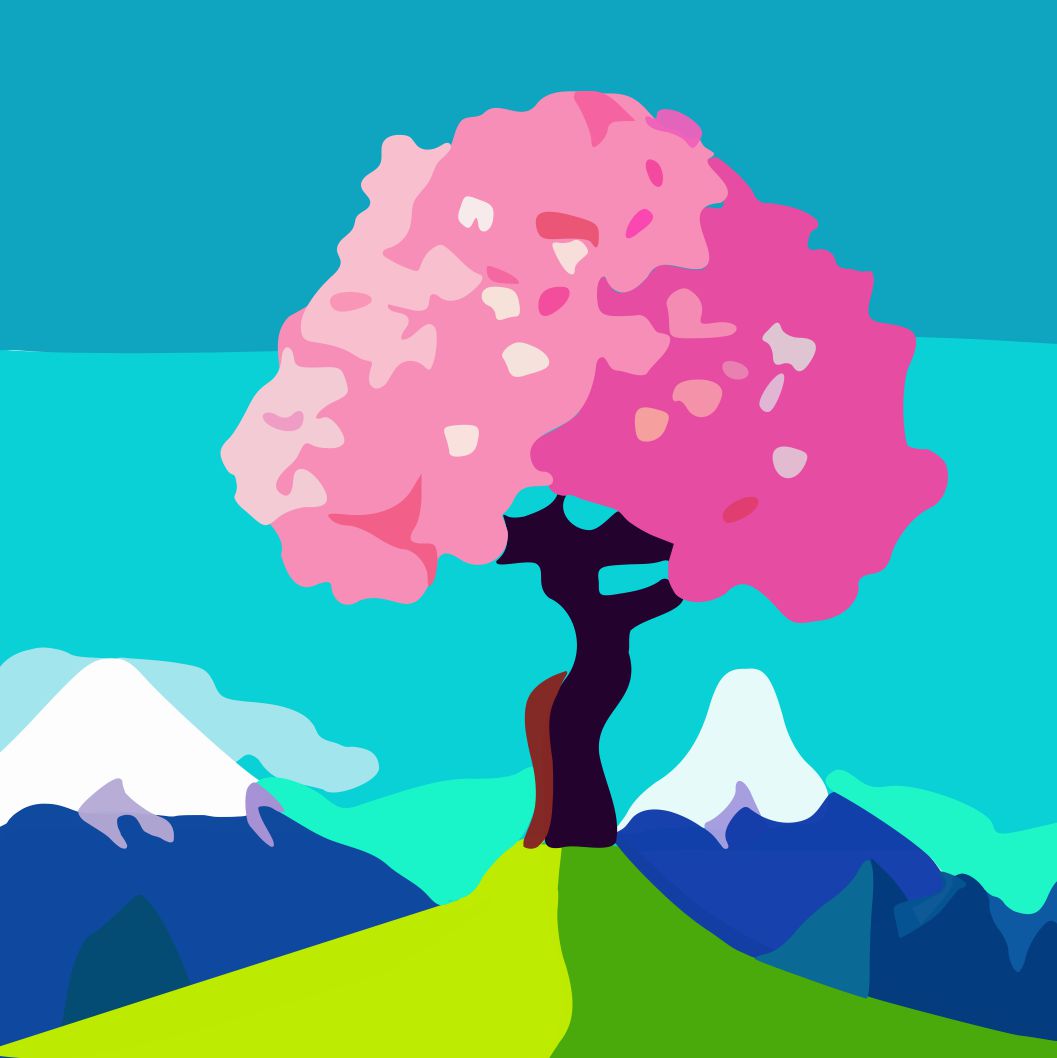} &
        \includegraphics[height=0.0975\textwidth,width=0.0975\textwidth]{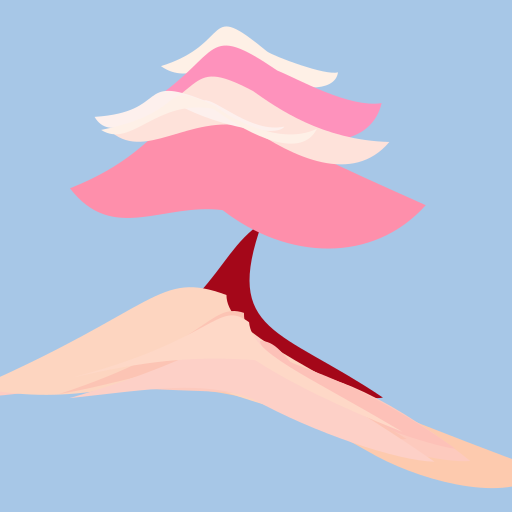} \\

        \multicolumn{2}{c}{``a green dragon  breathing fire''} &
        \multicolumn{2}{c}{``a dragon-cat hybrid''} \\
        \includegraphics[height=0.0975\textwidth,width=0.0975\textwidth]{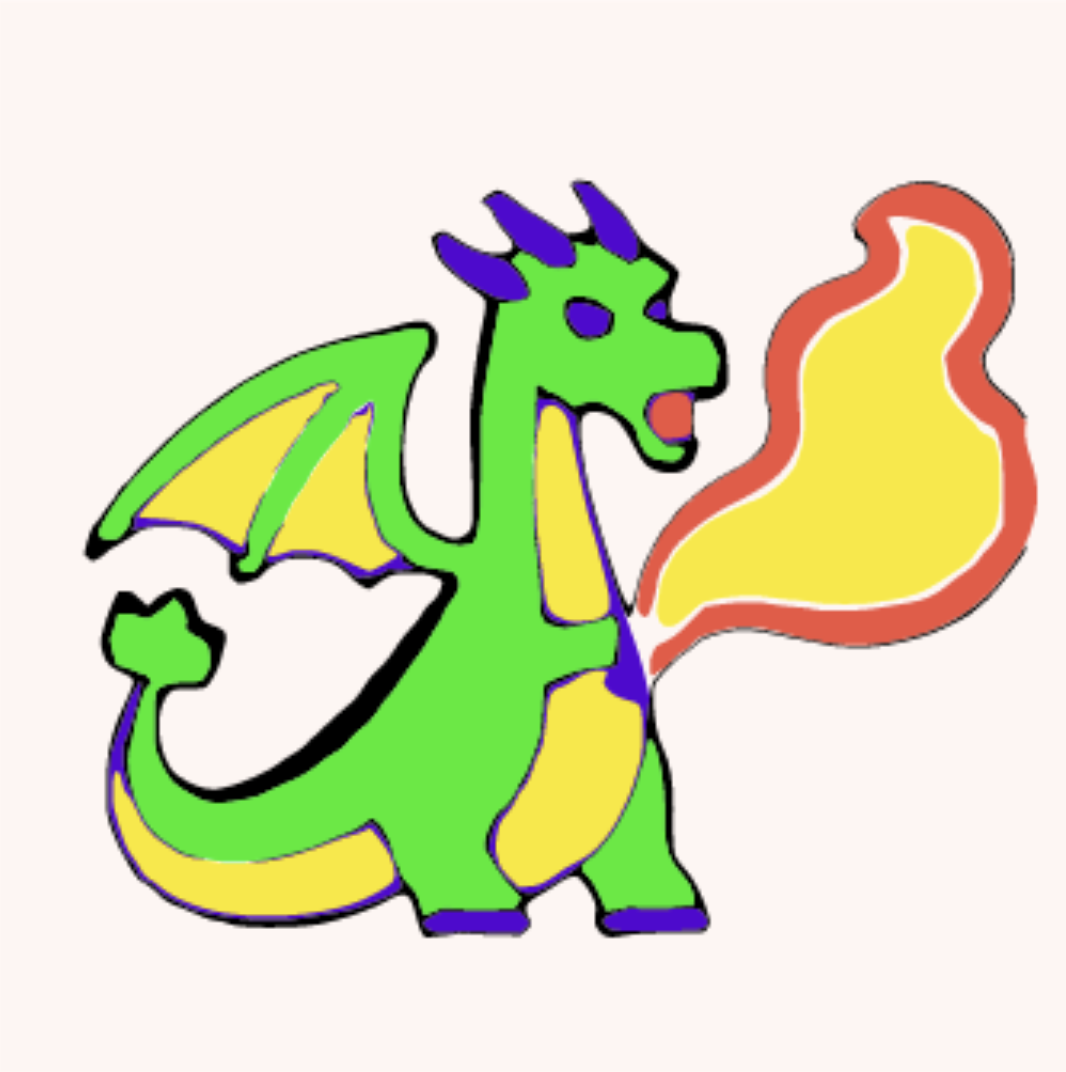} &
        \includegraphics[height=0.0975\textwidth,width=0.0975\textwidth]{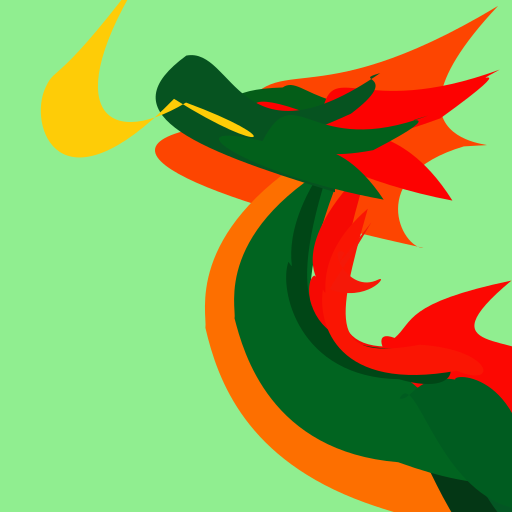} &
        \includegraphics[height=0.0975\textwidth,width=0.0975\textwidth]{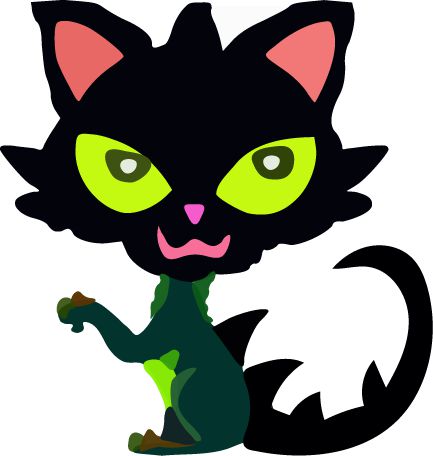} &
        \includegraphics[height=0.0975\textwidth,width=0.0975\textwidth]{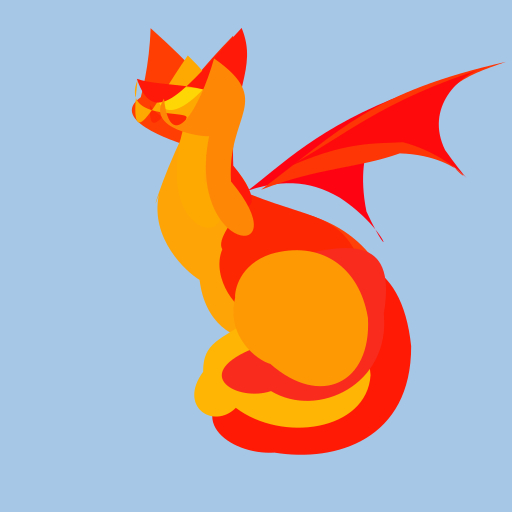} \\

        \multicolumn{2}{c}{``a 3D rendering of a temple''} &
        \multicolumn{2}{c}{\begin{tabular}{c} ``The Statue of Liberty with \\ the face of an owl'' \end{tabular}} \\
        \includegraphics[height=0.0975\textwidth,width=0.0975\textwidth]{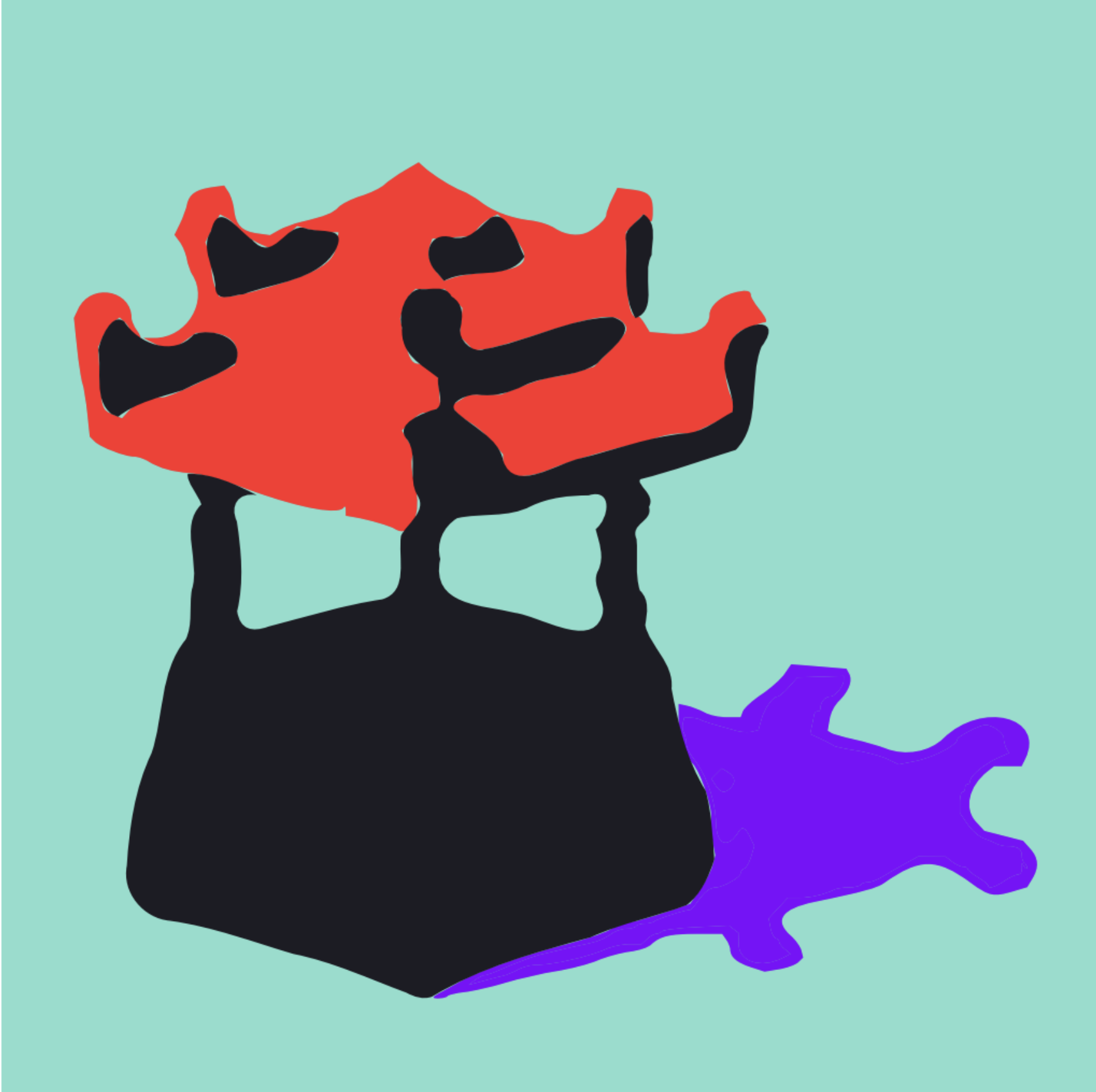} &
        \includegraphics[height=0.0975\textwidth,width=0.0975\textwidth]{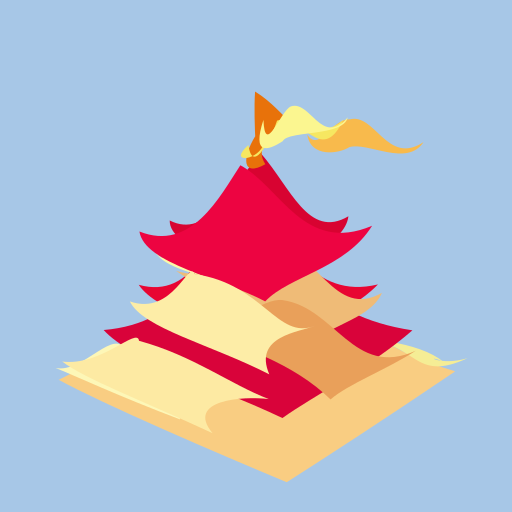} &
        \includegraphics[height=0.0975\textwidth,width=0.0975\textwidth]{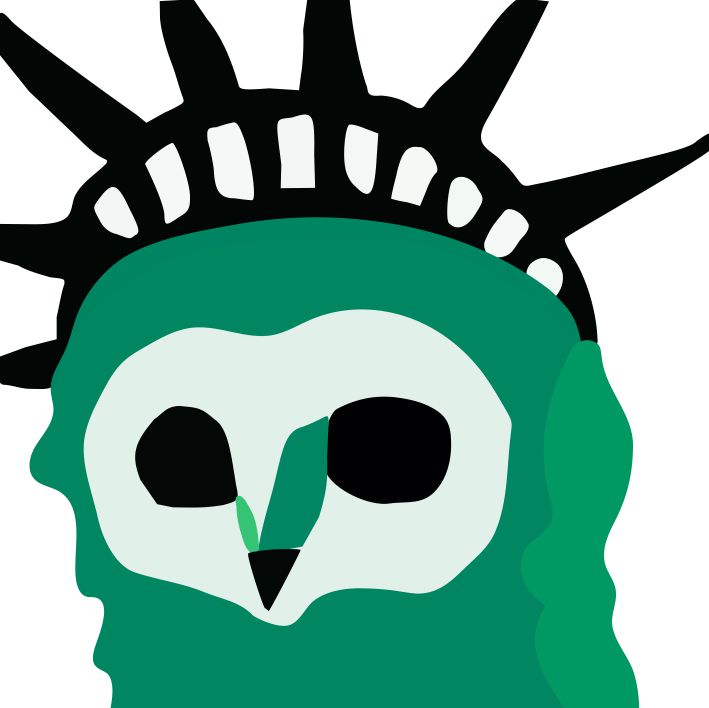} &
        \includegraphics[height=0.0975\textwidth,width=0.0975\textwidth]{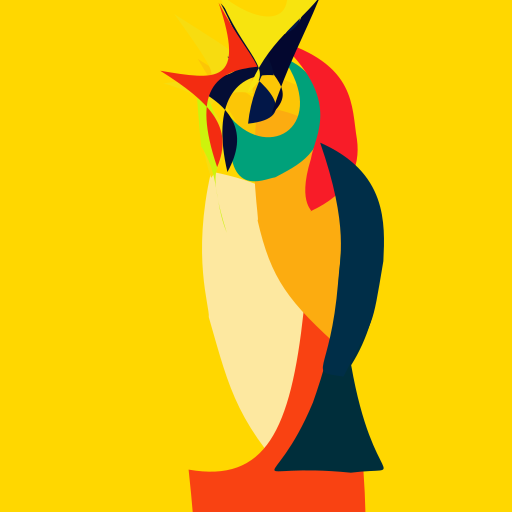} \\

        \begin{tabular}{c} NiVEL  \end{tabular} &
        \begin{tabular}{c} \textbf{NeuralSVG} \end{tabular} &
        \begin{tabular}{c} Text-to-Vector \end{tabular} &
        \begin{tabular}{c} \textbf{NeuralSVG} \end{tabular} \\
        
    \\[-0.6cm]        
    \end{tabular}
    }
    \caption{\textbf{Qualitative Comparisons.} As no code implementations are available, we provide visual comparisons to NIVeL~\cite{thamizharasan2024nivel} and Text-to-Vector~\cite{zhang2024text} using results shown in their paper. \\[-0.4cm]
    }
    \label{fig:comparisons_nivel_text2svg}
\end{figure}

Next, we compare NeuralSVG with more recent but closed-source techniques, as shown in~\Cref{fig:comparisons_nivel_text2svg}. First, when examining the results of NIVeL~\cite{thamizharasan2024nivel} artifacts are present, particularly along the black contours. This issue arises because NIVeL learns its implicit representation in pixel space and subsequently converts it to an SVG through a post-processing step, which results in pixel-like artifacts.
Additionally, a single layer in their implicit representation may encode multiple shapes, leading to potential errors when vectorizing the pixel layers.
We observe that the results of Text-to-Vector~\cite{zhang2024text} are comparable to those achieved with NeuralSVG. However, NeuralSVG learns ordered SVGs directly in a single training stage, whereas Text-to-Vector relies on a secondary post-processing step to decompose the SVG into a more editable format. Furthermore, the results presented here are taken directly from their published paper, which restricts our ability to thoroughly analyze the structure of their resulting SVG representations or even know how many shapes were used when rendering.

\subsection{Quantitative Comparisons}
\paragraph{\textbf{CLIP-Space Metrics}}
To quantitatively evaluate the methods, we follow prior work and employ two CLIP-space metrics. The first metric computes the CLIP-space cosine similarity between the embeddings of the generated SVGs and their corresponding input text prompts. We additionally report the R-Precision (R-Prec), which measures the percentage of generated SVGs that achieve maximal CLIP similarity with their correct prompt among all $128$ prompts. We average results across all prompts and five seeds.

\begin{table}
\small
\setlength{\tabcolsep}{4pt}
\addtolength{\belowcaptionskip}{-5pt}
\centering
\caption{\textbf{CLIP-Based Quantitative Comparisons}. We compute CLIP-space cosine similarities and R-Precision using the CLIP L/14 model on rasterized SVG results, optimized with varying numbers of shapes. \\[-0.65cm]} 
\begin{tabular}{l | c c | c c | c} 
    \toprule
    & \multicolumn{2}{c|}{VectorFusion} & \multicolumn{2}{c|}{SVGDreamer} & \textbf{NeuralSVG} \\
    Metric & $64$ & $16$ & $256$ & $16$ & $\mathbf{16}$ \\
    \midrule
    R-Precision $\uparrow$ & $83.46$ & $48.03$ & $85.03$ & $43.30$ & $67.18$ \\
    Text-Image Similarity $\uparrow$ & $26.33$ & $23.47$ & $26.58$ & $20.89$ & $26.94$ \\
    \bottomrule
\end{tabular}
\label{tb:clip_metrics}
\end{table}

Full results are presented in \Cref{tb:clip_metrics}. When constrained to the same number of shapes, NeuralSVG outperforms both VectorFusion and SVGDreamer across both metrics. This aligns with our visual comparisons, which show that competing methods struggle to generate organized shapes and interpretable scenes under the same constraints.
When VectorFusion and SVGDreamer use $64$ and $256$ shapes, all methods achieve comparable CLIP scores while VectorFusion and SVGDreamer attain higher R-Prec scores than NeuralSVG. 
However, our visual comparisons reveal that while these higher shape counts improve the image-based metrics, they result in highly disorganized outputs that are impractical for editing.
As such, NeuralSVG offers an appealing alternative by generating more organized shapes that create more editable scenes while using a small number of shapes.

\begin{figure}
    \centering
    \addtolength{\belowcaptionskip}{-5pt}
    \includegraphics[width=0.45\textwidth]{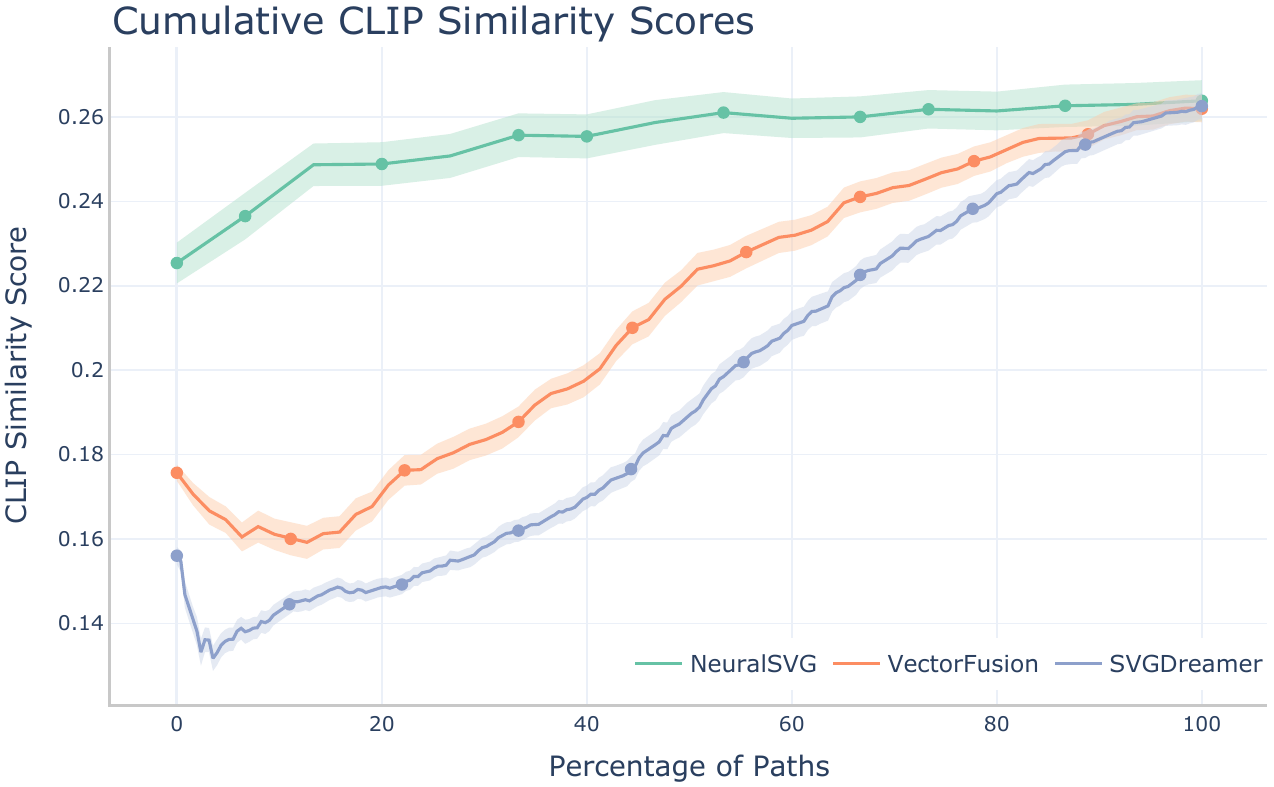} \\[-0.25cm]
    \caption{
    \textbf{Cumulative CLIP Similarities.} We show CLIP similarities obtained when using a subset of the learned shapes from each method, selected in rendering order. As shown, SVGs produced by NeuralSVG are much more recognizable when using a small percentage of the learned shapes.
    }
    \label{fig:cumulative_clip_scores}
\end{figure}

\paragraph{\textbf{Cumulative CLIP-Space Similarities}}
Next, considering the order-centric approach of NeuralSVG, it is important to examine whether the shapes learned by our method align better with CLIP than alternative approaches. To evaluate this, we compute CLIP-space similarities between input text prompts and generated SVGs using a subset of the learned shapes, selected in rendering order. The results in~\Cref{fig:cumulative_clip_scores} compare NeuralSVG to VectorFusion ($64$ shapes) and SVGDreamer ($256$ shapes). As illustrated, SVGs generated by NeuralSVG are significantly more recognizable when using a small fraction of the total shapes. This indicates the early shapes produced by NeuralSVG are semantically meaningful and have a more standalone meaning compared to those generated by alternative methods. Moreover, note that as the total shapes in VectorFusion and SVGDreamer are significantly higher, at $25\%$ and $6\%$, they already match the shape count used by our full method.

\begin{figure}
    \centering
    \setlength{\tabcolsep}{0.5pt}
    {\small
    \begin{tabular}{c c c c c c}

        \multicolumn{6}{c}{``a drawing of a cat''} \\
        \includegraphics[height=0.08\textwidth,width=0.08\textwidth]{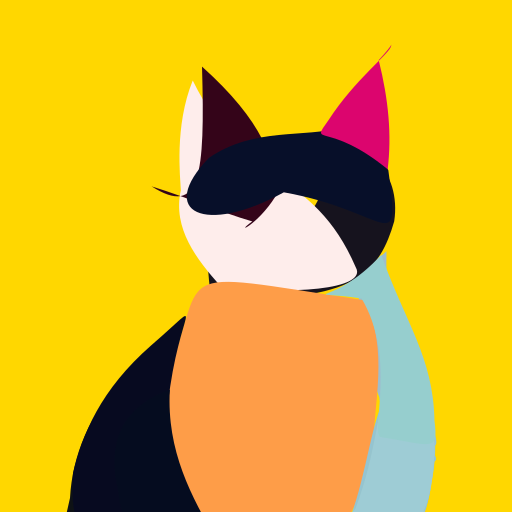} &
        \includegraphics[height=0.08\textwidth,width=0.08\textwidth]{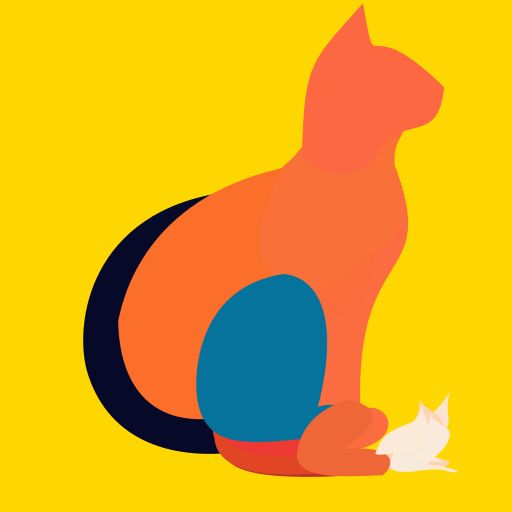} &
        \includegraphics[height=0.08\textwidth,width=0.08\textwidth]{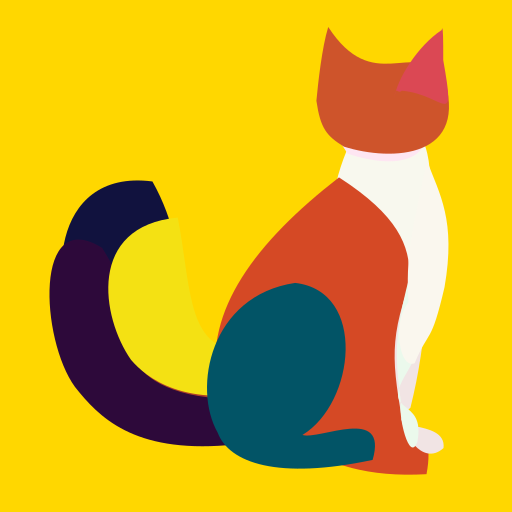} &
        \includegraphics[height=0.08\textwidth,width=0.08\textwidth]{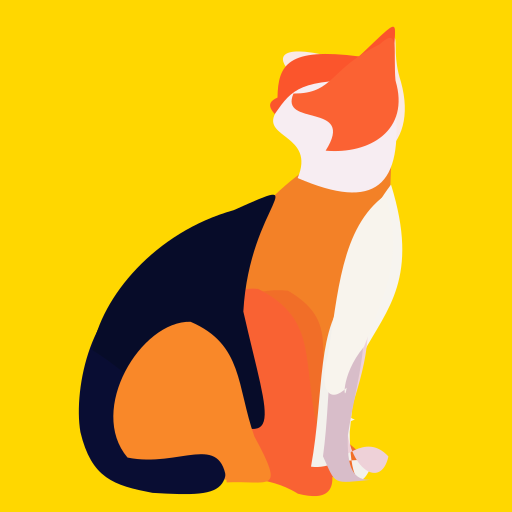} &
        \includegraphics[height=0.08\textwidth,width=0.08\textwidth]{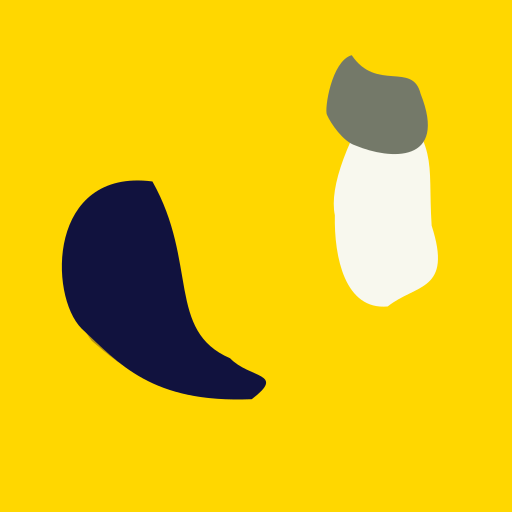} &
        \includegraphics[height=0.08\textwidth,width=0.08\textwidth]{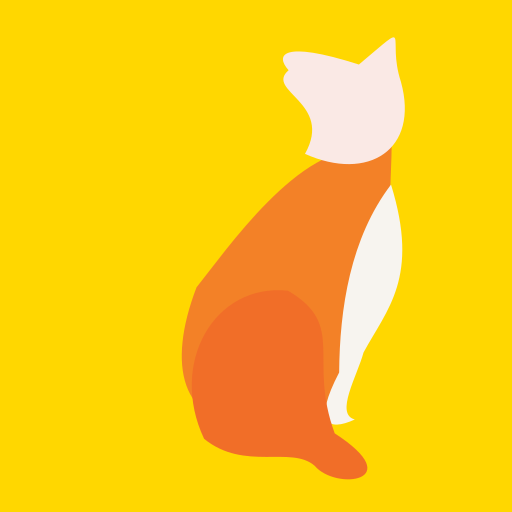} \\

        \multicolumn{6}{c}{``a man in an astronaut suit walking across a desert...''} \\
        \includegraphics[height=0.08\textwidth,width=0.08\textwidth]{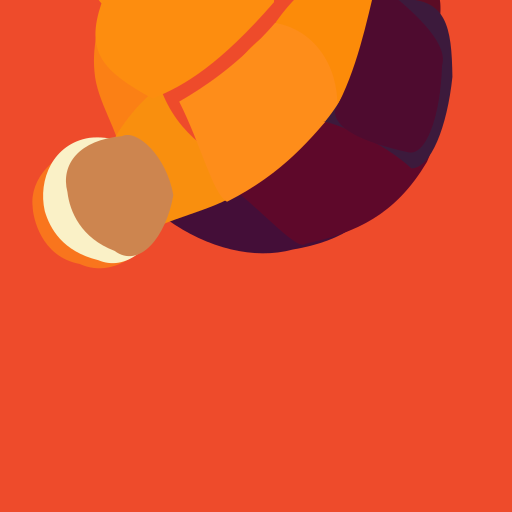} &
        \includegraphics[height=0.08\textwidth,width=0.08\textwidth]{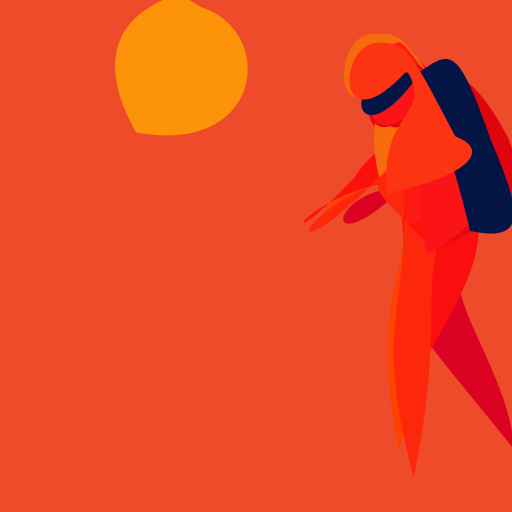} &
        \includegraphics[height=0.08\textwidth,width=0.08\textwidth]{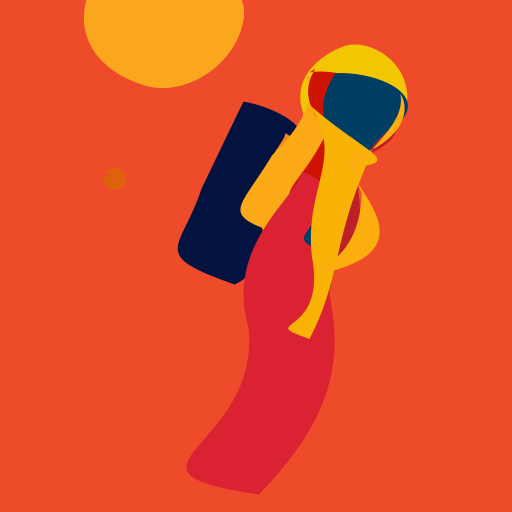} &
        \includegraphics[height=0.08\textwidth,width=0.08\textwidth]{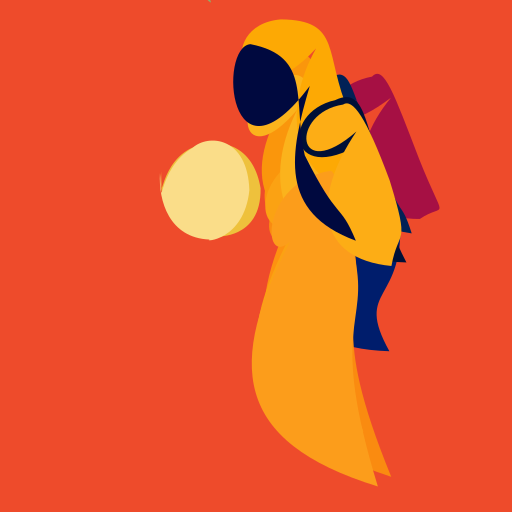} &
        \includegraphics[height=0.08\textwidth,width=0.08\textwidth]{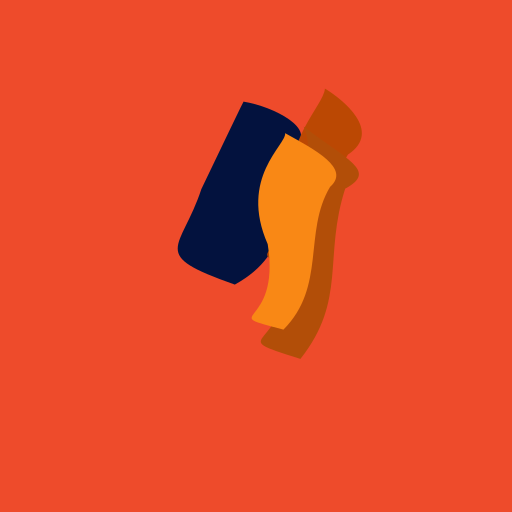} &
        \includegraphics[height=0.08\textwidth,width=0.08\textwidth]{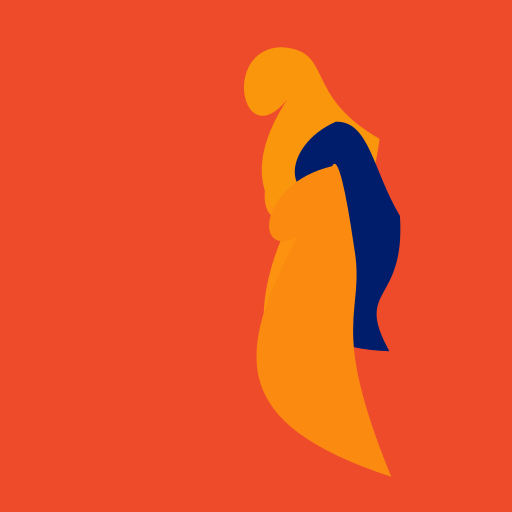} \\

        \multicolumn{6}{c}{``Pikachu, in pastel colors, childish and fun''} \\
        \includegraphics[height=0.08\textwidth,width=0.08\textwidth]{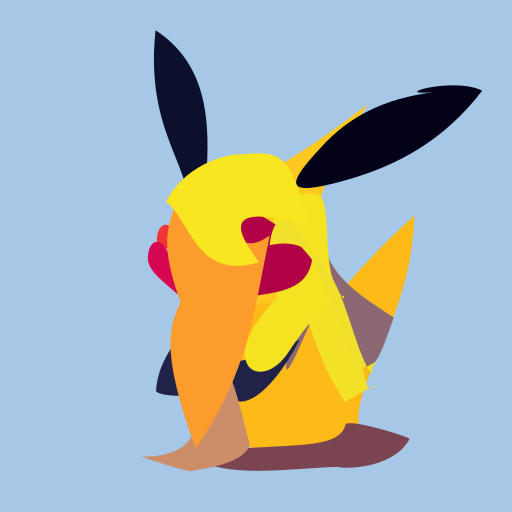} &
        \includegraphics[height=0.08\textwidth,width=0.08\textwidth]{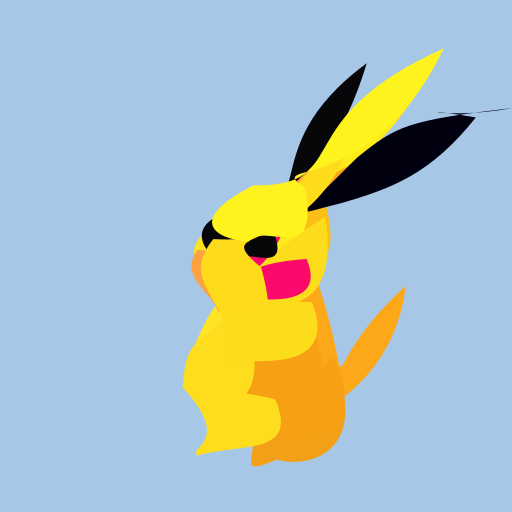} &
        \includegraphics[height=0.08\textwidth,width=0.08\textwidth]{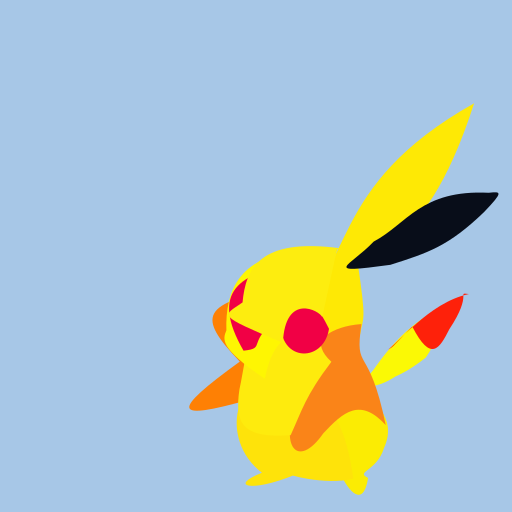} &
        \includegraphics[height=0.08\textwidth,width=0.08\textwidth]{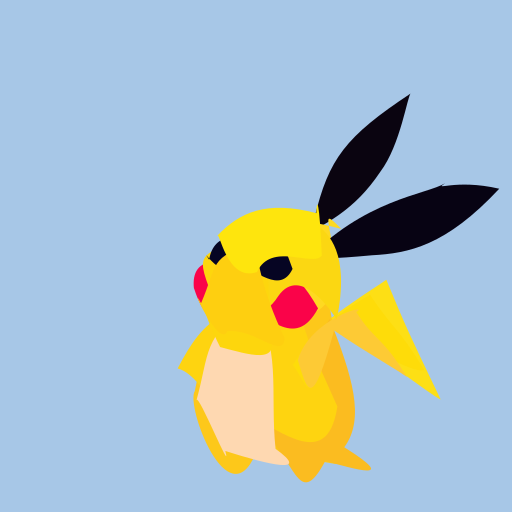} &
        \includegraphics[height=0.08\textwidth,width=0.08\textwidth]{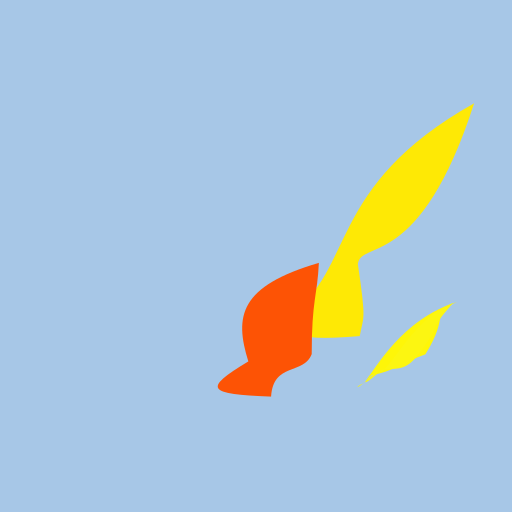} &
        \includegraphics[height=0.08\textwidth,width=0.08\textwidth]{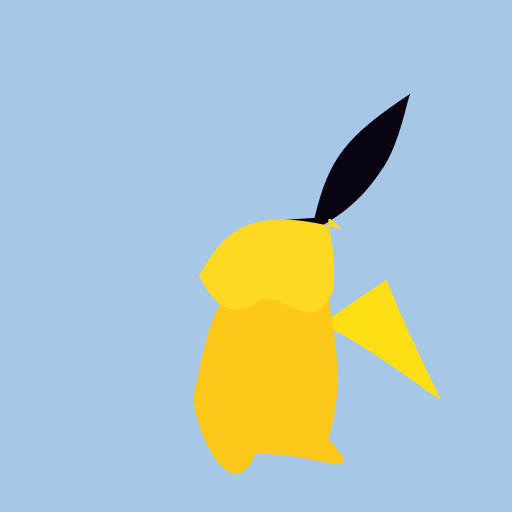} \\

        Direct Opt. & Joint MLP & w/o Drop & \textbf{NeuralSVG} &
        \begin{tabular}{c} w/o Drop \\ (4 Shapes) \end{tabular} & 
        \begin{tabular}{c} \textbf{NeuralSVG} \\ \textbf{(4 Shapes)} \end{tabular} \\
        
    \\[-0.65cm]        
    \end{tabular}
    }
    \caption{\textbf{Ablation Study.} We validate our key design choices: directly optimizing the shape primitives, using a single MLP network to learn both control point positions and colors, and omitting our ordered dropout technique. 
    The two rightmost columns illustrate results from NeuralSVG trained with and without dropout when rendering the first four learned shapes.
    }
    \label{fig:ablations}
\end{figure}

\subsection{Ablation Studies}\label{sec:ablations}
\begin{figure}
    \centering
    \addtolength{\belowcaptionskip}{-10pt}
    \includegraphics[width=0.45\textwidth]{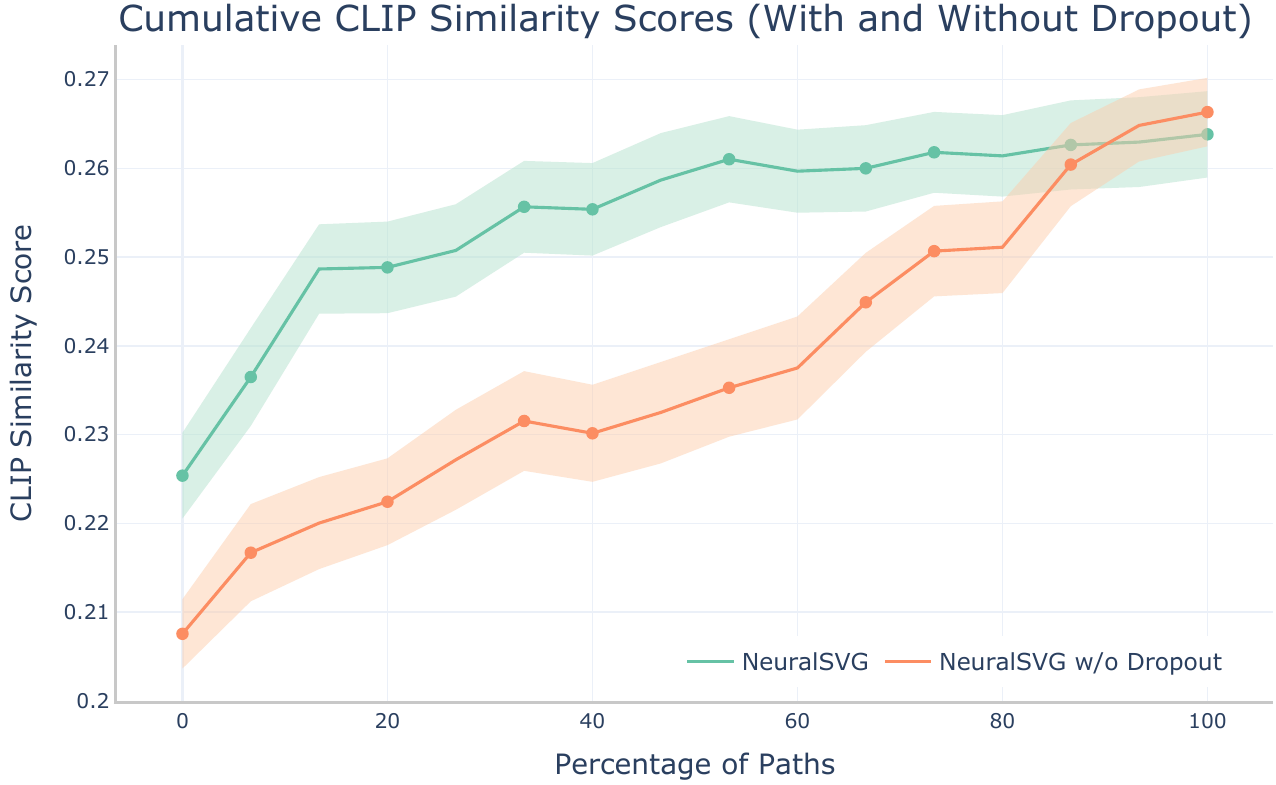} \\[-0.25cm]
    \caption{
    \textbf{Cumulative CLIP Similarities With and Without Dropout.} We show cumulative CLIP similarities achieved by NeuralSVG trained with and without dropout across 50 prompts, using 16 learnable shapes. Consistent~\Cref{fig:cumulative_clip_scores}, our dropout technique improves the recognizability of SVGs.
    }
    \label{fig:cumulative_clip_scores_no_dropout}
\end{figure}

Finally, we validate our key design choices, specifically the use of our dropout technique and the two MLP branches. Visual comparisons are presented in~\Cref{fig:ablations}. First, when attempting to directly optimize the shape primitives, the resulting SVGs often converge to non-smooth shapes and may fail to accurately adhere to the input prompt. This aligns with prior works that observe optimizing parameters via a neural network may assist in attaining smoother and more coherent results~\cite{vinker2023clipascene,Gal_2024_CVPR}. 
Next, using a single MLP to predict both the control point positions and colors leads to suboptimal results. For instance, in the second row, the astronaut is incorrectly colored the same as the background while in the first row, the cat appears almost entirely orange, lacking details such as its facial features. Finally, when dropout is omitted, the visual results are comparable to those of our full method, as is expected. However, as illustrated in the two rightmost columns, the learned shapes lack semantic meaning. As a result, when using a small number of shapes, the resulting SVGs are also not easily recognizable by CLIP (see~\Cref{fig:cumulative_clip_scores_no_dropout}). In contrast, NeuralSVG effectively captures the coarse structure of the scene even with a limited number of shapes thanks to our learned ordering. 

\vspace{0.2cm}

\subsection{Additional Controls}\label{sec:additional_apps}

\paragraph{\textbf{Color Palette Control}}
In~\Cref{fig:color_control}, we demonstrate our method's ability to dynamically adapt the color palette of the SVG using a single learned representation. Specifically, we show results obtained with colors unobserved during training, illustrating our ability to generalize to new palettes. This flexibility allows users to customize results based on personal preferences at inference, without requiring a dedicated optimization process for each modification.

\begin{figure}
    \centering
    \setlength{\tabcolsep}{1pt}
    \addtolength{\belowcaptionskip}{-5pt}
    {\small
    \begin{tabular}{c c c}

        & \multicolumn{2}{c}{``a dog''} \\
        \raisebox{0.3in}{\multirow{2}{*}{\includegraphics[height=0.12\textwidth,width=0.12\textwidth]{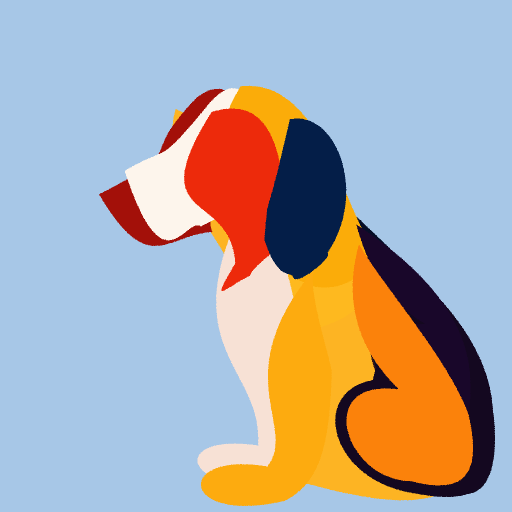}}} &
        \raisebox{0.045in}{\rotatebox{90}{\begin{tabular}{c}Naive \\ Squeeze \end{tabular}}} & \includegraphics[height=0.06\textwidth,width=0.24\textwidth]{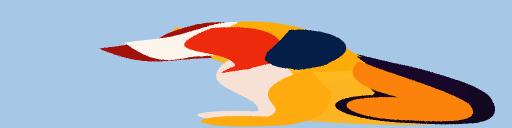} \\
        & \raisebox{0.045in}{\rotatebox{90}{\begin{tabular}{c} With \\ Control \end{tabular}}} & \includegraphics[height=0.06\textwidth,width=0.24\textwidth]{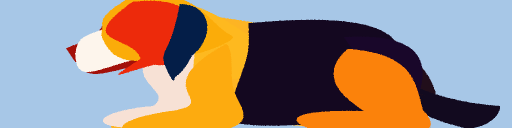} \\

        & \multicolumn{2}{c}{``a sports car''} \\
        \raisebox{0.3in}{\multirow{2}{*}{\includegraphics[height=0.12\textwidth,width=0.12\textwidth]{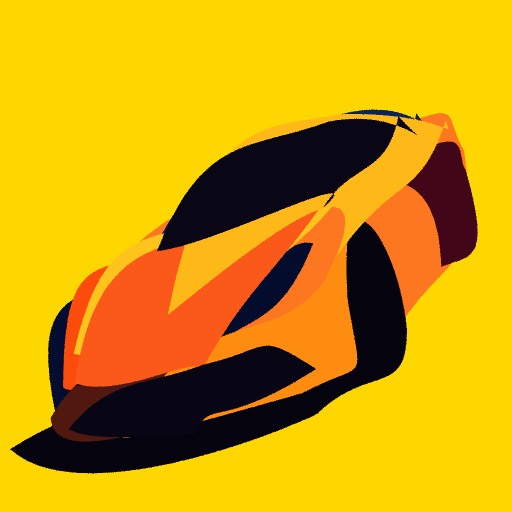}}} &
        \raisebox{0.045in}{\rotatebox{90}{\begin{tabular}{c}Naive \\ Squeeze \end{tabular}}} & \includegraphics[height=0.06\textwidth,width=0.24\textwidth]{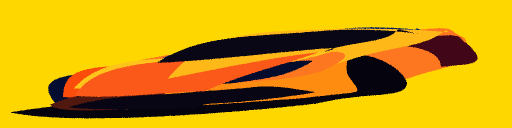} \\
        & \raisebox{0.045in}{\rotatebox{90}{\begin{tabular}{c} With \\ Control \end{tabular}}} & \includegraphics[height=0.06\textwidth,width=0.24\textwidth]{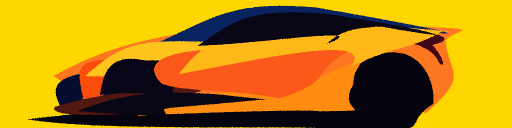} \\

    \\[-0.6cm]        
    \end{tabular}
    }
    \caption{\textbf{Controlling the Aspect Ratio.} We present results from optimizing NeuralSVG with aspect ratios of 1:1 and 4:1. In each pair, the top row shows the naive approach of squeezing the 1:1 output into a 4:1 ratio. The bottom row shows results where the trained network directly outputs the 4:1 ratio.
    }
    \label{fig:aspect_ratio}
\end{figure}

\paragraph{\textbf{Aspect Ratio Control}}
Another desired property for controlling SVGs at inference time is easily modifying their target aspect ratio. While one can technically modify the aspect ratio of the SVG manually, successfully generating a pleasing result for a target ratio can still be challenging. We show that by passing an encoding of the desired aspect ratio (e.g., 1:1 or 1:4) to our network and rendering accordingly, our method successfully learns to adapt the same SVG shapes to multiple aspect ratios in the same learned representation. We illustrate this in~\Cref{fig:aspect_ratio}, showing results obtained using aspect ratios of 1:1 and 4:1 when compared to the result one would achieve by automatically ``squeezing'' the 1:1 result.

\begin{figure}
    \centering
    \setlength{\tabcolsep}{1pt}
    \addtolength{\belowcaptionskip}{-10pt}
    {\small
    \begin{tabular}{c c c c c}

        \multicolumn{5}{c}{``a teapot''} \\
        \includegraphics[height=0.09\textwidth,width=0.09\textwidth]{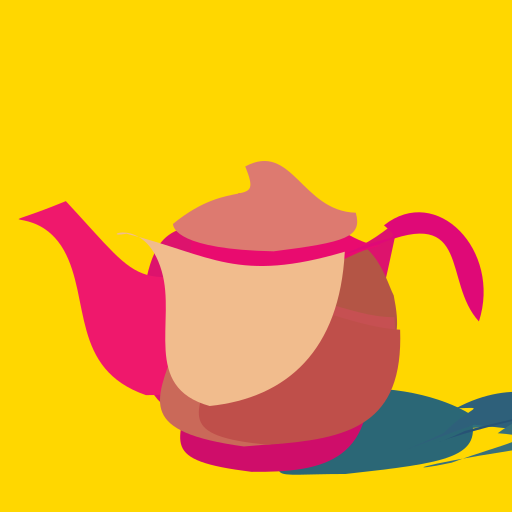} &
        \includegraphics[height=0.09\textwidth,width=0.09\textwidth]{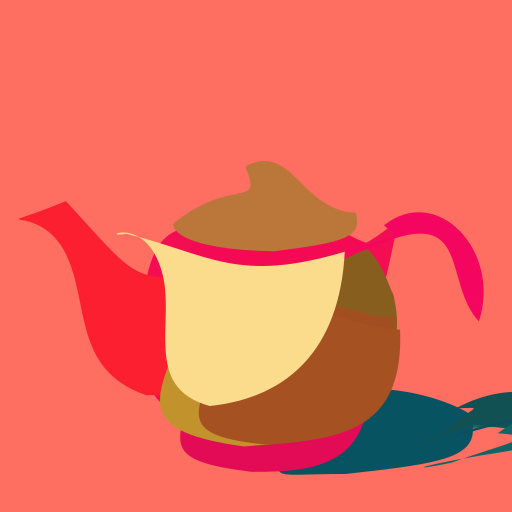} &
        \includegraphics[height=0.09\textwidth,width=0.09\textwidth]{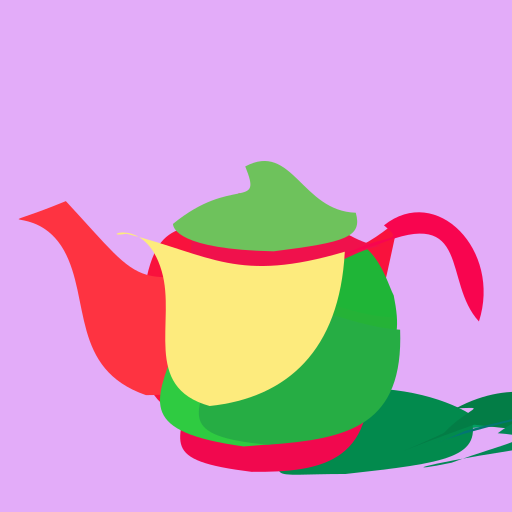} &
        \includegraphics[height=0.09\textwidth,width=0.09\textwidth]{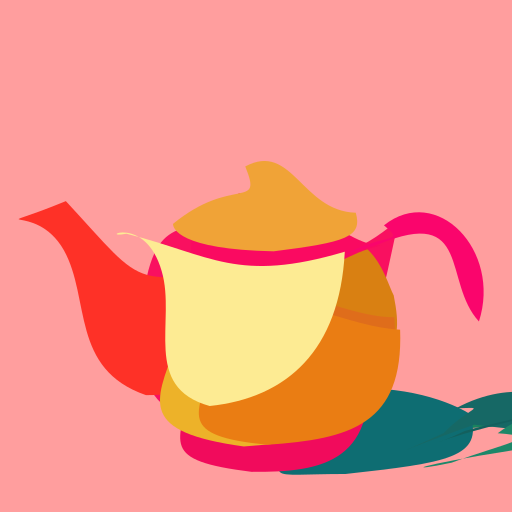} &
        \includegraphics[height=0.09\textwidth,width=0.09\textwidth]{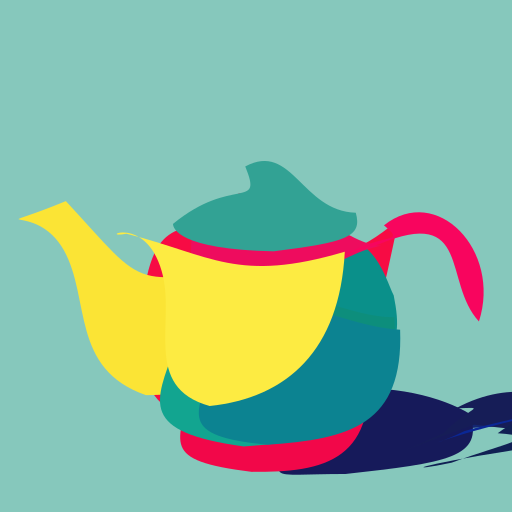} \\

        \multicolumn{5}{c}{``a knight holding a long sword''} \\
        \includegraphics[height=0.09\textwidth,width=0.09\textwidth]{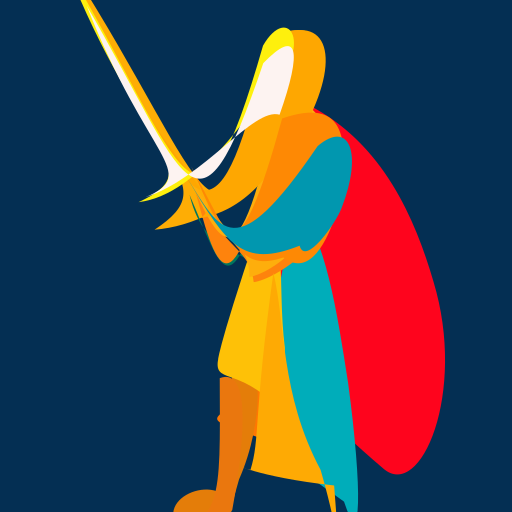} &
        \includegraphics[height=0.09\textwidth,width=0.09\textwidth]{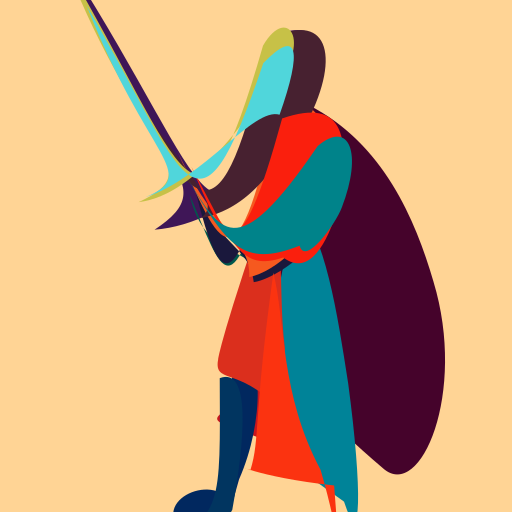} &
        \includegraphics[height=0.09\textwidth,width=0.09\textwidth]{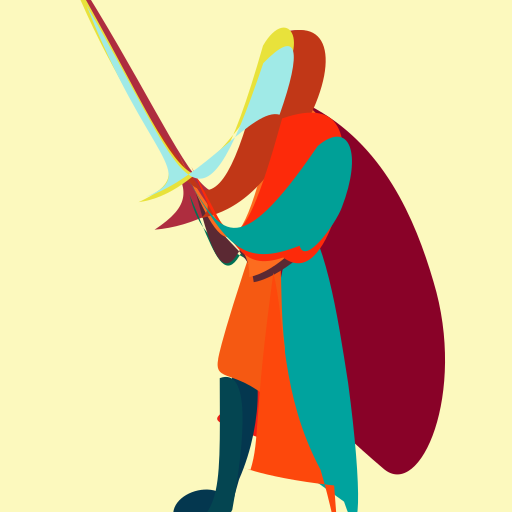} &
        \includegraphics[height=0.09\textwidth,width=0.09\textwidth]{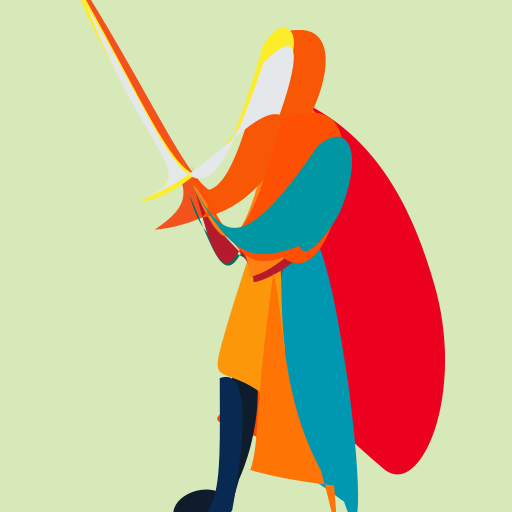} &
        \includegraphics[height=0.09\textwidth,width=0.09\textwidth]{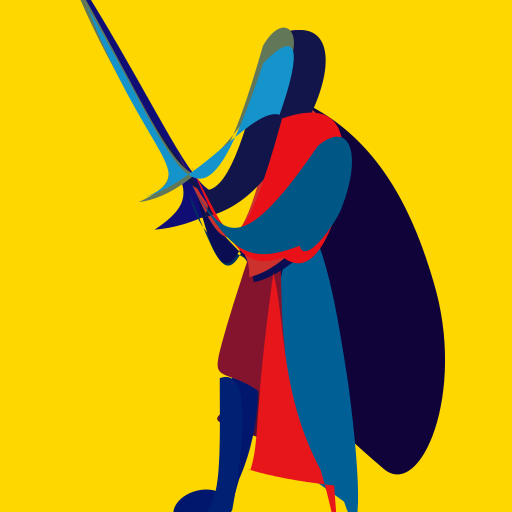} \\

        \multicolumn{5}{c}{``a peacock''} \\
        \includegraphics[height=0.09\textwidth,width=0.09\textwidth]{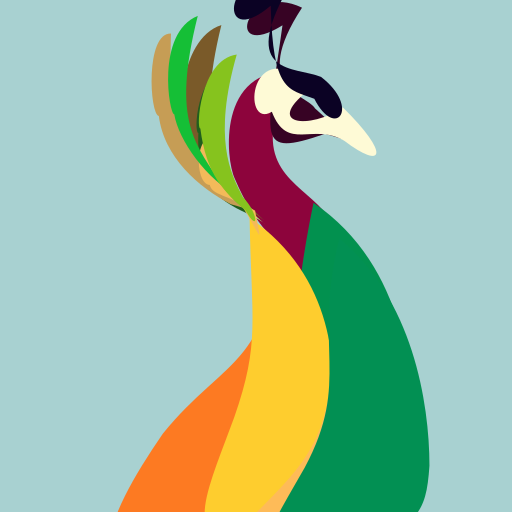} &
        \includegraphics[height=0.09\textwidth,width=0.09\textwidth]{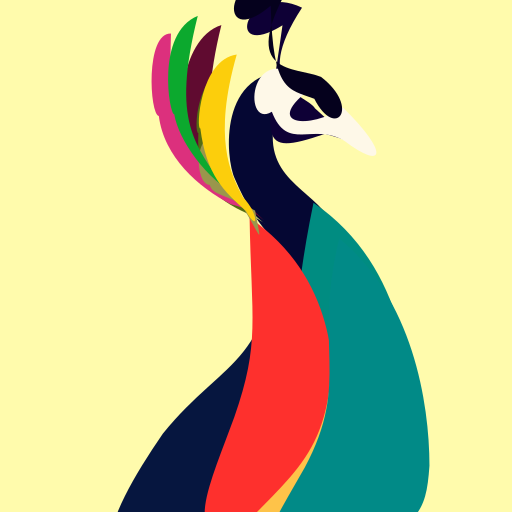} &
        \includegraphics[height=0.09\textwidth,width=0.09\textwidth]{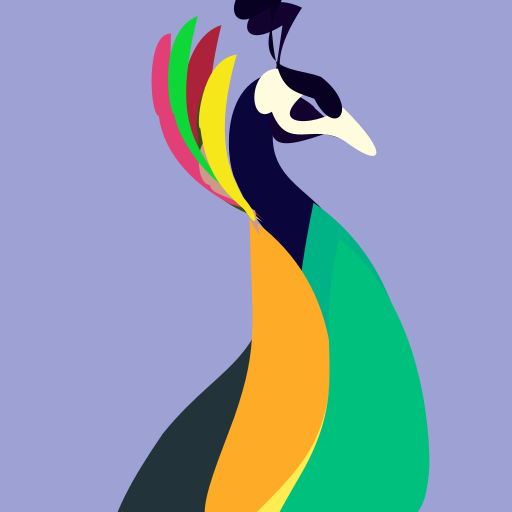} &
        \includegraphics[height=0.09\textwidth,width=0.09\textwidth]{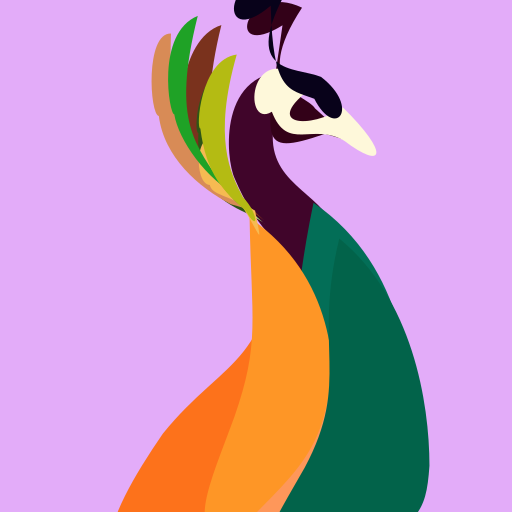} &
        \includegraphics[height=0.09\textwidth,width=0.09\textwidth]{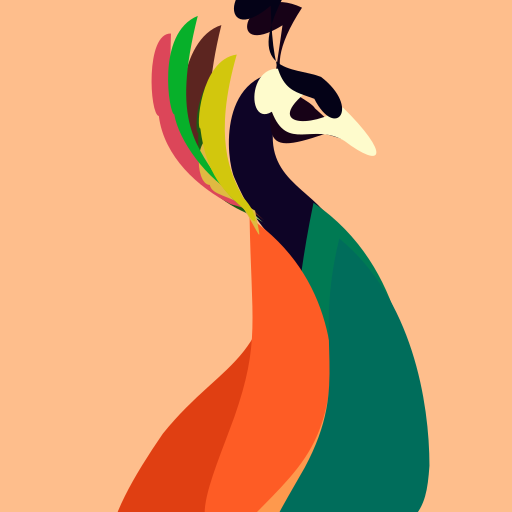} \\
        
        \multicolumn{5}{c}{``a cat as 3D rendered in Unreal Engine''} \\
        \includegraphics[height=0.09\textwidth,width=0.09\textwidth]{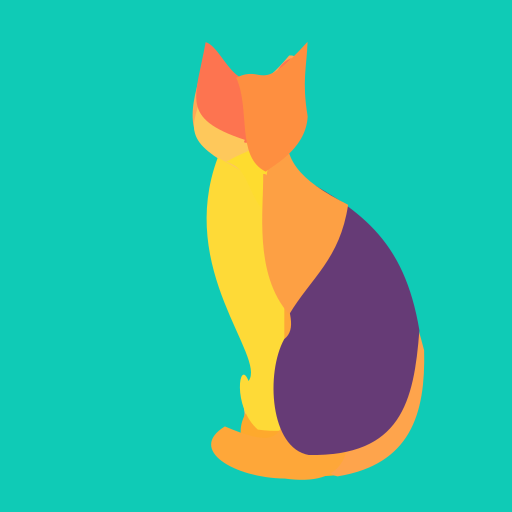} &
        \includegraphics[height=0.09\textwidth,width=0.09\textwidth]{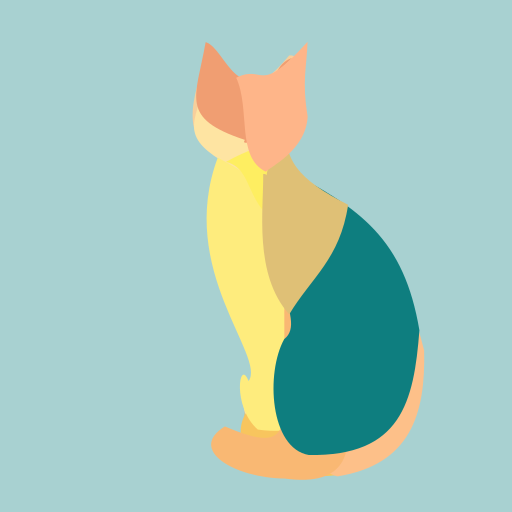} &
        \includegraphics[height=0.09\textwidth,width=0.09\textwidth]{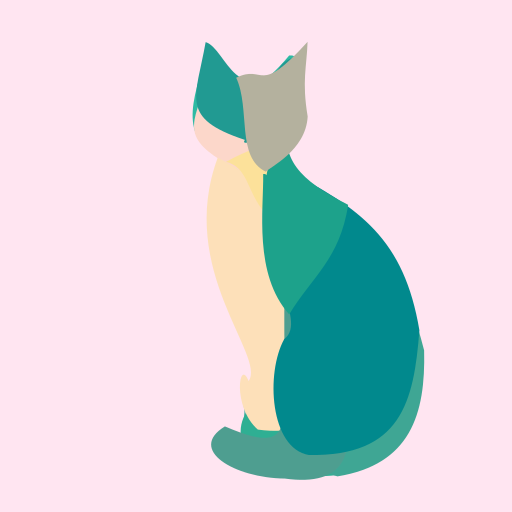} &
        \includegraphics[height=0.09\textwidth,width=0.09\textwidth]{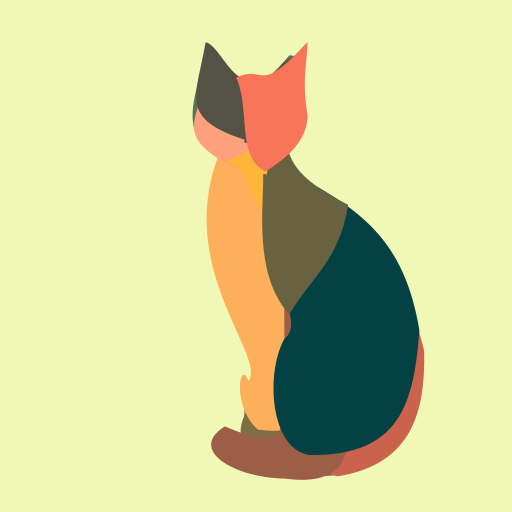} &
        \includegraphics[height=0.09\textwidth,width=0.09\textwidth]{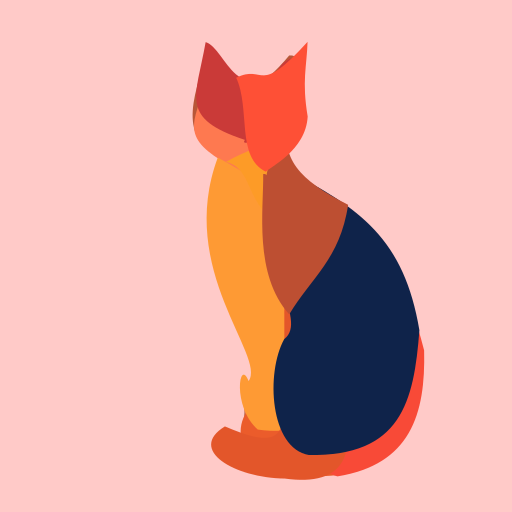} \\
        
    \\[-0.65cm]        
    \end{tabular}
    }
    \caption{\textbf{Controlling the Color Palette.} Given a learned representation, we render the result using different background colors specified by the user, resulting in varying color palettes in the resulting SVGs.
    }
    \label{fig:color_control}
\end{figure}

\begin{figure}
    \centering
    \setlength{\tabcolsep}{1pt}
    \addtolength{\belowcaptionskip}{-5pt}
    {\small
    \begin{tabular}{c c c c}

            \multicolumn{4}{c}{``a flamingo''} \\
            \includegraphics[width=0.185\linewidth,height=0.185\linewidth]{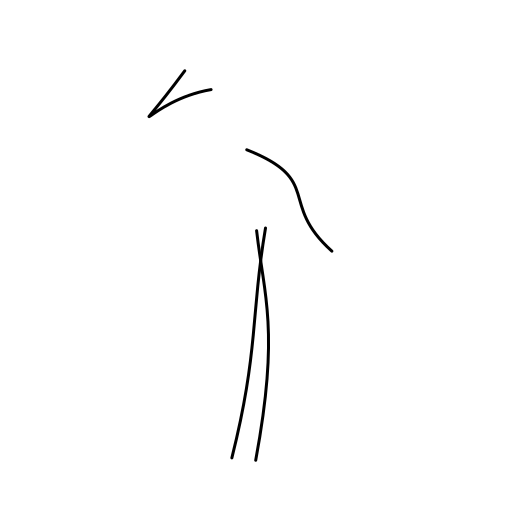} &
            \includegraphics[width=0.185\linewidth,height=0.185\linewidth]{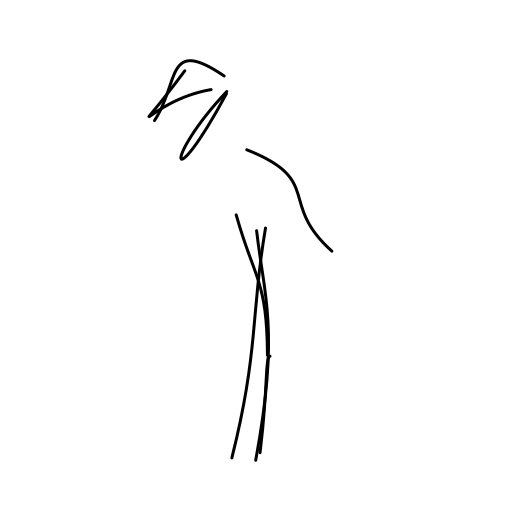} &
            \includegraphics[width=0.185\linewidth,height=0.185\linewidth]{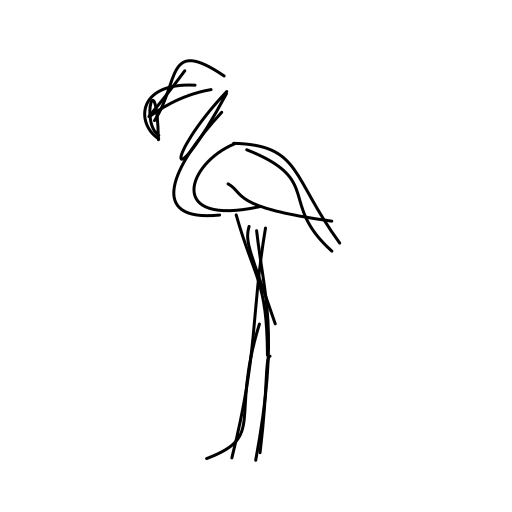} &
            \includegraphics[width=0.185\linewidth,height=0.185\linewidth]{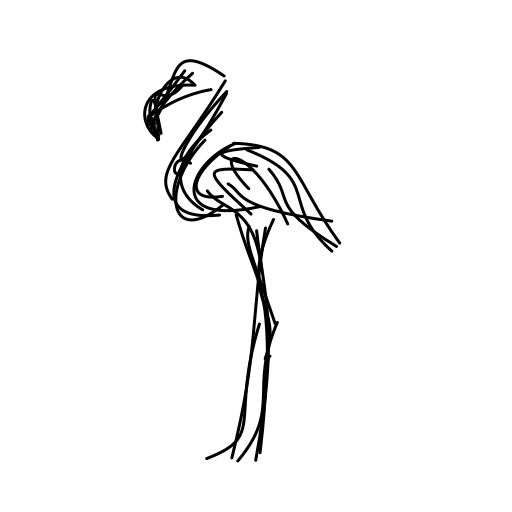} \\

            \multicolumn{4}{c}{``a rose''} \\
            \includegraphics[width=0.185\linewidth,height=0.185\linewidth]{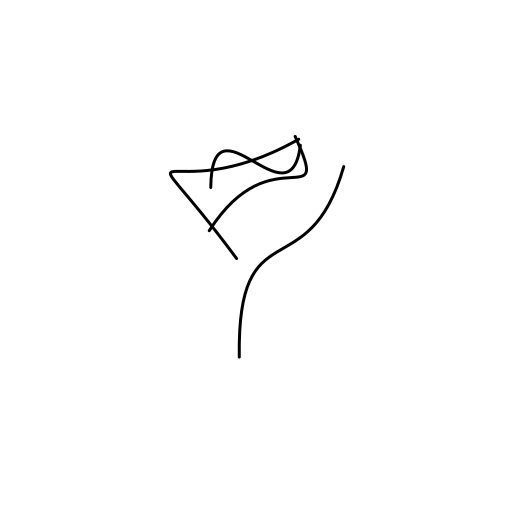} &
            \includegraphics[width=0.185\linewidth,height=0.185\linewidth]{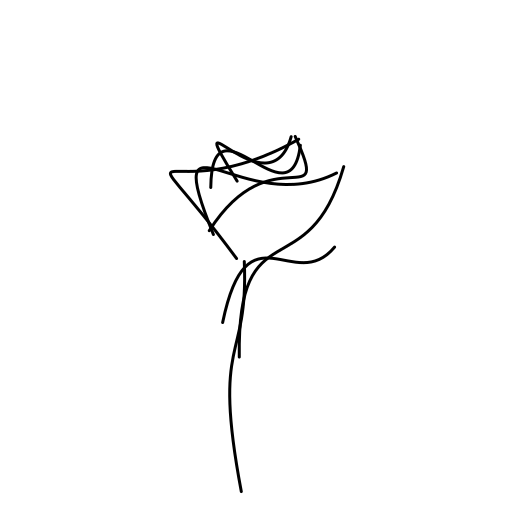} &
            \includegraphics[width=0.185\linewidth,height=0.185\linewidth]{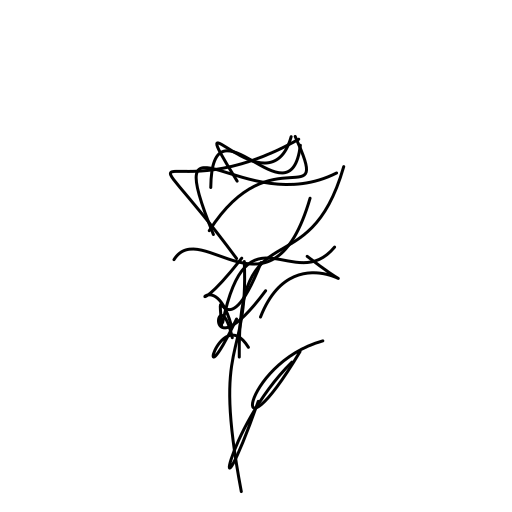} &
            \includegraphics[width=0.185\linewidth,height=0.185\linewidth]{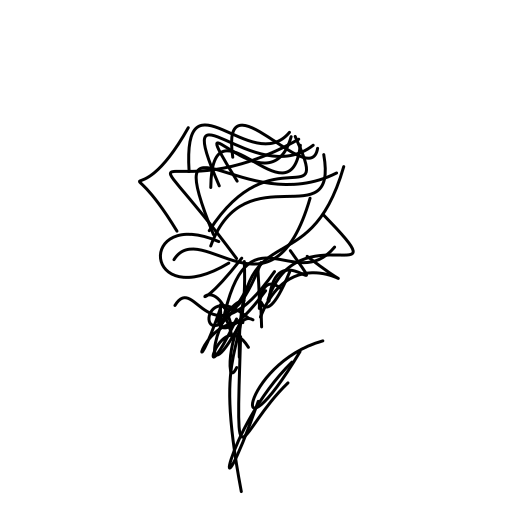} \\

            \multicolumn{4}{c}{``a vase''} \\
            \includegraphics[width=0.185\linewidth,height=0.185\linewidth]{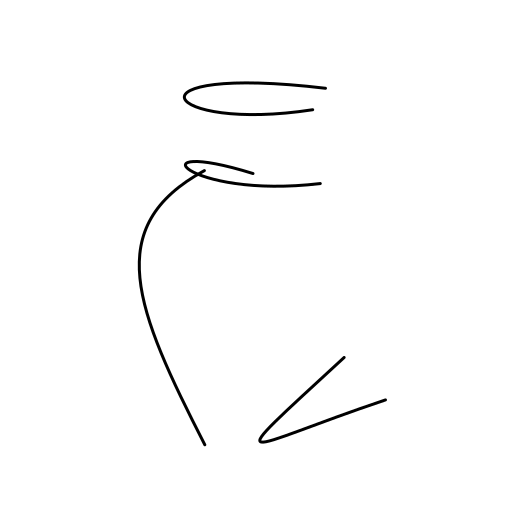} &
            \includegraphics[width=0.185\linewidth,height=0.185\linewidth]{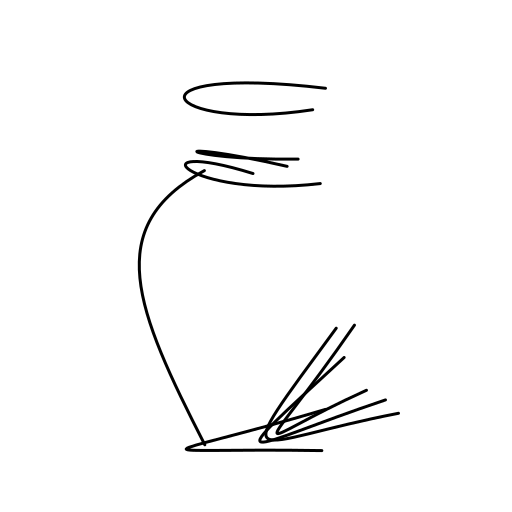} &
            \includegraphics[width=0.185\linewidth,height=0.185\linewidth]{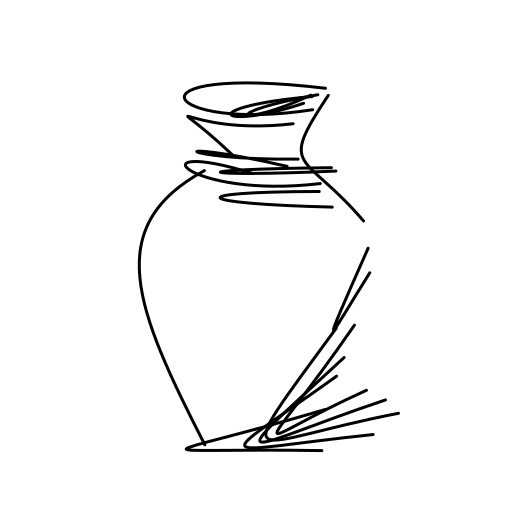} &
            \includegraphics[width=0.185\linewidth,height=0.185\linewidth]{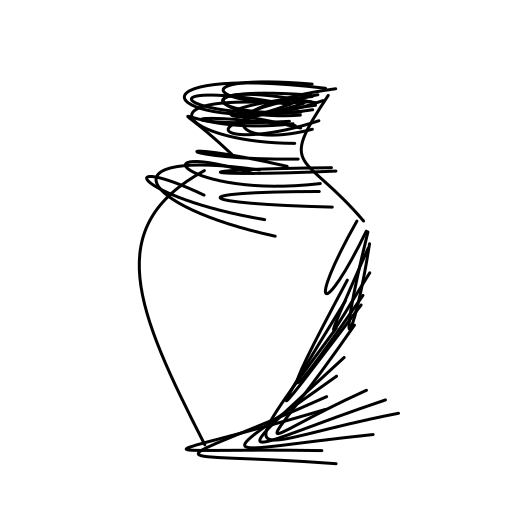} \\

            \multicolumn{4}{c}{``a camel''} \\
            \includegraphics[width=0.185\linewidth,height=0.185\linewidth]{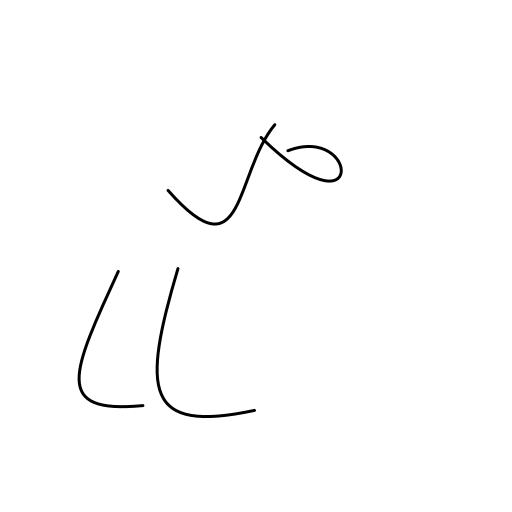} &
            \includegraphics[width=0.185\linewidth,height=0.185\linewidth]{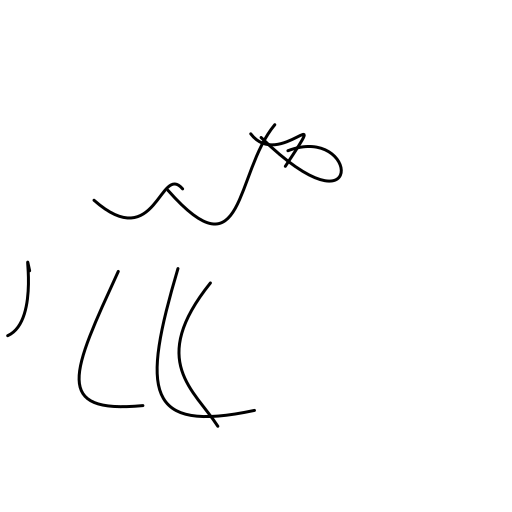} &
            \includegraphics[width=0.185\linewidth,height=0.185\linewidth]{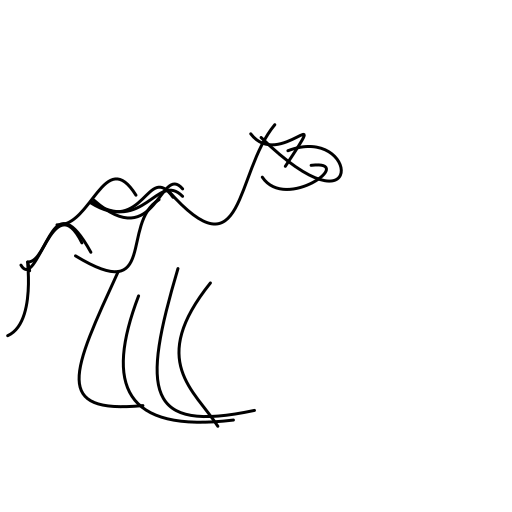} &
            \includegraphics[width=0.185\linewidth,height=0.185\linewidth]{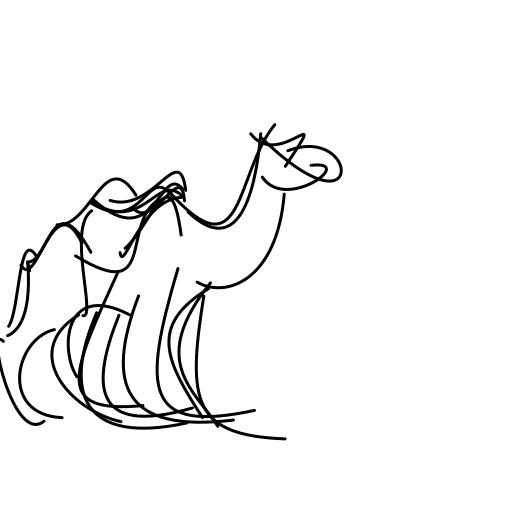} \\

        $4$ & $8$ & $16$ & $32$ \\

    \\[-0.6cm]        
    \end{tabular}
    }
    \caption{\textbf{Sketch Generation.} NeuralSVG can generate sketches with varying numbers of strokes using a single network, without requiring modifications to our framework.
    }
    \label{fig:sketches_dropout_inline}
\end{figure}

\subsection{Sketch Generation}
Our approach can also be applied to text-driven sketch generation, generating sketches with ordered strokes. As demonstrated in~\Cref{fig:sketches_dropout_inline}, the first strokes in the sketch depict the desired concept well, while adding more strokes adds details to the sketch. Notably other methods such as NIVeL~\cite{thamizharasan2024nivel} that use the pixel space as an intermediate stage during training, cannot enforce such stroke-based outputs. In contrast, our approach simply requires modifying the rendering parameters from closed shapes to open shapes when learning the representation.

\section{Conclusion}
We introduce NeuralSVG, a novel approach for generating vector graphics directly from text prompts while encouraging a layered structure essential for practical usability. NeuralSVG employs an implicit neural representation to encode the entire SVG within a compact network, optimized using Score Distillation Sampling (SDS). To address a key limitation of existing methods, our approach incorporates a dropout-based regularization technique, promoting the creation of semantically meaningful and well-ordered shapes. In addition to producing structured outputs, NeuralSVG offers enhanced inference-time control, enabling users to adapt the generated SVGs to their preferences, such as adjusting the color palette. We hope this work encourages further exploration into learning meaningful neural representations for vector graphics that are both practical for real-world design applications and provide users with greater flexibility through a more general representation.

\bibliographystyle{ACM-Reference-Format}
\bibliography{main}

\clearpage

\appendixpage

\section{Additional Details}

\paragraph{\textbf{Training Scheme}}
In the pretraining stage, we train the network for up to 300 steps using a constant learning rate of 0.01. For the full training process, we train for 4000 steps, employing a learning rate scheduler that features a linear warm-up from 0 to 0.018, followed by a cosine decay to a final value of 0.012. To improve training stability, we clip the gradients using a maximum norm of 0.1.

For computing the SDS loss, we utilize the Stable Diffusion 2.1 model from the diffusers library~\cite{diffusers}.

In all experiments, prompts are structured using the following format:
\begin{center}
    "A minimalist vector art of [object], isolated on a [color] background."
\end{center}
Here, [object] specifies the desired scene to be generated, and [color] represents the background color, which is either sampled during training or provided by the user at inference.

When applying our dropout technique, the indices are sampled as follows: with a probability of 0.7, all 16 shapes are rendered. Otherwise, the truncation index, between $1$ and $16$, is sampled from an exponential distribution with a temperature value of 3.

\paragraph{\textbf{LoRA Fine-Tuning}}
As detailed in the main paper, our SDS loss is applied with a LoRA adapter applied to Stable Diffusion 2.1. This adapter was pretrained on a high-quality dataset of vector art images. Specifically, the adapter was trained using 1,600 images spanning 145 different prompts, with minor variations between prompts (e.g., with different background colors). These images were generated using the Simple Vector Flux LoRA (see \texttt{renderartist/simplevectorflux} from diffusers.).

The LoRA adapter was trained for 15,000 steps with a rank of 4.

\newpage

\section{Additional Results and Comparisons}
\begin{figure*}[ht!]
    \centering
    \begin{minipage}[t]{0.5\linewidth} %
        \setlength{\tabcolsep}{0.5pt}
        \centering
        \small
\begin{tabular}{c c c c c}

            \multicolumn{5}{c}{``an owl standing on a wire''} \\
            \includegraphics[width=0.1875\linewidth,height=0.1875\linewidth]{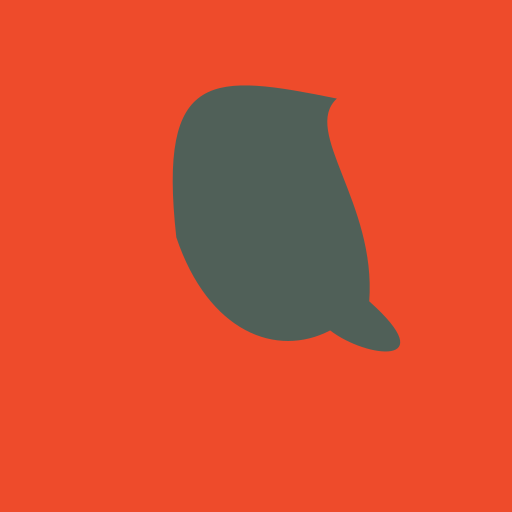} &
            \includegraphics[width=0.1875\linewidth,height=0.1875\linewidth]{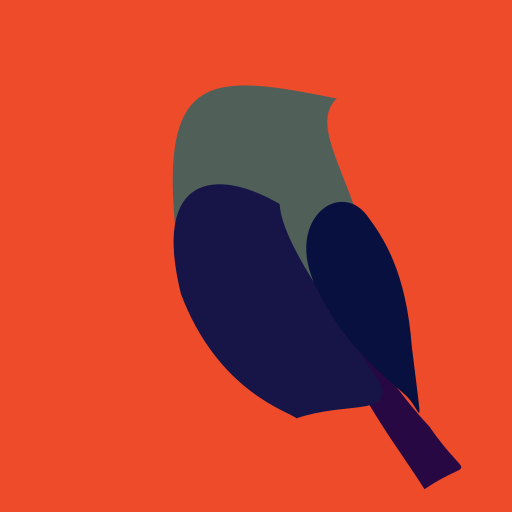} &
            \includegraphics[width=0.1875\linewidth,height=0.1875\linewidth]{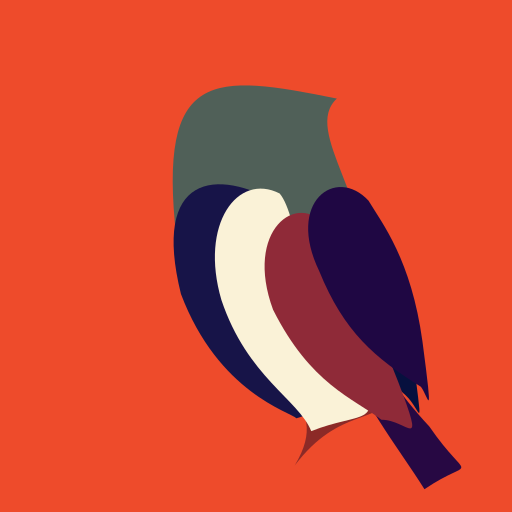} &
            \includegraphics[width=0.1875\linewidth,height=0.1875\linewidth]{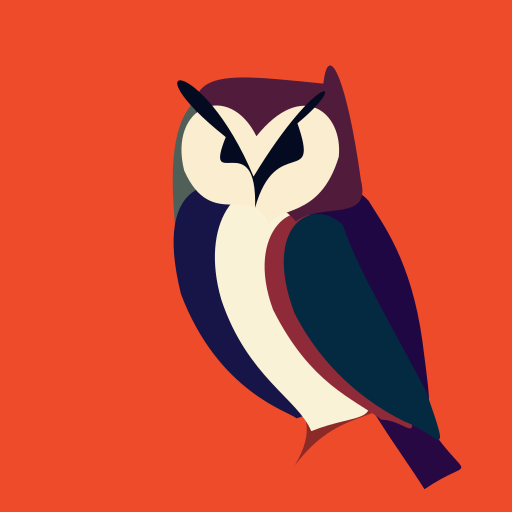} &
            \includegraphics[width=0.1875\linewidth,height=0.1875\linewidth]{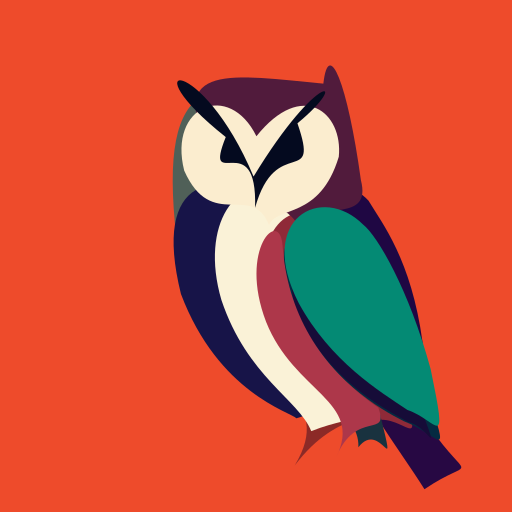} \\

            \multicolumn{5}{c}{``a knight holding a long sword''} \\
            \includegraphics[width=0.1875\linewidth,height=0.1875\linewidth]{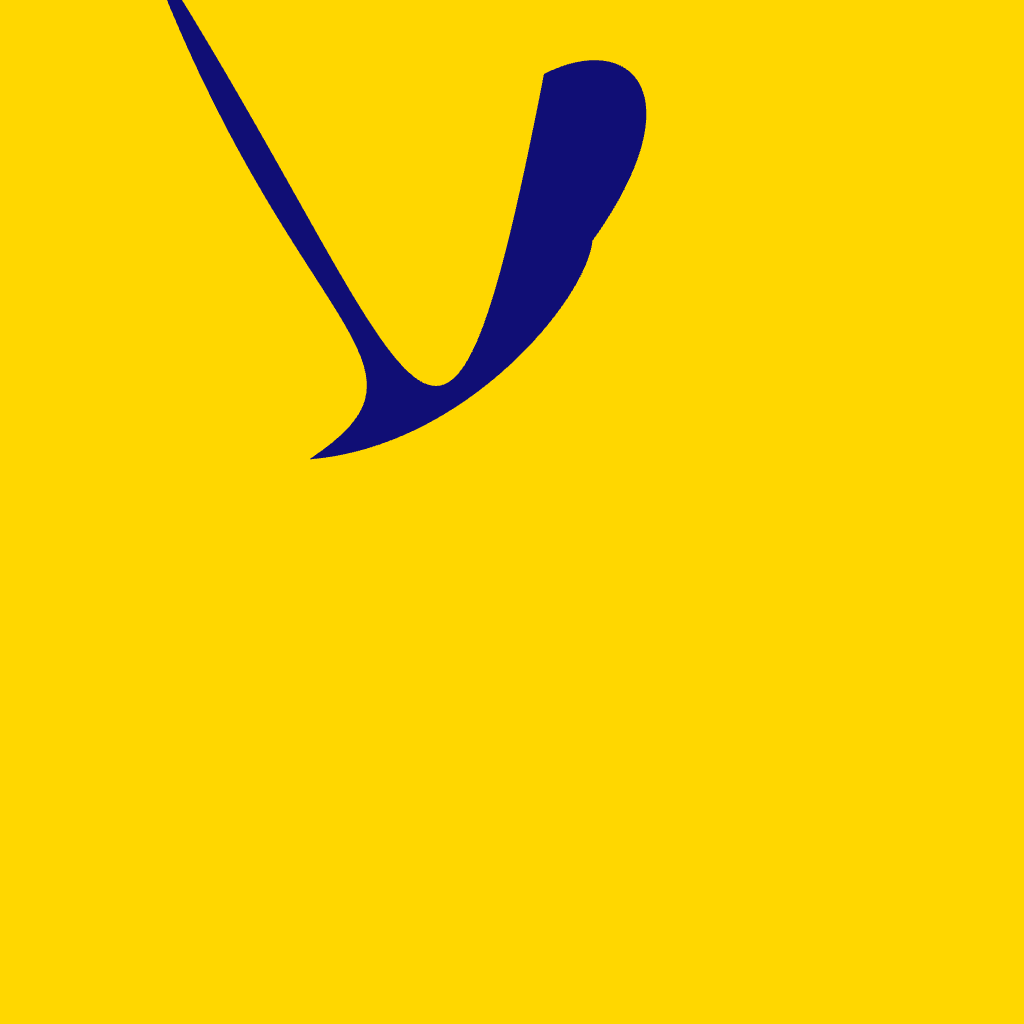} &
            \includegraphics[width=0.1875\linewidth,height=0.1875\linewidth]{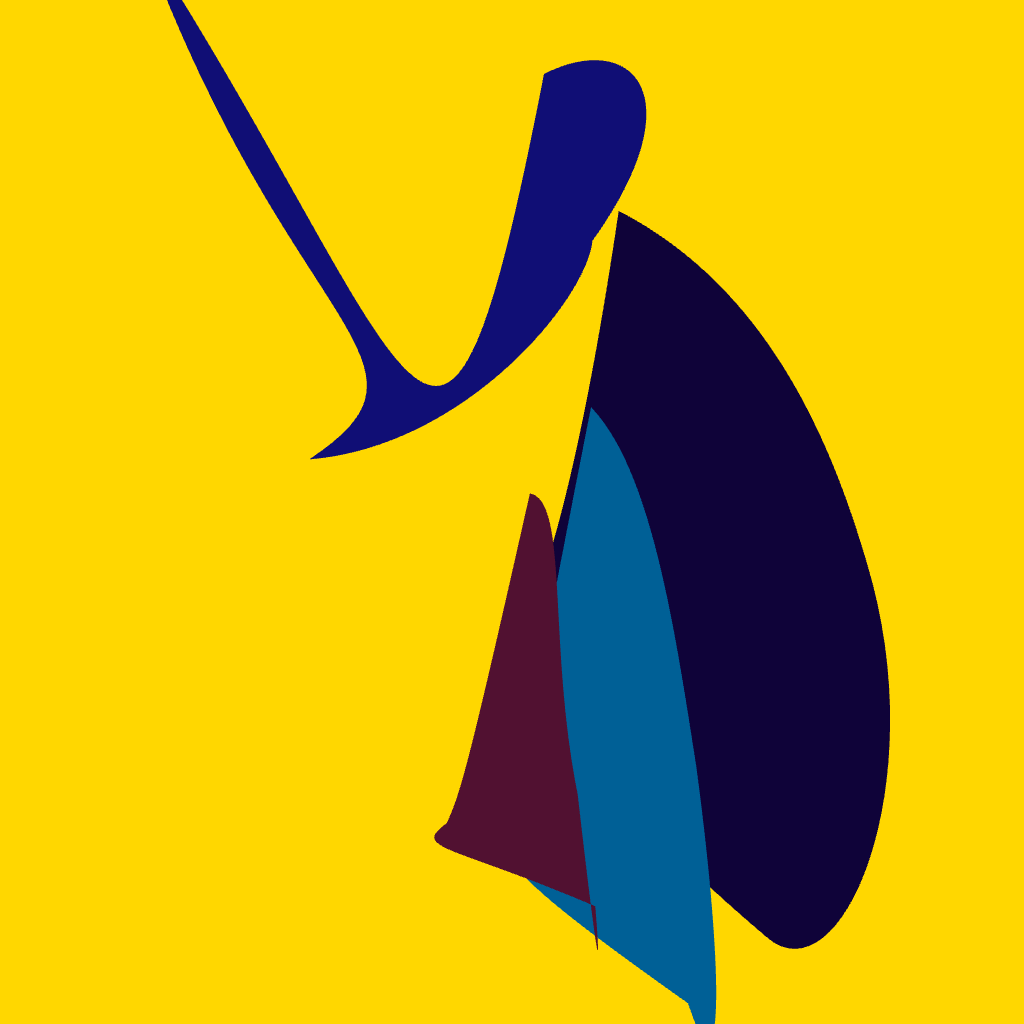} &
            \includegraphics[width=0.1875\linewidth,height=0.1875\linewidth]{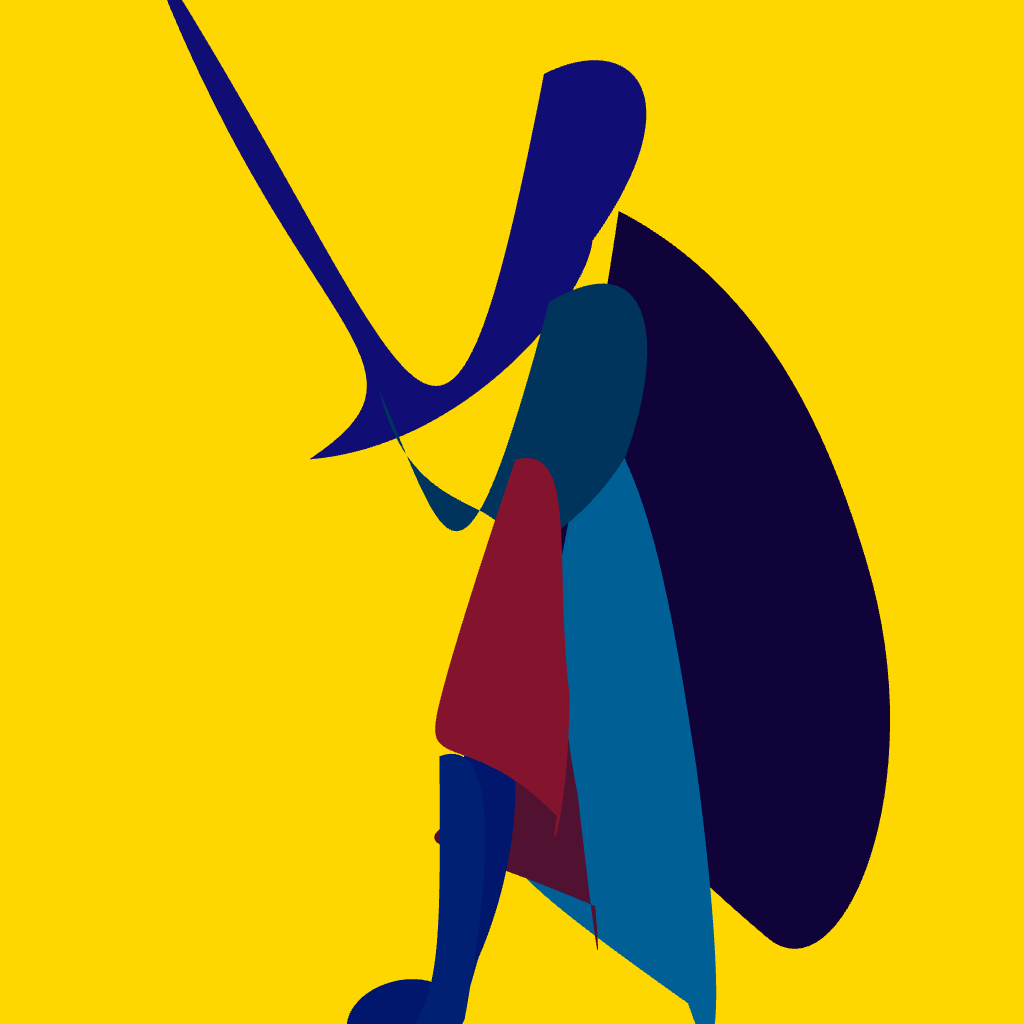} &
            \includegraphics[width=0.1875\linewidth,height=0.1875\linewidth]{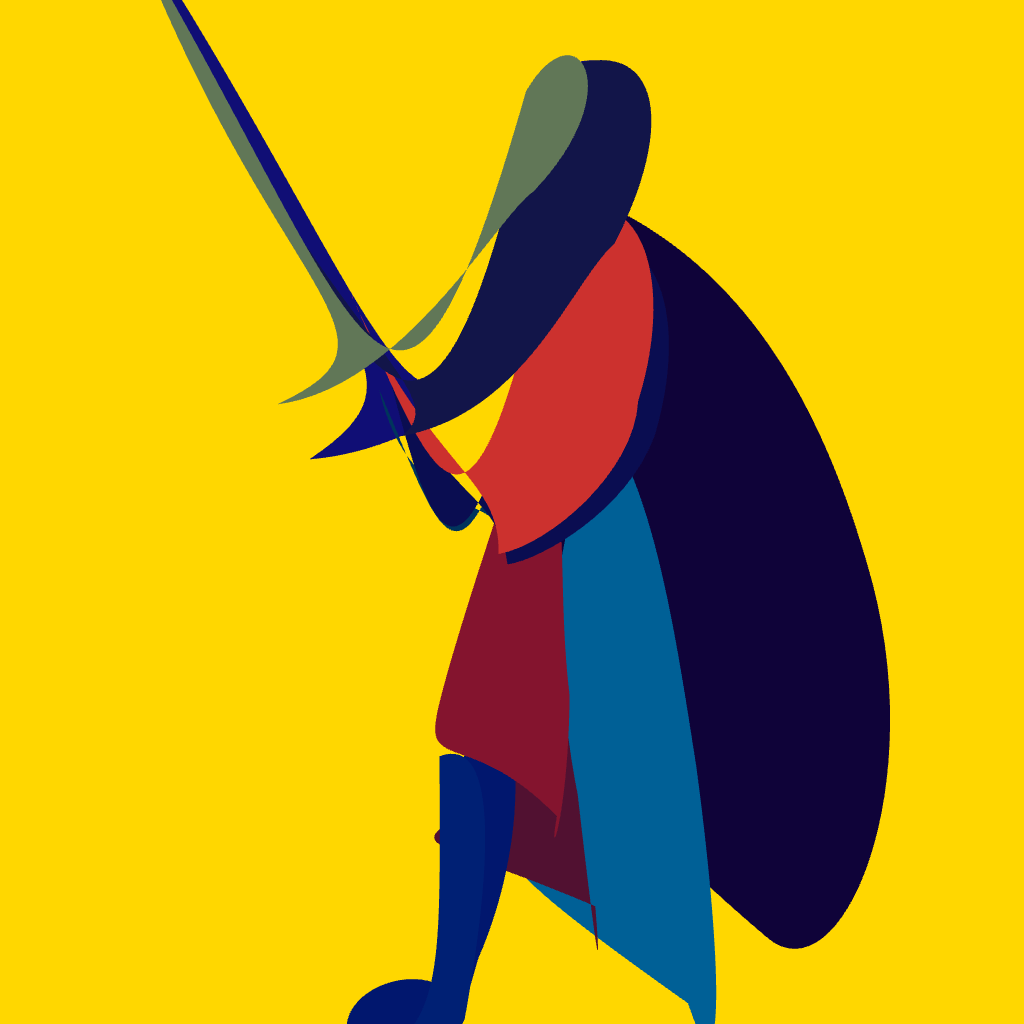} &
            \includegraphics[width=0.1875\linewidth,height=0.1875\linewidth]{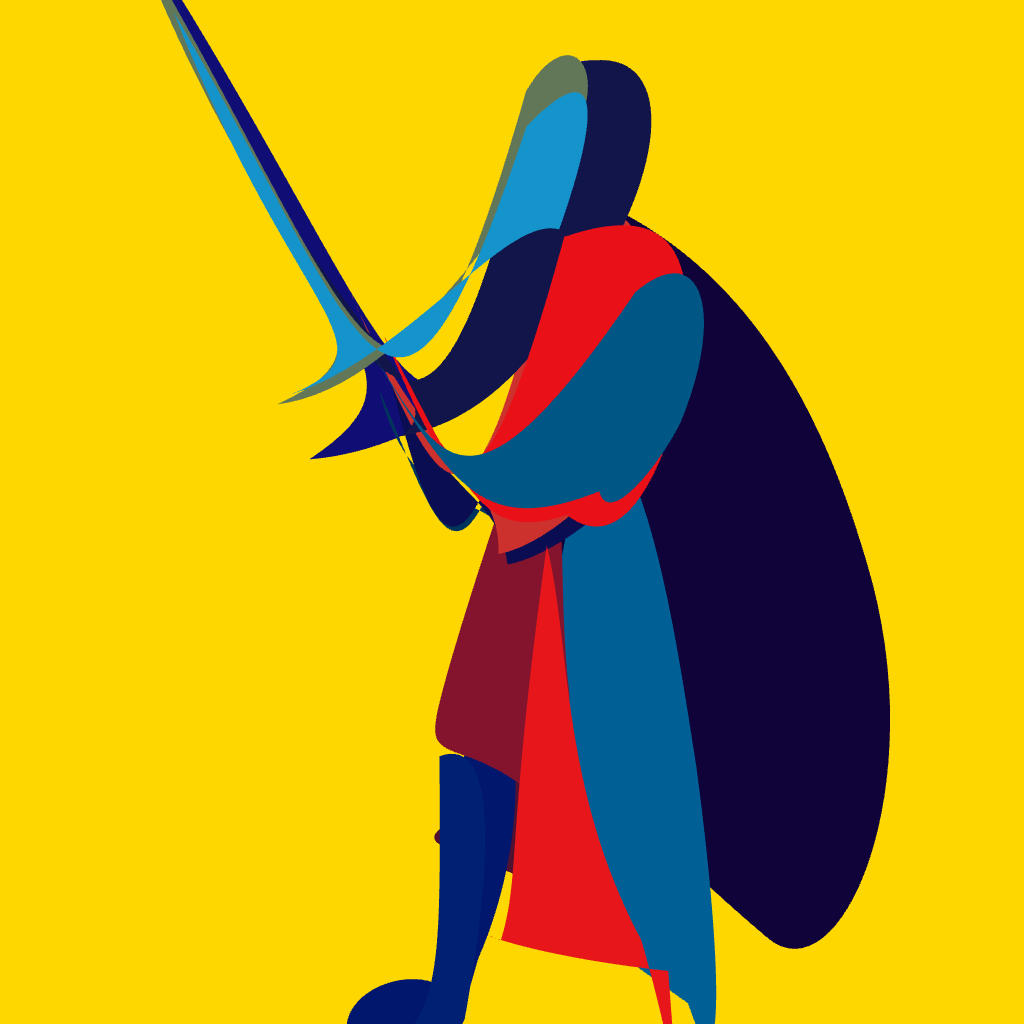} \\

            \multicolumn{5}{c}{``avocados''} \\
\end{tabular}
    \end{minipage}%
    \hfill %
    \begin{minipage}[t]{0.5\linewidth}
        \setlength{\tabcolsep}{0.5pt}
        \centering
        \small
        \begin{tabular}{c c c c c}
            \multicolumn{5}{c}{``a fox and a hare tangoing together''} \\
            \includegraphics[width=0.1875\linewidth,height=0.1875\linewidth]{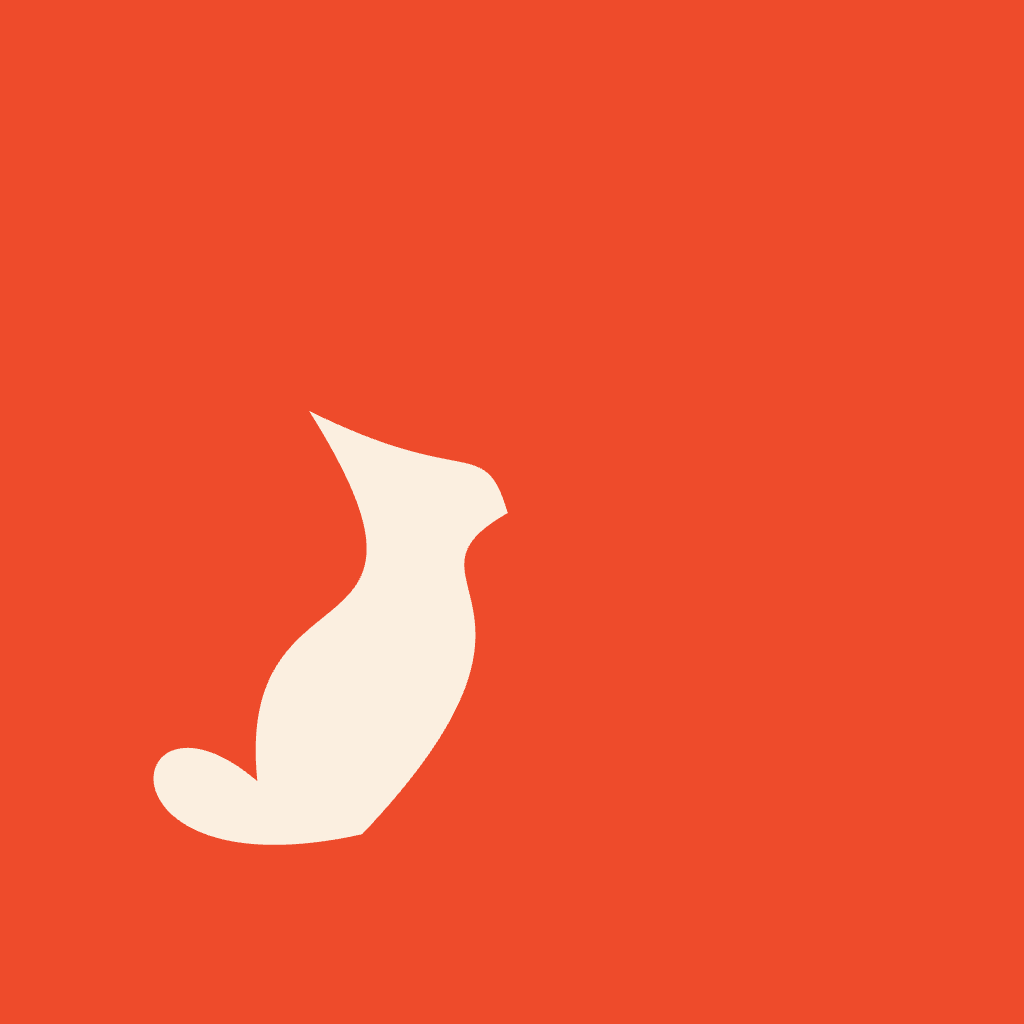} &
            \includegraphics[width=0.1875\linewidth,height=0.1875\linewidth]{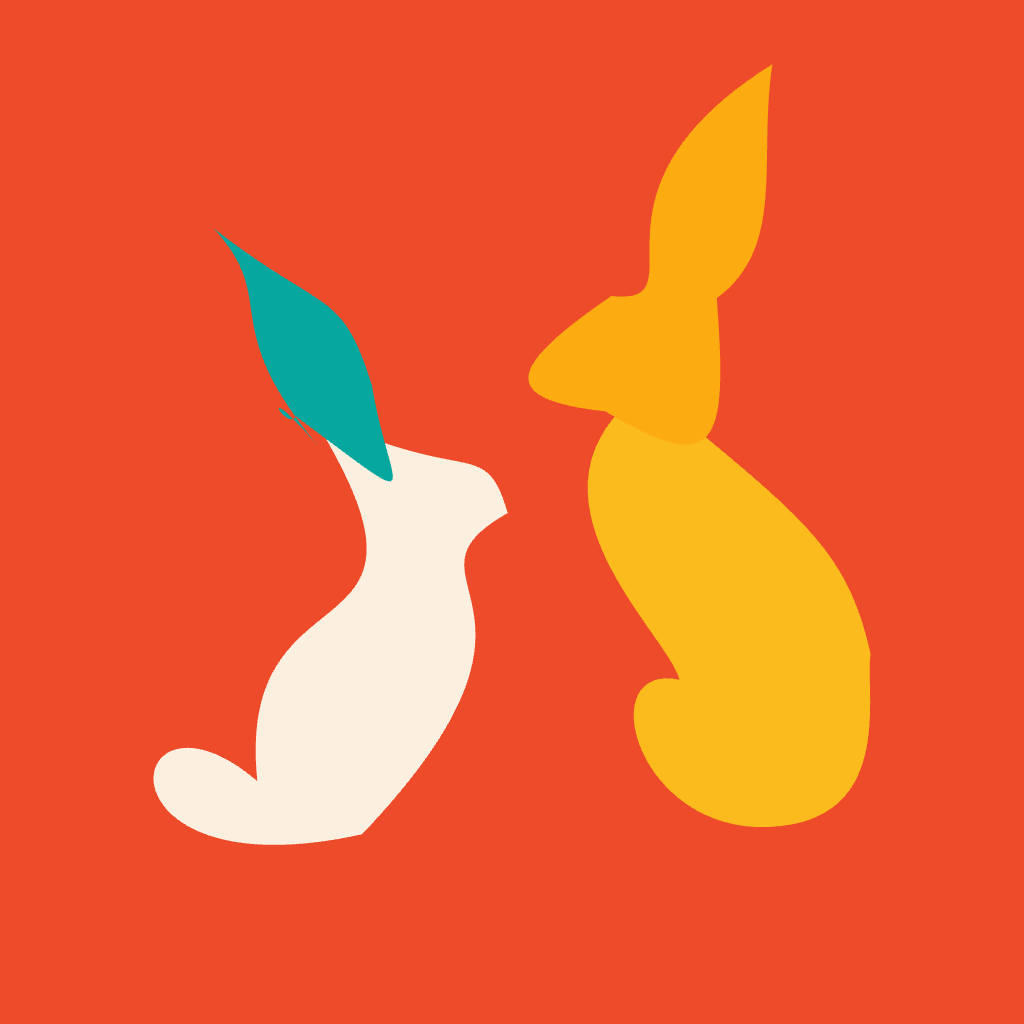} &
            \includegraphics[width=0.1875\linewidth,height=0.1875\linewidth]{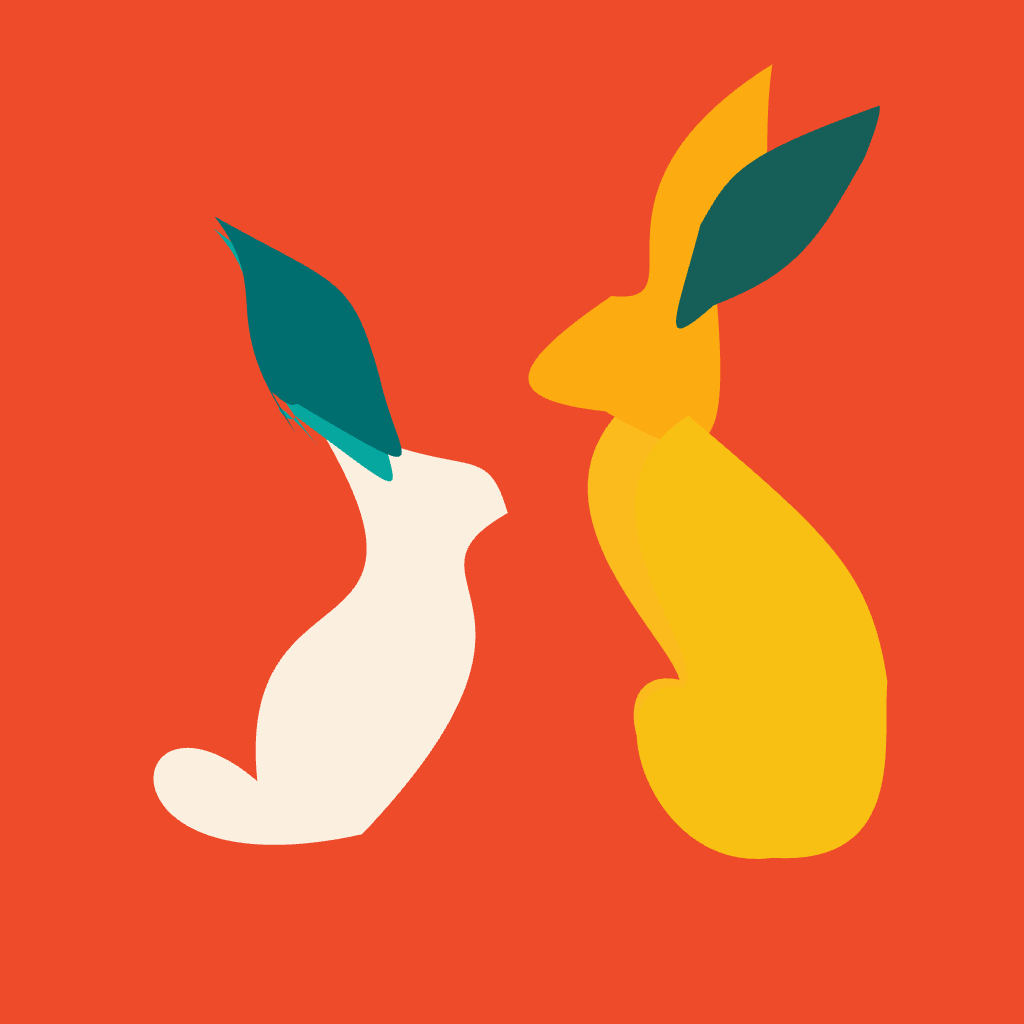} &
            \includegraphics[width=0.1875\linewidth,height=0.1875\linewidth]{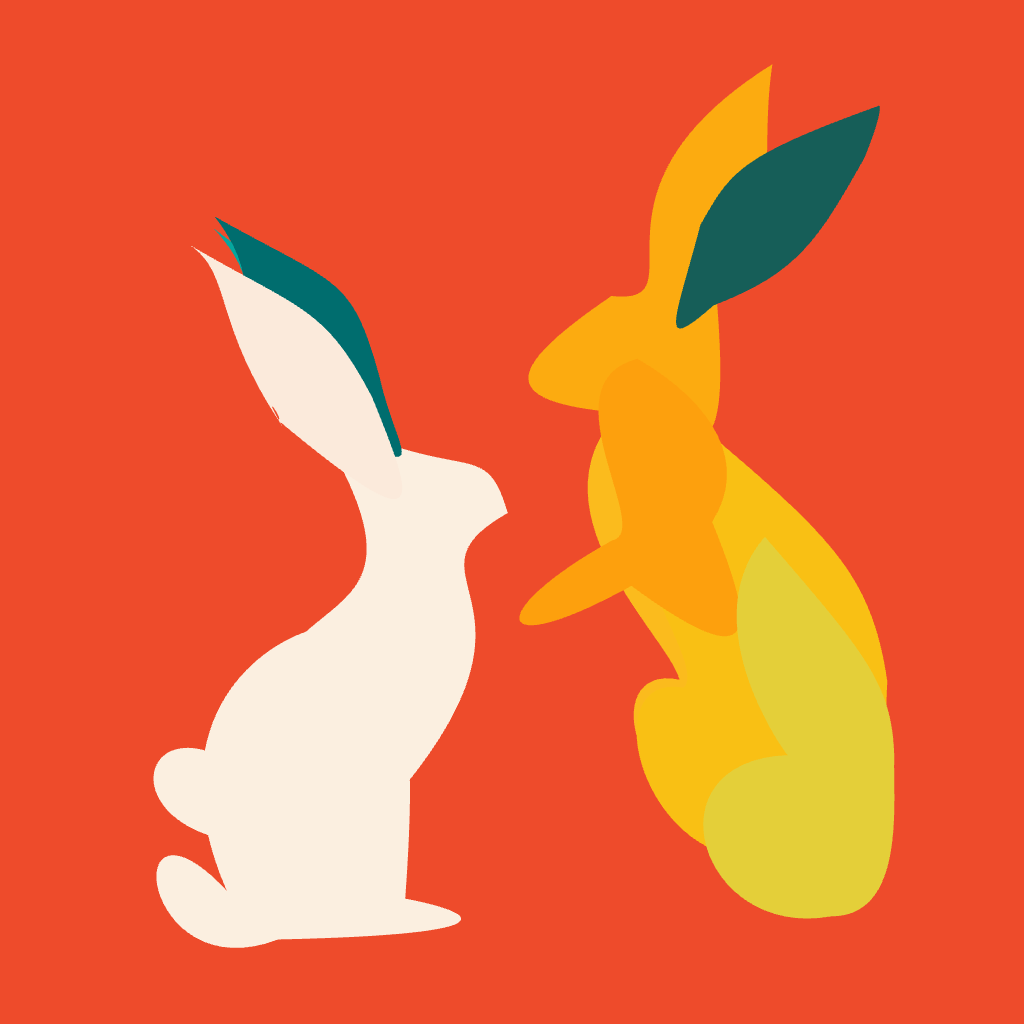} &
            \includegraphics[width=0.1875\linewidth,height=0.1875\linewidth]{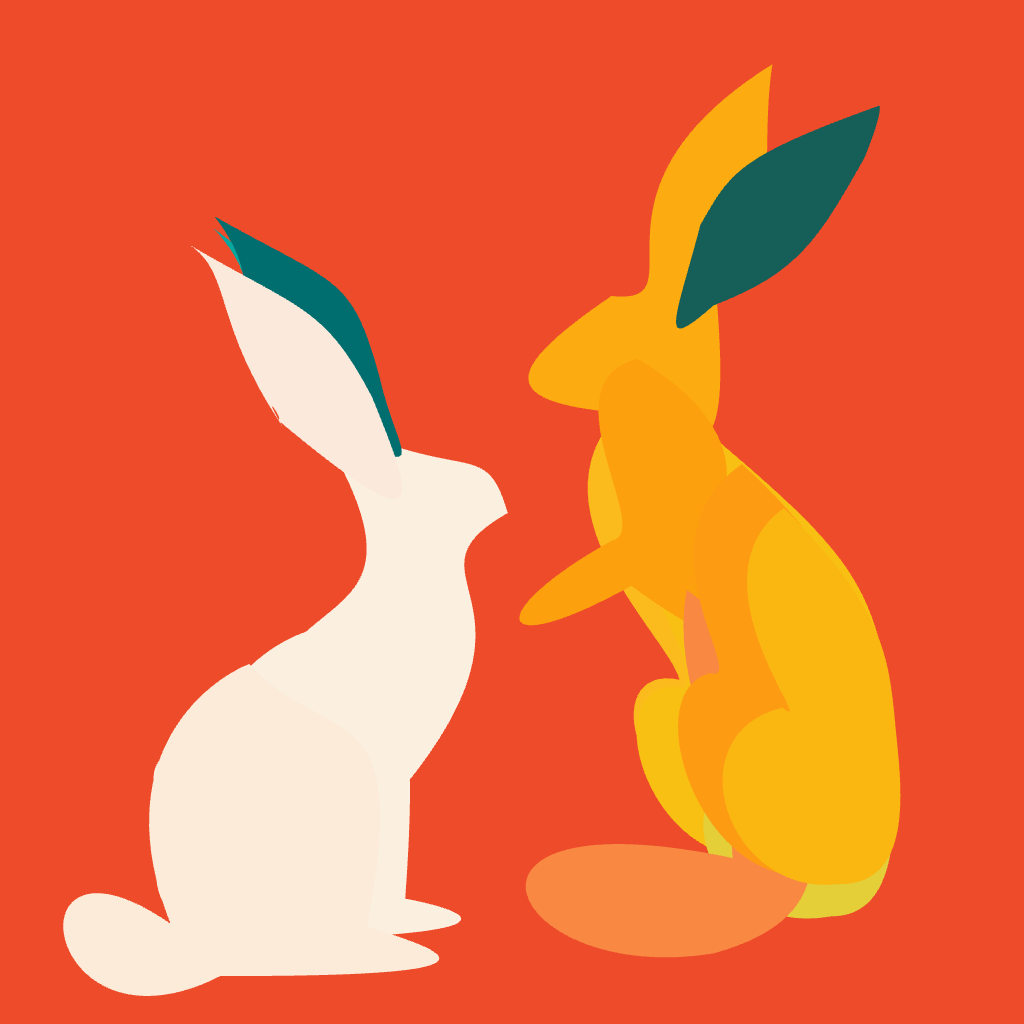} \\

            \multicolumn{5}{c}{``a fox playing the cello''} \\
            \includegraphics[width=0.1875\linewidth,height=0.1875\linewidth]{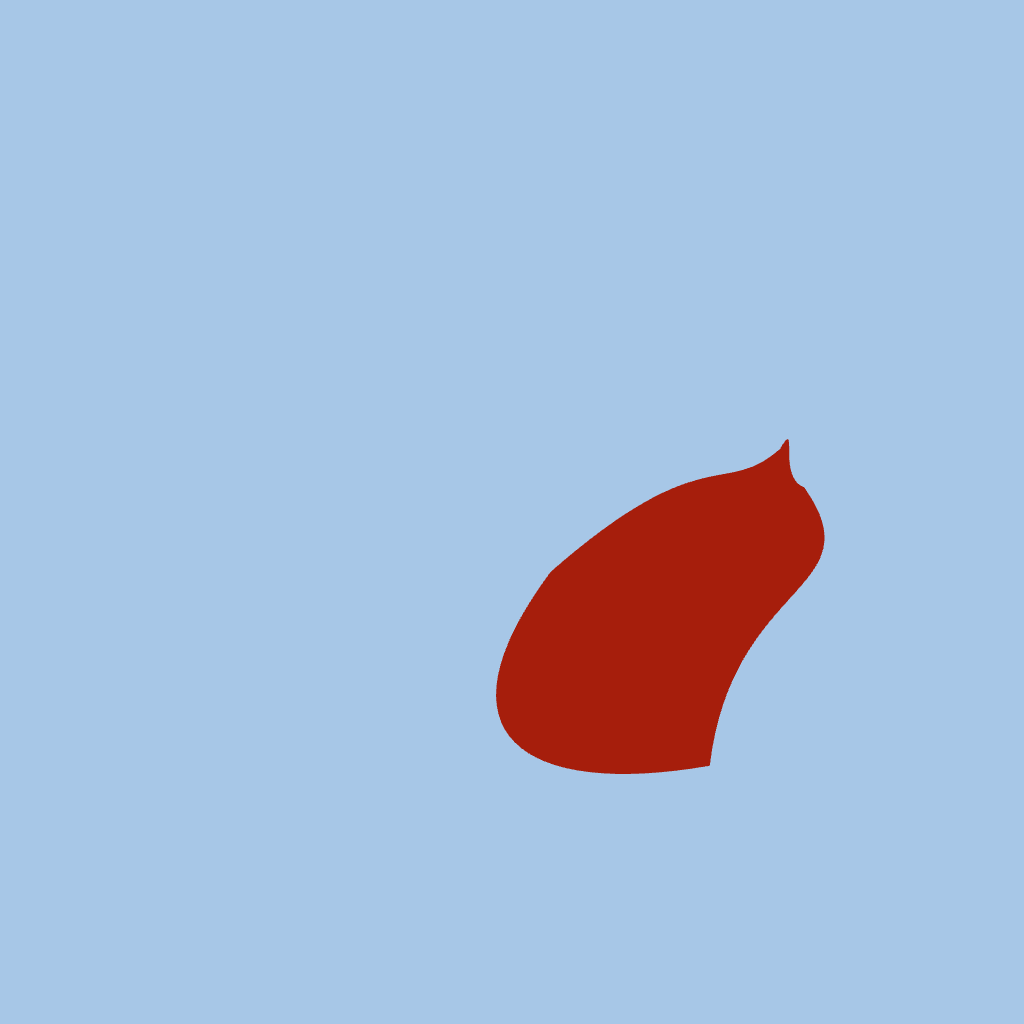} &
            \includegraphics[width=0.1875\linewidth,height=0.1875\linewidth]{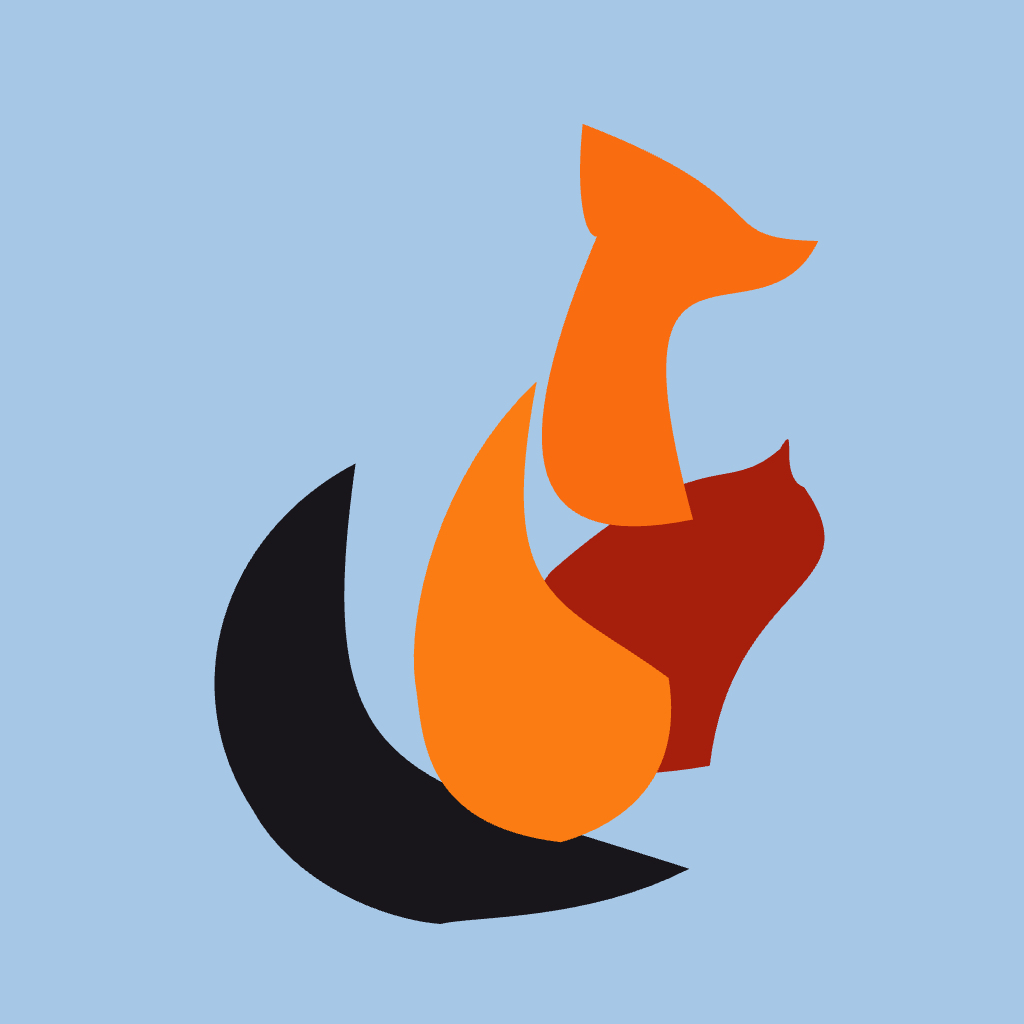} &
            \includegraphics[width=0.1875\linewidth,height=0.1875\linewidth]{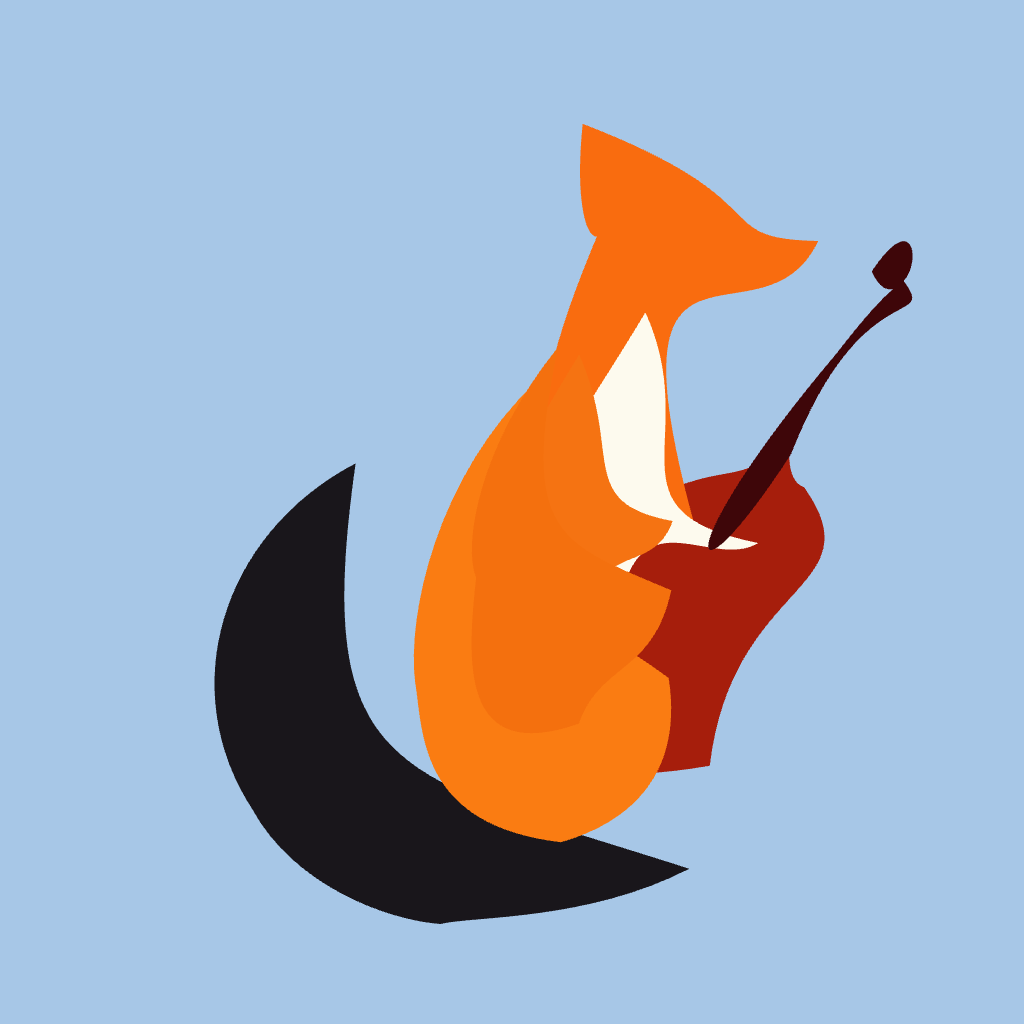} &
            \includegraphics[width=0.1875\linewidth,height=0.1875\linewidth]{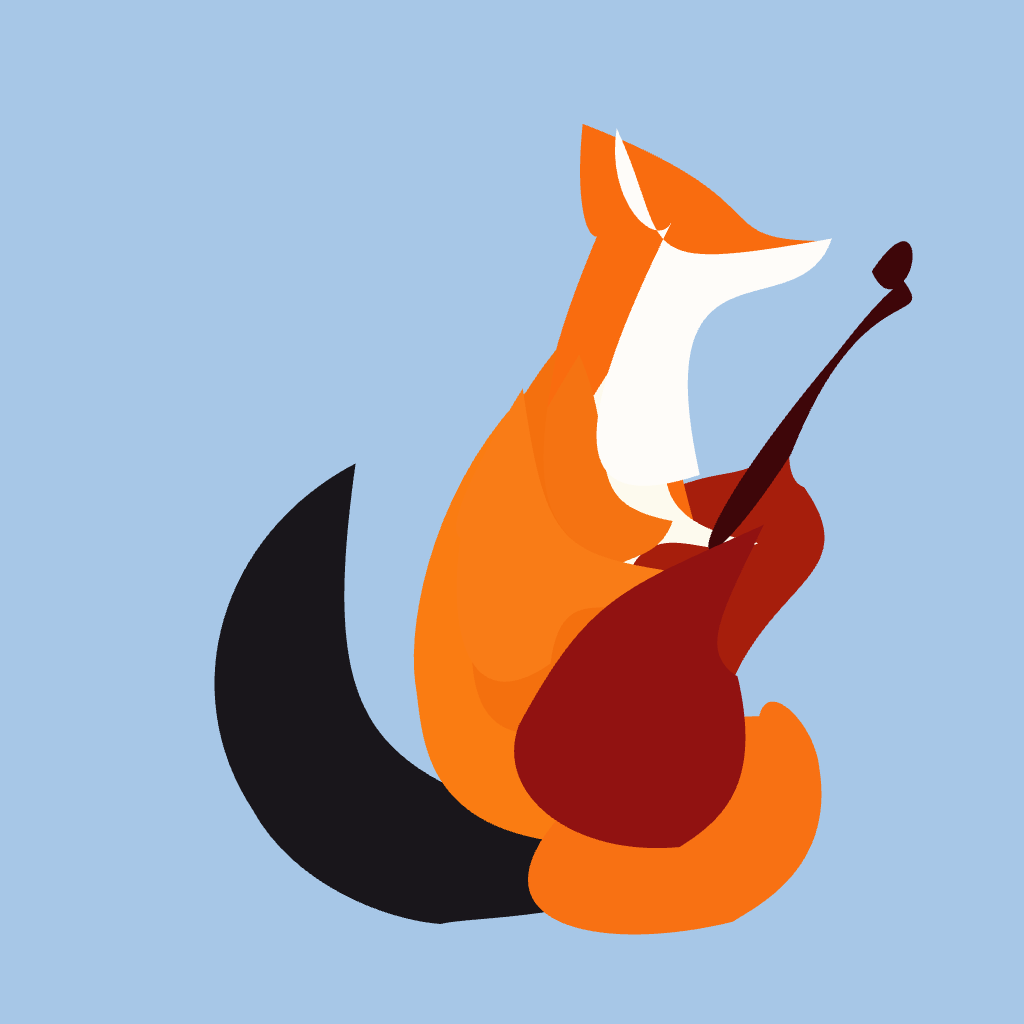} &
            \includegraphics[width=0.1875\linewidth,height=0.1875\linewidth]{images/dropout_results/fox_playing_cello/16.png} \\

            \multicolumn{5}{c}{``a girl with dress and a sun hat''} \\
\end{tabular}
    \end{minipage}
    \caption{\textbf{Additional Qualitative Results Obtained with NeuralSVG.} We show results generated by our method when keeping a varying number of learned shapes in the final rendering.}
    \label{fig:additional_results_dropout}
\end{figure*}

\begin{figure*}[ht!]
    \centering
    \begin{minipage}[t]{0.5\linewidth} %
        \setlength{\tabcolsep}{0.5pt}
        \centering
        \small
        \begin{tabular}{c c c c c}

            \multicolumn{5}{c}{``a great gray owl with a mouse in its beak''} \\
            \includegraphics[width=0.1875\linewidth,height=0.1875\linewidth]{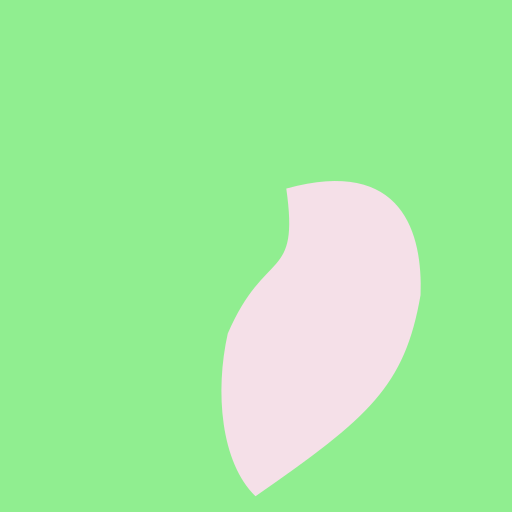} &
            \includegraphics[width=0.1875\linewidth,height=0.1875\linewidth]{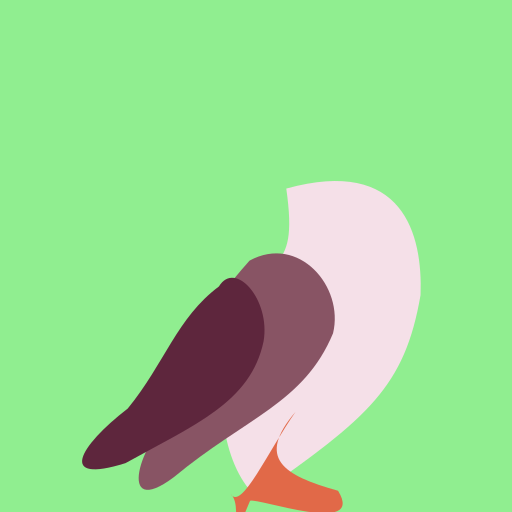} &
            \includegraphics[width=0.1875\linewidth,height=0.1875\linewidth]{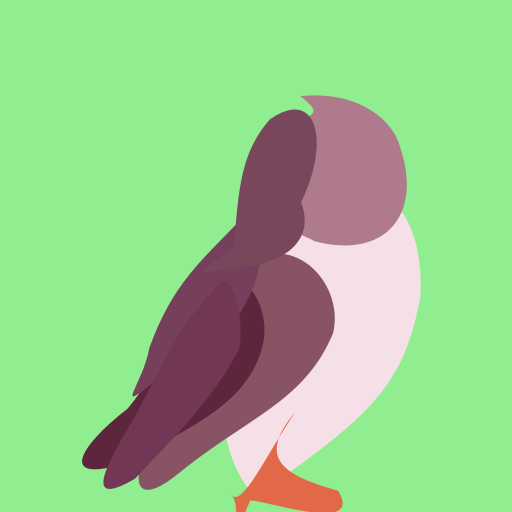} &
            \includegraphics[width=0.1875\linewidth,height=0.1875\linewidth]{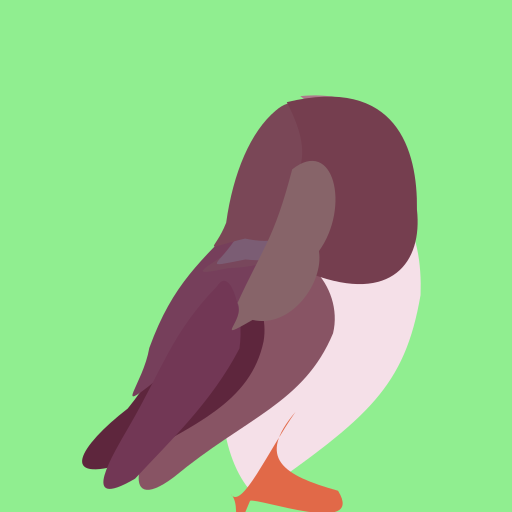} &
            \includegraphics[width=0.1875\linewidth,height=0.1875\linewidth]{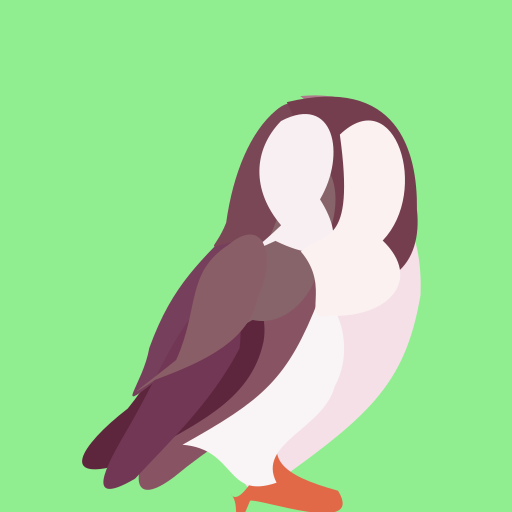} \\

            \multicolumn{5}{c}{``a sailboat''} \\
\end{tabular}
    \end{minipage}%
    \hfill %
    \begin{minipage}[t]{0.5\linewidth}
        \setlength{\tabcolsep}{0.5pt}
        \centering
        \small
        \begin{tabular}{c c c c c}

            \multicolumn{5}{c}{``a margarita''} \\
            \includegraphics[width=0.1875\linewidth,height=0.1875\linewidth]{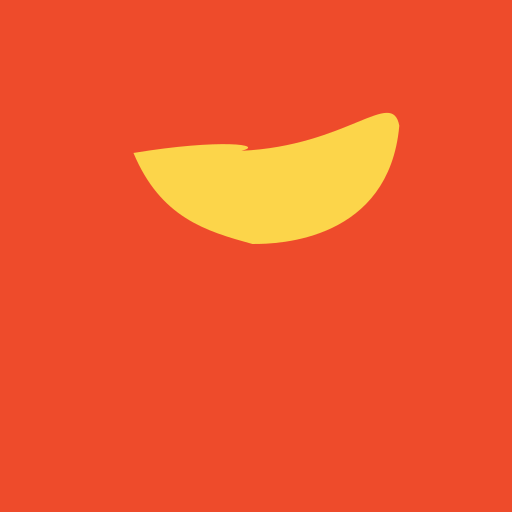} &
            \includegraphics[width=0.1875\linewidth,height=0.1875\linewidth]{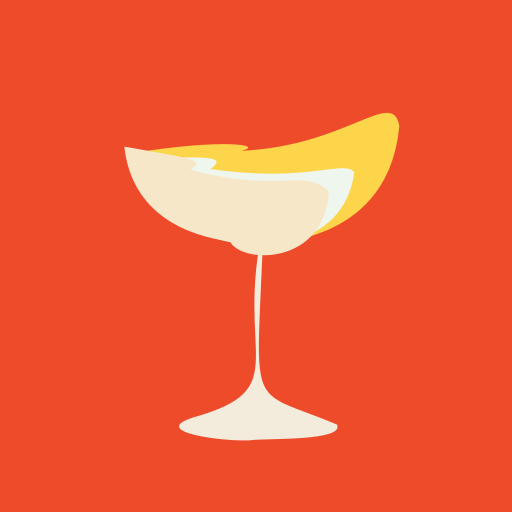} &
            \includegraphics[width=0.1875\linewidth,height=0.1875\linewidth]{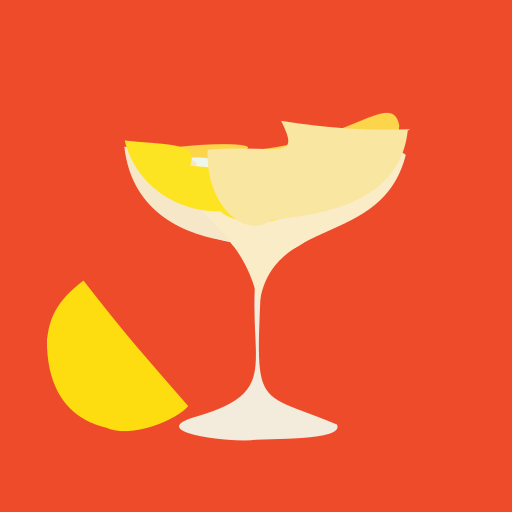} &
            \includegraphics[width=0.1875\linewidth,height=0.1875\linewidth]{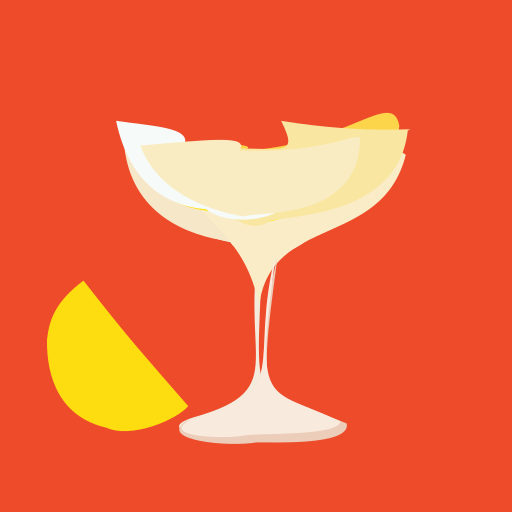} &
            \includegraphics[width=0.1875\linewidth,height=0.1875\linewidth]{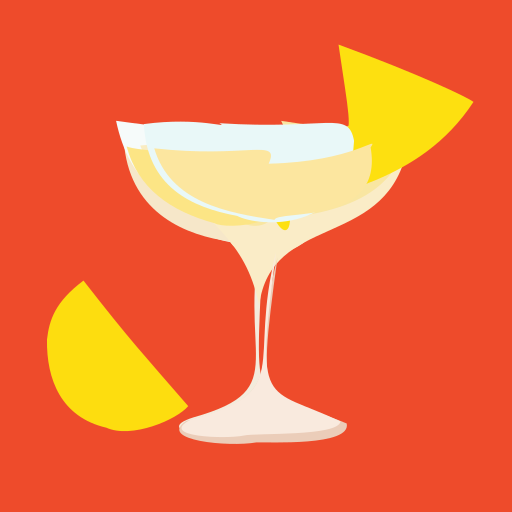} \\

            \multicolumn{5}{c}{``a peacock''} \\
            \includegraphics[width=0.1875\linewidth,height=0.1875\linewidth]{images/dropout_results/peacock/gold_1_shapes.png} &
            \includegraphics[width=0.1875\linewidth,height=0.1875\linewidth]{images/dropout_results/peacock/gold_4_shapes.png} &
            \includegraphics[width=0.1875\linewidth,height=0.1875\linewidth]{images/dropout_results/peacock/gold_8_shapes.png} &
            \includegraphics[width=0.1875\linewidth,height=0.1875\linewidth]{images/dropout_results/peacock/gold_12_shapes.png} &
            \includegraphics[width=0.1875\linewidth,height=0.1875\linewidth]{images/dropout_results/peacock/gold_16_shapes.png} \\

            \multicolumn{5}{c}{``a picture of a macaw''} \\
\end{tabular}
    \end{minipage}
    \caption{\textbf{Additional Qualitative Results Obtained with NeuralSVG.} We show results generated by our method when keeping a varying number of learned shapes in the final rendering.}
    \label{fig:additional_results_dropout_2}
\end{figure*}

\begin{figure*}
    \centering
    \setlength{\tabcolsep}{1pt}
    {\small
    \begin{tabular}{c c @{\hspace{1cm}} c c @{\hspace{0.2cm}} c c}

        \multicolumn{2}{c}{\begin{tabular}{c} ``a cake with chocolate \\ frosting and cherry'' \end{tabular}} &
        \multicolumn{2}{c}{``a boat''} &
        \multicolumn{2}{c}{\begin{tabular}{c} ``The Statue of Liberty \\ with the face of an owl'' \end{tabular}} \\
        \includegraphics[height=0.10\textwidth,width=0.10\textwidth]{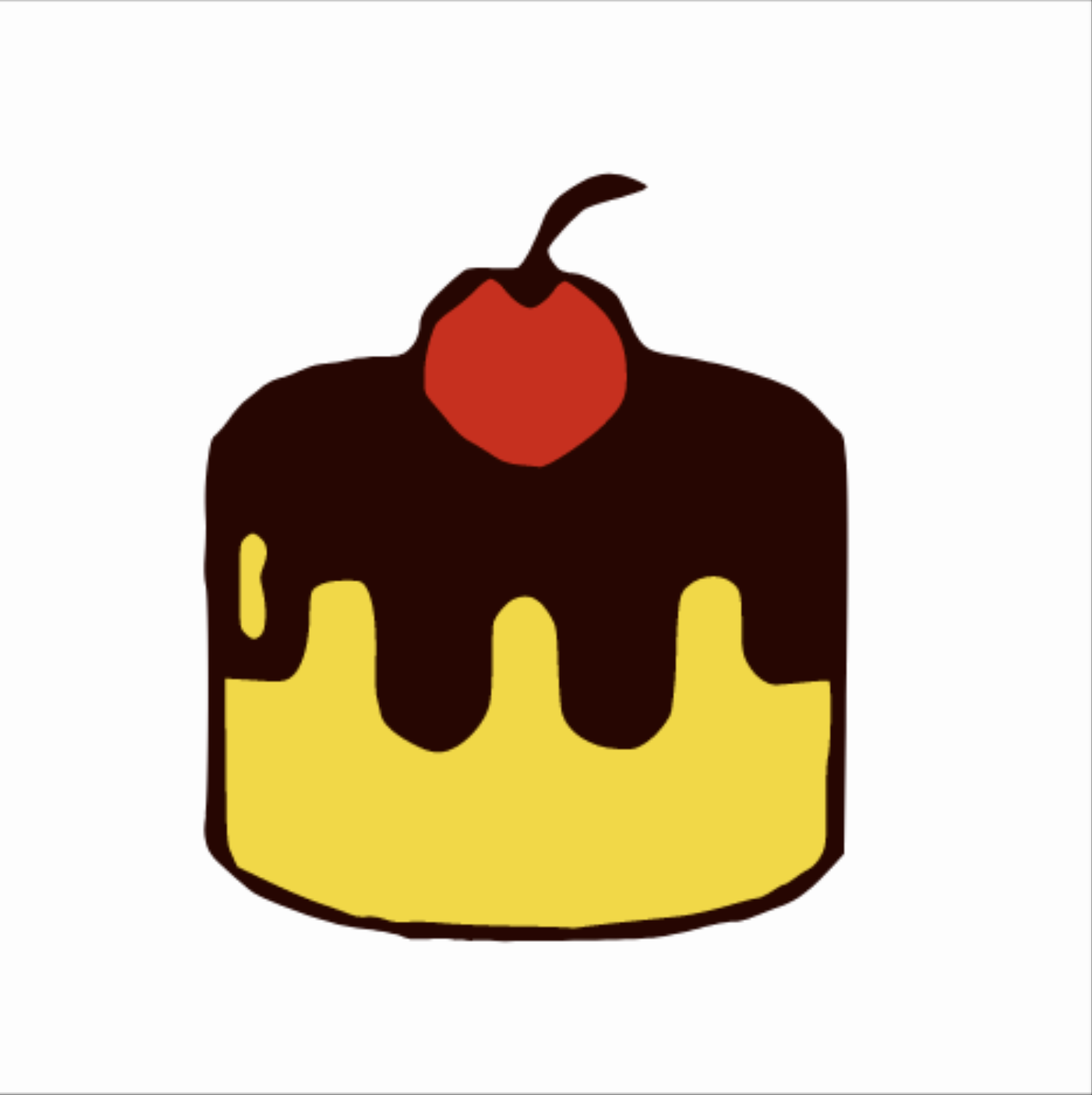} &
        \includegraphics[height=0.10\textwidth,width=0.10\textwidth]{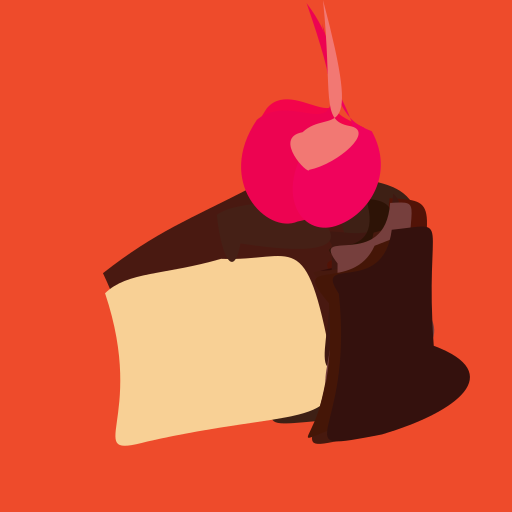} &
        \includegraphics[height=0.10\textwidth,width=0.10\textwidth]{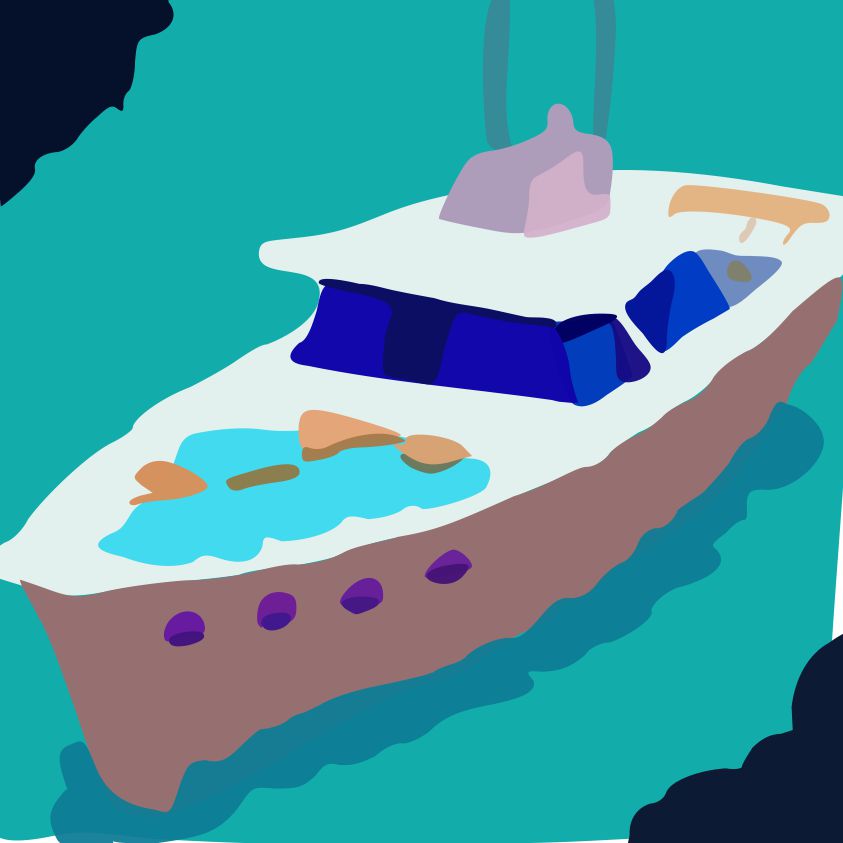} &
        \includegraphics[height=0.10\textwidth,width=0.10\textwidth]{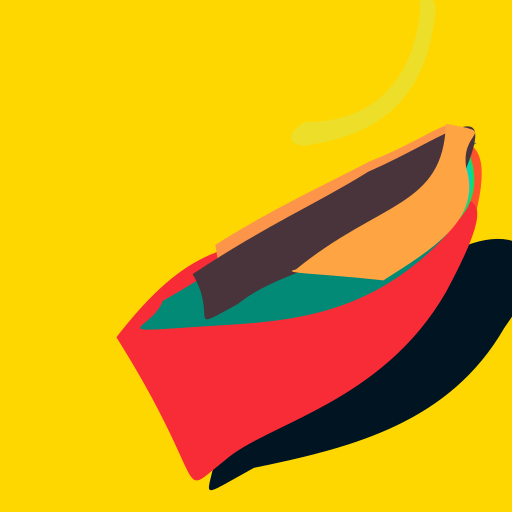} &
        \includegraphics[height=0.10\textwidth,width=0.10\textwidth]{images/text_to_vector/the_statue_of_liberty_with_the_face_of_an_owl.jpg} &
        \includegraphics[height=0.10\textwidth,width=0.10\textwidth]{images/ours_for_nivel_text2vector/owl_liberty.png} \\

        \multicolumn{2}{c}{``a 3D rendering of a temple''} &
        \multicolumn{2}{c}{``a crown''} &
        \multicolumn{2}{c}{``a torii gate''} \\
        \includegraphics[height=0.10\textwidth,width=0.10\textwidth]{images/nivel/temple_3D_rendering.png} &
        \includegraphics[height=0.10\textwidth,width=0.10\textwidth]{images/colors/temple/light-blue_16_shapes.png} &
        \includegraphics[height=0.10\textwidth,width=0.10\textwidth]{images/text_to_vector/a_crown.jpg} &
        \includegraphics[height=0.10\textwidth,width=0.10\textwidth]{images/ours_for_nivel_text2vector/crown.png} &
        \includegraphics[height=0.10\textwidth,width=0.10\textwidth]{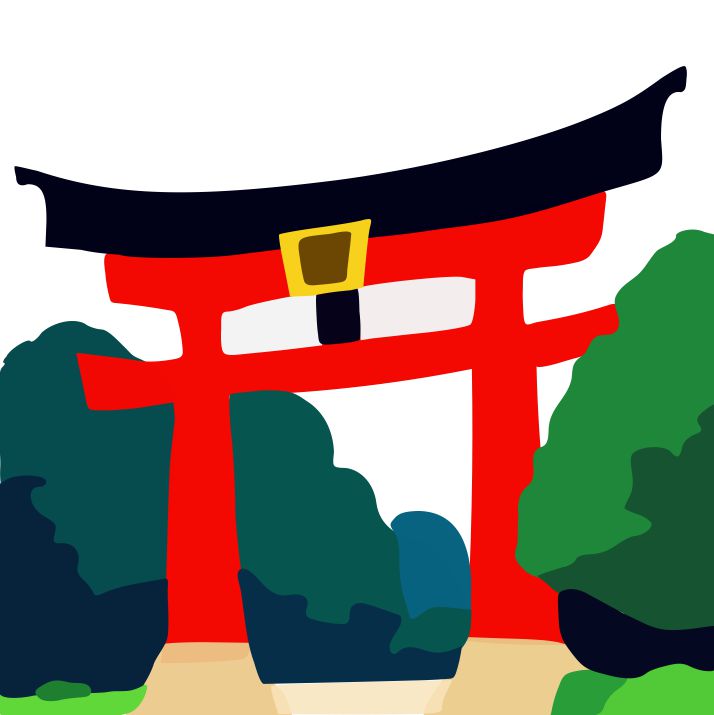} &
        \includegraphics[height=0.10\textwidth,width=0.10\textwidth]{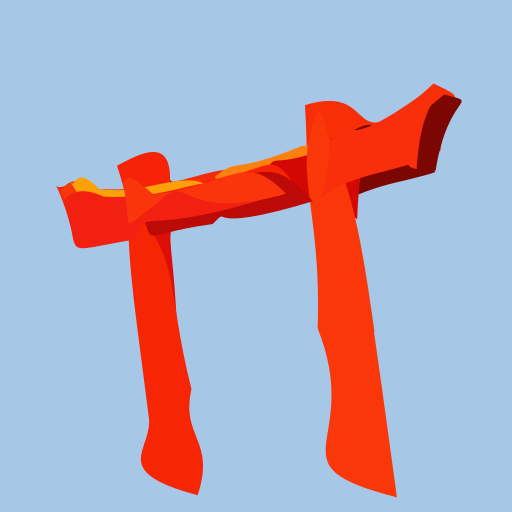} \\

        \multicolumn{2}{c}{``a green dragon breathing fire''} &
        \multicolumn{2}{c}{``a giraffe in street''} &
        \multicolumn{2}{c}{``Vincent Van Gogh''} \\
        \includegraphics[height=0.10\textwidth,width=0.10\textwidth]{images/nivel/green_dragon_breathing_fire.png} &
        \includegraphics[height=0.10\textwidth,width=0.10\textwidth]{images/ours_for_nivel_text2vector/green_dragon.png} &
        \includegraphics[height=0.10\textwidth,width=0.10\textwidth]{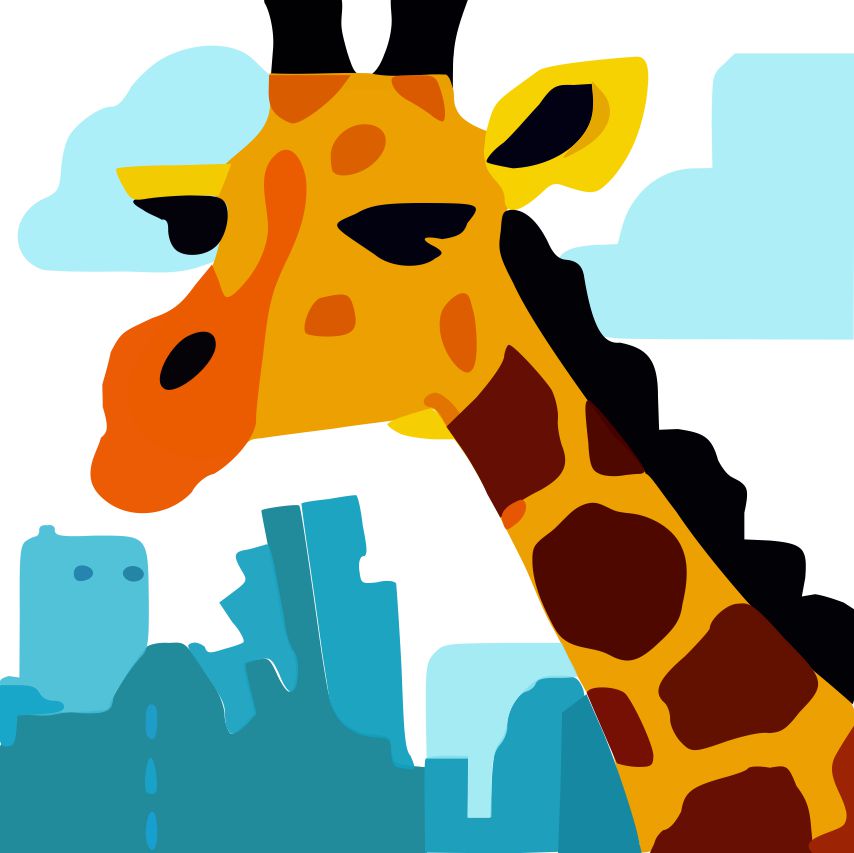} &
        \includegraphics[height=0.10\textwidth,width=0.10\textwidth]{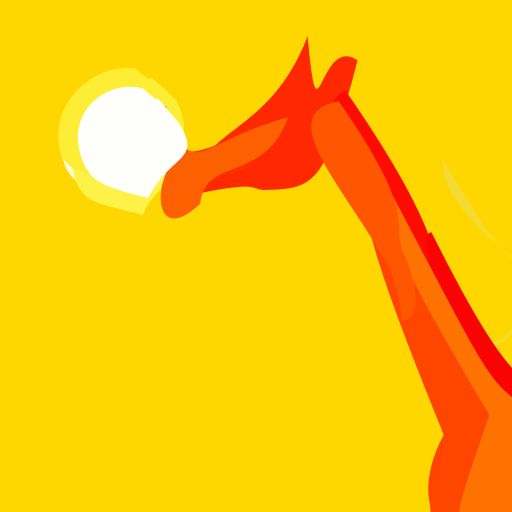} &
        \includegraphics[height=0.10\textwidth,width=0.10\textwidth]{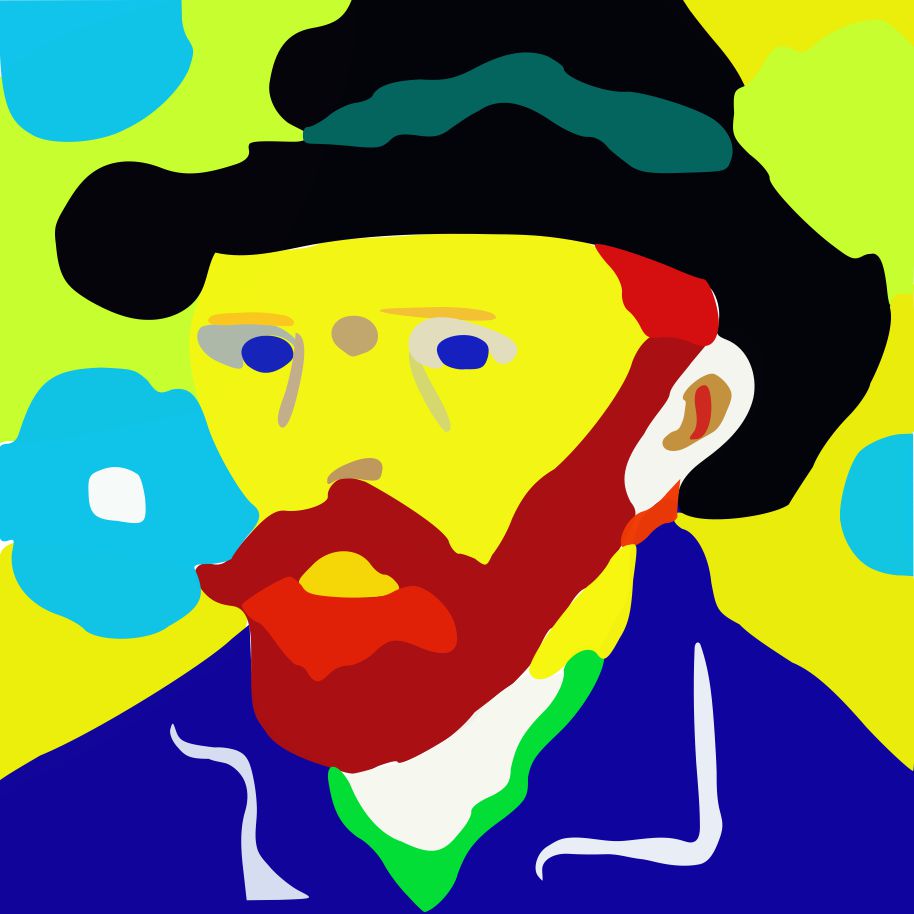} &
        \includegraphics[height=0.10\textwidth,width=0.10\textwidth]{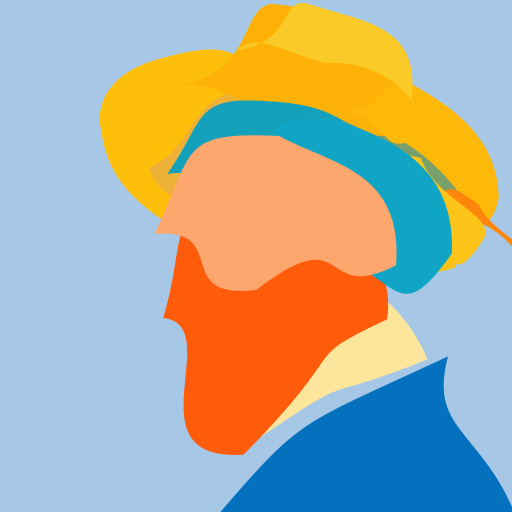} \\

        \multicolumn{2}{c}{``a walrus smoking a pipe''} &
        \multicolumn{2}{c}{``a Ming Dynasty vase''} &
        \multicolumn{2}{c}{``an erupting volcano''} \\
        \includegraphics[height=0.10\textwidth,width=0.10\textwidth]{images/nivel/walrus_smoking_pipe_12K_parameters.png} &
        \includegraphics[height=0.10\textwidth,width=0.10\textwidth]{images/colors/walrus/final_svg_gold.png} &
        \includegraphics[height=0.10\textwidth,width=0.10\textwidth]{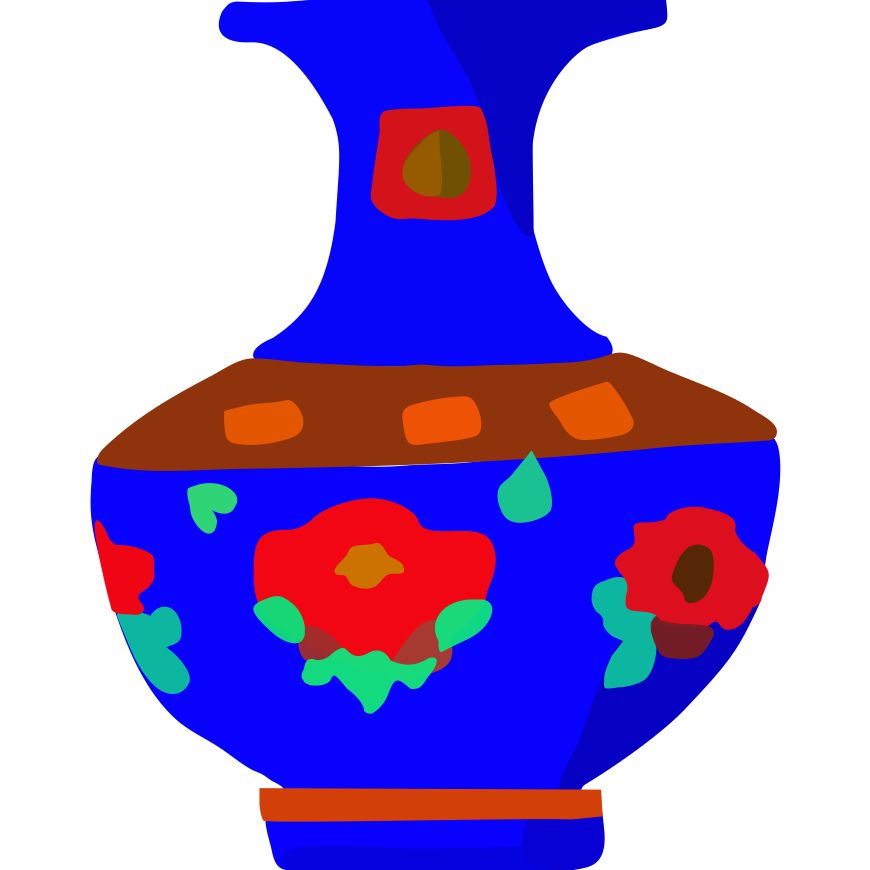} &
        \includegraphics[height=0.10\textwidth,width=0.10\textwidth]{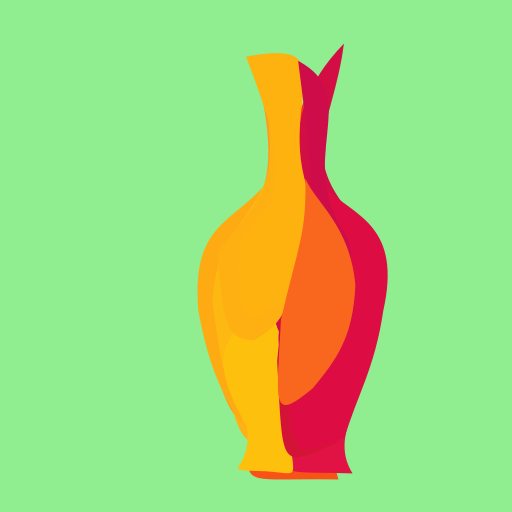} &
        \includegraphics[height=0.10\textwidth,width=0.10\textwidth]{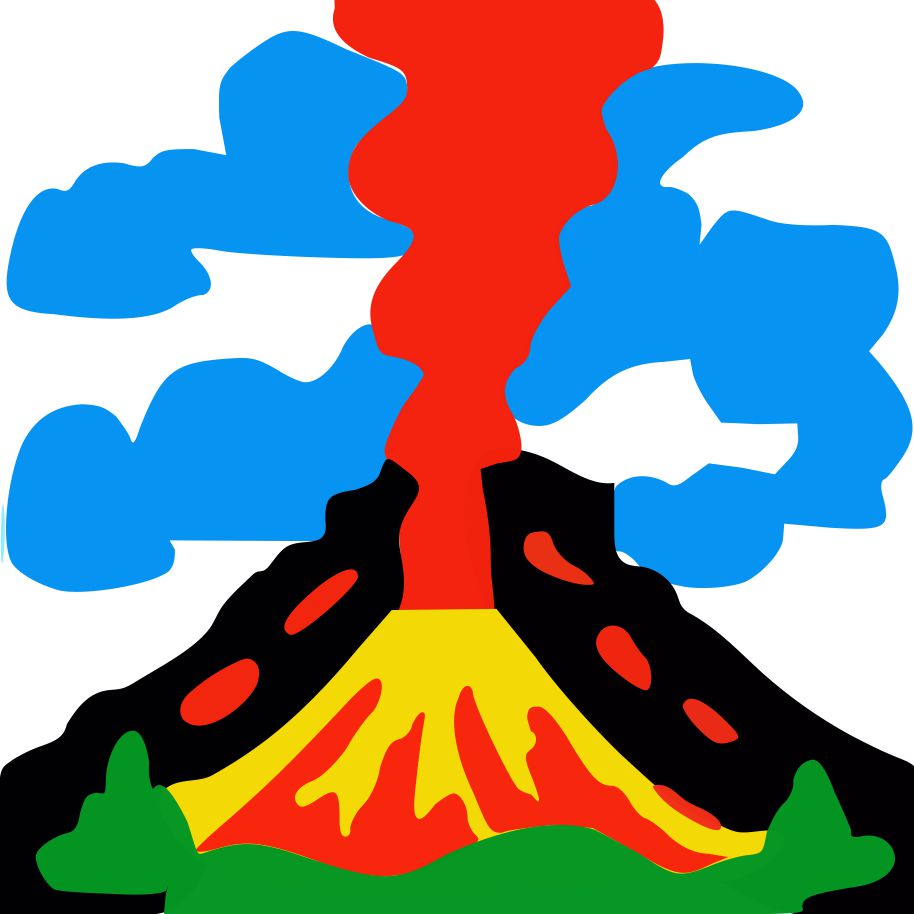} &
        \includegraphics[height=0.10\textwidth,width=0.10\textwidth]{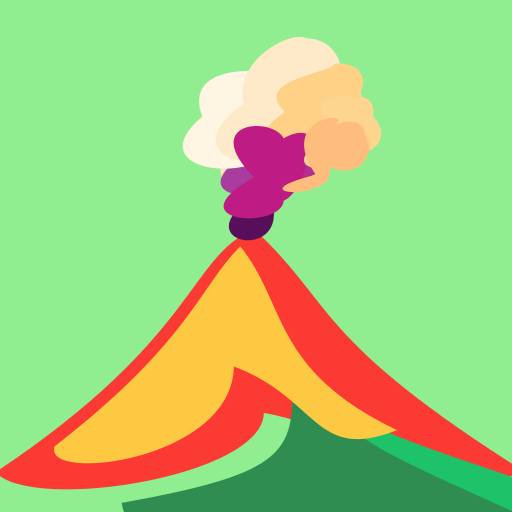} \\

        \multicolumn{2}{c}{``a vintage camera''} &
        \multicolumn{2}{c}{``a picture of Tokyo''} &
        \multicolumn{2}{c}{``a cruise ship''} \\
        \includegraphics[height=0.10\textwidth,width=0.10\textwidth]{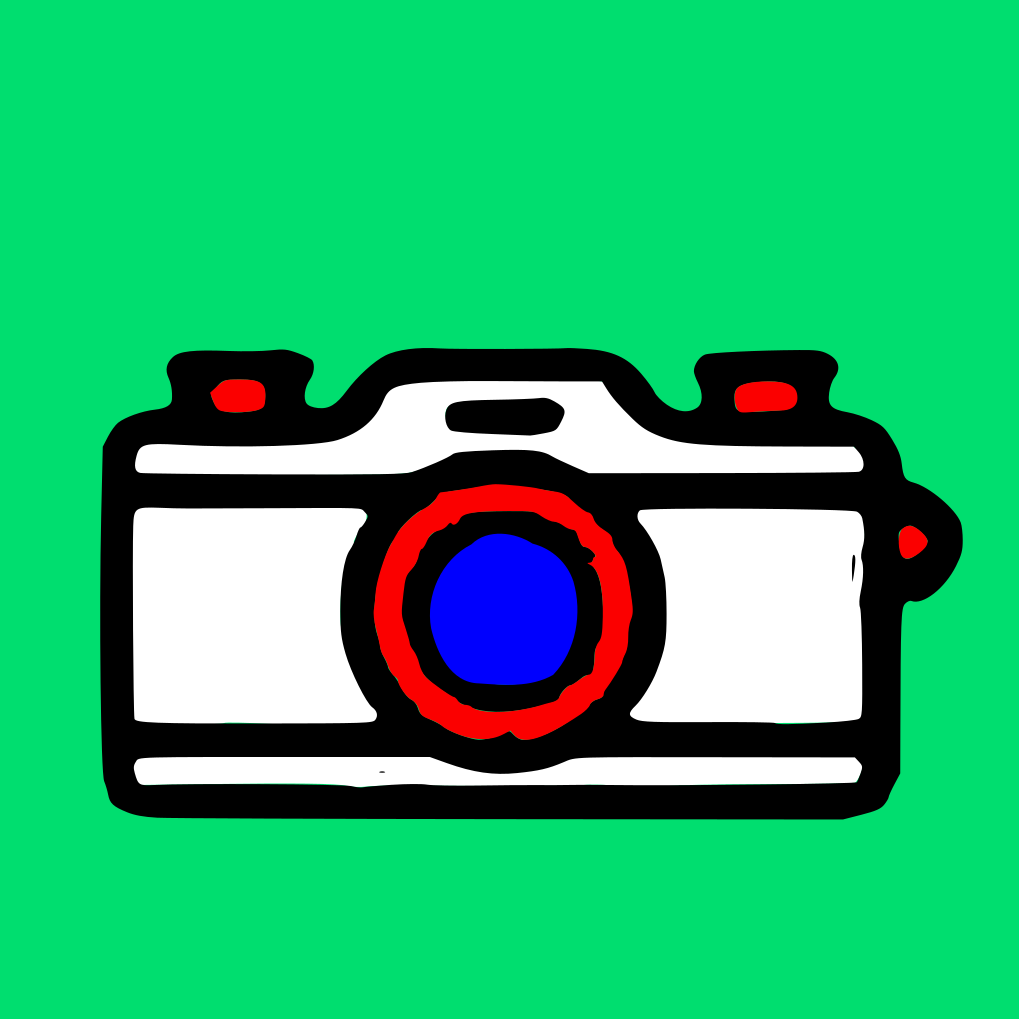} &
        \includegraphics[height=0.10\textwidth,width=0.10\textwidth]{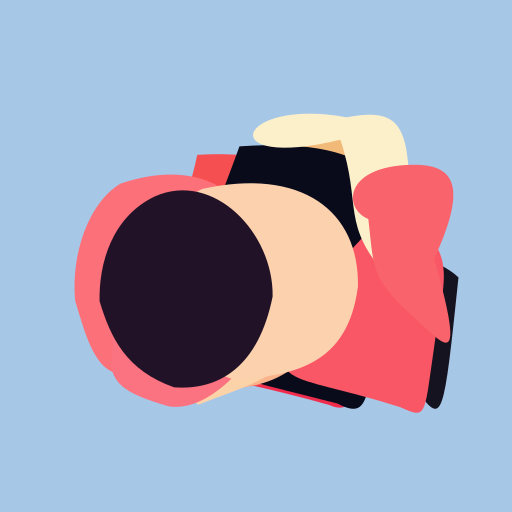} &
        \includegraphics[height=0.10\textwidth,width=0.10\textwidth]{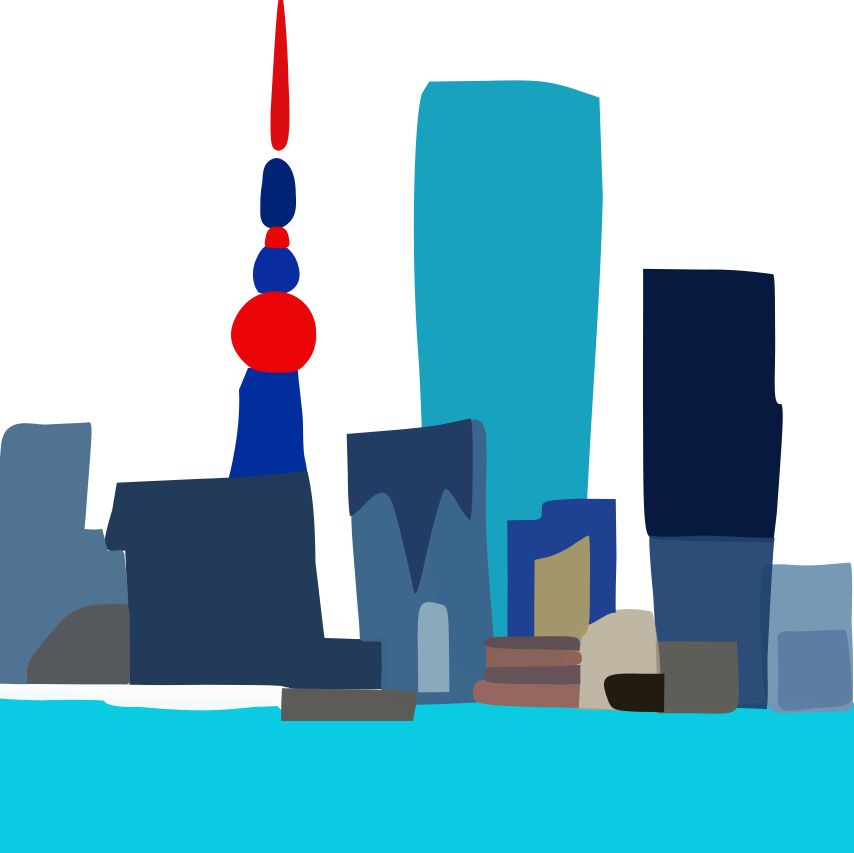} &
        \includegraphics[height=0.10\textwidth,width=0.10\textwidth]{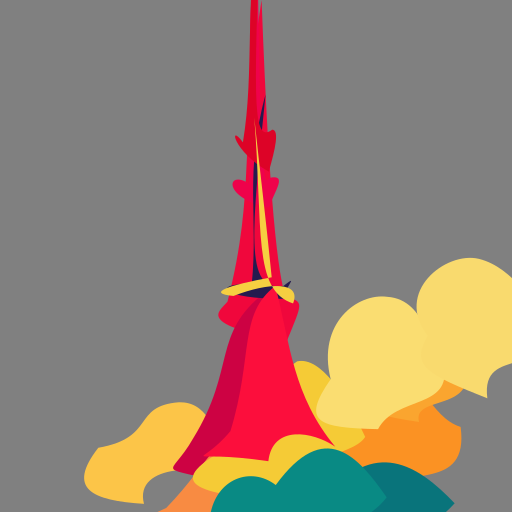} &
        \includegraphics[height=0.10\textwidth,width=0.10\textwidth]{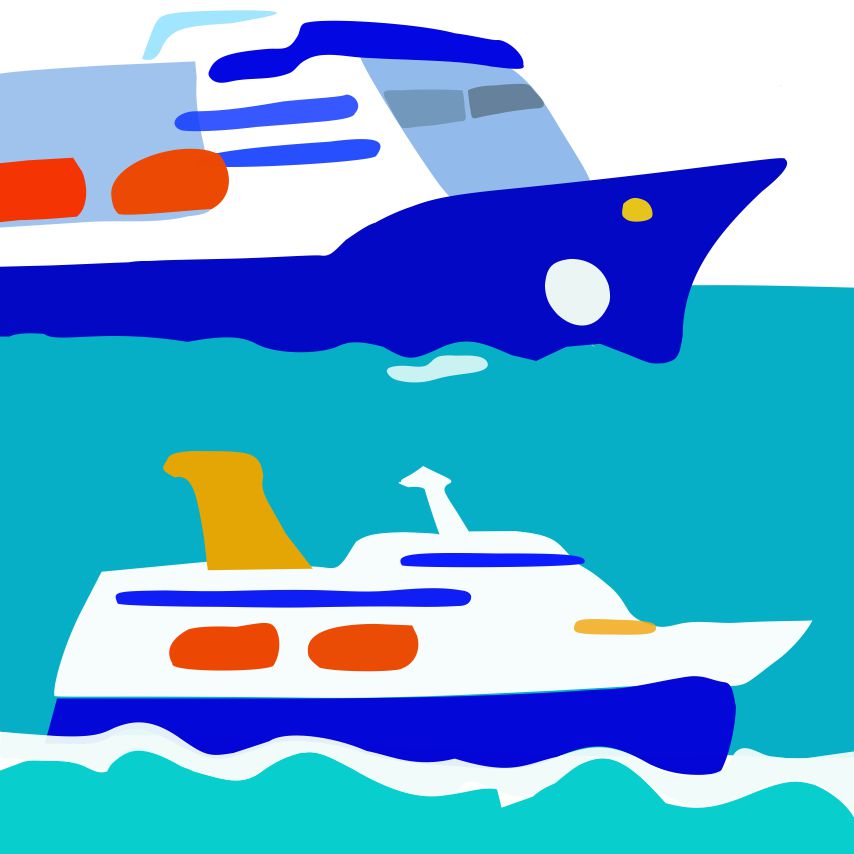} &
        \includegraphics[height=0.10\textwidth,width=0.10\textwidth]{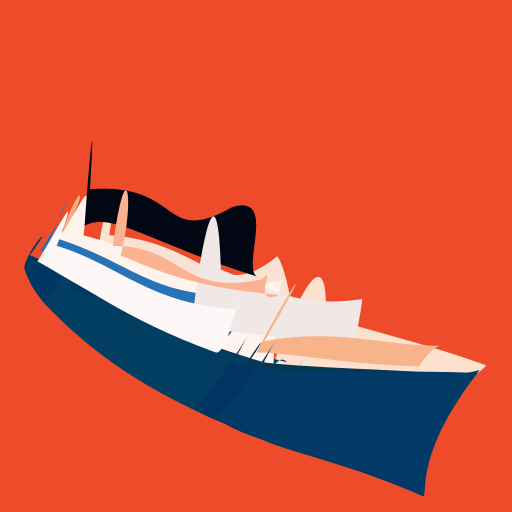} \\

        \multicolumn{2}{c}{\begin{tabular}{c}``a baby bunny on a \\ stack of pancakes'' \end{tabular}} &
        \multicolumn{2}{c}{\begin{tabular}{c}``a smiling sloth wearing \\ a jacket and cowboy hat''\end{tabular}} &
        \multicolumn{2}{c}{``a spaceship''} \\
        \includegraphics[height=0.10\textwidth,width=0.10\textwidth]{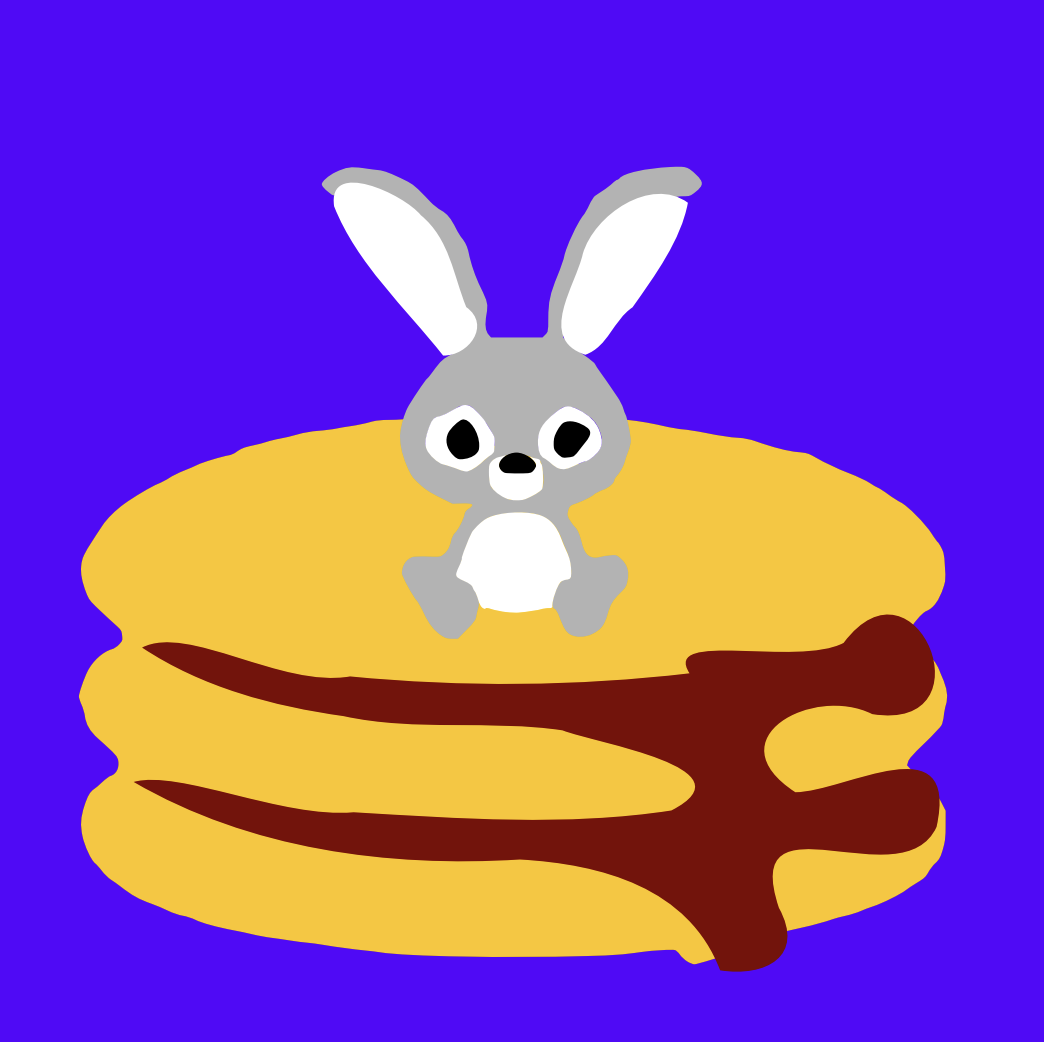} &
        \includegraphics[height=0.10\textwidth,width=0.10\textwidth]{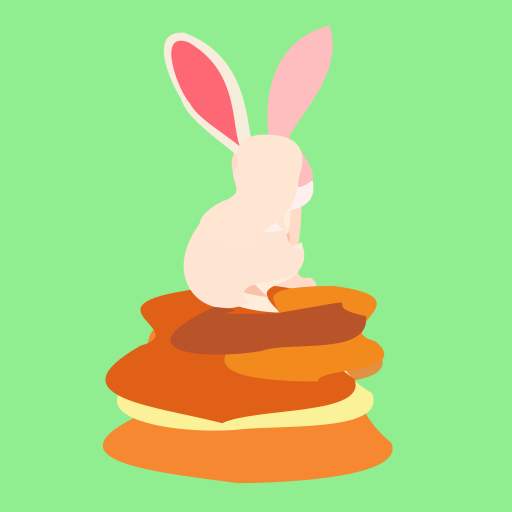} &
        \includegraphics[height=0.10\textwidth,width=0.10\textwidth]{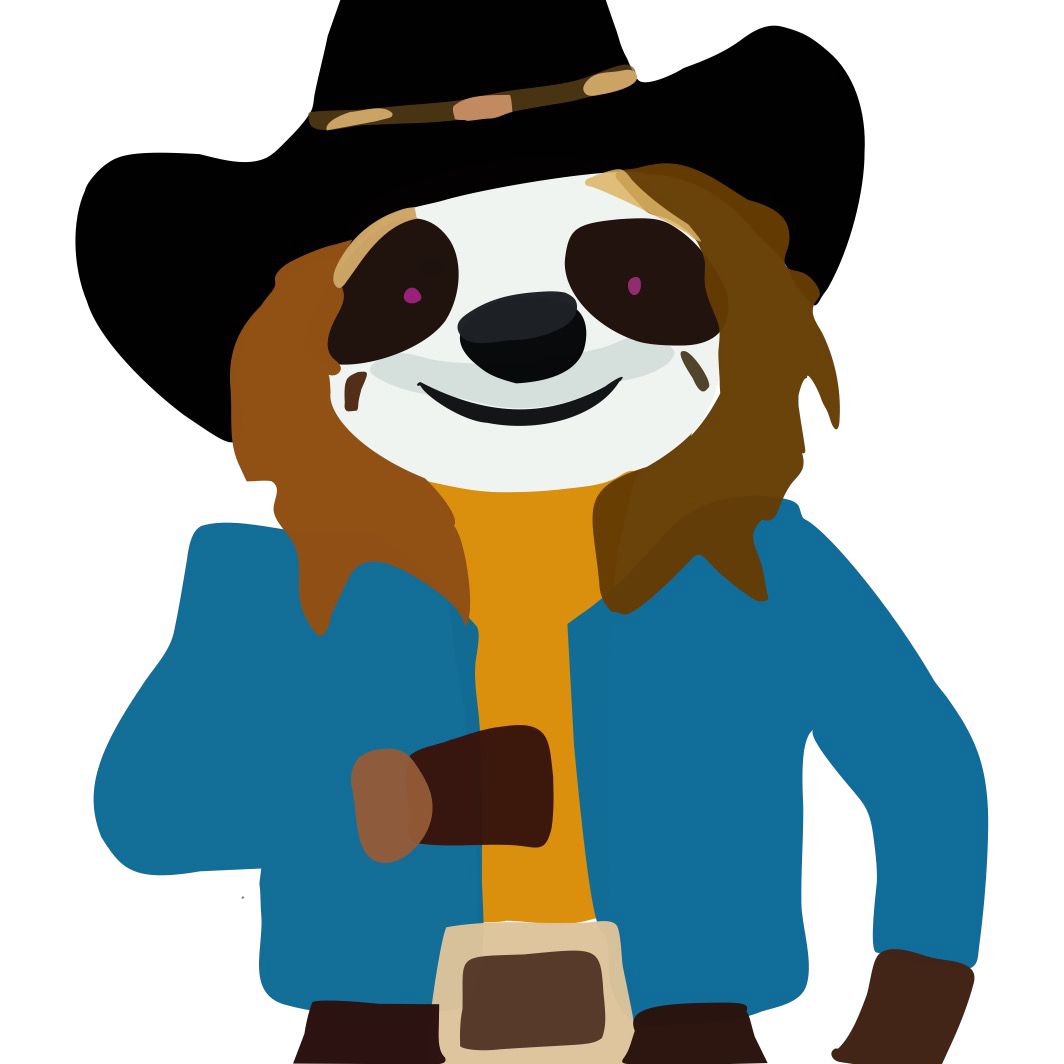} &
        \includegraphics[height=0.10\textwidth,width=0.10\textwidth]{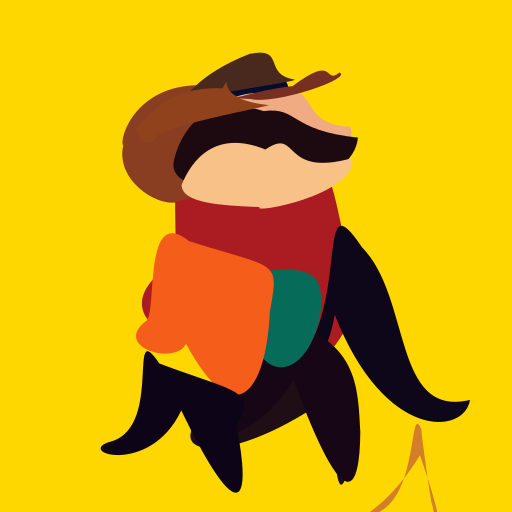} &
        \includegraphics[height=0.10\textwidth,width=0.10\textwidth]{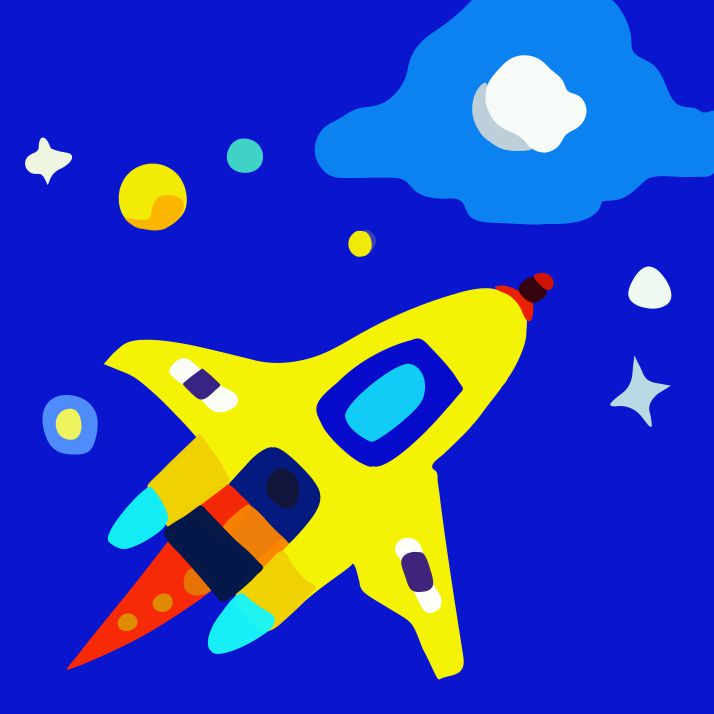} &
        \includegraphics[height=0.10\textwidth,width=0.10\textwidth]{images/ours_for_nivel_text2vector/spaceship.png} \\

        \multicolumn{2}{c}{``a spaceship''} &
        \multicolumn{2}{c}{``a dragon-cat hybrid''} &
        \multicolumn{2}{c}{``an espresso machine''} \\
        \includegraphics[height=0.10\textwidth,width=0.10\textwidth]{images/nivel/spaceship.png} &
        \includegraphics[height=0.10\textwidth,width=0.10\textwidth]{images/ours_for_nivel_text2vector/spaceship.png} &
        \includegraphics[height=0.10\textwidth,width=0.10\textwidth]{images/text_to_vector/dragon-cat_hybrid.jpg} &
        \includegraphics[height=0.10\textwidth,width=0.10\textwidth]{images/ours_for_nivel_text2vector/dragon_cat.png} &
        \includegraphics[height=0.10\textwidth,width=0.10\textwidth]{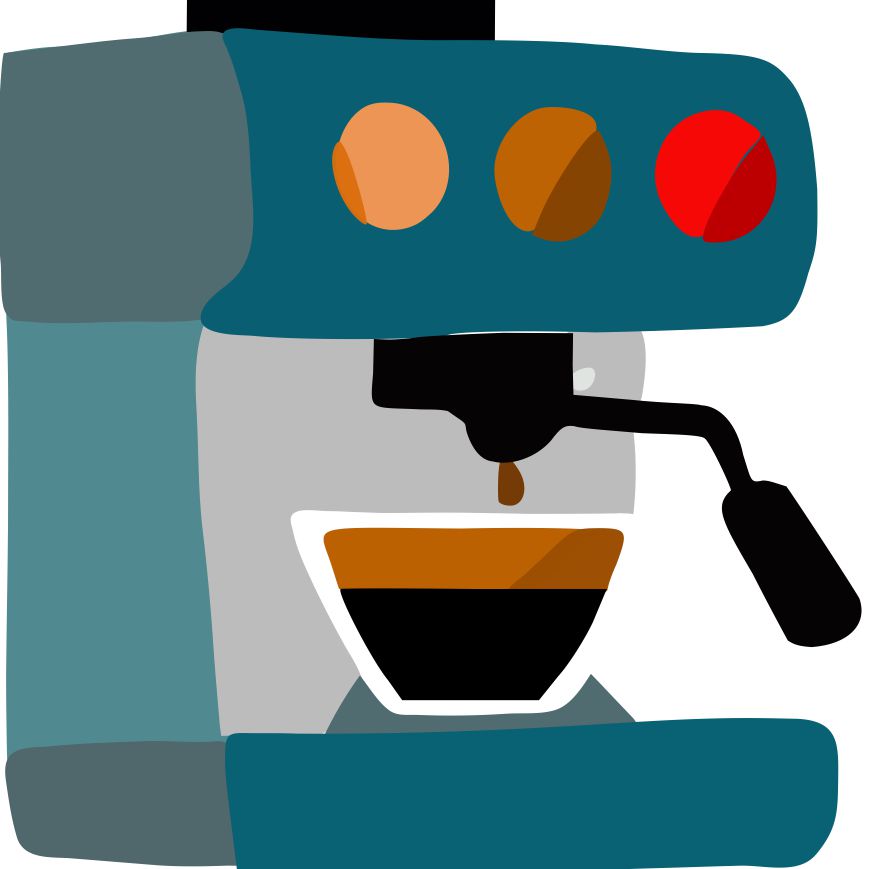} &
        \includegraphics[height=0.10\textwidth,width=0.10\textwidth]{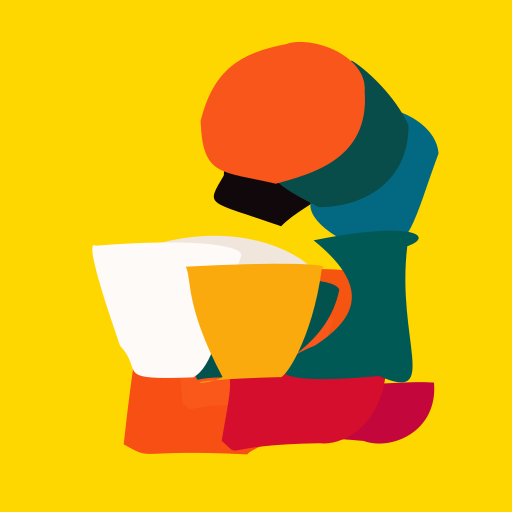} \\

        \multicolumn{2}{c}{``a stork playing a violin''} &
        \multicolumn{2}{c}{``a painting of the Mona Lisa''} &
        \multicolumn{2}{c}{``chocolate cake''} \\
        \includegraphics[height=0.10\textwidth,width=0.10\textwidth]{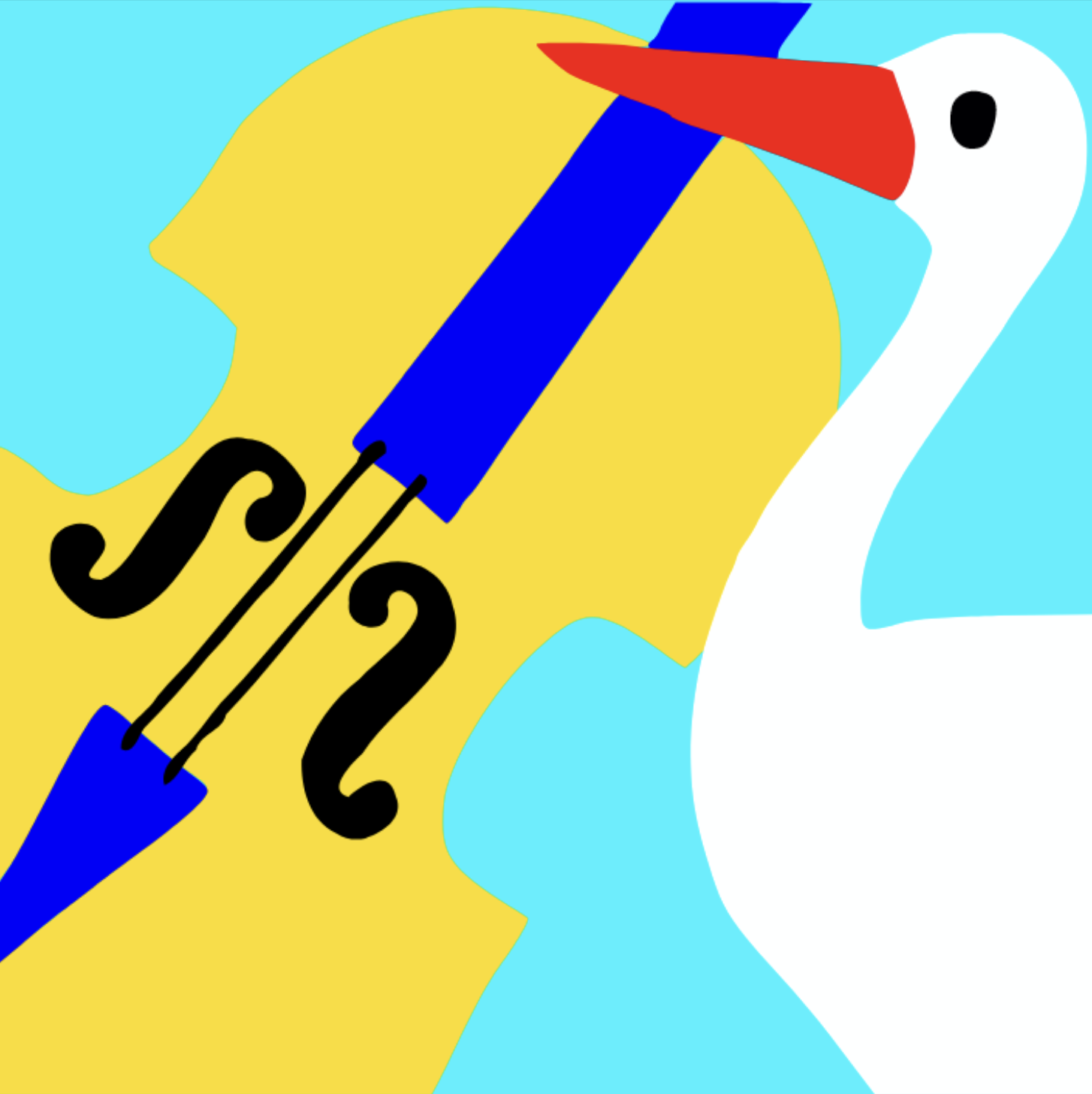} &
        \includegraphics[height=0.10\textwidth,width=0.10\textwidth]{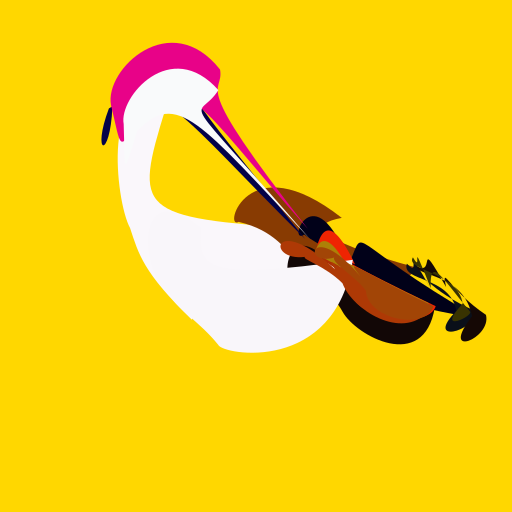} &
        \includegraphics[height=0.10\textwidth,width=0.10\textwidth]{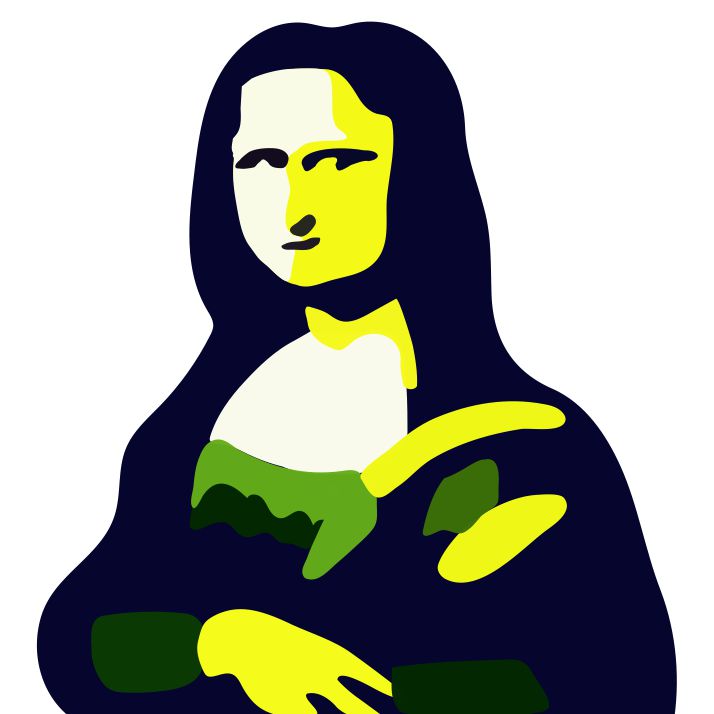} &
        \includegraphics[height=0.10\textwidth,width=0.10\textwidth]{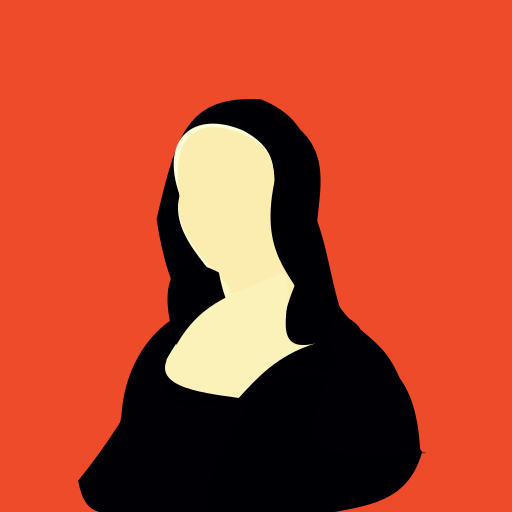} &
        \includegraphics[height=0.10\textwidth,width=0.10\textwidth]{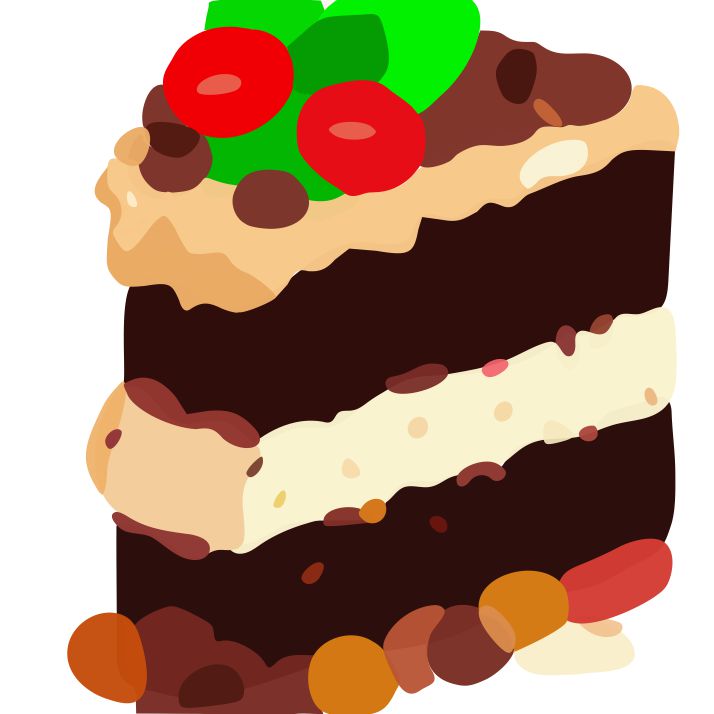} &
        \includegraphics[height=0.10\textwidth,width=0.10\textwidth]{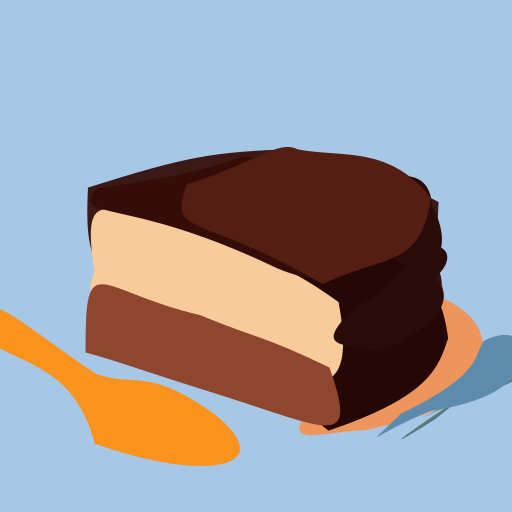} \\

        & & 
        \multicolumn{2}{c}{``A Japanese sakura tree on a hill''} &
        \multicolumn{2}{c}{``a Starbucks coffee cup''} \\
        & &
        \includegraphics[height=0.10\textwidth,width=0.10\textwidth]{images/text_to_vector/a_japanese_sakura_tree_on_a_hill.jpg} &
        \includegraphics[height=0.10\textwidth,width=0.10\textwidth]{images/dropout_results/sakura/light-blue_16_shapes.png} &
        \includegraphics[height=0.10\textwidth,width=0.10\textwidth]{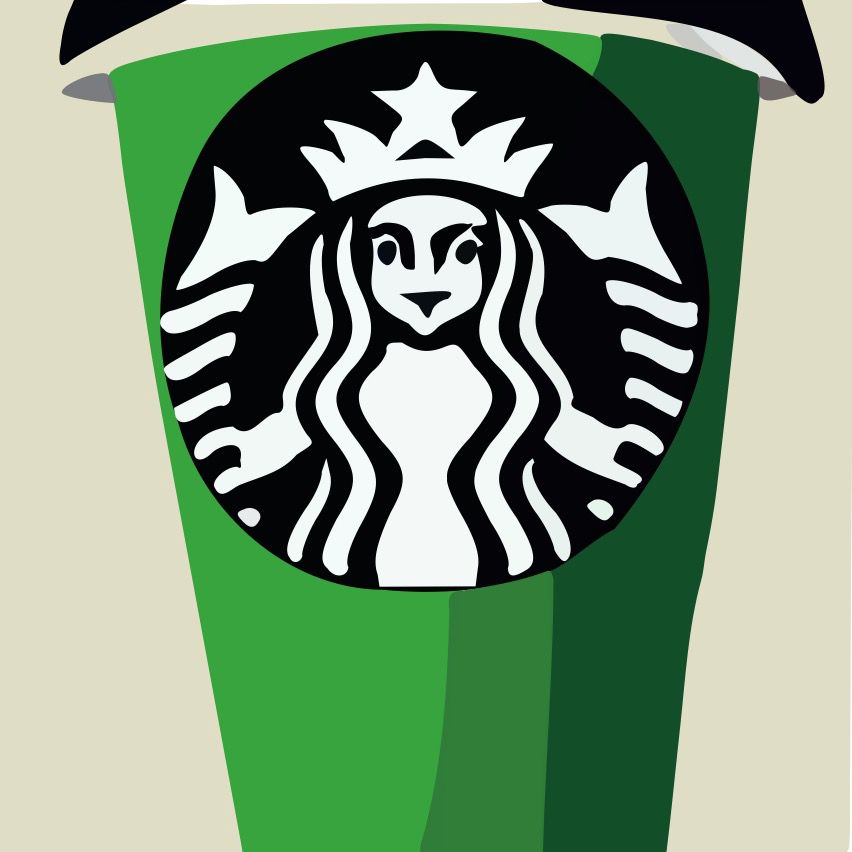} &
        \includegraphics[height=0.10\textwidth,width=0.10\textwidth]{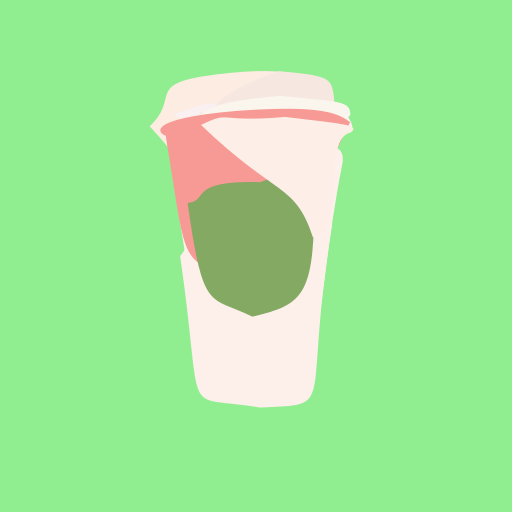} \\

\begin{tabular}{c} NiVEL  \end{tabular} &
        \begin{tabular}{c} \textbf{NeuralSVG} \end{tabular} &
        \begin{tabular}{c} Text-to-Vector \end{tabular} &
        \begin{tabular}{c} \textbf{NeuralSVG} \end{tabular} &
        \begin{tabular}{c} Text-to-Vector \end{tabular} &
        \begin{tabular}{c} \textbf{NeuralSVG} \end{tabular} \\
        
    \\[-0.5cm]        
    \end{tabular}
    }
    \caption{\textbf{Qualitative Comparisons.} As no code implementations are available, we provide visual comparisons to NIVeL~\cite{thamizharasan2024nivel} (left colunms) and Text-to-Vector~\cite{zhang2024text} (right columns) using results shown in their paper.
    }
    \label{fig:additional_comparisons_nivel_text2svg}
\end{figure*}

\begin{figure*}
    \centering
    \setlength{\tabcolsep}{0.5pt}
    {\small
    \begin{tabular}{c c c c c c c c c c}

        \multicolumn{10}{c}{``a boat''} \\
        \includegraphics[height=0.1\textwidth,width=0.1\textwidth]{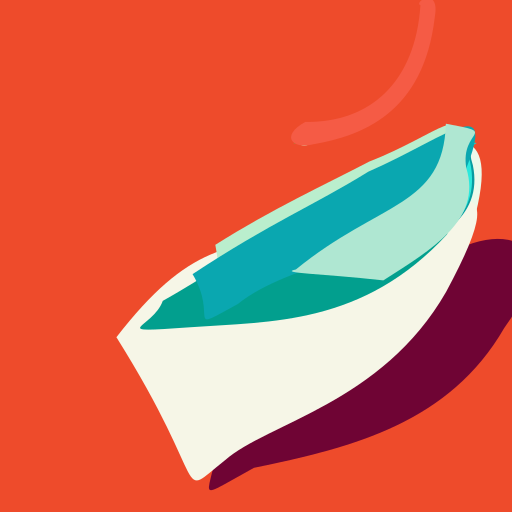} &
        \includegraphics[height=0.1\textwidth,width=0.1\textwidth]{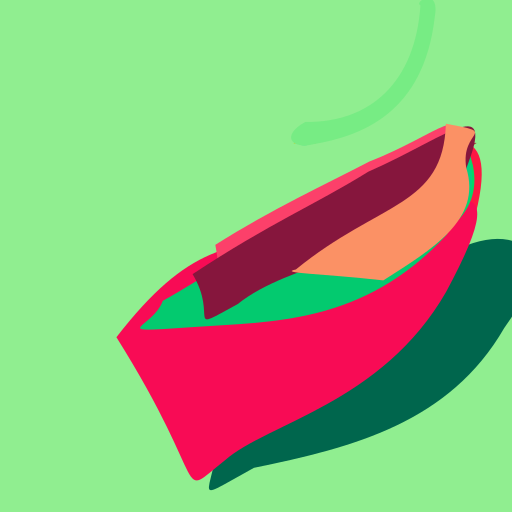} &
        \includegraphics[height=0.1\textwidth,width=0.1\textwidth]{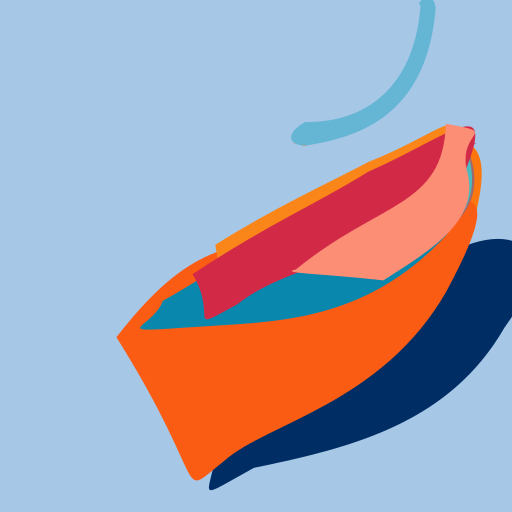} &
        \includegraphics[height=0.1\textwidth,width=0.1\textwidth]{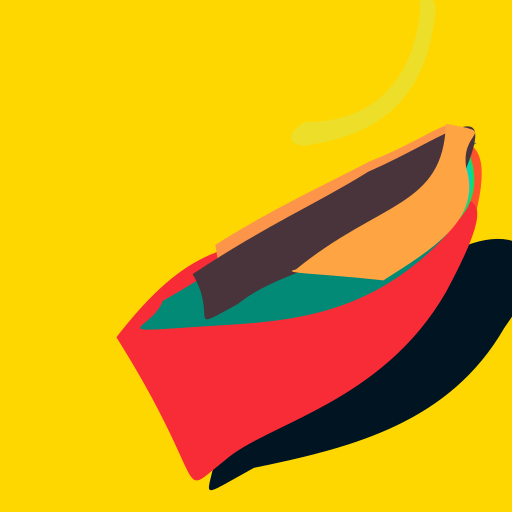} &
        \includegraphics[height=0.1\textwidth,width=0.1\textwidth]{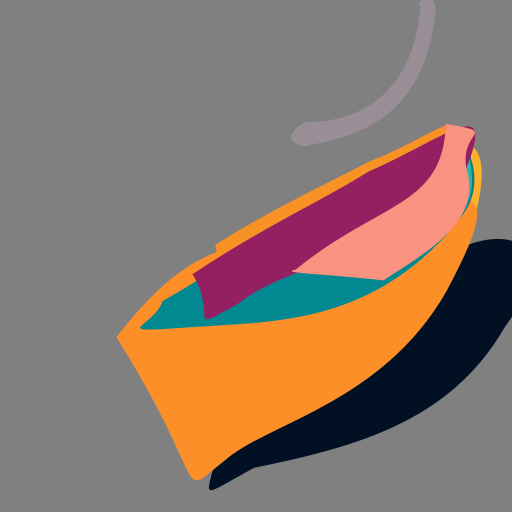} &
        \includegraphics[height=0.1\textwidth,width=0.1\textwidth]{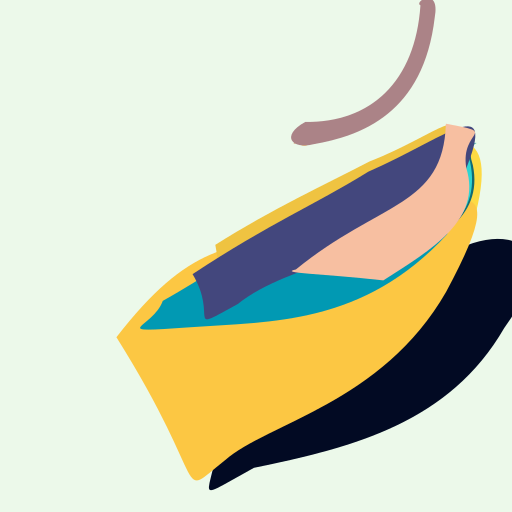} &
        \includegraphics[height=0.1\textwidth,width=0.1\textwidth]{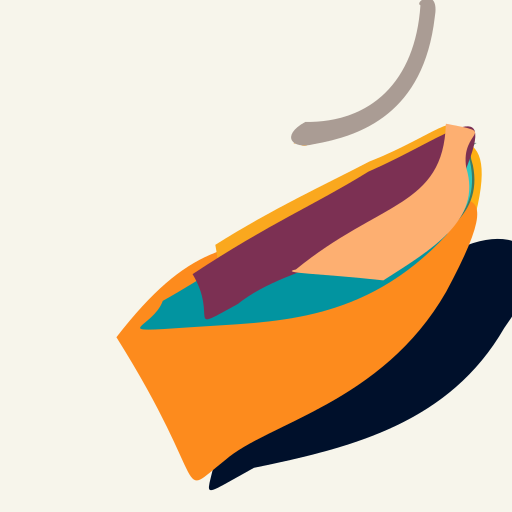} &
        \includegraphics[height=0.1\textwidth,width=0.1\textwidth]{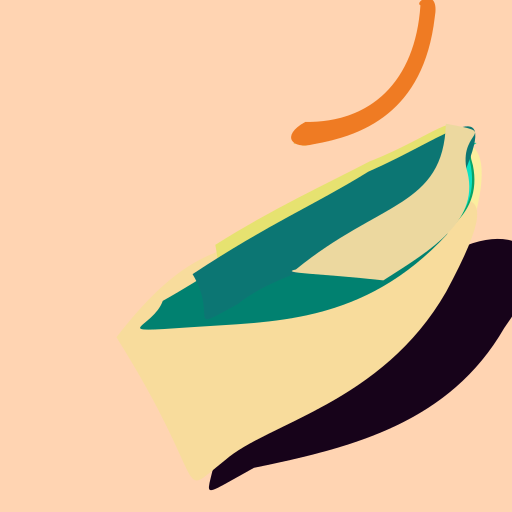} &
        \includegraphics[height=0.1\textwidth,width=0.1\textwidth]{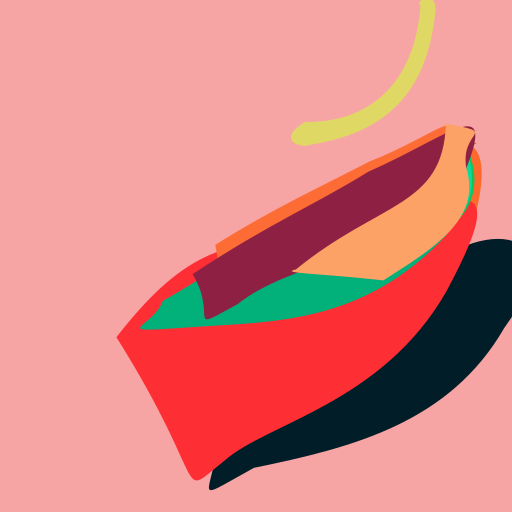} &
        \includegraphics[height=0.1\textwidth,width=0.1\textwidth]{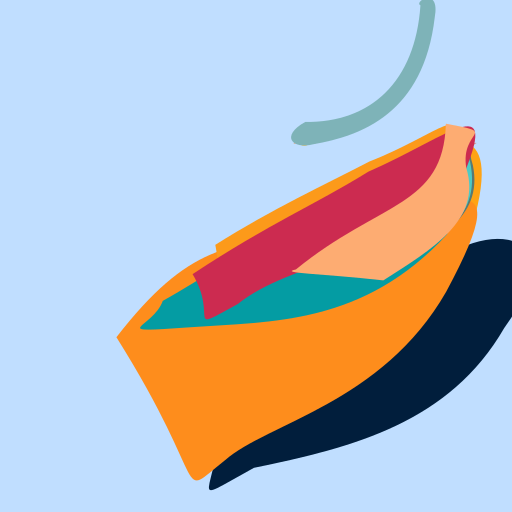} \\

        \multicolumn{10}{c}{``a 3D rendering of a temple''} \\
\end{tabular}
    }
    \caption{\textbf{Dynamically Controlling the Color Palette.} Given a learned representation, we render the result using different background colors specified by the user, resulting in varying color palettes in the resulting SVGs. The 5 leftmost columns show colors observed during training while the 5 rightmost columns show unobserved colors.}
    \label{fig:color_control_2}
\end{figure*}

\begin{figure*}
    \centering
    \setlength{\tabcolsep}{1pt}
    {\small
    %
}} & \includegraphics[height=0.10\textwidth,width=0.40\textwidth]{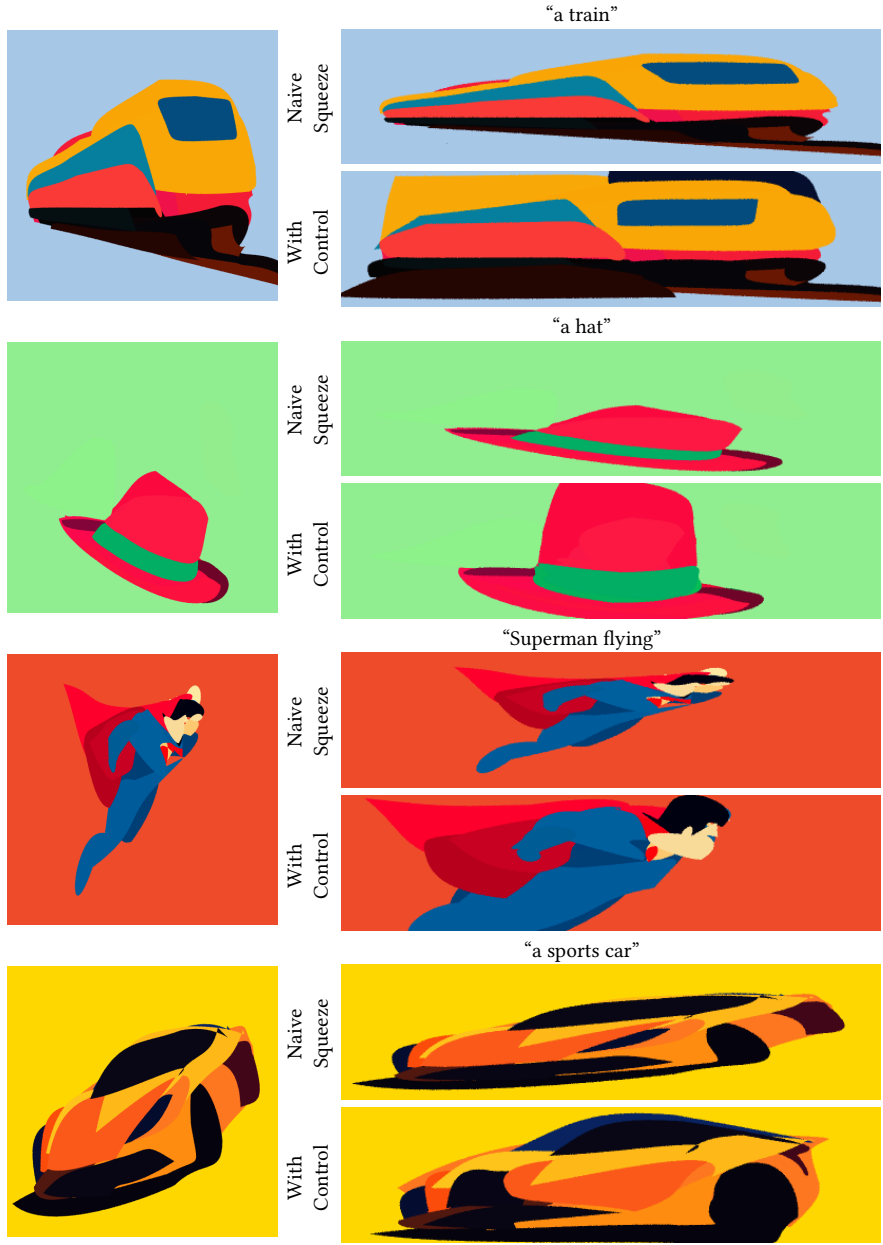} \\

    \\[-0.6cm]        
    \end{tabular}
    }
    \caption{\textbf{Dynamically Controlling the Aspect Ratio.} Additional results from optimizing NeuralSVG with aspect ratios of 1:1 and 4:1. In each pair of results, the top row shows the naive approach of squeezing the 1:1 output into a 4:1 aspect ratio. The bottom row shows the results where our trained network directly outputs the 4:1 aspect ratio.
    }
    \label{fig:additional_aspect_ratio}
\end{figure*}

\begin{figure*}[ht!]
    \centering
    \begin{minipage}[t]{0.5\linewidth} %
        \setlength{\tabcolsep}{0.5pt}
        \centering
        \small
        \begin{tabular}{c c c c c}

            \\ \\ \\

            \multicolumn{4}{c}{``a ballerina''} \\
            \includegraphics[width=0.25\linewidth,height=0.25\linewidth]{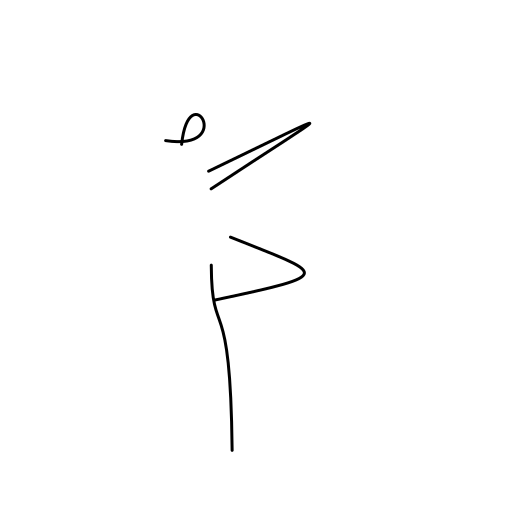} &
            \includegraphics[width=0.25\linewidth,height=0.25\linewidth]{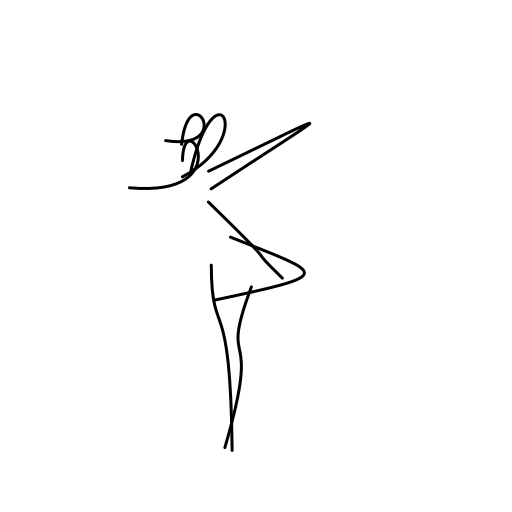} &
            \includegraphics[width=0.25\linewidth,height=0.25\linewidth]{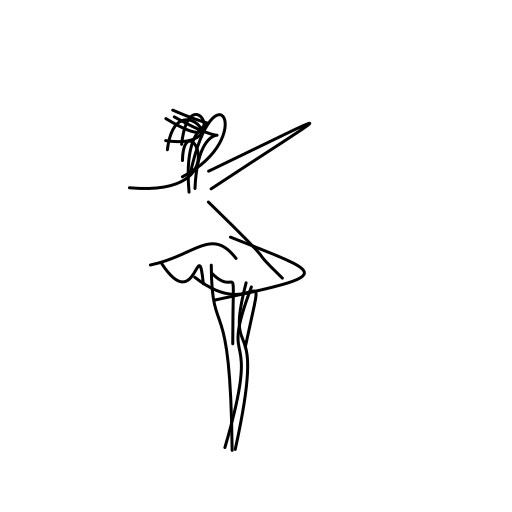} &
            \includegraphics[width=0.25\linewidth,height=0.25\linewidth]{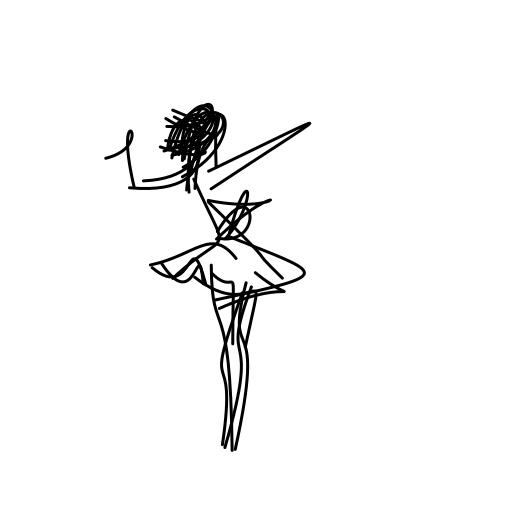} \\

            \multicolumn{4}{c}{``a boat''} \\
            \includegraphics[width=0.25\linewidth,height=0.25\linewidth]{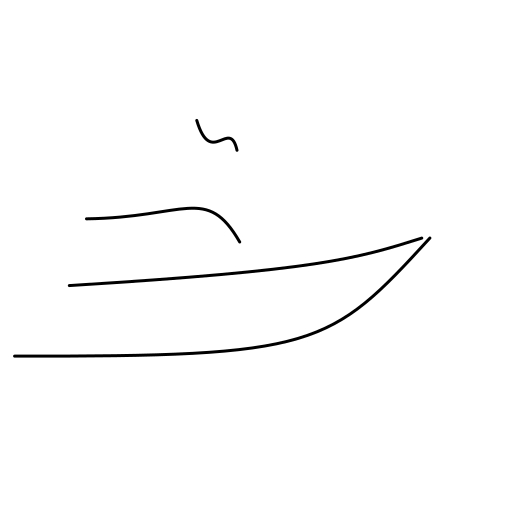} &
            \includegraphics[width=0.25\linewidth,height=0.25\linewidth]{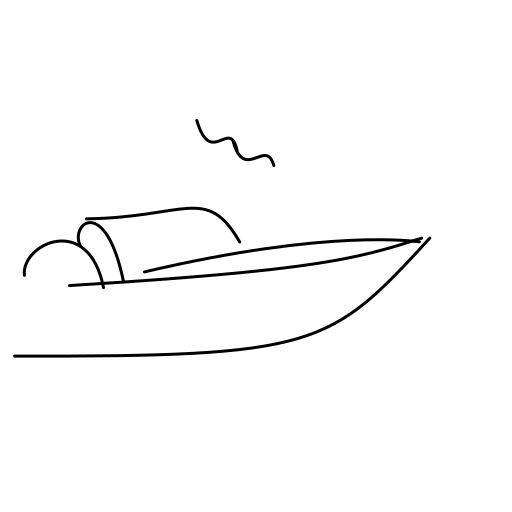} &
            \includegraphics[width=0.25\linewidth,height=0.25\linewidth]{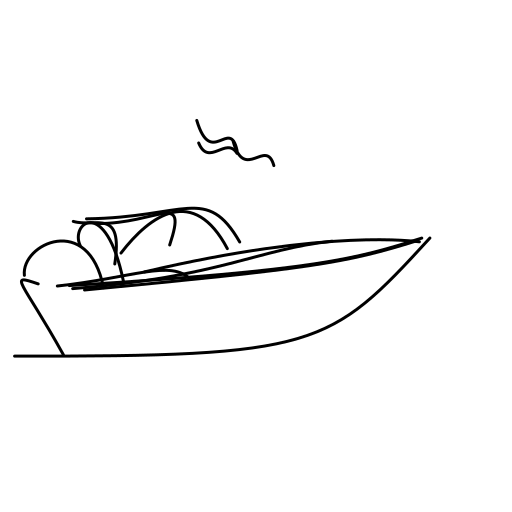} &
            \includegraphics[width=0.25\linewidth,height=0.25\linewidth]{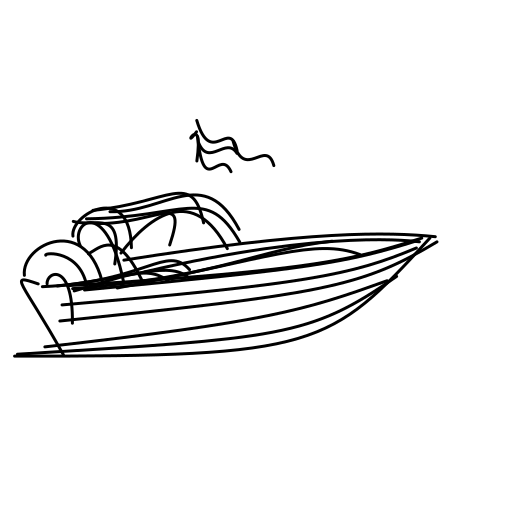} \\

            \multicolumn{4}{c}{``a cat''} \\
            \includegraphics[width=0.25\linewidth,height=0.25\linewidth]{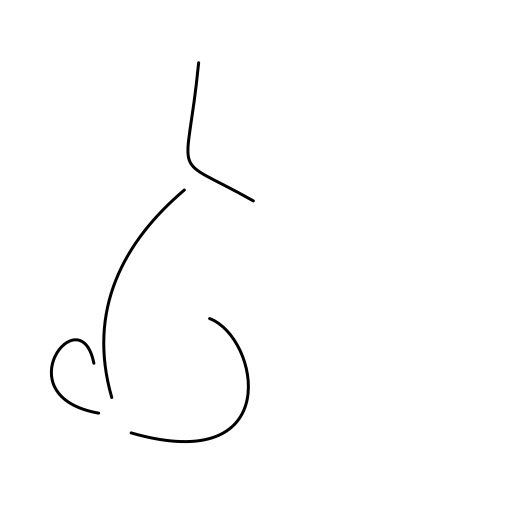} &
            \includegraphics[width=0.25\linewidth,height=0.25\linewidth]{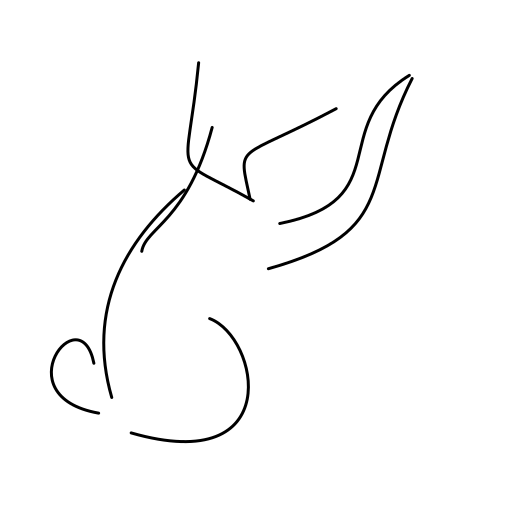} &
            \includegraphics[width=0.25\linewidth,height=0.25\linewidth]{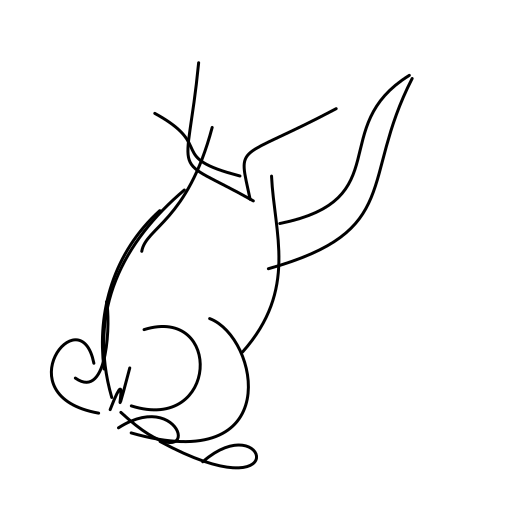} &
            \includegraphics[width=0.25\linewidth,height=0.25\linewidth]{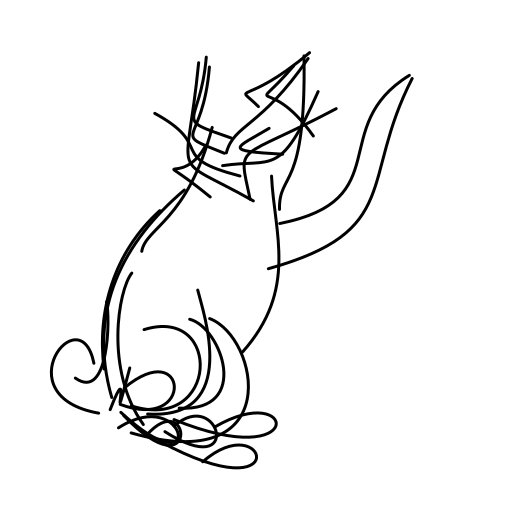} \\

            \multicolumn{4}{c}{``a giraffe''} \\
            \includegraphics[width=0.25\linewidth,height=0.25\linewidth]{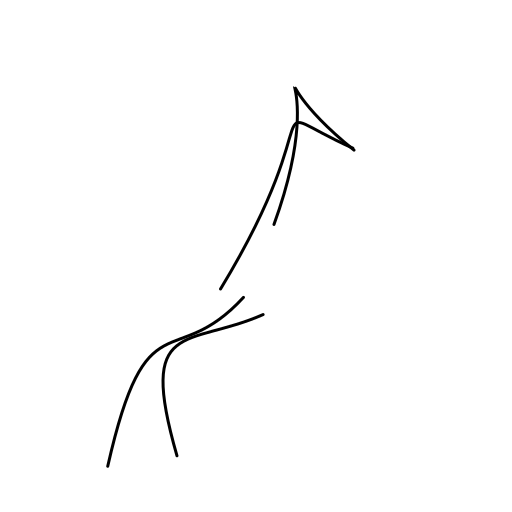} &
            \includegraphics[width=0.25\linewidth,height=0.25\linewidth]{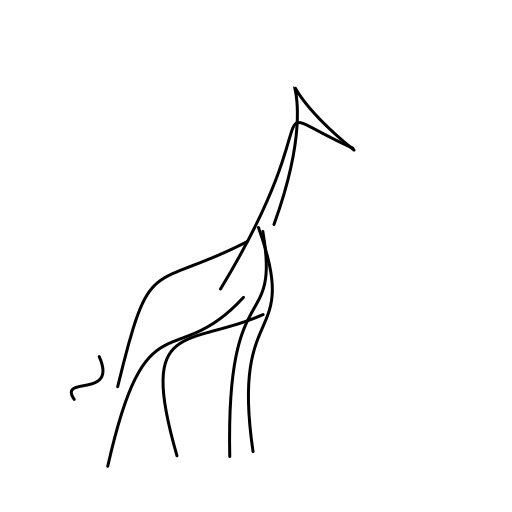} &
            \includegraphics[width=0.25\linewidth,height=0.25\linewidth]{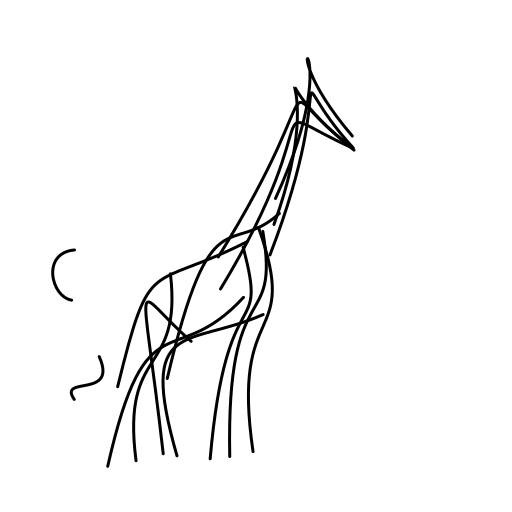} &
            \includegraphics[width=0.25\linewidth,height=0.25\linewidth]{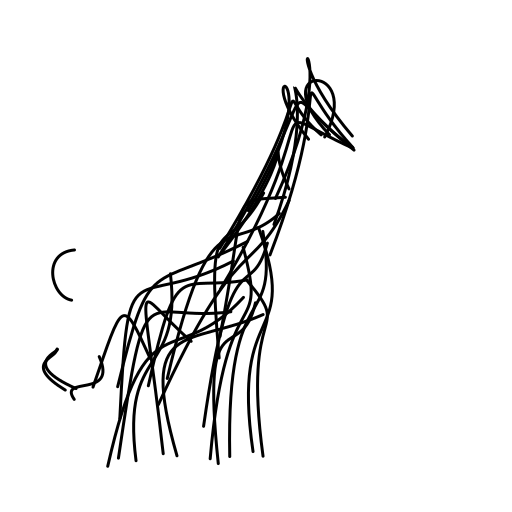} \\

            \multicolumn{4}{c}{``a rocket ship''} \\
            \includegraphics[width=0.25\linewidth,height=0.25\linewidth]{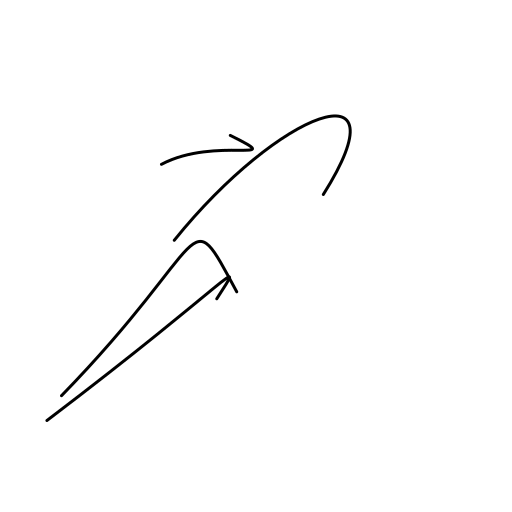} &
            \includegraphics[width=0.25\linewidth,height=0.25\linewidth]{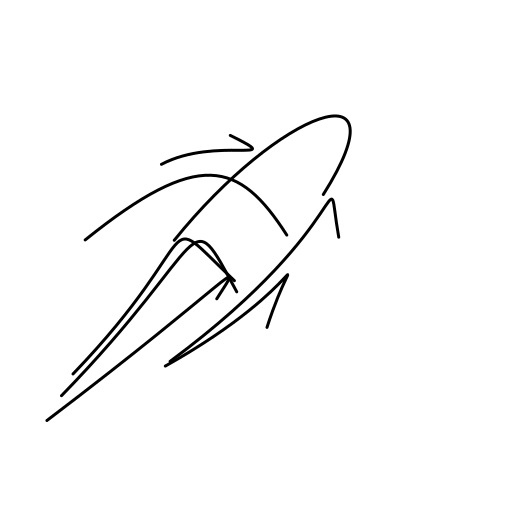} &
            \includegraphics[width=0.25\linewidth,height=0.25\linewidth]{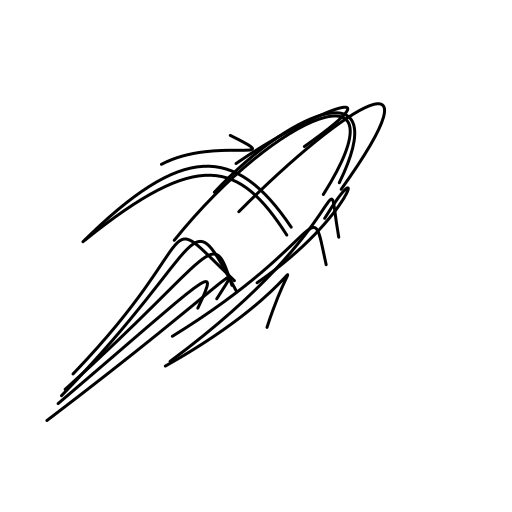} &
            \includegraphics[width=0.25\linewidth,height=0.25\linewidth]{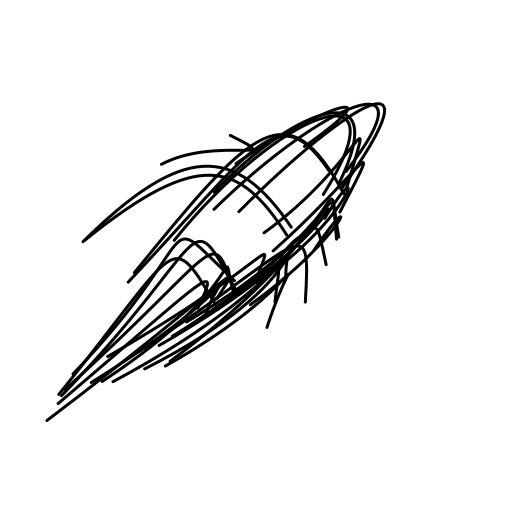} \\

            \multicolumn{4}{c}{``a rooster''} \\
            \includegraphics[width=0.25\linewidth,height=0.25\linewidth]{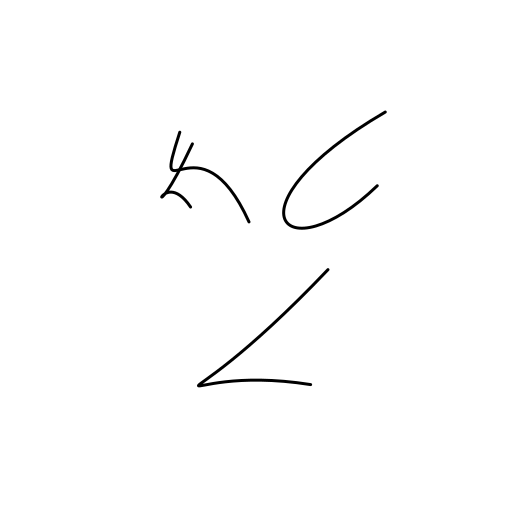} &
            \includegraphics[width=0.25\linewidth,height=0.25\linewidth]{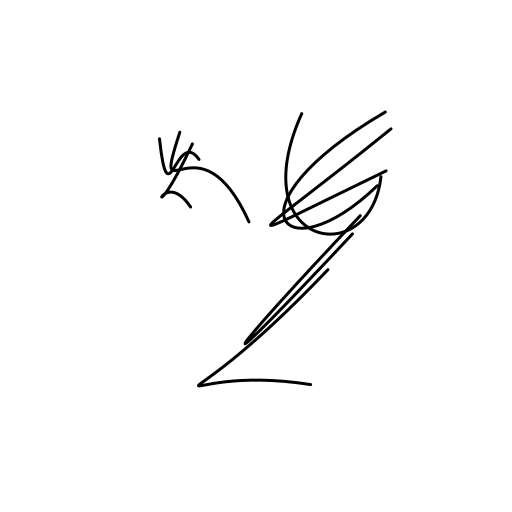} &
            \includegraphics[width=0.25\linewidth,height=0.25\linewidth]{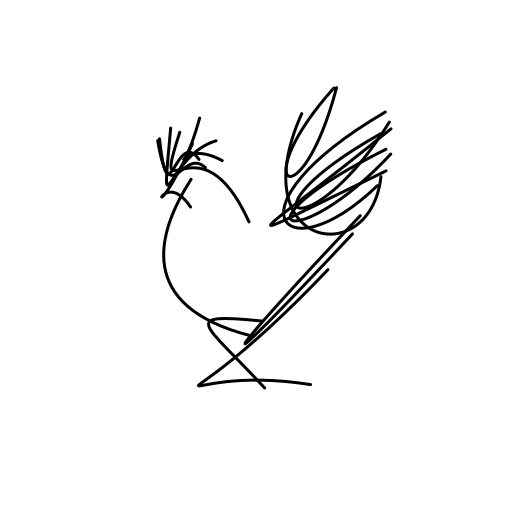} &
            \includegraphics[width=0.25\linewidth,height=0.25\linewidth]{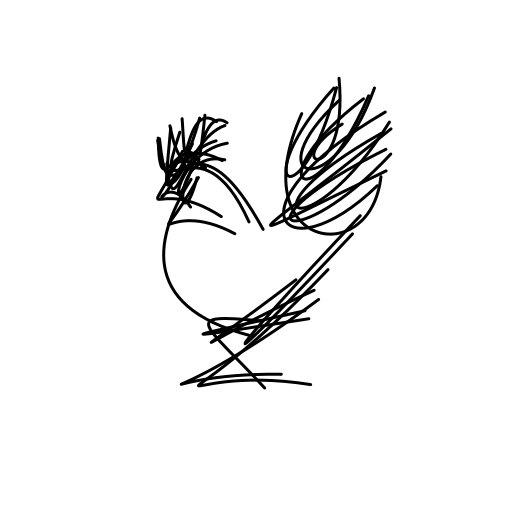} \\

            \multicolumn{4}{c}{``a strawberry''} \\
            \includegraphics[width=0.25\linewidth,height=0.25\linewidth]{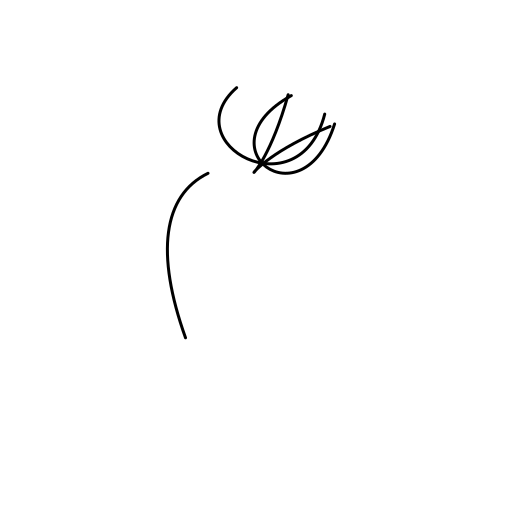} &
            \includegraphics[width=0.25\linewidth,height=0.25\linewidth]{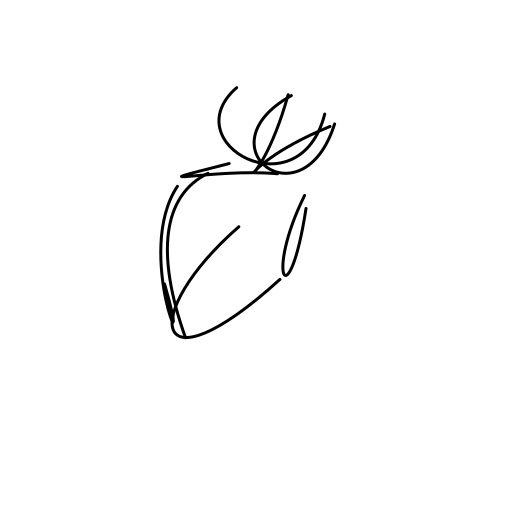} &
            \includegraphics[width=0.25\linewidth,height=0.25\linewidth]{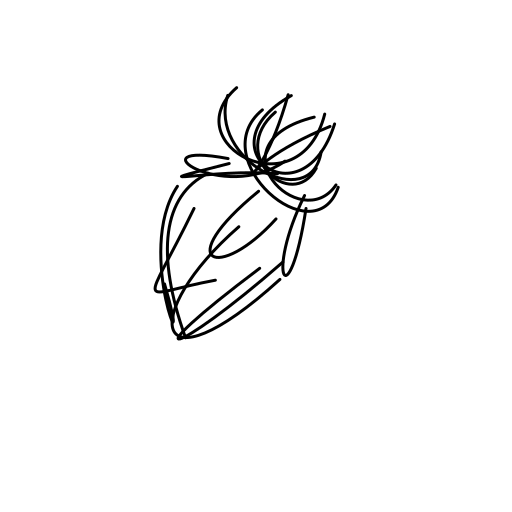} &
            \includegraphics[width=0.25\linewidth,height=0.25\linewidth]{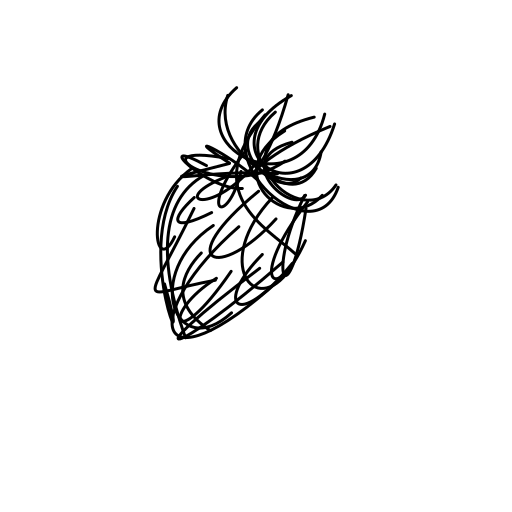} \\

            $4$ & $8$ & $16$ & $32$ \\

        \end{tabular}
    \end{minipage}%
    \hfill %
    \begin{minipage}[t]{0.5\linewidth}
        \setlength{\tabcolsep}{0.5pt}
        \centering
        \small
        \begin{tabular}{c c c c c}

            \\ \\ \\
        
            \multicolumn{4}{c}{``a bull''} \\
            \includegraphics[width=0.25\linewidth,height=0.25\linewidth]{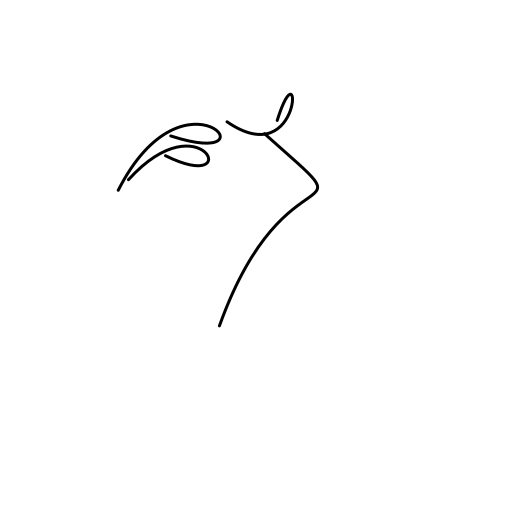} &
            \includegraphics[width=0.25\linewidth,height=0.25\linewidth]{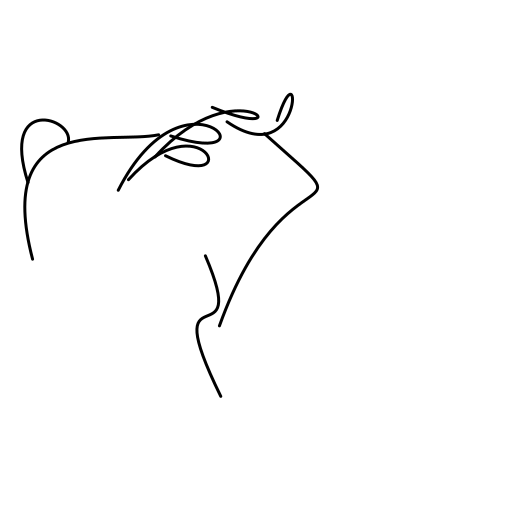} &
            \includegraphics[width=0.25\linewidth,height=0.25\linewidth]{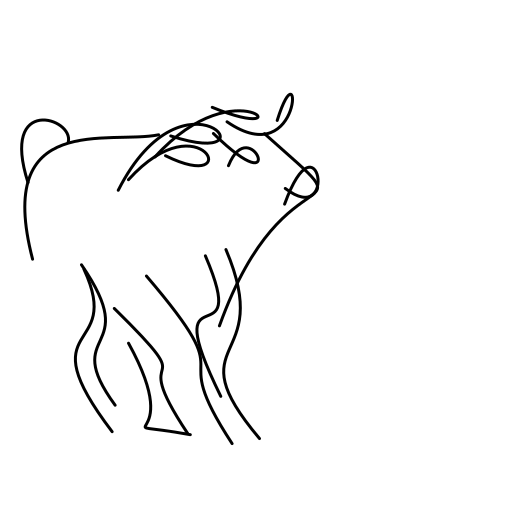} &
            \includegraphics[width=0.25\linewidth,height=0.25\linewidth]{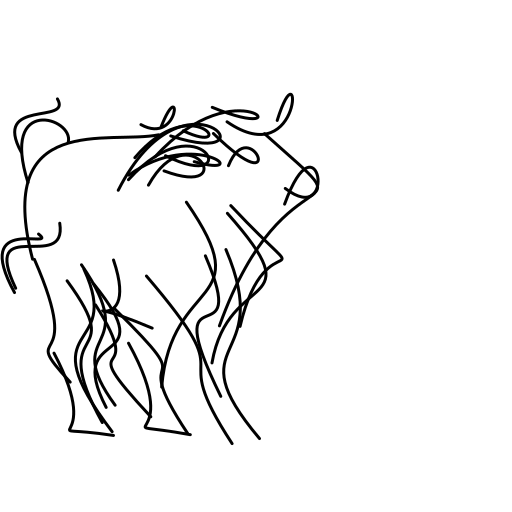} \\

            \multicolumn{4}{c}{``a baby penguin''} \\
            \includegraphics[width=0.25\linewidth,height=0.25\linewidth]{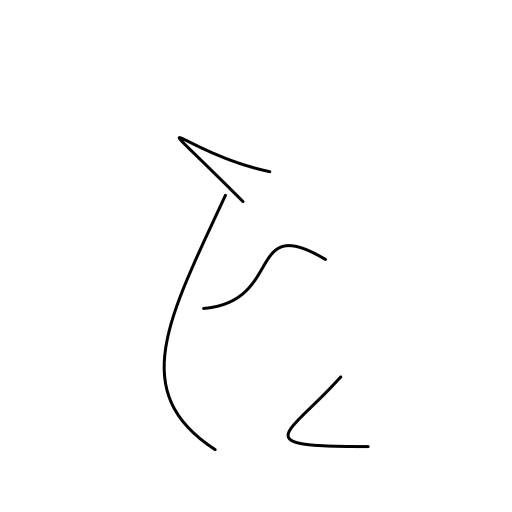} &
            \includegraphics[width=0.25\linewidth,height=0.25\linewidth]{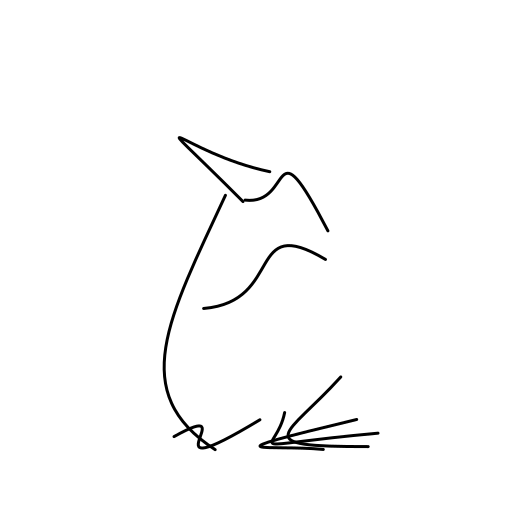} &
            \includegraphics[width=0.25\linewidth,height=0.25\linewidth]{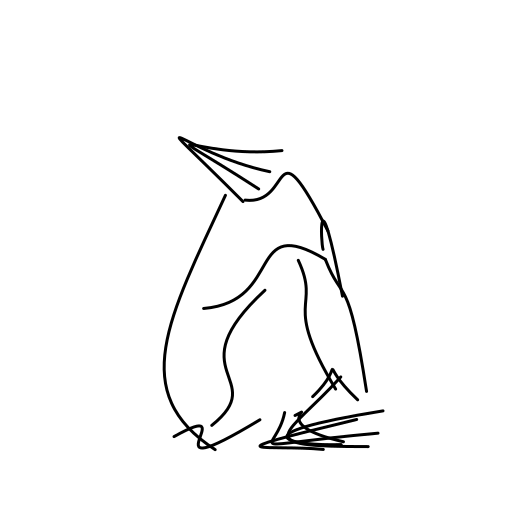} &
            \includegraphics[width=0.25\linewidth,height=0.25\linewidth]{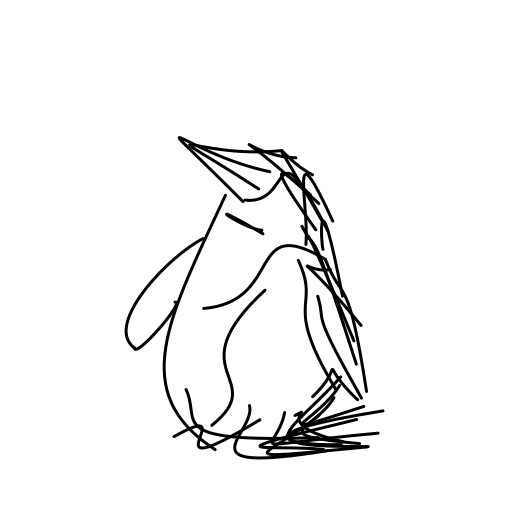} \\

            \multicolumn{4}{c}{``a sailboat''} \\
            \includegraphics[width=0.25\linewidth,height=0.25\linewidth]{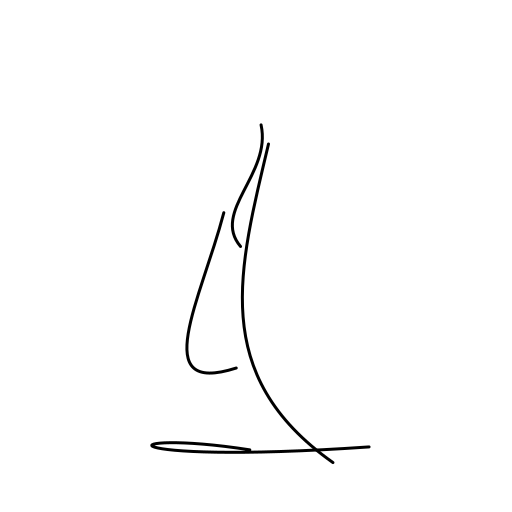} &
            \includegraphics[width=0.25\linewidth,height=0.25\linewidth]{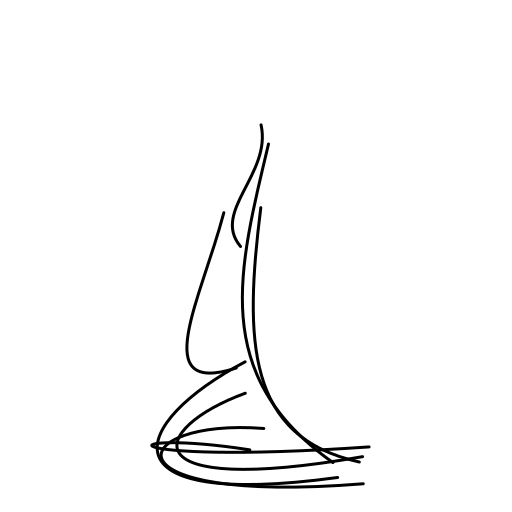} &
            \includegraphics[width=0.25\linewidth,height=0.25\linewidth]{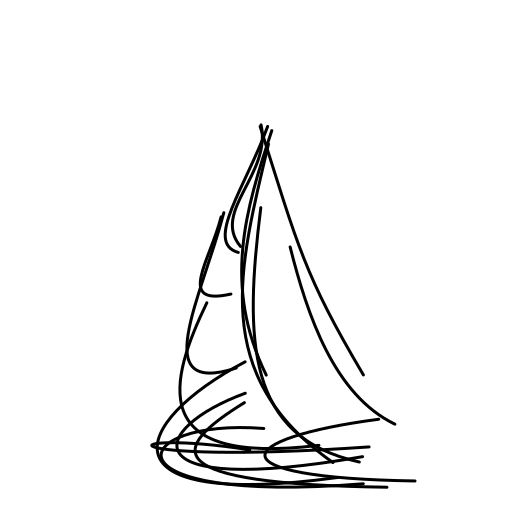} &
            \includegraphics[width=0.25\linewidth,height=0.25\linewidth]{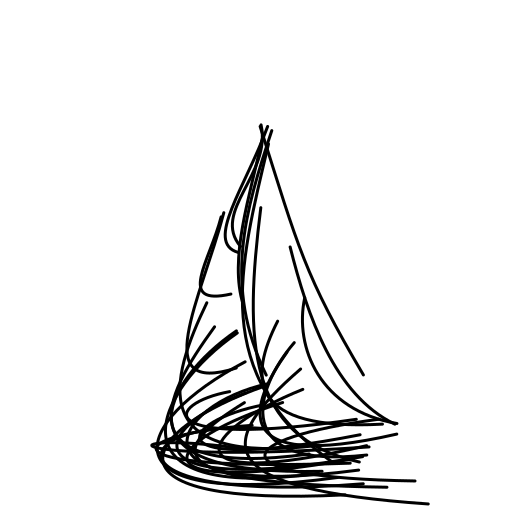} \\

            \multicolumn{4}{c}{``a lizard''} \\
            \includegraphics[width=0.25\linewidth,height=0.25\linewidth]{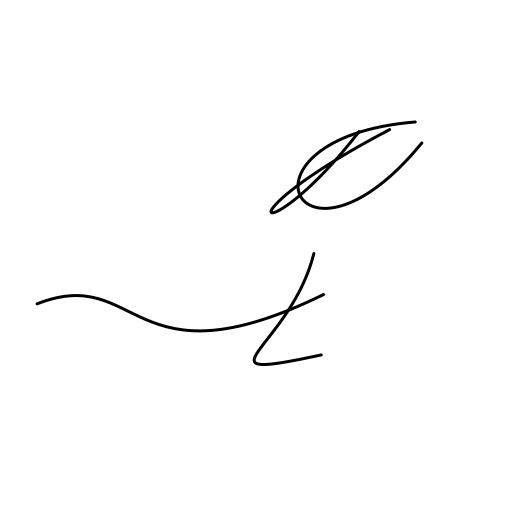} &
            \includegraphics[width=0.25\linewidth,height=0.25\linewidth]{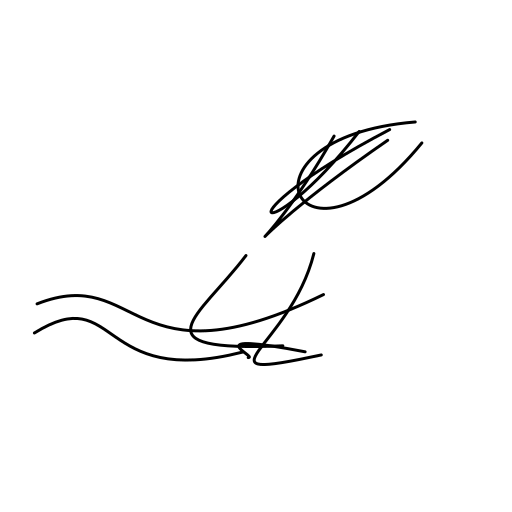} &
            \includegraphics[width=0.25\linewidth,height=0.25\linewidth]{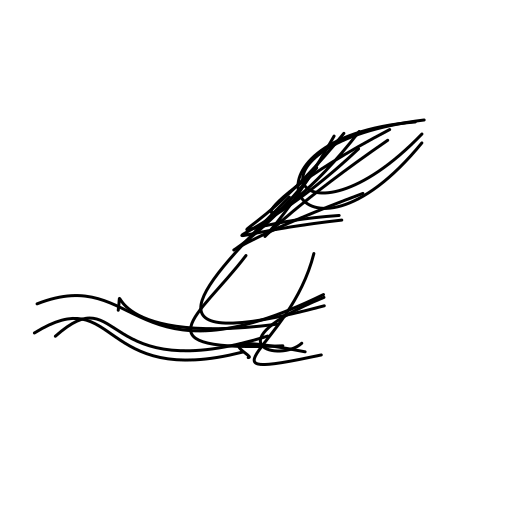} &
            \includegraphics[width=0.25\linewidth,height=0.25\linewidth]{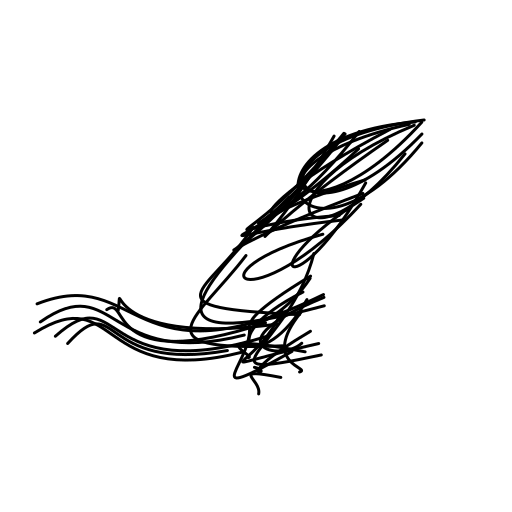} \\

            \multicolumn{4}{c}{``a margarita''} \\
            \includegraphics[width=0.25\linewidth,height=0.25\linewidth]{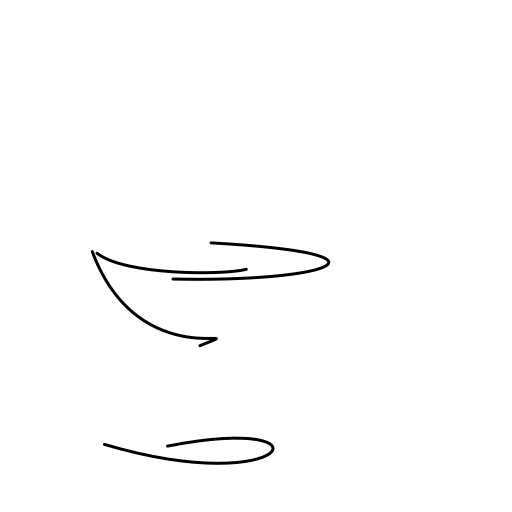} &
            \includegraphics[width=0.25\linewidth,height=0.25\linewidth]{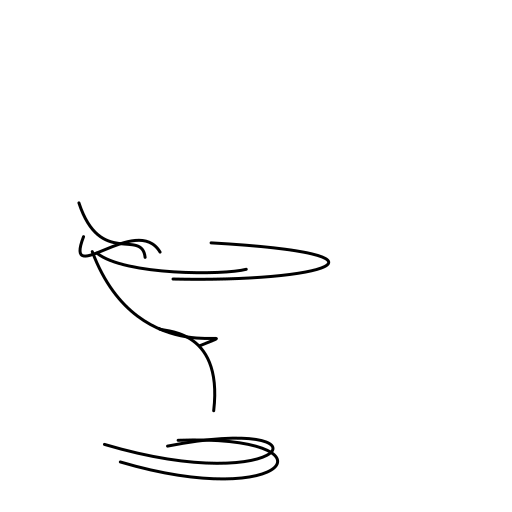} &
            \includegraphics[width=0.25\linewidth,height=0.25\linewidth]{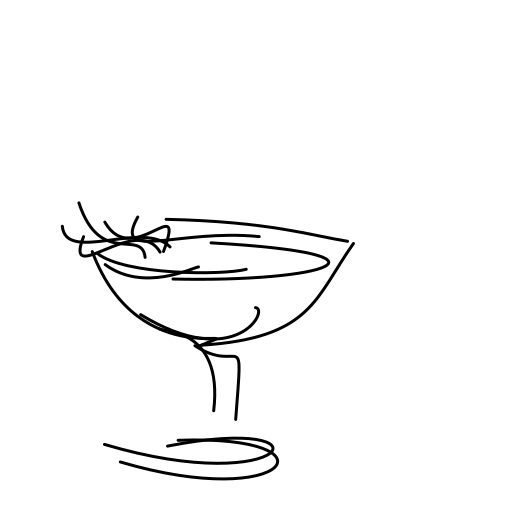} &
            \includegraphics[width=0.25\linewidth,height=0.25\linewidth]{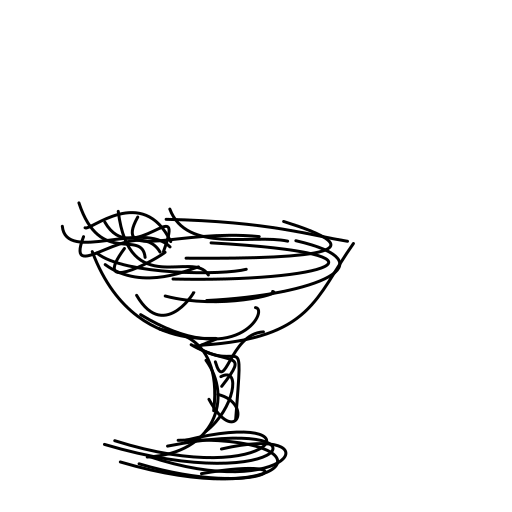} \\

            \multicolumn{4}{c}{``a glass of wine''} \\
            \includegraphics[width=0.25\linewidth,height=0.25\linewidth]{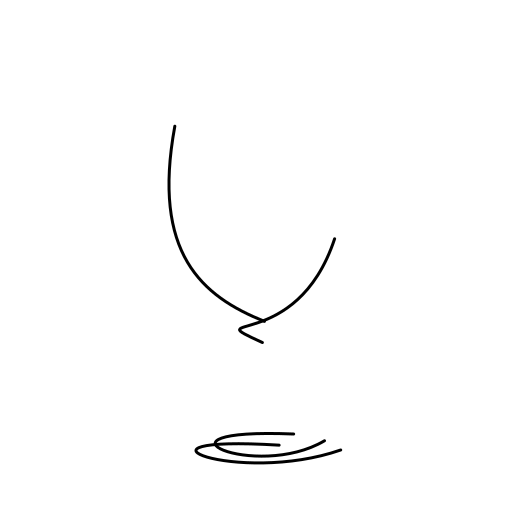} &
            \includegraphics[width=0.25\linewidth,height=0.25\linewidth]{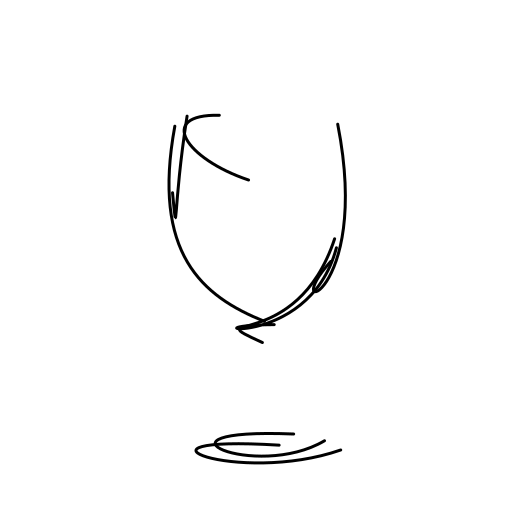} &
            \includegraphics[width=0.25\linewidth,height=0.25\linewidth]{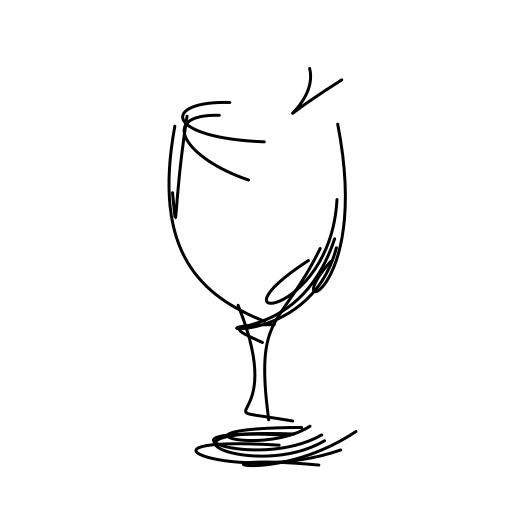} &
            \includegraphics[width=0.25\linewidth,height=0.25\linewidth]{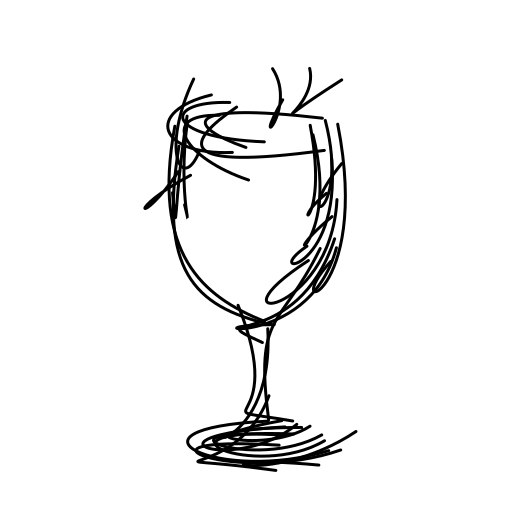} \\

            \multicolumn{4}{c}{``a teapot''} \\
            \includegraphics[width=0.25\linewidth,height=0.25\linewidth]{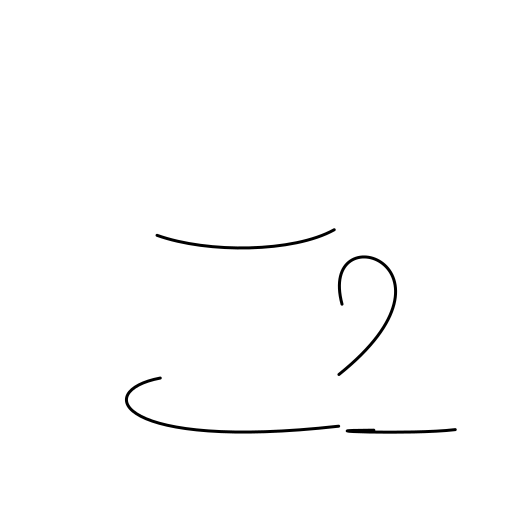} &
            \includegraphics[width=0.25\linewidth,height=0.25\linewidth]{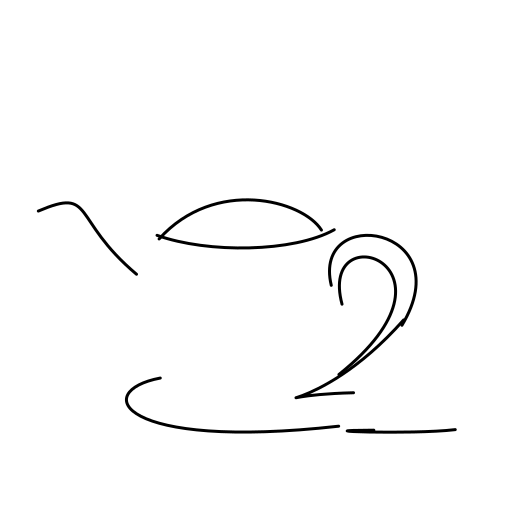} &
            \includegraphics[width=0.25\linewidth,height=0.25\linewidth]{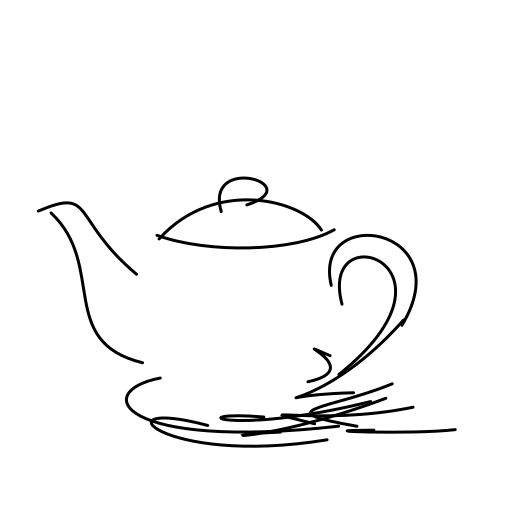} &
            \includegraphics[width=0.25\linewidth,height=0.25\linewidth]{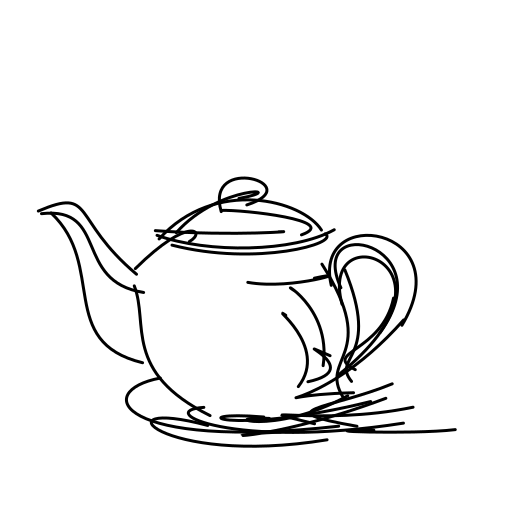} \\

        $4$ & $8$ & $16$ & $32$ \\
        \end{tabular}
    \end{minipage}
    \caption{\textbf{Additioanl Sketch Generation Results.} NeuralSVG can generate sketches with varying numbers of strokes using a single network, without requiring modifications to our framework.
    }
    \label{fig:sketches_dropout}
\end{figure*}

Below, we provide additional qualitative results and comparisons, as follows: 
\begin{enumerate}
    \item In~\Cref{fig:additional_results_dropout,fig:additional_results_dropout_2}, we present additional qualitative results produced by NeuralSVG when applying our dropout technique during inference. Specifically, we vary the number of learned shapes included in the final rendering, showing results with 1, 4, 8, 12, and all 16 shapes.
    \item In~\Cref{fig:additional_comparisons_vectorfusion_svgdream,fig:additional_comparisons_vectorfusion_svgdream_2}, we provide additional qualitative comparisons to open-source text-to-vector methods VectorFusion~\cite{jain2023vectorfusion} and SVGDreamer~\cite{xing2024svgdreamer}. 
    \item Following~\Cref{fig:additional_comparisons_vectorfusion_svgdream}, we provide corresponding outlines for the generated SVGs, showing that alternative methods have a tendency to produce nearly pixel-like shapes that are difficult to modify manually while NeuralSVG promotes individual shapes with more semantic meaning and order.
    \item In~\Cref{fig:additional_comparisons_nivel_text2svg}, we provide additional qualitative comparisons to closed-source techniques NIVeL~\cite{thamizharasan2024nivel} and Text-to-Vector~\cite{zhang2024text} using results presented in their respective papers. 
    \item Next, in~\Cref{fig:color_control,fig:color_control_2}, we show results obtained when rendering the learned SVG with different background colors at inference time, with both seen and unseen colors.
    \item In~\Cref{fig:additional_aspect_ratio}, we show additional results using our aspect ratio control, allowing us to generate SVGs at different aspect ratios using a single learned representation.
    \item Finally, in~\Cref{fig:sketches_dropout}, we show sketch generation results obtained using our NeuralSVG framework. Sketches are rendered using a varying number of strokes by modifying the truncation index at inference time. This approach enables a single learned representation to generate sketches at multiple levels of abstraction without modifying our text-to-vector framework.
\end{enumerate}

\begin{figure*}
    \centering
    \setlength{\tabcolsep}{1pt}
    {\small
    %
    }
    \caption{
    \textbf{Additional Qualitative Comparisons.} We provide additional visual comparisons to VectorFusion~\cite{jain2023vectorfusion} and SVGDreamer~\cite{xing2024svgdreamer} using a varying number of shapes.
    }
    \label{fig:additional_comparisons_vectorfusion_svgdream}
\end{figure*}
\begin{figure*}
    \centering
    \setlength{\tabcolsep}{1pt}
    {\small
    %
    }
    \caption{
    \textbf{Additional Qualitative Comparisons.} We provide additional visual comparisons to VectorFusion~\cite{jain2023vectorfusion} and SVGDreamer~\cite{xing2024svgdreamer} using a varying number of shapes.
    }
    \label{fig:additional_comparisons_vectorfusion_svgdream_2}
\end{figure*}
\begin{figure*}
    \centering
    \setlength{\tabcolsep}{1pt}
    {\small
    %
 \\
        
    \\[-0.5cm]        
    \end{tabular}
    }
    \caption{\textbf{Shape Outlines of the Generated SVGs.} We present the corresponding outlines of SVGs generated by NeuralSVG, VectorFusion, and SVGDreamer for the results shown in~\Cref{fig:additional_comparisons_vectorfusion_svgdream}. The alternative methods often produce nearly pixel-like shapes that are difficult to modify manually. In contrast, NeuralSVG generates cleaner SVGs, making them more editable and practical.}
    \label{fig:additional_outlines_comparisons_vectorfusion_svgdream}
\end{figure*}
\begin{figure*}
    \centering
    \setlength{\tabcolsep}{0.5pt}
    {\small
    \begin{tabular}{c c c c c c c c c c}

        \multicolumn{10}{c}{``a walrus smoking a pipe''} \\
        \includegraphics[height=0.1\textwidth,width=0.1\textwidth]{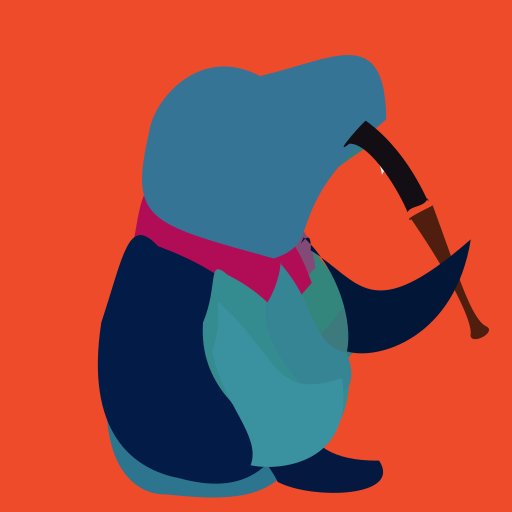} &
        \includegraphics[height=0.1\textwidth,width=0.1\textwidth]{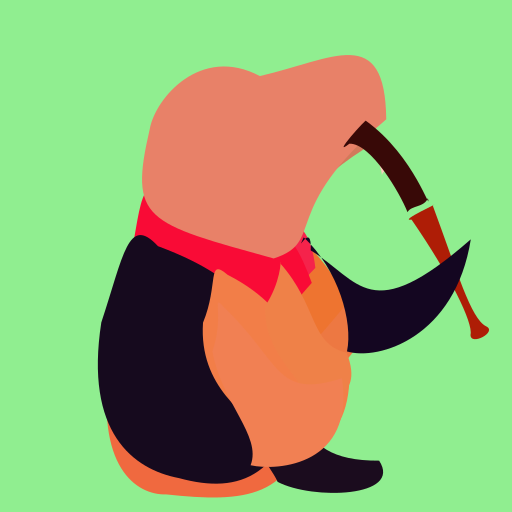} &
        \includegraphics[height=0.1\textwidth,width=0.1\textwidth]{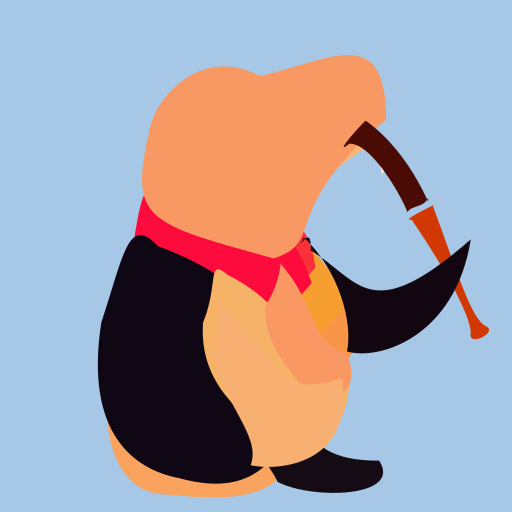} &
        \includegraphics[height=0.1\textwidth,width=0.1\textwidth]{images/colors/walrus/final_svg_gold.png} &
        \includegraphics[height=0.1\textwidth,width=0.1\textwidth]{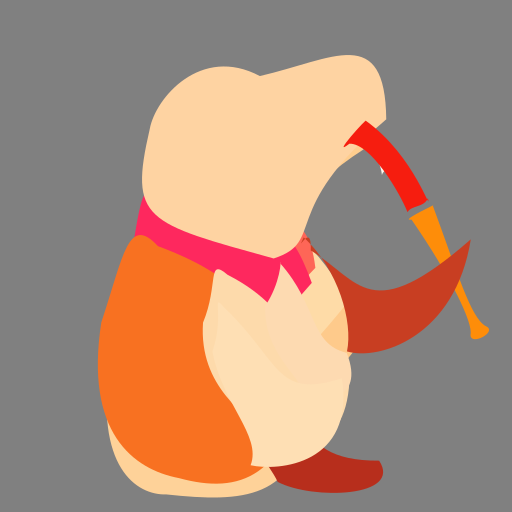} &
        \includegraphics[height=0.1\textwidth,width=0.1\textwidth]{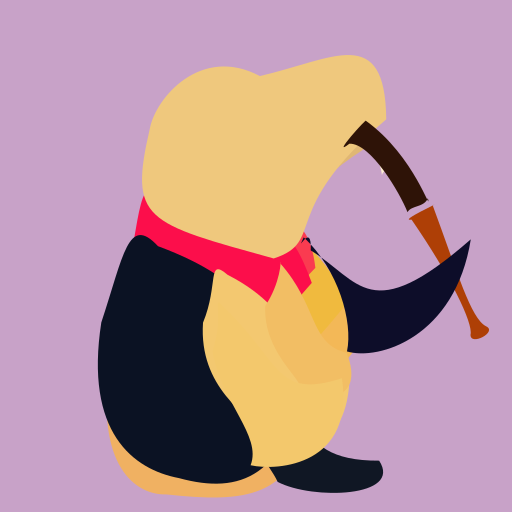} &
        \includegraphics[height=0.1\textwidth,width=0.1\textwidth]{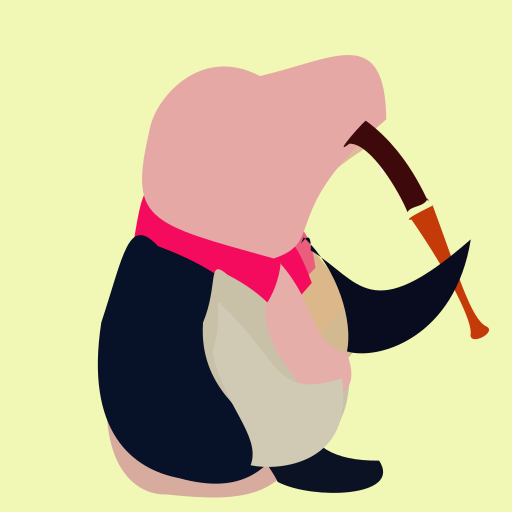} &
        \includegraphics[height=0.1\textwidth,width=0.1\textwidth]{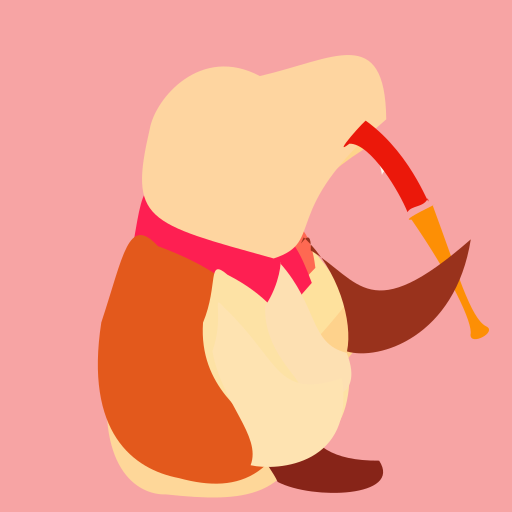} &
        \includegraphics[height=0.1\textwidth,width=0.1\textwidth]{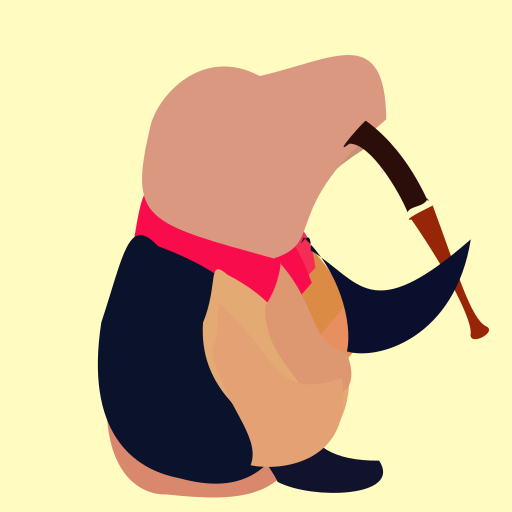} &
        \includegraphics[height=0.1\textwidth,width=0.1\textwidth]{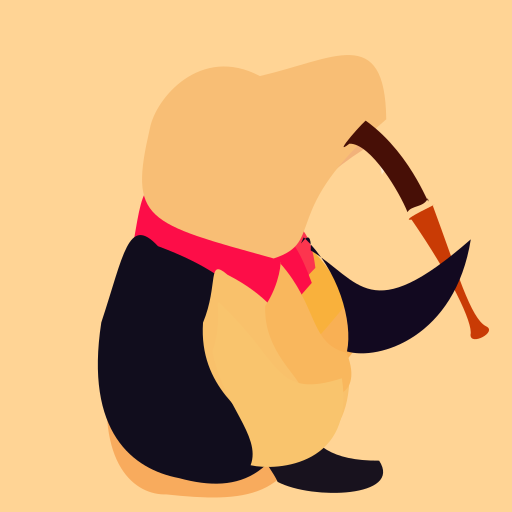} \\

        \multicolumn{10}{c}{``The Sydney Opera House''} \\
\end{tabular}
    }
    \caption{\textbf{Dynamically Controlling the Color Palette.} Given a learned representation, we render the result using different background colors specified by the user, resulting in varying color palettes in the resulting SVGs. The 5 leftmost columns show colors observed during training while the 5 rightmost columns show unobserved colors.}
    \label{fig:color_control}
\end{figure*}

\end{document}